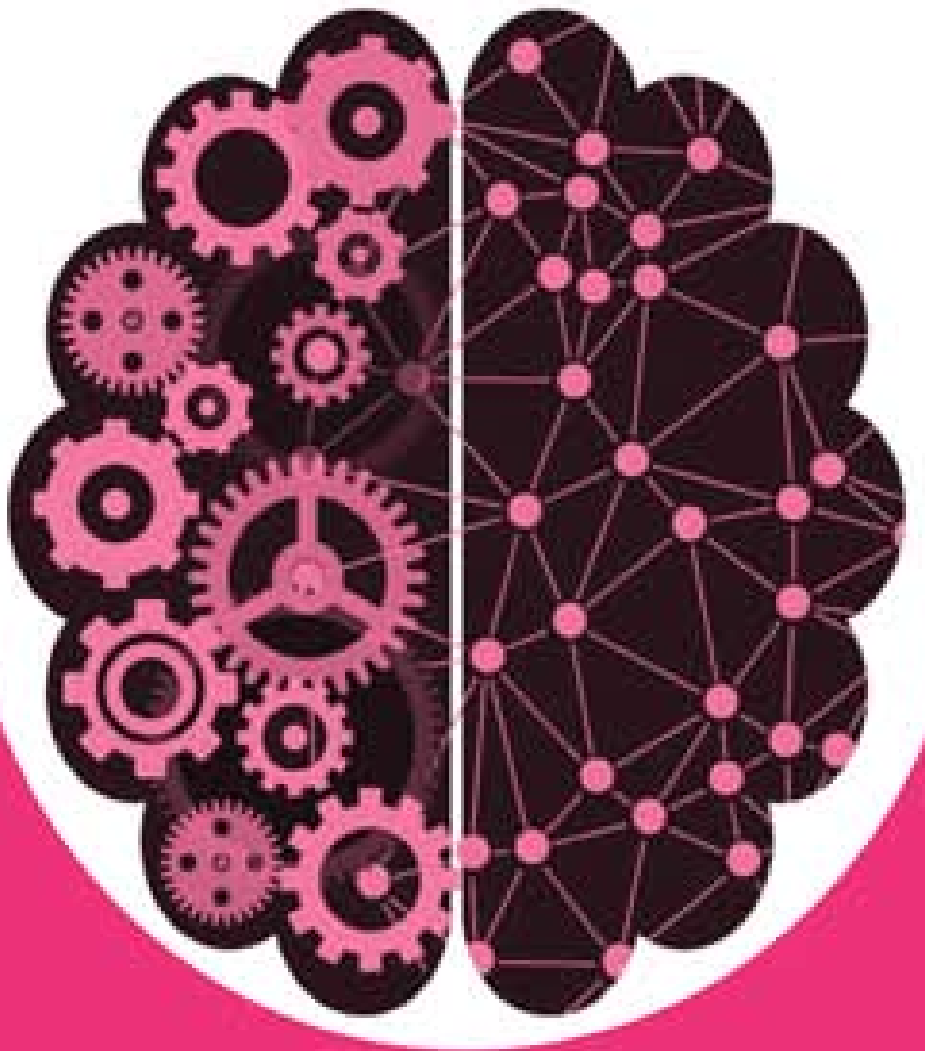

# یادگیری عمیق

از اصول اولیه تا ساخت شبکه های عصبی عمیق با پایتون

تالیف و گردآوری: میلاد وزان

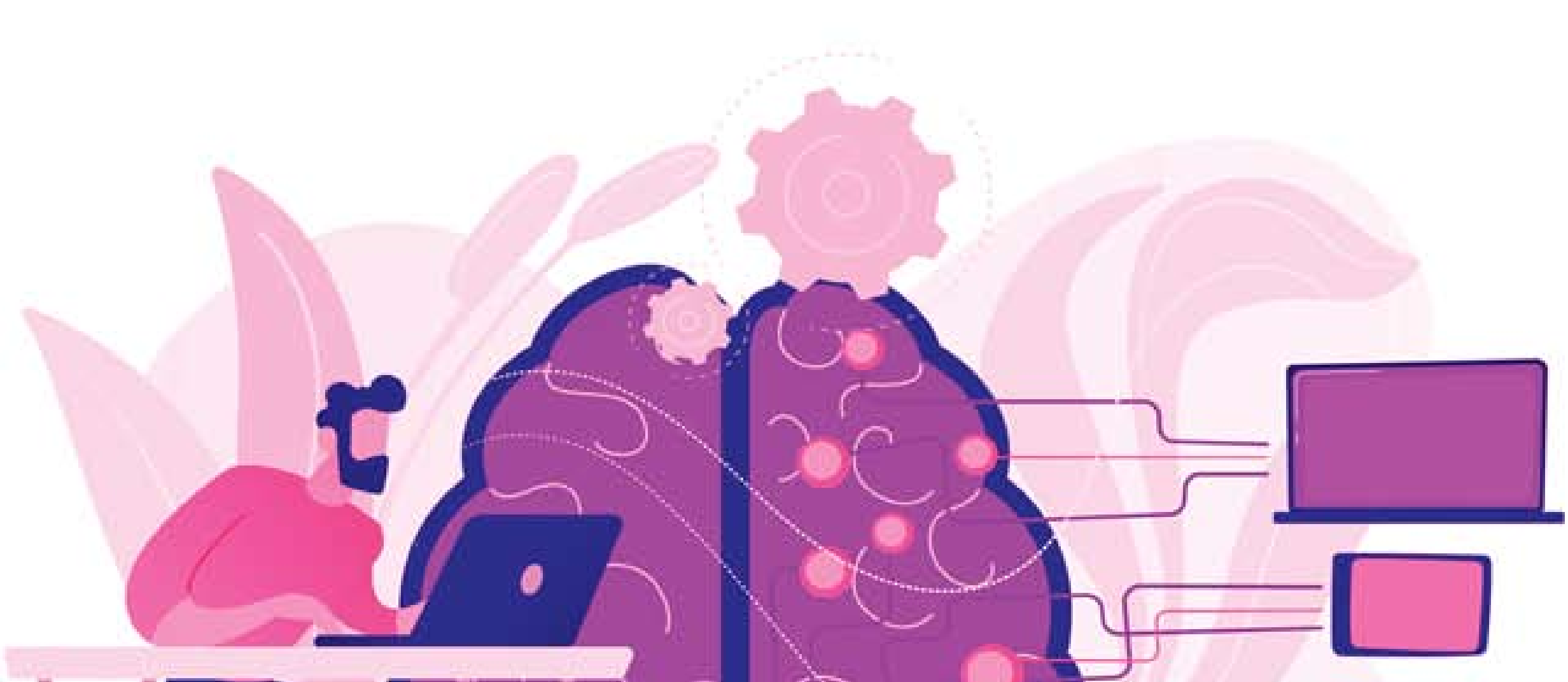

به نام خدا

# یادگیری عمیق

از اصول اولیه تا ساخت شبکه‌های عصبی عمیق با پایتون

تالیف و گردآوری:

میلاد وزان

---

# پیش‌گفتار

یادگیری عمیق یک فناوری جدید قدرتمند است که محبوبیت آن روز به روز در حوزه‌های مختلف در حالِ افزایش است. از این‌رو، بسیار مهم است که به یادگیری آن بپردازیم. این کتاب برای افراد مبتدی که هیچ آشنایی با یادگیری عمیق ندارند در نظر گرفته شده است تا خوانندگان را با یک دوره‌یِ فوقِ سریع در یادگیری عمیق آماده کند. تنها انتظار ما از خوانندگان این است که از قبل مهارت‌های برنامه‌نویسی اولیه در زبان پایتون را داشته باشند.

این راهنمایِ کوتاه، در نظر گرفته شده است تا شما را به عنوان یک مبتدی با درک درستی از موضوع، از جمله تجربه‌یِ عملیِ ملموس در توسعه مدل‌ها، مجهز کند. **اگر در حال حاضر بالاتر از سطح مبتدی هستید، این کتاب مناسب شما نیست!**

**میلاد وزان**

**کاشمر – بهار ۱۴۰۱**

09370174459

vazanmilad@gmail.com

# فهرست

## فصل اول: مقدمه‌ای بر یادگیری عمیق

## فصل دوم: پیش‌نیازها

## فصل سوم: شبکه‌های عصبی پیش‌خور

## فصل چهارم: شبکه‌های عصبی کانولوشنی

## فصل پنجم: شبکه‌های عصبی بازگشتی

## فصل ششم: شبکه متخاصم مولد

# مقدمه‌ای بر یادگیری عمیق

## اهداف یادگیری:

- یادگیری عمیق چیست؟
- علت محبوبیت یادگیری عمیق
- تفاوت آن با یادگیری ماشین
- آینده‌ی یادگیری عمیق
- کاربردهای یادگیری عمیق

# مقدمه

یادگیری عمیق زیرمجموعه‌ای از یادگیری ماشین است که بر استفاده از شبکه‌های عصبی برای حل مسائل پیچیده تمرکز دارد. امروزه به لطف پیشرفت‌های نرم‌افزاری و سخت‌افزاری که به ما امکان جمع‌آوری و پردازش مقادیر زیادی داده را می‌دهد، محبوبیت بیشتری پیدا کرده است. چراکه شبکه‌های عصبی عمیق برای عملکرد خوبی که ما از آن انتظار داریم به مقادیر زیادی داده و در نتیجه‌یِ آن به سخت‌افزار قدرتمند برای پردازش این حجم بزرگ داده نیاز دارد.

## یادگیری عمیق چیست؟

هوش مصنوعی اساسا شبیه‌سازی انسان‌ها و رفتارهای ذهنی آن‌ها توسط یک برنامه رایانه‌ای است که می‌تواند کارهایی را انجام دهند که معمولاً به هوش انسانی نیاز دارند. به عبارت ساده‌تر، سیستمی که می‌تواند رفتار انسان را تقلید کند. این رفتارها شامل حل مسئله، یادگیری و برنامه‌ریزی است که از طریق تجزیه و تحلیل داده‌ها و شناسایی الگوهای درون آن به منظور تکرار آن رفتارها بدست می‌آید.

کد، فناوری، الگوریتم یا هر سیستمی که بتواند مقوله فهم شناختی را تقلید کند که در خود یا در دستاوردهای آن پدیدار می‌شود، هوش مصنوعی است. این شامل یادگیری ماشین نیز می‌شود، جایی که ماشین‌ها می‌توانند با تجربه بیاموزند و بدون دخالت انسان مهارت‌هایی کسب کنند. از این‌رو، هوش مصنوعی سازنده‌ی یادگیری ماشین است. در واقع، یادگیری ماشین زیرمجموعه اصلی هوش مصنوعی است و می‌تواند ماشین‌ها را قادر سازد تا با استفاده از روش‌های آماری، تجربیات خود را با کیفیت‌تر و دقیق‌تر کنند. این امر به رایانه‌ها و ماشین‌ها امکان می‌دهد تا دستورات را بر اساس داده‌ها و یادگیری خود اجرا کنند. این برنامه‌ها یا الگوریتم‌ها به گونه‌ای طراحی شده‌اند که بتوانند در طول زمان اطلاعات بیشتری کسب کنند و با داده‌های جدید بهتر شوند و تطبیق پیدا کنند.

ایده اصلی اختراع یادگیری ماشین استدلال مبتنی‌بر نمونه است که فرآیند استدلال در مسئله مورد نظر با مراجعه به نمونه‌های مشابه قبلی ممکن می‌شود. مثال‌های قبلی که برای ایجاد ظرفیت استفاده می شود، نمونه‌های آموزشی نامیده شده و فرآیند انجام این کار یادگیری نامیده می‌شود. در سیستم‌های رایانه‌ای، تجربه در قالب داده‌ها وجود دارد و وظیفه اصلی یادگیری ماشین توسعه الگوریتم‌های یادگیری است که از داده‌ها مدل می‌سازد. با تغذیه داده‌های تجربی به الگوریتم یادگیری ماشین، ما مدلی را بدست می‌آوریم که می‌تواند پیش‌بینی‌هایی را در مشاهدات جدید انجام دهد.

یادگیری عمیق نیز زیرمجموعه‌ای از هوش مصنوعی و یادگیری ماشین است که در آن شبکه‌های عصبی مصنوعی، الگوریتم‌هایی که از مغز انسان الهام گرفته شده‌اند، از مقادیر زیادی داده توانایی یادگیری بدست می‌آورند. الگوریتم یادگیری عمیق مشابه نحوه یادگیری ما از تجربه‌ها و نمونه‌ها، یک کار را به طور مکرر انجام می‌دهد و هر بار کمی آن را تغییر می‌دهد تا نتیجه را بهبود بخشد. با انجام این کار، به رایانه‌ها کمک می‌کند تا ویژگی‌ها از داده‌ها پیدا کنند و با تغییرات سازگار شوند. قرار گرفتن مکرر در معرض مجموعه داده‌ها به ماشین‌ها کمک می‌کند تا تفاوت‌ها و منطق داده‌ها را درک کنند و به یک نتیجه‌گیری قابل اعتماد برسند. در ساده‌ترین حالت، یادگیری عمیق را می‌توان راهی برای خودکارسازی **تجزیه و تحلیل پیش‌گویانه** (**predictive analytics**) در نظر گرفت.

**تعریف ۱.۱** **یادگیری عمیق**

یادگیری عمیق، مجموعه‌ای از الگوریتم‌هایی است که "از طریق لایه‌ها یاد می‌گیرند". به عبارت دیگر، شامل یادگیری از طریق لایه‌هایی است که الگوریتم را قادر می‌سازد تا سلسله‌مراتبی از مفاهیم پیچیده را از مفاهیم ساده‌تر ایجاد کند.

برای درک بهتر یادگیری عمیق، کودک نوپایی را تصور کنید چیزی که در حال یاد گرفتن آن است گربه می‌باشد. کودک نوپا با اشاره به اشیاء و گفتن کلمه گربه یاد می‌گیرد که چه چیزی گربه است و چه چیزی گربه نیست. والدین می‌گویند: "بله، آن گربه است" یا "نه، آن گربه نیست". همان‌طور که کودک نوپا همچنان به اشیاء اشاره می‌کند، از ویژگی‌هایی که همه گربه‌ها دارند بیشتر و بیشتر آگاه می‌شود؛ کاری که کودک نوپا بدون این‌که بداند انجام می‌دهد. به این طریق است که یک انتزاع پیچیده (مفهوم گربه) را با ساختن سلسله مراتبی که در آن هر سطح از انتزاع با دانشی که از لایه قبلی سلسله مراتب بدست آمده، ایجاد می‌کند تا برایش این انتزاع پیچیده، ساده و روشن شود.

## پیدایش یادگیری عمیق؟

از آغاز عصر رایانه‌ها، محققان در مورد هوش ماشینی، نظریه‌پردازی می‌کردند و رویای داشتن رایانه‌های هوشمندی را در سر می پروراندند که بتوانند راه حل‌های مسائل پیچیده را بیاموزند و درک کنند. در دهه‌های ۱۹۵۰ و ۶۰، اولین شبکه‌های عصبی، از جمله الگوریتم پرسپترون برای طبقه‌بندی تصاویر، پدیدار شدند. با این حال، این موارد اولیه بسیار ساده بودند و نتوانستند محبوبیت گسترده‌ای داشته باشند.

شبکه‌های عصبی در دهه ۱۹۸۰ زمانی که محققان روش‌هایی را برای پس‌انتشار پارامترها به منظور ساخت و آموزش شبکه‌های عصبی چند سطحی توسعه دادند، دوباره ظهور کردند.

با پیشرفت‌ها در دهه ۲۰۰۰، تکنیک‌هایی پدیدار گشتند تا امکان افزایش لایه‌های شبکه‌های عصبی را فراهم کنند. این شبکه‌های چندلایه باعث شد که این حوزه از تحقیقات هوش مصنوعی "**یادگیری عمیق**" نامگذاری شود، چراکه الگوریتم‌ها داده‌ها را در چندین لایه پردازش می‌کنند تا به پاسخ برسند.

در سال ۲۰۱۲، شبکه های عصبی عمیق شروع به عملکردی بهتر از الگوریتم‌های طبقه‌بندی سنتی، از جمله الگوریتم‌های یادگیری ماشین کردند. این افزایش کارایی تا حد زیادی بدلیل افزایش عملکرد پردازنده‌های رایانه (GPU) و حجم انبوه داده‌ای است که اکنون در دسترس است. دیجیتالی شدن سریع منجر به تولید داده‌های در مقیاس بزرگ شده است و این داده‌ها اکسیژنی برای آموزش مدل‌های یادگیری عمیق هستند. از آن زمان، هر سال، یادگیری عمیق همچنان در حال بهتر شدن و تبدیل به بهترین رویکرد برای حل مشکلات در بسیاری از حوزه‌های مختلف شده است.

این انفجارِ محبوبیت و استفاده از یادگیری عمیق تا حد زیادی به لطف پیشرفت در سخت‌افزار و مجموعه داده‌های برچسب‌گذاری شده عظیم است که به مدل‌های یادگیری عمیق اجازه می‌دهد تا به سرعت در طول زمان بهبود یابند.

## علت محبوبیت یادگیری عمیق ؟

صنعت نرم‌افزار امروزه به سمت هوش ماشینی حرکت می‌کند و این یادگیری ماشین است که راهی برای هوشمندسازی ماشین‌ها بوجود آورده است. به بیان ساده، یادگیری ماشین مجموعه‌ای از الگوریتم‌هایی است که داده‌ها را تجزیه می‌کنند، از آنها یاد می‌گیرند و سپس آنچه را که یاد گرفته‌اند برای تصمیم‌گیری هوشمندانه به کار می‌گیرند. نکته‌ای که در مورد الگوریتم‌های یادگیری ماشین سنتی وجود دارد، این است که آن‌ها هر چقدر که پیچیده به نظر برسند، هم‌چنان شبیه ماشین هستند. به عبارت دیگر، آن‌ها برای بدست آوردن یادگیری به کارشناسان حوزه نیاز دارند. برای کارشناسان هوش مصنوعی، اینجا نقطه‌ای است که یادگیری عمیق نویدبخش می‌شود. چراکه شبکه‌های عصبی عمیق بدون نیاز به مداخله‌ی انسانی، ویژگی‌های سطح بالا را از داده‌ها به صورت افزایشی (سلسله‌مراتبی) یاد می‌گیرند. این امر نیاز به کارشناسان دامنه و استخراج ویژگی‌ها به صورت دستی را از بین می‌برد. انتخاب ویژگی برای یک مجموعه داده تاثیر بسیار زیادی در موفقیت یک مدل یادگیری ماشین دارد، حال آن که این استخراج ویژگی‌ها به‌صورت دستی فرآیندی زمان‌بر و پیچیده خواهد بود.

امروزه علاوه بر شرکت‌ها و سازمان‌ها، حتی افرادِ به سمتِ جنبه‌های فناوری، به یادگیری عمیق، تمایل دارند و هم‌چنان تعداد این شرکت‌ها و افراد در استفاده از یادگیری عمیق روز به روز در حال افزایش هستند. برای درک این دلیل، باید به مزایایی که می‌توان با استفاده از رویکرد

یادگیری عمیق بدست آورد، نگاه کرد. به‌طور خلاصه می‌توان مزیت‌های کلیدی که هنگام استفاده از این فناوری وجود دارد را به‌صورت زیر فهرست کرد:

- **عدم نیاز به مهندسی ویژگی‌ها:** در یادگیری ماشین، مهندسی ویژگی یک کار اساسی و مهم است. چراکه دقت را بهبود می‌بخشد و گاهی اوقات این فرآیند می‌تواند به دانش دامنه در مورد یک مساله خاص نیاز داشته باشد. یکی از بزرگترین مزایای استفاده از رویکرد یادگیری عمیق، توانایی آن در اجرای مهندسی ویژگی به صورت خودکار است. در این رویکرد، یک الگوریتم داده‌ها را اسکن می‌کند تا ویژگی‌های مرتبط را شناسایی کند و سپس آن‌ها را برای ارتقای سریع‌تر یادگیری، بدون اینکه به طور صریح به او گفته شود، ترکیب می‌کند. این توانایی به دانشمندان داده کمک می‌کند تا مقدار قابل توجهی در زمان صرفه‌جویی کرده و به دنبال آن نتایج بهتری را نیز بدست آورند.
- **حداکثر استفاده از داده‌های بدون ساختار:** تحقیقات نشان می‌دهد که درصد زیادی از داده‌های یک سازمان بدون ساختار هستند، زیرا اکثر آن‌ها در قالب‌های مختلفی همانند تصویر، متن و غیره هستند. برای اکثر الگوریتم‌های یادگیری ماشین، تجزیه و تحلیل داده‌های بدون ساختار دشوار است. از این‌رو، اینجاست که یادگیری عمیق مفید می‌شود. چراکه می‌توانید از قالب‌های داده‌ای مختلف برای آموزش الگوریتم‌های یادگیری عمیق استفاده کنید و همچنان بینش‌های مرتبط با هدف آموزش را بدست آورید. برای مثال، می‌توانید از الگوریتم‌های یادگیری عمیق برای کشف روابط موجود بین تجزیه و تحلیل صنعت، گفتگوی رسانه‌های اجتماعی و موارد دیگر برای پیش‌بینی قیمت‌های سهام آینده یک سازمان استفاده کنید.
- **ارائه نتایج با کیفیت بالا:** انسان‌ها گرسنه یا خسته می‌شوند و گاهی اوقات اشتباه می‌کنند. در مقابل، وقتی صحبت از شبکه‌های عصبی می‌شود، این‌طور نیست. هنگامی که یک مدل یادگیری عمیق بدرستی آموزش داده شود، می‌تواند هزاران کار معمولی و تکراری را در مدت زمان نسبتا کوتاه‌تری در مقایسه با آن‌چه که برای یک انسان لازم است، انجام دهد. علاوه بر این، کیفیت کار هرگز کاهش نمی‌یابد، مگر اینکه داده‌های آموزشی حاوی داده‌های خامی باشد که نشان‌دهنده مساله‌ای نیست که می‌خواهید آن راحل کنید.
- **یادگیری انتقالی:** یادگیری عمیق دارای چندین مدل از پیش‌آموزش دیده با وزن‌ها و سوگیری‌های ثابت است که برخی از این الگوریتم‌ها در پیش‌بینی بسیار عالی هستند.

- **دقت بالای نتایج:** هنگامی که یادگیری عمیق با حجم عظیمی از داده آموزش داده می‌شود، می‌تواند دقت خوبی در مقایسه با الگوریتم‌های یادگیری ماشین سنتی داشته باشد.

با در نظر گرفتن مزایای فوق و استفاده بیشتر از رویکرد یادگیری عمیق، می‌توان گفت که تاثیر قابل توجه یادگیری عمیق در فناوری‌هایِ مختلفِ پیشرفته همانند اینترنت اشیا در آینده بدیهی است. یادگیری عمیق، راهِ درازی را طی کرده است و به سرعت در حال تبدیل شدن به یک فناوری حیاتی است که به‌طور پیوسته توسط مجموعه‌ای از کسب‌وکارها، در صنایع مختلف مورد استفاده قرار می‌گیرد.

با این حال باید توجه داشت که یادگیری عمیق همچنین ممکن است بهترین انتخاب بر اساس داده‌ها نباشد. به عنوان مثال، اگر مجموعه داده کوچک باشد، گاهی اوقات مدل‌های یادگیری ماشینِ خطیِ ساده‌تر ممکن است نتایج دقیق‌تری به همراه داشته باشند. هرچند، برخی از متخصصان یادگیری ماشین استدلال می‌کنند که یک شبکه‌یِ عصبیِ عمیقِ آموزش‌دیده‌یِ مناسب، همچنان می‌تواند با مقادیرِ کمِ داده، عملکرد خوبی داشته باشد.

## یادگیری عمیق چگونه کار می‌کند؟

مدل‌های یادگیری عمیق با تجزیه و تحلیل مداوم داده‌ها و با کشف ساختارهای پیچیده در داده‌ها توانایی یادگیری بدست می‌آورند. فرآیند یادگیری با ساخت مدل‌های محاسباتی به نام شبکه‌های عصبی که از ساختار مغز الهام گرفته شده، حاصل می‌شود. هسته اصلی این یادگیری به روشی تکراری در راستای آموزش ماشین‌ها برای تقلید از هوش انسانی متکی است. یک شبکه عصبی مصنوعی این روش تکراری را از طریق چندین سطح سلسله مراتبی انجام می‌دهد و در این ساختار با رفتن به لایه‌های سطح بعدی، قادر به حل مفاهیم پیچیده‌تری از مسئله می‌شود. سطوح اولیه به ماشین‌ها کمک می‌کنند تا اطلاعات ساده را بیاموزند. با رفتن به هر سطح جدید، ماشین‌ها اطلاعات بیشتری را جمع‌آوری کرده و آن‌ها را با آنچه در آخرین سطح آموخته بوده‌اند ترکیب می‌کنند. در پایان فرآیند، سیستم یک قطعه اطلاعات نهایی را جمع‌آوری می‌کند که یک ورودی ترکیبی است. این اطلاعات از چندین سلسله مراتب عبور می‌کند و شبیه به تفکرِ منطقیِ پیچیده است.

بیایید با کمک یک مثال آن را بیشتر تجزیه کنیم. دستیار صوتی مانند الکسا یا سیری را در نظر بگیرید تا ببینید چگونه از یادگیری عمیق برای تجربیات مکالمه طبیعی استفاده می‌کند. در سطوح اولیه شبکه عصبی، زمانی که دستیار صوتی با داده‌ها تغذیه می‌شود، سعی می‌کند صداها و موارد دیگر را شناسایی کند. در سطوح بالاتر، اطلاعات مربوط به واژگان را می‌گیرد و یافته‌های

سطوح قبلی را به آن اضافه می‌کند. در سطوح بعدی، اعلانات (فرمان‌ها) را تجزیه و تحلیل می‌کند و تمام نتایج خود را ترکیب می‌کند. برای بالاترین سطحِ ساختارِ سلسله مراتبی، دستیار صوتی به اندازه کافی آموخته است که بتواند یک دیالوگ را تجزیه و تحلیل کند و بر اساس آن ورودی، اقدام مربوط را ارائه دهد.

در یادگیری عمیق، نیازی به برنامه‌نویسیِ صریحِ همه‌چیز نداریم. آن‌ها می‌توانند به‌طور خودکار بازنمایی‌هایی را از داده‌هایی مانند تصاویر، ویدیو یا متن، بدون معرفی قوانین دستی یاد بگیرند. معماری‌هایِ بسیار انعطاف‌پذیر آن‌ها می‌توانند مستقیما از داده‌های خام یاد بگیرند و در صورت ارائه داده‌های بیشتر می‌توانند عملکرد خود را افزایش دهند.

## معایب و چالش‌های یادگیری عمیق

اگرچه اهمیت و پیشرفت‌های یادگیری عمیق در حال افزایش است، اما چند جنبه منفی یا چالش وجود دارد که برای توسعه یک مدل یادگیری عمیق باید با آن‌ها مقابله کرد. بزرگترین محدودیت مدل‌های یادگیری عمیق این است که آن‌ها از طریق مشاهدات یاد می‌گیرند. این بدان معنی است که آنها فقط می‌دانند که در داده‌هایی که در آن آموزش داده‌اند چه چیزی وجود دارد و تنها در نگاشت بین ورودی و خروجی بسیار خوب هستند. به عبارت دیگر، آن‌ها از زمینه‌یِ داده‌هایی که از آن‌ها استفاده می‌کنند چیزی نمی‌دانند. در حقیقت، کلمه "عمیق" در یادگیری عمیق بیشتر اشاره به مرجع معماری فناوری و تعداد لایه‌های پنهان است که در ساختار آن قرار دارد نه درک عمیقی از آنچه که در حال انجام آن است.

مدل‌های یادگیری عمیق یکی از دادهِ خوارترین مدل‌های داده در دنیای یادگیری ماشین هستند. آن‌ها به حجم عظیمی از داده ها نیاز دارند تا به عملکرد مطلوب خود برسند و به قدرتی که از آن‌ها انتظار داریم به ما خدمت کنند. با این حال، داشتن این مقدار داده همیشه آسان نیست. علاوه بر این، در حالی‌که می‌توانیم مقادیر زیادی داده در مورد یک موضوع داشته باشیم، اغلب اوقات برچسب‌گذاری نمی‌شود، بنابراین نمی‌توانیم از آن برای آموزش هر نوع الگوریتم یادگیری بانظارت استفاده کنیم. به‌طور خلاصه، اگر کاربر مقدار کمی داده داشته باشد این مدل‌ها به روشی قابل تعمیم یاد نمی‌گیرند. یادگیری عمیق زمانی می‌تواند بهترین عملکرد را داشته باشد که حجم زیادی از داده‌های با کیفیت در دسترس باشد. با افزایش داده‌های موجود، عملکرد سیستم یادگیری عمیق نیز رشد می‌کند.

زمانی که داده‌های با کیفیت به سیستم وارد نمی‌شوند، یک سیستم یادگیری عمیق می‌تواند با شدت شکست مواجه شود.

موضوع **سوگیری‌ها** (biases) نیز یک مشکل عمده برای مدل‌های یادگیری عمیق است. اگر یک مدل بر روی داده‌هایی آموزش ببیند که دارای سوگیری هستند، مدل آن سوگیری‌ها را در پیش‌بینی‌های خود بازتولید می‌کند.

اگرچه مدل‌های یادگیری عمیق بسیار کارآمد هستند و می‌توانند یک راه‌حل مناسب برای یک مشکل خاص پس از آموزش با داده‌ها فرموله کنند، اما برای یک مساله مشابه قادر به انجام این کار نیستند و نیاز به آموزش مجدد دارند. برای نشان دادن این موضوع، یک الگوریتم یادگیری عمیق را در نظر بگیرید که یاد می‌گیرد اتوبوس‌های مدرسه همیشه زرد هستند، اما ناگهان اتوبوس‌های مدرسه آبی می‌شوند. از این‌رو، باید دوباره آموزش داده شود. برعکس، یک کودک پنج ساله مشکلی برای تشخیص وسیله نقلیه به عنوان یک اتوبوس مدرسه آبی ندارد. علاوه بر این، آن‌ها همچنین در موقعیت‌هایی که ممکن است کمی متفاوت با محیطی باشد که با آن تمرین کرده‌اند، عملکرد موثری ندارند. برای مثال DeepMind گوگل سیستمی را برای شکست ٤٩ بازی آتاری آموزش داد. با این حال، هر بار که سیستم یک بازی را شکست می‌داد، باید برای شکست دادن بازی بعدی دوباره آموزش داده می‌شد. این ما را به محدودیت دیگری در یادگیری عمیق می‌رساند، یعنی در حالی که مدل ممکن است در نگاشت ورودی‌ها به خروجی‌ها فوق‌العاده خوب باشد، اما ممکن است در درک زمینه داده‌هایی که آنها مدیریت می‌کنند خوب نباشد.

الگوی یادگیری عمیق یا به‌طور کلی‌تر الگوریتم‌های یادگیری ماشینِ فعلی به صورت مجزا یاد می‌گیرد: با توجه به مجموعه داده‌های آموزشی، الگوریتم یادگیری ماشین را روی مجموعه داده اجرا می‌کند تا یک مدل تولید کند و هیچ تلاشی برای حفظِ دانشِ آموخته شده و استفاده از آن در یادگیریِ آینده انجام نمی‌دهد. اگرچه این الگویِ یادگیریِ مجزا بسیار موفقیت‌آمیز بوده است، اما به تعداد زیادی نمونه آموزشی نیاز دارد و فقط برای کارهایی که به خوبی تعریف شده و محدود هستند مناسب است. با دردسترس قرار گرفتن مجموعه داده‌های بزرگ‌تر و کاهش هزینه‌های محاسباتی، مدل‌هایی که قادر به حل وظایف بزرگ‌تر هستند نیز در دسترس شدند. با این حال، آموزش یک مدل هر بار که نیاز به یادگیری یک کار جدید دارد، ممکن است غیرممکن باشد. چراکه ممکن است داده‌های قدیمی‌تر در دسترس نباشند، داده‌های جدید به دلیل مشکلات حفظ حریم خصوصی نتوانند ذخیره شوند یا دفعاتی که سیستم باید در آن بروزرسانی شود، نمی‌تواند از آموزش یک مدل جدید با تمام داده‌ها به اندازه کافی مکرر پشتیبانی کند. وقتی شبکه‌های عصبی عمیق وظایف جدیدی را یاد می‌گیرند، در صورت عدم استفاده از معیارهای خاص، دانش جدید بر دانش قدیمی‌تر اولویت داده می‌شود و معمولا باعث فراموشی دانش دوم می‌شود. این موضوع معمولا به عنوان **فراموشی فاجعه‌آمیز** (**catastrophic forgetting**) شناخته می‌شود (شکل ۱ـ۱ را ببینید). فراموشی فاجعه‌آمیز زمانی اتفاق می‌افتد که یک شبکه عصبی آموزش‌دیده، زمانی که برای انجام وظایف جدید تطبیق داده می‌شود قادر به حفظ توانایی خود برای انجام وظایفی که قبلا آموخته است، نیست.

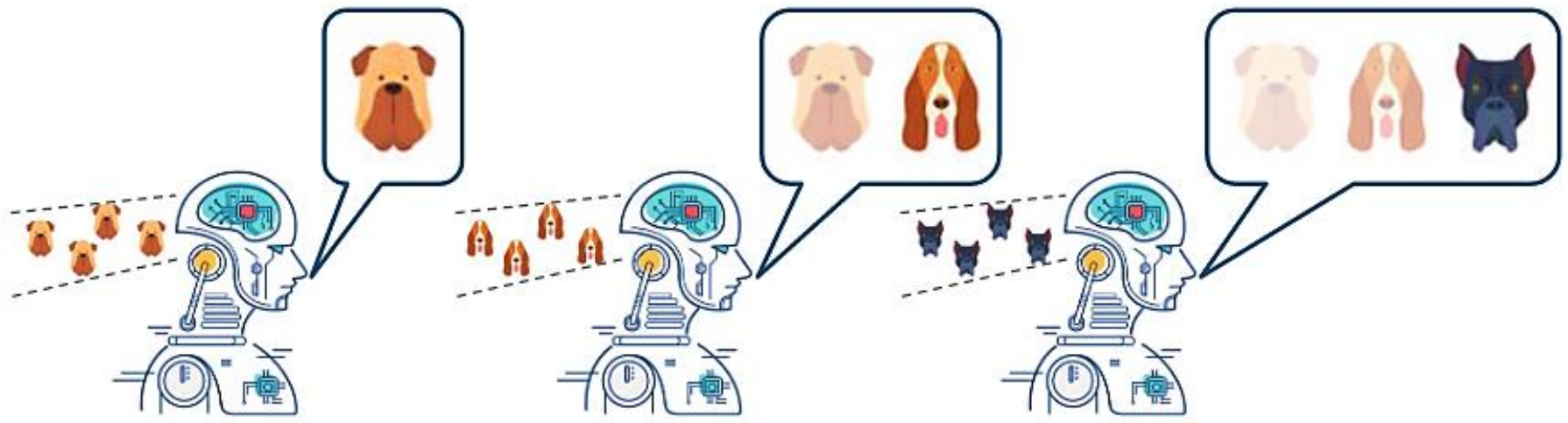

**شکل ۱ـ۱. تصویری از فراموشی فاجعه‌آمیز.** دانشِ آموخته شده‌یِ قبلی هنگام یادگیری کلاس‌های جدیدی که برای مدتی دیده نشده، فراموش می‌شود (به صورت تدریجی محو می‌شود).

مشکل دیگر شبکه‌های عصبی عمیق این است که آن‌ها اغلب با فرضیات جهان بسته آموزش داده می‌شوند، یعنی فرض می‌شود که توزیع داده‌های آزمایشی مشابه توزیع داده‌های آموزشی است. با این حال، هنگامی که در کارهای دنیای واقعی بکار گرفته می‌شوند، این فرض درست نیست و منجر به کاهش قابل توجهی از عملکرد آن‌ها می‌شود. هنگامی که شبکه‌های عصبی عمیق داده‌هایی را پردازش می‌کنند که شبیه توزیع مشاهده شده در زمان آموزش نیستند که به اصطلاح **خارج ازتوزیع نامیده** (**Out-of-distribution**) می‌شوند، اغلب پیش‌بینی‌های اشتباهی انجام می‌دهند و این کار را با اطمینان بیش از حد انجام می‌دهند (شکل ۱ـ۲ را ببینید). در این موارد خروجی شبکه یک تناظر مستقیم با راه‌حل مساله دارد، یعنی احتمال برای هر کلاس. با این حال، جمع نمایشِ بردارِ خروجی مجبور است همیشه به یک برسد. این بدان معناست که وقتی به شبکه یک ورودی نشان داده می‌شود که بخشی از توزیع آموزشی نیست، بازهم احتمال را به نزدیک‌ترین کلاس می‌دهد تا جمع احتمالات به یک برسد. این پدیده منجر به مشکل شناخته شده‌ی شبکه‌های عصبی زیاد مطمئن (overconfident) به محتوایی شده است که هرگز ندیده‌اند.

اگرچه این افت عملکرد برای کاربردهای همانند توصیه‌گرهای محصول قابل قبول است، اما استفاده از چنین سیستم‌هایی در حوزه‌های همانند پزشکی و رباتیک خانگی خطرناک است، چرا که می‌توانند باعث بروز حوادث جدی شوند. یک سیستم هوش مصنوعی ایده‌آل باید در صورت امکان به نمونه‌های خارج از توزیع تعمیم پیدا کند. بنابراین، توانایی تشخیصِ خارج از توزیع برای بسیاری از برنامه‌های کاربردی دنیای واقعی بسیار مهم است و برای تضمین قابلیت اطمینان و ایمنی سیستم‌های یادگیری ماشین ضروری است. به عنوان مثال، در رانندگی خودران، ما می‌خواهیم سیستم رانندگی زمانی که صحنه‌های غیرعادی یا اشیایی را که قبلا ندیده است و نمی‌تواند تصمیم ایمن بگیرد را تشخیص دهد، هشدار بدهد و کنترل را به انسان واگذار کند.

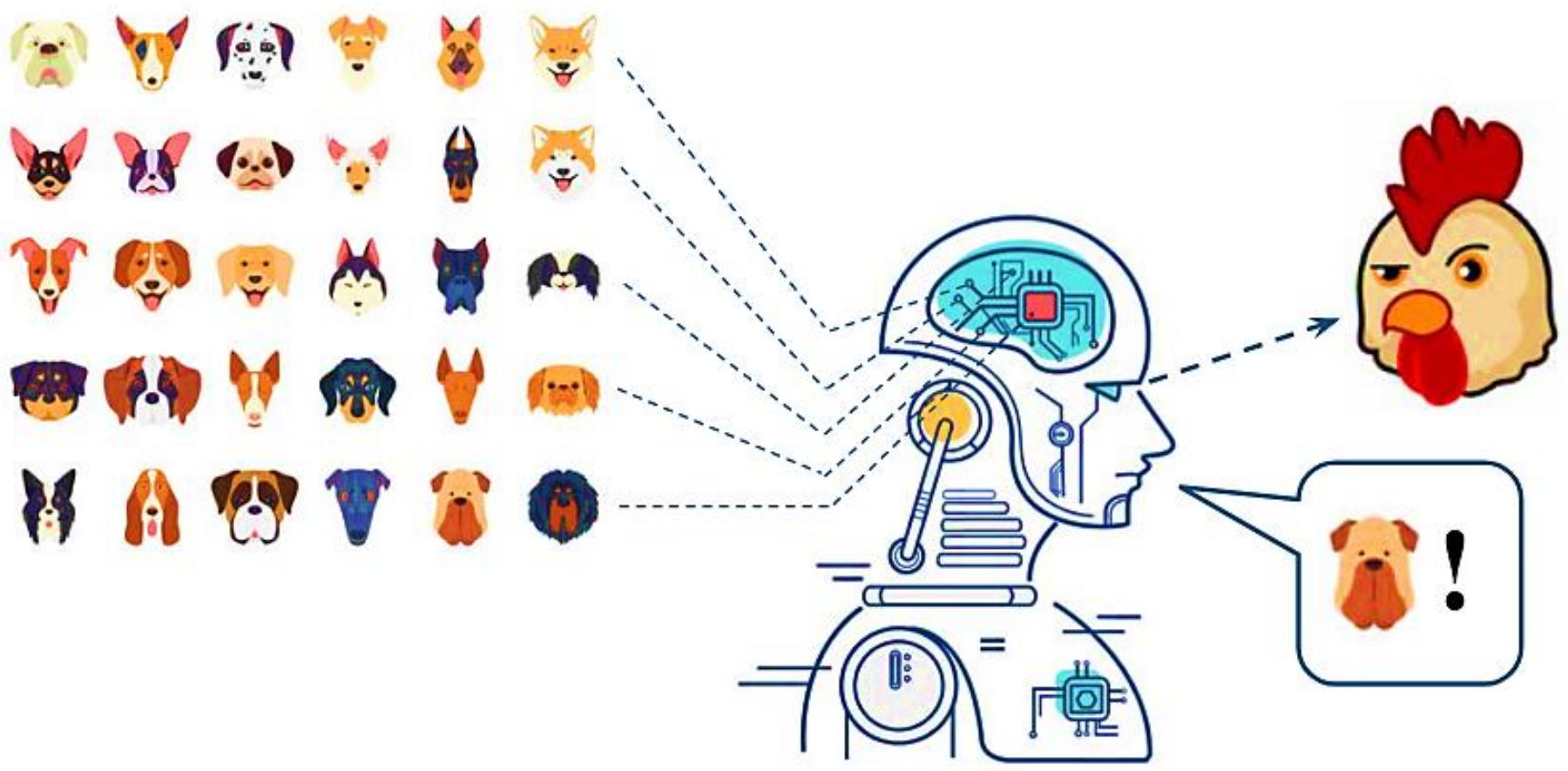

**شکل ۱ـ۲.** هنگامی که یک نمونه‌ی جدیدِ خارج از توزیعِ آموخته شده ارائه شود، شبکه‌های عصبی عمیق یک کلاس را از توزیع آموخته‌شده با اطمینان زیادی پیش‌بینی می‌کنند.

در نهایت، شناخته‌شده‌ترین نقطه ضعف شبکه‌های عصبی ماهیت **عدم شفافیت** آنهاست. در حالی‌که، تصمیمات گرفته شده توسط مدل‌های مبتنی بر قاعده را می‌توان توسط دستورات $if$ و $else$ ردیابی کرد، در یادگیری عمیق چنین چیزی امکان پذیر نخواهد بود. این عدم شفافیت همان چیزی است که در یادگیری عمیق از آن به عنوان "**جعبه سیاه**" یاد می‌شود.

به عبارت ساده، شما نمی‌دانید چگونه و یا چرا شبکه عصبی شما خروجی خاصی را بدست آورده است. به عنوان مثال، وقتی تصویری از یک گربه را به یک شبکه عصبی تغذیه می‌کنید و آن را یک ماشین پیش‌بینی می‌کند، درک اینکه چه چیزی باعث شده است که به این پیش‌بینی برسد بسیار سخت است. این سناریو در تصمیمات تجاری مهم خواهند بود. آیا می‌توانید تصور کنید که مدیر عامل یک شرکت بزرگ تصمیمی در مورد میلیون‌ها دلار بگیرد بدون اینکه بفهمد چرا باید این کار را انجام دهد؟ فقط به این دلیل که "رایانه" می‌گوید او باید این کار را انجام دهد؟ در مقایسه، الگوریتم‌هایی مانند درخت تصمیم بسیار قابل تفسیر هستند.

الگوریتم‌های یادگیری عمیق، الگوها و همبستگی‌ها را از طریق داده‌های تغذیه شده به آن پیدا می‌کنند و گاهی تصمیماتی می‌گیرند که حتی برای مهندسانی که آن‌ها را ایجاد کرده‌اند، گیج کننده است. این امر زمانی که یادگیری عمیق کاری با اهمیت کم را انجام می‌دهد، مشکلی را بوجود نخواهد آورد. اما وقتی تصمیم به سرنوشت یک متهم در دادگاه یا معالجه پزشکی بیمار باشد، بسیار سرنوشت‌ساز خواهد بود. چراکه اشتباهات می‌تواند عواقب زیادی را به‌همراه داشته باشد. طبق گفته گری مارکوس:

**"مسئله شفافیت هنوز حل نشده است، کاربران هنگام استفاده از یادگیری عمیق برای کار در حوزه‌های تشخیص پزشکی و تجارت مالی، دوست دارند درک کنند که چگونه یک سیستم مشخص یک تصمیم مشخصی را گرفته است."**

روی هم رفته، به گفته اندرو انگ، یادگیری عمیق راهی عالی برای "ساخت جامعه‌ای مبتنی‌بر هوش مصنوعی" است و غلبه بر این کاستی‌ها با کمک سایر فن‌آوری‌ها، راه درست برای رسیدن به این هدف است.

## آینده‌ی یادگیری عمیق؟

یادگیری عمیق در حال حاضر موثرترین فناوری هوش مصنوعی برای کاربردهای متعدد است. با این حال، نظرات متفاوتی در مورد توانایی‌هایِ یادگیری عمیق وجود دارد. در حالی که برخی محققان یادگیری عمیق بر این باورند که همه مشکلات را می‌توان با یادگیری عمیق حل کرد، دانشمندان زیادی وجود دارند که به نقص‌هایی در یادگیری عمیق اشاره می‌کنند.

گری مارکوس روانشناس پژوهشی یکی از پیشگامان در حوزه‌یِ یادگیری عمیق!!!، روش‌های جدیدی را برای بهبود راه‌حل‌های یادگیری عمیق پیشنهاد می‌کند. این روش‌ها شامل معرفی استدلال یا دانشِ قبلی به یادگیری عمیق، یادگیری خودنظارتی، شبکه‌های کپسولی و غیره هستند. گری مارکوس، تاکید می‌کند که تکنیک‌های یادگیری عمیق، داده‌ِ خوار و شکننده هستند و توانایی آن‌ها برای تعمیم محدود است.

یان لکان، دانشمند رایانه می‌گوید: "هیچ یک از تکنیک‌های هوش مصنوعی که ما در اختیار داریم نمی‌توانند از طریق ساختار و یا از طریق یادگیری، بازنمایی‌هایی از جهان بسازد که به چیزی نزدیک به آنچه در حیوانات و انسان‌ها مشاهده می‌کنیم باشد". از این‌رو تکنیک‌های فعلی هوش مصنوعی مانند یادگیری عمیق هنوز هم قادر به ایجاد یک هوش مصنوعی عمومی که دارای هوش قابل مقایسه با حیوانات یا انسان است را ندارند. با این حال، لکان معتقد است که هوش مصنوعی می‌تواند به سمت توسعه هوش عمومی مبتنی بر یادگیری عمیق بدون‌نظارت پیشرفت کند. پیشرفتی که اخیرا نیاز انسان به ارائه داده‌های برچسب‌دار دستی را که ماشین‌ها از آن‌ها یاد می‌گیرند، برطرف می‌کند.

گری مارکوس که از مدافعانِ رویکرد ترکیبی برای یادگیری عمیق است، یک برنامه‌یِ چهار مرحله‌ای را برای آینده‌ی یادگیری عمیق پیشنهاد می‌کند:

۱. **اتصال به دنیای هوش مصنوعی کلاسیک.** مارکوس رهایی از یادگیری عمیق را پیشنهاد نمی‌کند، بلکه ادعا می‌کند که ما باید از سایر رویکردهای هوش مصنوعی مانند دانش قبلی، استدلال و مدل‌های شناختیِ غنی همراه با یادگیری عمیق برای تغییرات دگرگون‌کننده استفاده کنیم.

۲. **ساخت چارچوب‌های شناختی غنی و پایگاه‌های اطلاعاتیِ دانش در مقیاس بزرگ.** سیستم‌های یادگیری عمیق به‌طور عمده فقط با همبستگی بین چیزهای خاص پر شده‌اند. بنابراین به دانش زیادی نیاز دارند.

۳. **ابزارهایی برای استدلال انتزاعی برای تعمیم مؤثر.** ما باید بتوانیم در مورد این چیزها استدلال کنیم. فرض کنید در مورد اشیاء فیزیکی و موقعیت آن‌ها در جهان اطلاعات داریم، برای مثال یک فنجان. فنجان حاوی مداد است. سپس سیستم‌های یادگیری عمیق باید بتوانند متوجه این موضوع شوند که اگر سوراخی در ته فنجان ایجاد کنیم، ممکن است مدادها بیفتند. انسان‌ها همیشه این نوع استدلال را انجام می‌دهند، اما سیستم‌های یادگیری عمیق یا به‌طور کلی‌تر هوش مصنوعی فعلی، این توانایی را ندارند.

۴. **مکانیسم‌هایی برای بازنمایی و القای مدل‌های شناختی.**

یادگیری عمیق هنوز راه درازی در پیش دارد تا بتواند به ظرفیت‌های همتایان انسانی خود دست یابد.

## استدلال نمادین (symbolic reasoning)

تاریخچه‌یِ تحقیقات هوش مصنوعی گاهی اوقات به عنوان یک کشمکش بین دو رویکرد مختلفِ استدلال نمادین و یادگیری ماشین توصیف می‌شود. در دهه اول، استدلال نمادین غالب بود، اما یادگیری ماشین در دهه ۱۹۹۰ شروع به نفوذ کرد و با انقلاب یادگیری عمیق، این حوزه را فراگرفت. با این حال، به نظر می‌رسد، استدلال نمادین تنها مجموعه‌یِ دیگری از روش‌هاست که ممکن است به گسترش و قدرتمندتر کردنِ یادگیری عمیق منجر شود.

نیکو استروم می گوید: "**شبکه‌های دگرگون‌ساز (Transformer networks)** چیزی به نام **توجه (attention)** دارند. بنابراین می‌توان یک بردار در شبکه داشت و شبکه می‌تواند بیشتر از همه‌یِ اطلاعات دیگر به آن بردار بپردازد. از این‌رو اگر پایگاه دانشی از اطلاعات دارید، می‌توانید آن را با بردارهایی که نشان دهنده حقیقت در آن پایگاه دانش هستند، از قبل پر کنید و سپس می‌توانید از شبکه بخواهید که بسته به ورودی، به دانش درست توجه کند. به این ترتیب می‌توانید سعی کنید دانشِ ساختار یافته‌یِ جهان را با سیستم یادگیری عمیق ترکیب کنید.

شبکه‌های عصبی گراف نیز وجود دارند که می‌توانند دانشِ درباره‌یِ جهان را نشان دهند. شما گره‌ها و یال‌هایی بین این گره‌ها دارید که روابط بین آن‌ها را نشان می‌دهند. بنابراین، برای مثال، می‌توانید موجودیت‌هایی را در گره‌ها و سپس روابط بین موجودیت‌ها را نشان دهید. می‌توانیم از توجه در بخشی از گراف دانش که برای زمینه یا سوال فعلی مهم است استفاده کنیم.

به نظر می‌رسد که می‌توان تمام دانش را در یک گراف نشان داد. با این حال، نکته‌ی مهم اسن است که چگونه می‌توان آن را به روشی کارآمد و مناسب انجام داد؟

"هینتون خیلی وقت پیش این ایده را داشت. او آن را **بردار فکر (thought vector)** نامید. هر فکری که می‌توانید داشته باشید، می‌توانیم با یک بردار نمایش دهیم. دلیل جالب بودن این

است که ما می‌توانیم هر چیزی را در گراف نشان دهیم، اما برای اینکه این کار به خوبی هماهنگ با یک مدل یادگیری عمیق باشد، باید از طرف دیگر چیزی نیز داشته باشیم که بتوانیم هر چیزی را با آن نشان دهیم و اتفاقا این همان بردار است. بنابراین ما می‌توانیم بین این دو یک نگاشت ایجاد کنیم."

## کاربردهای یادگیری عمیق؟

امروزه یادگیری عمیق به ابزاری قدرتمند و محبوب برای حل مشکلات انسانی در هر حوزه‌ای در حال استفاده است. یادگیری عمیق برای صدها مشکل، از بینایی رایانه تا پردازش زبان طبیعی، بکار گرفته شده است. در بسیاری از موارد، کارایی یادگیری عمیق بهتر از کارهای قبلی بوده است. یادگیری عمیق به شدت در دانشگاه‌ها برای مطالعه‌یِ هوش و در صنعت در ساختن سیستم‌های هوشمند برای کمک به انسان در کارهای مختلف استفاده می‌شود. در ادامه این بخش، نگاهی گذار به بخشی از کاربردهای یادگیری عمیق که مطمئناً شما را شگفت زده خواهد کرد، خواهیم داشت. البته باید گفت که برنامه‌های مختلف بسیار زیادی وجود دارد و این لیست به هیچ وجه جامع نیست.

### دستیاران مجازی

محبوب‌ترین برنامه یادگیری عمیق دستیاران مجازی الکسا، سیری و دستیار گوگل است. هر تعامل با این دستیارها فرصتی برای آن‌ها فراهم می‌کند تا در مورد صدا و لهجه شما بیشتر بیاموزند و در نتیجه یک تجربه ثانویه از تعامل انسانی را برای شما به ارمغان می‌آورند. دستیاران مجازی از یادگیری عمیق برای دانستن بیشتر در مورد موضوعات خود استفاده می‌کنند، از ترجیحات غذاخوری شما گرفته تا مکان‌های پربازدید یا موسیقی‌های مورد علاقه شما. آن‌ها یاد می‌گیرند که دستورات شما را با ارزیابی زبان طبیعی انسان برای اجرای آن‌ها درک کنند.

یکی دیگر از قابلیت‌هایی که به دستیاران مجازی اعطا شده است، ترجمه گفتار شما به متن، یادداشت‌برداری برای شما و رزرو قرار ملاقات است. دستیاران مجازی به معنای واقعی کلمه در خدمت شما هستند، چراکه می‌توانند همه‌یِ کارها را از انجام وظایف تا پاسخگویی خودکار به تماس‌های خاص شما گرفته تا هماهنگی وظایف بین شما و اعضای تیمتان انجام دهند.

### تجمیع اخبار و کشف اخبار تقلب

اکنون راهی برای فیلتر کردن همه اخبار بد و زشت از گزیده‌ی اخبار شما وجود دارد. استفاده گسترده از یادگیری عمیق در جمع‌آوری اخبار، تلاش‌ها را برای سفارشی‌سازی اخبار طبق نظر خوانندگان تقویت می‌کند. اگرچه این ممکن است جدید به نظر نرسد، سطوح جدیدتری از

پیچیدگی برای تعریف شخصیت‌های خواننده برای فیلتر کردن اخبار بر اساس پارامترهای جغرافیایی، اجتماعی، اقتصادی همراه با ترجیحات فردی خواننده انجام شده است. از سوی دیگر، کشف اخبار کلاهبرداری، یک دارایی مهم در دنیای امروزی است که در آن اینترنت به منبع اصلی همه اطلاعات واقعی و جعلی تبدیل شده است.

تشخیص اخبار جعلی بسیار سخت می‌باشد، چراکه ربات‌ها آن را به طور خودکار در کانال‌ها تکرار می‌کنند. یادگیری عمیق به توسعه طبقه‌بندهایی کمک می‌کند که می‌توانند اخبار جعلی یا مغرضانه را شناسایی کرده و آن‌ها را از گزیده‌ی اخبار شما حذف کند و در مورد نقض احتمالی حریم خصوصی به شما هشدار دهند.

## هوش عاطفی

در حالی‌که رایانه‌ها ممکن است نتوانند احساسات انسانی را تکرار کنند، به لطف یادگیری عمیق، درک بهتری از حالات ما بدست می‌آورند. الگوهایی مانند تغییر در لحن، اخم‌های خفیف یا هق‌هق همگی سیگنال‌های داده ارزشمندی هستند که می‌توانند به هوش مصنوعی کمک کنند و حالات روحی ما را تشخیص دهند. از چنین برنامه‌هایی می‌توان برای کمک به شرکت‌ها برای ارتباط دادن داده‌های احساسات برای تبلیغات استفاده کرد یا حتی پزشکان را در مورد وضعیت عاطفی بیمار آگاه کرد.

## مراقبت‌های بهداشتی

پزشکان نمی‌توانند ۲۴ ساعت شبانه روز را در کنار بیماران خود باشند، اما تنها چیزی که همه ما تقریبا همیشه همراه خود داریم تلفن‌هایمان است. به لطف یادگیری عمیق، ابزارهای پزشکی می‌توانند داده‌هایِ عکس‌هایی که می‌گیریم و داده‌های حرکتی را برای تشخیص مشکلات سلامتی بالقوه بررسی کنند. نرم‌افزار بینایی رایانه Robbie.AI از این داده‌ها برای ردیابی الگوهای حرکتی بیمار برای پیش‌بینی سقوط و همچنین تغییرات در وضعیت روانی کاربر استفاده می‌کند. یادگیری عمیق همچنین برای تشخیص سرطان پوست از طریق تصاویر ثابت شده است.

یادگیری عمیق همچنین در تحقیقات بالینی برای یافتن راه‌حل‌هایی برای بیماری‌های غیرقابل درمان به‌شدت مورد استفاده قرار می‌گیرد، اما شک و تردید پزشکان و فقدان مجموعه داده‌های گسترده هنوز چالش‌هایی را برای استفاده از یادگیری عمیق در پزشکی ایجاد می‌کند.

## شناسایی کلاهبرداری

از آنجایی که بانک‌ها به دیجیتالی کردن فرآیند تراکنش‌های خود ادامه می‌دهند، احتمال کلاهبرداری‌های دیجیتالی در حال افزایش است، بنابراین برای جلوگیری از این نوع کلاهبرداری،

یادگیری عمیق نقش مهمی ایفا می‌کند. تشخیص کلاهبرداری را می‌توان به سرعت از طریق تکنیک‌های یادگیری عمیق انجام داد.

## خلاصه فصل

- **هوش مصنوعی سیستمی است که می‌تواند رفتار انسان را تقلید کند.**
- **در سیستم‌های رایانه‌ای، تجربه در قالب داده‌ها وجود دارد.**
- **الگوریتم یادگیری عمیق مشابه نحوه یادگیری ما از تجربه‌ها و نمونه‌ها، یک کار را به طور مکرر انجام می‌دهد و هر بار کمی آن را تغییر می‌دهد تا نتیجه را بهبود بخشد.**
- **مدل‌های یادگیری عمیق یکی از دادهِ خوارترین مدل‌های داده در دنیای یادگیری ماشین هستند.**
- **شناخته‌شده‌ترین نقطه ضعف شبکه‌های عصبی ماهیت عدم شفافیت آنهاست.**

## آزمونک

**۱. هوش مصنوعی، یادگیری ماشین و یادگیری عمیق را تعریف کنید.**
**۲. فراموشی فاجعه‌آمیز در شبکه‌های عصبی چه زمانی اتفاق می‌افتد؟**
**۳. مزیت مهم یادگیری عمیق در مقایسه با یادگیری ماشین در چیست؟**
**۴. چند مورد از محدودیت و چالش‌های یادگیری عمیق را شرح دهید.**

# ۲ پیش‌نیازها

## اهداف یادگیری:

- داده چیست؟
- انواع داده‌ها
- آشنایی با رویکردهای متفاوت یادگیری ماشین
- مزایا و معایب هر یک از رویکردها
- انتخاب و ارزیابی مدل چیست؟

# مقدمه

یادگیری عمیق زیرمجموعه‌ای از روش‌های یادگیری ماشین است. از این‌رو، مروری بر مفاهیم یادگیری ماشین قبل از فراگیری یادگیری عمیق مفید خواهد بود. این فصل جزئیات مفاهیمی که برای درک یادگیری عمیق مورد نیاز است را به گونه‌ای ارائه می‌دهد که پیش‌نیازها حداقل شود و اطلاعات کاملی در مورد داده‌ها، ابزارهای مورد نیاز و مبانی یادگیری ماشین ارائه می‌دهد. به‌طور کلی، این فصل به عنوان پایه‌ای برای یادگیری کامل مطالب ارائه شده در این کتاب است.

## داده (Data)

کلمه "**داده**" از کلمه لاتین dare گرفته شده است که به معنای "**چیزی داده شده**" است؛ یک مشاهده یا یک واقعیت در مورد یک موضوع. داده‌ها اشکال و قالب‌های مختلفی دارند، اما به‌طور کلی می‌توان آن را نتیجه آزمایش تصادفی تصور کرد؛ آزمایشی که نتیجه آن را نمی‌توان از قبل تعیین کرد، اما عملکرد آن هنوز در معرض تجزیه و تحلیل است. داده‌های یک آزمایش تصادفی اغلب در یک جدول یا صفحه گسترده ذخیره می‌شود. یک قرارداد آماری این است که متغیرها را که اغلب ویژگی‌ها نامیده می‌شود، به عنوان ستون و موارد منفرد را به عنوان ردیف نشان داده شوند.

**تعریف ۱.۲ داده**

داده ها به قطعات متمایزِ اطلاعاتی اطلاق می شود که معمولاً به گونه ای قالب‌بندی و ذخیره می شوند که با هدف خاصی مطابقت داشته باشد. داده ها می توانند به اشکال مختلف وجود داشته باشند، به صورت اعداد یا متن بر روی کاغذ، به صورت بیت ها یا بایت های ذخیره شده در حافظه الکترونیکی یا به عنوان حقایقی که در ذهن یک فرد وجود دارند. با این حال، در علوم رایانه ، داده ها معمولاً به اطلاعاتی اشاره دارند که به صورت الکترونیکی منتقل یا ذخیره می شوند و به شکلی ترجمه شده‌اند که برای جابجایی یا پردازش کارآمد باشند.

برای آموزش الگوریتم یادگیری عمیق ما به داده‌های زیادی برای ورودی نیاز داریم. هر نقطه داده را می‌توان یک نمونه یا یک مثال نیز نامید. اگر مسئله یادگیری بانظارت داشته باشیم، هر ورودی (بردار) یک خروجی (بردار) مرتبط خواهد داشت. می‌توانید این‌ها را به عنوان متغیرهای مستقل و وابسته‌ی مرتبط در نظر بگیرید.

## داده‌های قابل خواندن توسط ماشین در مقابل قابل خواندن توسط انسان

داده‌ها یک دارایی ضروری برای همه‌یِ سازمان‌ها هستند و اساس بسیاری از توسعه برنامه‌ها را تشکیل می‌دهند و همچنین کاربرد الگوریتم به صحت داده‌ها و صحت منبع بستگی دارد. نمایش داده‌ها در قالب صحیح بدون از بین بردن مقدار اصلی، بخش مهمی از سیستم مدیریت داده است.

همه داده‌ها را می‌توان به عنوان قابل خواندن توسط ماشین، قابل خواندن توسط انسان یا هر دو دسته‌بندی کرد. داده‌های قابل خواندن توسط انسان از قالب‌های زبان طبیعی (مانند یک فایل متنی حاوی کدهای ASCII یا سند PDF) استفاده می‌کنند، در حالی که داده‌های قابل خواندن توسط ماشین از زبان‌های رایانه‌ای با ساختار رسمی برای خواندن توسط سیستم‌ها یا نرم‌افزارهای رایانه‌ای استفاده می‌کنند. برخی از داده‌ها هم توسط ماشین‌ها و هم توسط انسان قابل خواندن هستند، مانند CSV، HTML یا JSON.

قالب قابل خواندن توسط ماشین برای دستگاه‌ها و ماشین‌ها طراحی شده است. درک این قالب برای انسان پیچیده است و همچنین ابزارهای تخصصی برای خواندن محتوای داده‌های قابل خواندن توسط ماشین ضروری است. داده‌های ارائه شده در قالب قابل خواندن توسط ماشین را می‌توان به‌طور خودکار استخراج کرد و برای پردازش و تجزیه و تحلیل بیشتر بدون دخالت انسان استفاده کرد.

داده‌های قابل خواندن توسط انسان می‌تواند توسط انسان درک و تفسیر شود. تفسیر داده‌ها به تجهیزات یا دستگاه‌های تخصصی نیاز ندارد. این زبان دارای یک زبان طبیعی است (به عنوان مثال، فارسی، انگلیسی، فرانسوی و غیره) و نمایش داده‌ها بدون ساختار است. نمونه‌ای از قالب‌های قابل خواندن توسط انسان یک سند PDF است. اگرچه PDF یک رسانه دیجیتال است اما نمایش داده‌های آن نیازی به هیچ تجهیزات تخصصی یا رایانه‌ای برای تفسیر ندارد. علاوه بر این، اطلاعات موجود در سند PDF معمولاً برای انسان‌ها در نظر گرفته شده است، نه ماشین‌ها.

## عبارت داده در فناوری

داده‌ها به جلودار بسیاری از گفتگوهای اصلی در مورد فناوری تبدیل شده‌اند. نوآوری‌های جدید به‌طور دائم بر روی داده‌ها، نحوه استفاده و تجزیه و تحلیل آن‌ها تفسیر می‌شود. در نتیجه، زبان رایج فناوری شامل تعدادی از عبارات جدید و قدیمی شده است:

- **کلان داده (Big data):** حجم عظیمی از داده‌های ساختاریافته و بدون‌ساختار که به سرعت تولید و از منابع مختلفی بدست آمده و سبب افزایش بینش و تصمیم‌گیری می‌شود

- **تجزیه و تحلیل کلان داده (Big data analytics):** فرآیند جمع‌آوری، سازماندهی و ترکیب مجموعه‌های بزرگ داده برای کشف الگوها یا سایر اطلاعات مفید.
- **یکپارچگی داده‌ها (Data integrity):** اعتبار داده‌ها که می‌تواند به طرق مختلفی از جمله خطای انسانی یا خطاهای انتقال به خطر بیفتد.
- **فراداده (Metadata):** اطلاعات خلاصه در مورد یک مجموعه داده.
- **داده‌ها خام (Raw data):** اطلاعاتی که جمع‌آوری شده‌اند، اما قالب‌بندی یا تجزیه و تحلیل نشده‌اند.
- **داده‌های ساختاریافته (Structured data):** هر داده‌ای که در یک فیلد ثابت در یک رکورد یا فایل قرار دارد، از جمله داده‌های موجود در پایگاه‌های داده رابطه‌ای و صفحات گسترده. به عبارت ساده‌تر، داده‌های ساختاریافته شامل انواع داده‌های به وضوح تعریف شده با الگوهایی است که آن‌ها را براحتی قابل جستجو می‌کند.
- **داده‌های بدون ساختار (Unstructured data):** اطلاعاتی که در یک پایگاه داده به صورت ردیف و ستونی سنتی مانند داده‌های ساختاریافته نیستند. به عبارت دیگر، داده‌های بدون‌ساختار شامل داده‌هایی است که معمولا به راحتی قابل جستجو نیست، از جمله فرمت‌هایی مانند صوت، ویدئو و پست‌های رسانه‌های اجتماعی.

## انواع داده‌ها

داده‌ها در قالب‌ها و انواع مختلفی می‌آیند. درک خصوصیات هر ویژگی یا جنبه، اطلاعاتی را در مورد اینکه چه نوع عملیاتی را می‌توان بر روی آن ویژگی انجام داد، ارائه می‌دهد. به عنوان مثال، دما در داده‌های آب و هوا را می‌توان به صورت یکی از فرمت‌های زیر بیان کرد:

1. **درجه سانتی‌گراد عددی (۲۵ درجه سانتی‌گراد)، فارنهایت و یا در مقیاس کلوین**
2. **برچسب‌گذاری بر اساس هوای گرم، سرد و یا ملایم**
3. **تعداد روزهای یک سال زیر صفر درجه سانتی‌گراد (۲۰ روز در سال زیر صفر)**

همه‌یِ این ویژگی‌ها دمای یک منطقه را نشان می‌دهند، اما هر کدام دارای انواع داده‌ای متفاوتی هستند.

### عددی (Numeric) یا پیوسته (Continuous)

دمایی که بر حسب سانتی‌گراد یا فارنهایت بیان می‌شود عددی و پیوسته است، زیرا می‌توان آن را با اعداد نشان داد و بین ارقام بی‌نهایت مقدار را گرفت. عدد صحیح شکل خاصی از نوع داده عددی است که دارای اعشار در مقدار نیست یا به‌طور دقیق‌تر دارای مقادیر بی‌نهایت بین اعداد متوالی نیست. معمولا تعداد چیزی، تعداد روزهای با دمای کم‌تر از ۰ درجه سانتیگراد، تعداد

سفارشات، تعداد فرزندان یک خانواده و غیره را نشان می‌دهند. اگر نقطه صفر تعریف شود، داده‌های عددی به یک نسبت یا نوع داده واقعی تبدیل می‌شوند. به عنوان مثال می‌توان به دما در مقیاس کلوین، موجودی حساب بانکی و درآمد اشاره کرد.

### رسته‌ای (Categorical) یا اسمی (Nominal)

داده‌های رسته‌ای یا اسمی به عنوان داده‌هایی تعریف می‌شوند که برای نام‌گذاری یا برچسب‌گذاری متغیرها، بدون هیچ مقدار کمی استفاده می‌شوند. معمولا هیچ ترتیب ذاتی برای داده‌های اسمی وجود ندارد. به‌عنوان مثال، رنگ یک تلفن هوشمند را می‌توان به‌عنوان نوع داده اسمی در نظر گرفت. زیرا نمی‌توانیم یک رنگ را با رنگ‌های دیگر مقایسه کنیم. به عبارت دیگر، نمی‌توان بیان کرد که "قرمز" بزرگ‌تر از "آبی" است. به‌عنوان مثالی دیگر، رنگ چشم یک متغیر اسمی است که دارای چند دسته (آبی، سبز، قهوه‌ای) است و راهی برای مرتب‌سازی این دسته‌ها از بالاترین به کمترین وجود ندارد.

## مجموعه داده

یک مجموعه داده، مجموعه‌ای از داده‌ها است که معمولاً به صورت جدول ارائه می‌شود. هر ستون نشان‌دهنده یک متغیر خاص (ویژگی) است. هر ردیف مربوط به یک عضو معین از مجموعه داده مورد نظر است و مقادیر را برای هر یک از متغیرها می‌کند. هر مقدار به عنوان یک داده شناخته می‌شود.

**تعریف ۱.۲** مجموعه داده

مجموعه داده‌ها را اغلب می‌توان به عنوان مجموعه‌ای از اشیاء داده با ویژگی های یکسان در نظر گرفت. نام‌های دیگر برای یک شی داده عبارتند از: رکورد، نقطه، بردار، الگو، رویداد، مورد، نمونه، مثال، مشاهده یا موجودیت.

**تعریف ۱.۲** ویژگی

یک ویژگی، مشخصه داده است. ویژگی را می توان به عنوان یک متغیر توضیحی در نظر گرفت. این ویژگی ممکن است عددی باشد (ارتفاع درخت ۳ متر) یا ممکن است توصیفی باشد (رنگ چشم آبی). اغلب اگر توصیفی باشد، برای انجام دست‌ورزی های ریاضی باید به آن یک برچسب عددی بدهید.

هر نقطه داده اغلب با یک بردار ویژگی نشان داده می‌شود، هر ورودی در بردار نشان‌دهنده یک ویژگی است.

# رویکردهای یادگیری ماشین

در زمینه یادگیری ماشین، دو نوع رویکرد اصلی وجود دارد: بانظارت و بدون‌نظارت. تفاوت اصلی بین این دو نوع رویکرد این است که یادگیری بانظارت با استفاده از یک حقیقت یادگیری را بدست می‌آورد یا به عبارت دیگر، ما از مقدار خروجی نمونه‌های خود، آگاهی قبلی داریم. بنابراین، هدف از یادگیری بانظارت، یادگیری تابعی است که با در نظر گرفتن نمونه‌ای از داده‌ها و خروجی‌های مورد نظر، به بهترین وجه رابطه بین ورودی و خروجی قابل مشاهده در داده‌ها را تخمین می‌زند. یادگیری بانظارت معمولاً در زمینه طبقه‌بندی انجام می‌شود، زمانی که می‌خواهیم ورودی را به برچسب‌های خروجی نگاشت کنیم، یا رگرسیون، زمانی که می‌خواهیم ورودی را به یک خروجی پیوسته نگاشت کنیم.

از سوی دیگر، در یادگیری بدون‌نظارت، خروجی‌های برچسب‌گذاری شده‌ای وجود ندارد، بنابراین هدف آن استنباط ساختار طبیعی موجود در مجموعه‌ای از نقاط داده و کشف الگوها بدون هیچ راهنمایی است. متداول‌ترین وظایف در یادگیری بدون‌نظارت عبارتند از خوشه‌بندی و کاهش ابعاد. در این موارد، ما می‌خواهیم ساختار ذاتی داده‌های خود را بدون استفاده از برچسب‌های ارائه شده یاد بگیریم.

در یک مدل یادگیری بانظارت، الگوریتم روی یک مجموعه داده برچسب‌گذاری شده یاد می‌گیرد و دارای یک کلید پاسخ است که الگوریتم می‌تواند از آن برای ارزیابی دقت خود در داده‌های آموزشی استفاده کند. در مقابل، یک مدل بدون‌نظارت، از داده‌های بدون برچسب استفاده می‌کند و الگوریتم سعی می‌کند با استخراج ویژگی‌ها و الگوها به تنهایی آن‌ها به ساختار ذاتی داده‌ها پی ببرد.

## یادگیری بانظارت

یادگیری بانظارت یکی از پرکاربردترین شاخه‌های یادگیری ماشین است که از داده‌های آموزشی برچسب‌گذاری شده برای کمک به مدل‌ها در پیش‌بینی دقیق استفاده می‌کند. داده‌های آموزشی در اینجا به‌عنوان سرپرست و معلم برای ماشین‌ها عمل می‌کنند، از این رو به این نام اشاره می‌شود. یادگیری بانظارت مبتنی‌بر تولید خروجی از تجربیات گذشته (داده‌های برچسب دار) است. در یادگیری بانظارت، یک متغیر ورودی ($x$) با کمک یک تابع نگاشت که توسط یک مدل یادگیری ماشین آموخته می‌شود، به متغیر خروجی ($y$) نگاشت می‌شود.

$$y = f(x)$$

در اینجا مدل تابعی را ایجاد می‌کند که دو متغیر را با هدف نهایی بهم متصل می‌کند تا برچسب صحیح داده‌های ورودی را پیش‌بینی کند.

یک الگوریتم یادگیری بانظارت همیشه دارای یک متغیر هدف یا نتیجه (یا متغیر وابسته) است که از مجموعه‌ای از پیش‌بینی کننده‌های ارائه شده (متغیرهای مستقل) شناسایی می‌شود. الگوریتم از این مجموعه متغیرها برای ایجاد تابعی استفاده می‌کند که ورودی‌ها را به خروجی‌های دلخواه نگاشت می‌کند. این فرآیند آموزشی تا زمانی که تکرار می‌شود که مدل به سطح بالایی از دقت دست یابد.

مجموعه داده برچسب‌گذاری شده به این معنی است که هر نمونه در مجموعه داده آموزشی با پاسخی که الگوریتم باید به خودی خود ارائه کند برچسب‌گذاری می‌شود. بنابراین، مجموعه داده برچسب‌گذاری شده از تصاویر گل به مدل می‌گوید که کدام عکس‌ها مربوط به گل رز، گل بابونه و نرگس است. هنگامی که یک تصویر جدید به مدل نشان داده می‌شود، مدل آن را با نمونه‌های آموزشی مقایسه می‌کند تا برچسب صحیح را پیش‌بینی کند.

## طبقه‌بندی (Classification)

طبقه‌بندی یک فرآیند بانظارت است، یعنی الگوریتم یادگیرنده بر اساس داده‌های آموزشی که از قبل برچسب خورده‌اند سعی در یافتن یک ارتباط بین داده‌ها و برچسب‌ها دارد.

در طبقه‌بندی کلاس‌ها از قبل مشخص هستند و اغلب با عنوان هدف، برچسب یا دسته نامیده می‌شوند.

داده‌های دارای برچسب برای آموزش دسته‌بند استفاده می‌شود تا بتواند بر روی داده‌های ورودی جدید به‌خوبی عمل کند و بتواند کلاس درست آن نمونه را پیش‌بینی کند. به عبارت دیگر، هدف این است که یک تقریب خوب برای $f(x)$ پیدا شود تا بتواند برای داده‌های دیده‌نشده در فرآیند آموزش پیش‌بینی انجام دهد و بگوید که نمونه جدید به کدام یکی از کلاس‌ها تعلق دارد.

مسائل طبقه‌بندی را می‌توان از دو دیدگاه متفاوت تقسیم‌بندی کرد. از دیدگاه تعداد برچسب که به دو دسته **طبقه‌بندی تک‌برچسبی** (**single-label classification**) و **طبقه‌بندی چندبرچسبی** (**multi-label classification**) قابل تقسیم هستند و از دیدگاه تعداد کلاس‌ها به دو دسته **طبقه‌بندی دودویی** (**Binary Classification**) و **طبقه‌بندی چندکلاسی** (**Multi-Class Classification**) تقسیم‌بندی می‌شوند.

طبقه‌بندی دودویی که در آن هر نمونه تنها به یکی از دو کلاس از پیش تعریف‌شده اختصاص داده می‌شود، ساده‌ترین نوع طبقه‌بندی است. طبقه‌بندی دودویی با تعریف کلاس‌های بیشتر به طبقه‌بندی چندکلاسی گسترش می‌یابد.

### طبقه‌بندی تک برچسبی

در طبقه‌بندی داده‌های تک‌برچسبی، هر نمونه تنها می‌تواند با یک برچسب ارتباط داشته باشد و الگوریتم طبقه‌بندی در مرحله آموزش برای هر نمونه جدید تنها یک برچسب را پیش‌بینی می‌کند. به طور کلی، مسائل طبقه‌بندی تک‌برچسبی را می‌توان به دو گروه اصلی تقسیم کرد: مسائل دودویی و چندکلاسی.

### طبقه‌بندی دودویی

مساله دسته‌بندی دودویی ساده‌ترین حالت از مسائل طبقه‌بندی است که در آن مجموعه کلاس‌ها تنها به دو مورد محدود می‌شود. در این زمینه، ما بین کلاس مثبت و کلاس منفی تمایز قائل می‌شویم. یک مثال ساده از مسائل دسته‌بندی دودویی زمانی است که یک زن به پزشک مراجعه می‌کند تا از باردار بودن خود مطلع شود. نتیجه آزمایش ممکن است مثبت یا منفی باشد.

**به‌طور اساسی طبقه‌بندی دودویی نوعی پیش‌بینی است که به این موضوع می‌پردازد که یک نمونه به کدام یک از دو گروه کلاس تعلق دارد.**

### طبقه‌بندی چندکلاسی

طبقه‌بندی چندکلاسی یا چندگانه، طبقه‌بندی عناصر به کلاس‌های مختلف است. برخلاف طبقه‌بندی دودویی که محدود به تنها دو کلاس است، محدودیتی در تعداد کلاس‌ها ندارد و می‌تواند طبقه‌بندی بیش‌از دو کلاس را انجام دهد. به عنوان مثال، طبقه‌بندی اخبار در دسته‌های مختلف، طبقه‌بندی کتاب‌ها براساس موضوع و طبقه‌بندی حیوانات مختلف در یک تصویر نمونه‌های از طبقه‌بندی چندکلاسی هستند.

### طبقه‌بندی چندبرچسبی

در مسائل طبقه‌بندی سنتی، هر نمونه با یک برچسب کلاس مرتبط است. با این حال، در بسیاری از سناریوهای دنیای واقعی، یک نمونه ممکن است با چندین برچسب مرتبط باشد. به عنوان مثال، در طبقه‌بندی اخبار، بخشی از اخبار مربوط به عرضه آیفون جدید توسط اپل، هم با برچسب تجارت و هم با برچسب فناوری مرتبط است. به عبارت دیگر، هر نمونه به جای تنها یک برچسب، با مجموعه‌ای از برچسب‌ها مرتبط است. یادگیری چندبرچسبی یک زمینه یادگیری ماشین است که به یادگیری از داده‌های چندبرچسبی اشاره دارد که در آن هر نمونه با چندین برچسب بالقوه مرتبط است.

طبقه‌بندی چندبرچسبی یکی از مسائل مهم در زمینه پردازش زبان طبیعی، به خصوص در طبقه‌بندی متون است. علاوه بر این، از آن در بسیاری مسائل دنیای واقعی همانند، بازیابی

اطلاعات، تشخیص بیماری و بیوانفورماتیک استفاده می‌شود. تفاوت طبقه‌بندی چندبرچسبی با طبقه‌بندی چندکلاسی، در تعداد برچسب‌هایی است که می‌تواند به یک نمونه اختصاص یابد.

طبقه‌بندی چندبرچسبی حالت تعمیم یافته‌ای از طبقه‌بندی تک‌برچسبی است، چرا که در آن هر نمونه می‌تواند به جای یک برچسب با مجموعه‌ای از برچسب‌ها در ارتباط باشد.

## رگرسیون

تفاوت اصلی بین مدل‌های رگرسیون و طبقه‌بندی این است که الگوریتم‌های رگرسیون برای پیش‌بینی مقادیر پیوسته (نمرات آزمون) استفاده می‌شوند، در حالی که الگوریتم‌های طبقه‌بندی مقادیر گسسته (مذکر/مونث، درست/نادرست) را پیش‌بینی می‌کنند. *رگرسیون یک فرآیند آماری است که رابطه معناداری بین متغیرهای وابسته و مستقل پیدا می‌کند.* به عنوان یک الگوریتم، یک عدد پیوسته را پیش‌بینی می‌کند. به عنوان مثال، ممکن است از یک الگوریتم رگرسیون برای تعیین نمره‌ی آزمون دانش‌آموزان بسته به تعداد ساعات مطالعه آن‌ها در هفته استفاده کنید. در این شرایط ساعات مطالعه شده به متغیر مستقل تبدیل می‌شود و نمره نهایی آزمون دانشجو متغیر وابسته است. می‌توانید خطی از بهترین تناسب را از طریق نقاط داده مختلف ترسیم کنید تا پیش‌بینی‌های مدل را هنگام معرفی ورودی جدید نشان دهید. از همین خط می‌توان برای پیش‌بینی نمرات آزمون بر اساس عملکرد دانش‌آموز دیگر نیز استفاده کرد.

به عنوان یک مثال دیگر، فرض کنید می‌خواهیم سیستمی داشته باشیم که بتواند قیمت یک خودروی دست دوم را پیش‌بینی کند. ورودی‌ها، ویژگی‌های خودرو همانند برند، سال، مسافت پیموده شده و اطلاعات دیگری که به اعتقاد ما بر ارزش خودرو تاثیر می‌گذارد و خروجی قیمت خودرو است. یا ناوبری یک ربات متحرک (اتوموبیل خودران) را در نظر بگیرید؛ خروجی زاویه‌ای است که در هر بار فرمان باید بچرخد تا بدون برخورد به موانع و انحراف از مسیر پیشروی کند و ورودی‌ها توسط حسگرهای برروی اتوموبیل همانند دوربین فیلم‌برداری، GPS و غیره ارائه می‌شوند.

## مزایا و معایب یادگیری بانظارت

### مزایا

- یادگیری بانظارت فرآیند ساده ای است که می توانید آن را درک کنید.
- پس از اتمام فرآیند آموزش، لزومی نیست که داده های آموزشی را در حافظه خود نگه دارید.
- نتیجه آن در مقابل روش یادگیری بدون نظارت از دقت بیشتری برخوردار است.
- از آن جایی داده های برچسب‌دار وجود دارند، می توانید براحتی مدل خود را تست و اشکال زدایی کنید.

**معایب**

- یادگیری با نظارت از جنبه‌های مختلف محدود است به‌طوری که نمی‌تواند برخی از وظایف پیچیده در یادگیری ماشین را انجام دهد.
- اگر ورودی بدهیم که از هیچ یک از کلاس‌های داده آموزشی نباشد، ممکن است خروجی، یک برچسب کلاس اشتباه باشد. به‌عنوان مثال، فرض کنید یک طبقه‌بند تصویر را با داده‌های گربه و سگ آموزش داده‌اید. سپس اگر تصویر زرافه را بدهید، خروجی ممکن است گربه یا سگ باشد، که درست نیست.
- برای آموزش به زمان محاسباتی زیادی نیاز دارد.
- جمع‌آوری و برچسب‌گذاری داده‌ها پرهزینه و زمان‌بر است.

## یادگیری بدون‌نظارت

یادگیری بدون‌نظارت در یادگیری ماشین زمانی است که به هیچ وجه دسته‌بندی یا برچسب‌گذاری داده‌ها وجود ندارد. وظیفه این است که اطلاعات گروه‌بندی نشده را بر اساس برخی شباهت‌ها و تفاوت‌ها بدون هیچ گونه راهنمایی، مرتب کند. به عبارت دیگر، از ماشین انتظار می‌رود الگوها و ساختار پنهان در داده‌های بدون برچسب را به تنهایی پیدا کند. به همین دلیل است که به آن بدون‌نظارت می‌گویند. چراکه هیچ راهنمایی وجود ندارد که به ماشین یاد دهد چه چیزی درست و چه چیزی نادرست است. در این رویکرد، ماشین این را نمی‌داند به دنبال چه چیزی است، اما می‌تواند به‌طور مستقل داده‌ها را مرتب کند و الگوهای قانع کننده‌ای پیدا کند.

یادگیری بدون نظارت، اطلاعات مرتب نشده را بر اساس شباهت‌ها و تفاوت‌ها گروه‌بندی می‌کند، حتی اگر هیچ دسته‌ای ارائه نشده باشد.

یکی از ویژگی‌های مهم این مدل‌ها این است، در حالی‌که مدل می‌تواند روش‌های مختلفی را برای گروه‌بندی یا سفارش داده‌های شما پیشنهاد دهد، این به شما بستگی دارد که تحقیقات بیشتری برروی این مدل‌ها انجام دهید تا از چیز مفیدی رونمایی کنید.

یادگیری بدون‌نظارت در تحلیل اکتشافی بسیار مفید است، چراکه می‌تواند به‌طور خودکار ساختار ذاتی در داده‌ها را شناسایی کند. به عنوان مثال، اگر یک تحلیلگر سعی می‌کند مصرف‌کنندگان را بخش‌بندی کند، روش‌های خوشه‌بندی بدون‌نظارت نقطه شروع خوبی برای تحلیل آن‌ها خواهد بود.

## خوشه‌بندی

خوشه‌بندی فرآیند تخصیص نمونه داده‌ها به تعداد معینی از خوشه‌ها است (شکل ۲ـ۱) به گونه‌ای که نقاط داده متعلق به یک خوشه دارای ویژگی‌های مشابه باشند. به عبارت ساده‌تر، خوشه‌ها چیزی نیستند جز گروه‌بندی نقاط داده به گونه‌ای که فاصله بین نقاط داده درون خوشه‌ها حداقل باشد. هدف تحلیل خوشه‌ای (در حالت ایده‌آل) یافتن خوشه‌هایی است که نمونه‌های درون هر خوشه کاملا شبیه یکدیگر باشند، در حالی که هر خوشه‌ای با یکدیگر کاملا متفاوت باشد.

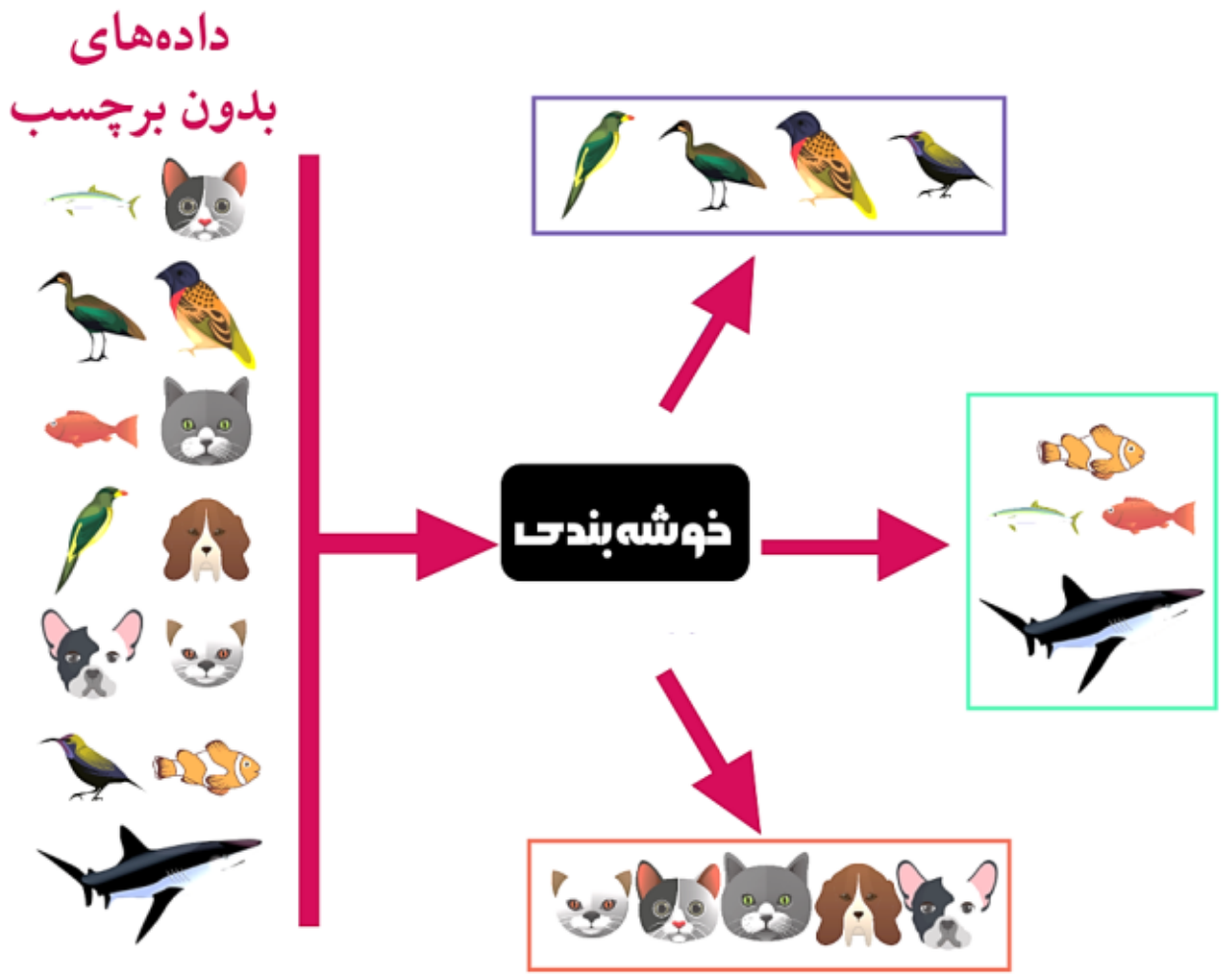

شکل ۲ـ۱. خوشه‌بندی.

از آن‌جایی که خوشه‌بندی توسط الگوریتم انجام می‌شود، این احتمال وجود دارد که بتوانید همبستگی‌های ناشناخته قبلی را در داده‌ها کشف کنید که می‌تواند به شما در برخورد با یک چالش تجاری از دیدگاه جدید کمک کند.

## کاهش ابعاد

کاهش ابعاد به فرآیند کاهش تعداد ویژگی‌ها در مجموعه داده‌ها اشاره دارد، در حالی که تا آنجا که ممکن است تغییرات در مجموعه داده اصلی حفظ شود. فرآیند کاهش ابعاد اساسا داده‌ها را از فضای ویژگی‌های با ابعاد بالا به فضای ویژگی‌های با بعد کم‌تر تبدیل می‌کند. به طور همزمان، مهم است که ویژگی‌های معنی دار موجود در داده‌ها در طول تبدیل از بین نروند.

کاهش ابعاد یک مرحله پیش‌پردازش داده است. به این معنا که قبل از آموزش مدل، کاهش ابعاد را انجام می‌دهیم.

## مقایسه یادگیری بانظارت با یادگیری بدون‌نظارت

فرض کنید می‌خواهیم به یک کودک یک زبان جدید، به عنوان مثال انگلیسی، آموزش دهیم. اگر این کار را طبق اصل یادگیری بانظارت انجام دهیم، به سادگی یک فرهنگ لغت با کلمات انگلیسی و ترجمه به زبان مادری‌اش، به عنوان مثال فارسی، به او می‌دهیم. شروع یادگیری برای کودک نسبتاً آسان خواهد بود و احتمالا با به‌خاطر سپردن ترجمه‌ها می‌تواند خیلی سریع پیشرفت کند. با این حال، او در خواندن و درک متون به زبان انگلیسی مشکل خواهد داشت، چراکه او فقط ترجمه‌های کلمات فارسی‌ـ‌انگلیسی و نه ساختار دستوری جملات انگلیسی را یاد گرفته است.

طبق اصل یادگیری بدون‌نظارت، سناریو کاملاً متفاوت به نظر می‌رسد. به‌عنوان مثال پنج کتاب انگلیسی را به کودک تقدیم می‌کردیم و یادگیری را به خودش بدست آورد. البته این کار بسیار پیچیده‌تر است!! به‌عنوان مثال، با کمک "داده‌ها"، کودک می‌تواند تشخیص دهد که کلمه "من" به طور نسبتاً مکرر در متون وجود دارد و در بسیاری از موارد در ابتدای یک جمله نیز آمده است و از آن نتیجه‌گیری کند.

این مثال تفاوت بین یادگیری بانظارت و بدون‌نظارت را نشان می‌دهد. یادگیری بانظارت در بسیاری از موارد الگوریتم ساده‌تری است. با این حال، مدل فقط زمینه‌هایی را می‌آموزد که به صراحت در مجموعه داده‌های آموزشی وجود دارند و به عنوان ورودی به مدل داده شده‌اند. برای مثال، کودکی که انگلیسی را یاد می‌گیرد، می تواند کلمات فارسی را به خوبی به انگلیسی ترجمه کند، اما خواندن و درک متون انگلیسی را یاد نگرفته است.

از سوی دیگر، یادگیری بدون‌نظارت، با کار بسیار پیچیده‌تری مواجه است، زیرا باید ساختارها را به‌طور مستقل شناسایی و یاد بگیرد. در نتیجه زمان و تلاش تمرین نیز بیشتر است. با این حال، مزیت این است که مدل آموزش‌دیده، زمینه‌هایی را نیز تشخیص می‌دهد که به صراحت آن‌ها را فرانگرفته است. کودکی که با کمک پنج رمان انگلیسی، زبان انگلیسی را به خود آموزش داده است، احتمالاً می‌تواند متون انگلیسی را بخواند، تک تک کلمات را به فارسی ترجمه کند و همچنین گرامر انگلیسی را درک کند.

## چرا یادگیری بدون‌نظارت؟

یادگیری بانظارت در بهینه‌سازی عملکردِ وظایفی با مجموعه داده‌هایی با برچسب‌های فراوان، کارآیی بسیار خوبی از خود نشان می‌دهد. به عنوان مثال، مجموعه داده‌یِ بسیار بزرگی از تصاویری از اشیا را در نظر بگیرید که هر تصویر برچسب‌گذاری شده است. اگر مجموعه داده به اندازه کافی بزرگ باشد، اگر آن را به اندازه کافی با استفاده از الگوریتم‌های یادگیری ماشین

مناسب و با رایانه‌ای قدرتمند آموزش دهیم، می‌توانیم یک مدل طبقه‌بند تصویرِ مبتنی‌بر یادگیری بانظارتِ بسیار خوب بسازیم.

همان‌طور که الگوریتم بانظارت بر روی داده‌ها آموزش می‌بیند، می‌تواند عملکرد خود را (از طریق تابع هزینه) با مقایسه برچسب تصویر پیش‌بینی‌شده خود با برچسب تصویر واقعی که در مجموعه داده داریم، اندازه‌گیری کند. الگوریتم، به صورت صریح سعی می‌کند این تابع هزینه را به حداقل برساند؛ به‌طوری که خطای آن در تصاویری که قبلا دیده نشده است (مجموعه آزمون) تا حد امکان کم باشد. به همین دلیل است که برچسب‌ها بسیار قدرتمند هستند. آن‌ها با ارائه یک معیار خطا به هدایت الگوریتم کمک می‌کنند. الگوریتم از معیار خطا برای بهبود عملکرد خود در طول زمان استفاده می‌کند. بدون چنین برچسب‌هایی، الگوریتم نمی‌داند که چقدر در طبقه‌بندی درست تصاویر موفق است یا نه. با این حال، گاهی اوقات هزینه‌ی برچسب‌گذاری دستی یک مجموعه داده بسیار بالا است.

علاوه بر این، به همان اندازه‌ای که مدل‌های یادگیری بانظارت قدرتمند هستند، در تعمیم‌دهیِ دانشِ فراتر از موارد برچسب‌گذاری شده‌ای که روی آن‌ها آموزش دیده‌اند نیز، محدود هستند. از آن‌جایی که اکثر داده‌های جهان بدون‌برچسب هستند، با استفاده از یادگیری بانظارت، توانایی هوش مصنوعی برای گسترش عملکرد خود به نمونه‌هایی که قبلا دیده نشده‌اند، محدود است. به عبارت دیگر، یادگیری بانظارت در حل مسائل **هوش مصنوعی محدود** (**Narrow AI**) عالی است، اما در حل مسائل از نوع هوش مصنوعی قوی، چندان خوب نیست.

در مقابل، برای مسائلی که الگوها ناشناخته هستند یا به‌طور دائم در حال تغییر هستند یا مجموعه داده‌های برچسب‌گذاری‌شده کافی برای آن‌ها نداریم، یادگیری بدون‌نظارت واقعا می‌درخشد. یادگیری غیرنظارتی، به جای هدایت شدن توسط برچسب‌ها، با یادگیریِ ساختارِ زیربنایی داده‌هایی که روی آنها آموزش دیده است، کار می‌کند. یادگیری بدون‌نظارت این کار را با تلاش برای بازنمایی از داده‌هایی که روی آن آموزش می‌بیند با مجموعه‌ای از پارامترها انجام می‌دهد. با انجام این **یادگیری بازنمایی** (**representation learning**)، یادگیری بدون‌نظارت می‌تواند الگوهای متمایزی را در مجموعه داده شناسایی کند.

**تعریف ۱.۲ یادگیری بازنمایی**

یادگیری بازنمایی زیرمجموعه‌ای از یادگیری ماشین است که هدف آن بدست آوردن ویژگی های خوب و مفید از داده ها به‌طور خودکار، بدون آن که یک مهندس ویژگی درگیر مساله باشد. در این رویکرد، ماشین داده های خام را به عنوان ورودی می گیرد و به‌طور خودکار بازنمایی های مورد نیاز برای شناسایی ویژگی را کشف می کند و سپس به‌طور خودکار ویژگی های جدید را یاد می گیرد و آن را اعمال می کند. به عبارت دیگر، هدف یادگیری بازنمایی یافتن تبدیلی است که داده های خام را به بازنمایی که برای یک وظیفه یادگیری ماشین مناسب‌تر است (به عنوان مثال طبقه‌بندی) نگاشت می کند. از آن جا که این روش می تواند به عنوان یادگیری ویژگی های معنادار تفسیر شود، به آن یادگیری ویژگی نیز گفته می شود.

در مثال مجموعه داده تصویر (این بار بدون‌برچسب)، یادگیری بدون‌نظارت ممکن است بتواند تصاویر را بر اساس شباهت آن‌ها به یکدیگر و تفاوت آن‌ها با بقیه شناسایی و گروه‌بندی کند. به عنوان مثال، تمام تصاویری که شبیه صندلی هستند باهم و همه تصاویری که شبیه به گربه هستند با هم گروه‌بندی می‌شوند. البته، خود یادگیری بدون‌نظارت نمی‌تواند این گروه‌ها را به عنوان "صندلی" یا "گربه" برچسب‌گذاری کند. با این حال، اکنون که تصاویر مشابه با هم گروه‌بندی شده‌اند، انسان وظیفه برچسب‌گذاری بسیار ساده‌تری دارد. به‌جای برچسب‌گذاری میلیون‌ها تصویر با دست، انسان‌ها می‌توانند به صورت دستی همه‌یِ گروه‌های مجزا را برچسب‌گذاری کنند و این برچسب‌ها برای همه اعضای هر گروه اعمال شوند .

از این‌رو، یادگیری بدون‌نظارت، مسائل حل نشدنی قبلی را قابل حل‌تر می‌کند و در یافتنِ الگوهایِ پنهان، هم در داده‌های گذشته‌یِ در دسترسِ برای آموزش و هم در داده‌هایِ آینده، بسیار چابک‌تر عمل می‌کند. حتی اگر یادگیری بدون‌نظارت در حل مسائل خاص (مسائل محدود هوش مصنوعی) مهارت کم‌تری نسبت به یادگیری بانظارت دارد، اما در مقابله با مشکلاتِ بازتر از نوع هوش مصنوعی قوی و تعمیم این دانش بهتر است. مهم‌تر از آن، یادگیری بدون‌نظارت می‌تواند بسیاری از مشکلات رایجی را که دانشمندان داده هنگام ساخت راه حل‌های یادگیری ماشین با آن مواجه می‌شوند، برطرف کند.

## مزایا و معایب یادگیری بدون‌نظارت

**مزایا**

- می‌تواند آنچه که ذهن انسان نمی‌تواند تصور کند را ببیند.
- بدست آوردن داده‌های بدون برچسب نسبتاً ساده‌تر است.

**معایب**

- هزینه بیشتری دارد زیرا ممکن است نیاز به مداخله انسانی برای درک الگوها و ارتباط آنها با دانش حوزه داشته باشد.
- سودمندی و مفید بودن نتایج همیشه قابل تایید نیست، زیرا هیچ برچسب یا معیار خروجی برای تایید مفید بودن آن وجود ندارد.
- نتایج اغلب دقت کم‌تری دارند.

## یادگیری تقویتی

یادگیری تقویتی ریشه در روانشناسی یادگیری حیوانات دارد و در مورد یادگیری رفتار بهینه در یک محیط برای بدست آوردن حداکثر پاداش است. این رفتار بهینه از طریق تعامل با محیط و

مشاهدات نحوه واکنش آن آموخته می‌شود. به یادگیرنده (عامل) برای اقدامات صحیح جایزه داده می‌شود و برای اعمال اشتباه تنبیه در نظر گرفته می‌شود.

در غیاب ناظر، یادگیرنده به دنبال یک خط‌مشی موثر برای حل یک وظیفه‌یِ تصمیم‌گیری است. چنین خط‌مشی دیکته می‌کند که چگونه عامل باید در هر حالتی که ممکن است با آن مواجه شود رفتار کند تا با آزمون و خطا در تعامل با یک محیط پویا، کل پاداش مورد انتظار را بیشینه کند (یا تنبیه را کمینه کند). از آنجایی که می‌تواند اقداماتی را که منجر به موفقیت نهایی در یک محیط دیده‌نشده می‌شود بدون کمک ناظر بیاموزد، یادگیری تقویتی یک الگوریتم بسیار قدرتمند است.

# انتخاب و ارزیابی مدل

انتخاب مدل در زمینه یادگیری ماشین می‌تواند معانی متفاوتی داشته باشد. ممکن است علاقه‌مند به انتخاب بهترین ابرپارامترها برای یک روش یادگیری ماشین باشیم. ابرپارامترها، پارامترهای روش یادگیری هستند که باید آن‌ها را به صورت **پیشینی** مشخص کنیم، **یعنی قبل از برازش مدل**. در مقابل، پارامترهای مدل، پارامترهایی هستند که در نتیجه برازش ایجاد می‌شوند. به عنوان مثال، در یک مدل شبکه عصبی، تعداد نورون‌های لایه پنهان و تعداد لایه‌های پنهان یک ابرپارامتر است که باید قبل از برازش مشخص شود، در حالی که وزن‌های مدل پارامترهای مدل هستند. یافتن ابرپارامترهای مناسب برای یک مدل می‌تواند برای کارایی مدل بسیار مهم است.

در مواقع دیگر، ممکن است بخواهیم بهترین روش یادگیری را از مجموعه روش‌های یادگیری ماشین (الگوریتم‌ها) واجد شرایط انتخاب کنیم.

**تعریف ۲.۲** **انتخاب مدل (Model selection)**

**انتخاب مدل تکنیکی برای انتخاب بهترین مدل پس از ارزیابی تک تک مدل ها براساس معیار های مورد نیاز است.**

قبل از پرداختن رویکردها برای انتخاب مدل یک چیز دیگر وجود دارد که باید در مورد آن بحث کنیم: **ارزیابی مدل (model evaluation)**. هدف ارزیابی مدل تخمین خطای تعمیم مدل انتخاب شده است، یعنی اینکه مدل انتخاب‌شده تا چه اندازه روی داده‌های دیده نشده عمل می‌کند. بدیهی است که یک مدل یادگیری ماشین خوب، مدلی است که نه تنها روی داده‌هایی که در طول آموزش دیده شده عملکرد خوبی دارد (در غیر این صورت، یک مدل یادگیری ماشین می‌تواند به سادگی داده‌های آموزشی را به خاطر بسپارد)، بلکه باید روی داده‌های دیده نشده نیز عملکرد خوبی داشته باشد. بنابراین، قبل از استقرار یک مدل به تولید، باید نسبتاً مطمئن باشیم که عملکرد مدل در مواجهه با داده‌های جدید کاهش نمی‌یابد.

اما چرا ما به تمایز بین **انتخاب مدل** و **ارزیابی مدل** نیاز داریم؟ دلیل آن بیش‌برازش است. اگر خطای تعمیم مدل انتخابی خود را بر روی همان داده‌هایی که برای انتخاب مدل برنده استفاده کرده‌ایم (با فرض اینکه انتخاب مدل براساس مجموعه آموزشی صورت گرفته باشد) تخمین بزنیم، یک تخمین خوش‌بینانه بدست خواهیم آورد. چرا؟ پاسخ ساده است**!!** مدل یادگیری ماشین توانسته است به سادگی داده‌های آموزشی را به خاطر بسپارد. از این‌رو ارزیابی کارایی و جلوگیری از چنین مسائلی، ما به داده‌های کاملا مستقل برای تخمین خطای تعمیم یک مدل نیاز داریم.

استراتژی پیشنهادی برای انتخاب مدل به مقدار داده‌های موجود بستگی دارد. اگر داده‌های زیادی در دسترس باشد، ممکن است داده‌ها را به چند بخش تقسیم کنیم که هر کدام هدف خاصی را دنبال می‌کنند. به عنوان مثال، برای تنظیم ابرپارامتر، ممکن است داده‌ها را به سه مجموعه تقسیم کنیم: **آموزش / اعتبارسنجی / آزمایش**. مجموعه آموزشی برای آموزش مدل‌های مختلف به تعداد ترکیب‌های مختلف ابرپارامترهای مدل استفاده می‌شود. سپس این مدل‌ها بر روی مجموعه اعتبارسنجی ارزیابی می‌شوند و مدلی که بهترین عملکرد را در این مجموعه اعتبارسنجی داشته باشد به عنوان مدل برنده انتخاب می‌شود. سپس، مدل بر روی داده‌های آموزشی به همراه داده‌های اعتبارسنجی با استفاده از مجموعه ابرپارامترهای انتخابی بازآموزی می‌شود و عملکرد تعمیم با استفاده از مجموعه آزمون برآورد می‌شود. اگر این خطای تعمیم مشابه خطای اعتبارسنجی باشد، بر این باور هستیم که مدل روی داده‌های دیده نشده عملکرد خوبی خواهد داشت.

آنچه ما به‌طور ضمنی در طول بحث بالا فرض کرده‌ایم این است که داده‌های آموزش، اعتبارسنجی و آزمون از یک توزیع نمونه‌برداری شده‌اند. اگر این‌طور نباشد، همه تخمین‌ها کاملاً اشتباه خواهند بود. به همین دلیل **ضروری است که قبل از ساخت مدل اطمینان حاصل شود که توزیع داده‌ها تحت تأثیر تقسیم‌بندی داده‌های شما قرار نمی‌گیرد.**

اما چه می‌شود اگر داده‌های کم، تنها چیزی است که ما داریم؟ چگونه انتخاب و ارزیابی مدل را در این مورد انجام دهیم؟ پاسخ این است ارزیابی مدل تغییر نمی‌کند چراکه ما هنوز به یک مجموعه آزمایشی نیاز داریم که بر اساس آن بتوانیم خطای تعمیم مدل نهایی انتخاب‌شده را تخمین بزنیم. از این‌رو، داده‌ها را به دو مجموعه، یک مجموعه آموزشی و یک مجموعه آزمایشی تقسیم می‌کنیم. آنچه در مقایسه با روش قبلی تغییر می‌کند، نحوه استفاده ما از مجموعه آموزشی است. یکی از این تکنیک‌ها اعتبارسنجی متقابل است که در ادامه بخش به آن پرداخته خواهد شد. به‌طور خلاصه اما می‌توان گفت، *اعتبارسنجی متقابل تکنیکی است که مجموعه آموزشی اصلی را به دو مجموعه داده آموزشی و آزمایشی (اعتبارسنجی) تقسیم می‌کند.*

هنگام برخورد با داده‌های سری زمانی که یک مسئله‌یِ پیش‌بینی است، مجموعه‌های آموزش، اعتبارسنجی و آزمایش باید با تقسیم داده‌ها در امتداد محور زمانی انتخاب شوند. یعنی "قدیمی‌ترین" داده‌ها برای آموزش، جدیدتر برای اعتبارسنجی و جدیدترین داده برای آزمایش استفاده می‌شود. نمونه گیری تصادفی در این مورد معنی ندارد.

## تقسیم‌بندی داده‌ها

هر چند الگوریتم‌های یادگیری ماشین بانظارت ابزارهای شگفت‌انگیز و قدرتمند در پیش‌بینی و دسته‌بندی هستند، اما این سوال به‌وجود می‌آید که این پیش‌بینی‌ها تا چه اندازه دقیق هستند و آیا راهی برای سنجش میزان کارایی مدل وجود دارد؟ از آنجایی که این الگوریتم‌ها دارای نمونه‌های برچسب خورده می‌باشند، این پرسش را می‌توان با تقسیم نمونه‌های آموزشی به چندین بخش، پاسخ داد.

با تقسیم‌بندی داده‌ها ابتدا، آموزش را روی بخشی از داده‌ها انجام داده، سپس برای سنجش میزان کارای مدل و قابلیت **تعمیم‌دهی** آن از داده‌های آزمایشی استفاده می‌کنیم. تعمیم‌دهی نشان‌دهنده میزان عملکرد مدل، در برخورد با داده‌هایی است، که تاکنون مدل آن‌ها را در فرآیند آموزش مشاهده نکرده است. البته در طراحی مدل‌های یادگیری ماشین در بیشتر اوقات مجموعه داده‌های مسئله مورد نظر را علاوه بر داده‌های آموزشی و آزمایشی به بخش دیگری نیز تقسیم می‌کنیم، نحوه‌یِ این تقسیم‌بندی به‌صورت زیر می‌باشد:

- **مجموعه آموزشی:** به‌طور معمول بزرگ‌ترین در بین این سه دسته داده‌ها می‌باشد و برای یافتن پارامترهای مدل مورد استفاده قرار می‌گیرد. مجموعه داده آموزشی رابطه اساسی بین داده‌ها و برچسب‌های آن را به بهترین وجه ممکن توضیح می‌دهد.
- **مجموعه آزمایشی:** اندازه‌گیری عملکرد یک مدل را براساس توانایی مدل در پیش‌بینی داده‌هایی که در فرآیند یادگیری نقشی نداشته می‌سنجند، مجموعه آزمایشی همان داده‌های دیده نشده در فرآیند یادگیری هستند. این مجموعه کارایی مدل نهایی را می‌سنجد. اگر مدلی عملکرد خوبی در مجموعه آموزشی داشته باشد و همچنین متناسب با مجموعه آزمون باشد، یعنی برچسب درست را برای تعداد زیادی از داده‌های ورودی نادیده پیش‌بینی کند، حداقل بیش‌برازش صورت

گرفته است. لازم به ذکر است که مجموعه آزمون معمولاً فقط یک بار به محض مشخص شدن کامل پارامترها و ابرپارامترهای مدل برای ارزیابی عملکرد تعمیم‌دهی مدل استفاده می‌شود. با این حال، برای تقریب عملکرد پیش‌بینی یک مدل در طول آموزش، از یک مجموعه اعتبارسنجی استفاده می‌شود.

- **مجموعه اعتبارسنجی:** در ارزیابی انواع مختلف مدل‌ها و الگوریتم‌ها برای مسئله مورد نظر از مجموعه اعتبارسنجی استفاده می‌شود. از این داده‌ها برای تنظیم ابرپارامترها و جلوگیری از بیش‌برازش مدل استفاده تا بهترین مدل انتخاب شود.

هیچ قانون کلی در مورد نحوه تقسیم داده‌ها وجود ندارد. با این حال، مجموعه اعتبارسنجی باید به اندازه کافی بزرگ باشد تا بتوانیم تفاوت عملکردی را که می‌خواهیم بدست آوریم، اندازه‌گیری کنیم.

مجموعه اعتبارسنجی برای بدست آوردن مقادیر ابرپارامترهای بهینه (بهینه‌سازی ابرپارمترها) و کمک به انتخاب مدل استفاده می‌شود و مجموعه آزمون برای ارزیابی عملکرد مدل نهایی در نمونه‌های دیده شده در فرآیند یادگیری استفاده می‌شود.

## موازنه سوگیری و واریانس (Bias-VarianceTrade-Off)

در یادگیری ماشین، رابطه‌ِی بین پیچیدگی مدل، خطای آموزش و آزمون، نتیجه‌ِی دو ویژگی رقابتی سوگیری و واریانس است. سوگیری به خطایی اشاره دارد که هنگام تلاش برای استفاده از یک مدل ساده برای حل یک مشکل پیچیده دنیای واقعی معرفی می‌شود. به عبارت دیگر، این ناتوانی یک مدل یادگیری ماشین، در به تصویر کشیدن رابطه واقعی در داده‌ها است. به عنوان مثال، اگر بخواهیم از رگرسیون خطی برای تخمین یک رابطه غیرخطی استفاده کنیم، مدل بایاس بالایی خواهد داشت. این به این دلیل است که یک خط مستقیم هرگز نمی‌تواند به اندازه کافی انعطاف‌پذیر باشد تا یک رابطه غیرخطی را به تصویر بکشد.

در مقابل، واریانس تفاوت در **برازش** (**fit**) بین مجموعه‌های داده است. برای مثال، زمانی که یک مدل بیش‌برازش می‌کند، واریانس بالایی دارد. چراکه خطای پیش‌بینی برای مجموعه آموزشی و آزمایشی بسیار متفاوت است.

به طور کلی، ما دوست داریم تا حد امکان سوگیری و واریانس کم‌تری داشته باشیم. با این حال، این معیارها اثرات متضادی دارند و نمی‌توان سوگیری را بدون افزایش واریانس کاهش داد. به منظور یافتن تعادل بهینه بین سوگیری و واریانس، چندین مدل را ارزیابی می‌کنیم تا بهترین پارامترها را برای مدل پیدا کنیم. به عنوان مثال، گاهی اوقات یک مجموعه داده را به دو

بخش تقسیم می‌کنیم: مجموعه آموزشی و آزمایشی. هنگام ارزیابی نحوه عملکرد یک مدل ساخته شده بر روی مجموعه آموزشی، هم در مجموعه آموزشی و هم در مجموعه آزمایشی، می‌خواهیم خطای پیش‌بینی تا حد امکان کم باشد. اگر مدل دارای خطای پیش‌بینی کم در مجموعه آموزشی باشد، اما خطای پیش‌بینی بالا در مجموعه آزمایشی داشته باشد، گفته می‌شود که مدل دارای واریانس بالایی است و در نتیجه داده‌ها منجر به بیش‌برازش شده است. به طور کلی، مدل‌های پیچیده‌تر واریانس بالاتری دارند. این به این دلیل است که یک مدل پیچیده می‌تواند داده‌های خاصی را که با آن مطابقت دارد را با دقت بیشتری دنبال کند. با این حال، از آنجایی که یک مدل پیچیده داده‌ها را با دقت بیشتری دنبال می‌کند، به احتمال زیاد رابطه واقعی را در داده‌های آموزشی نشان می‌دهد و در نتیجه سوگیری کم‌تری دارد. از این‌رو، انتخاب مدلی با سوگیری نسبتا کم‌تر تنها با هزینه واریانس بالاتر قابل دستیابی است.

از سوی دیگر، اگر مدل دارای خطای پیش‌بینی بالا در هر دو مجموعه آموزشی و آزمایشی باشد، گفته می‌شود که مدل دارای سوگیری بالایی است و در نتیجه داده‌ها را نادیده می‌گیرد. به‌طور خلاصه، اگر مدل مبتنی بر داده‌های آموزشی بسیار پیچیده باشد، خطای پیش‌بینی در داده‌های آموزشی کم خواهد بود. به عبارت دیگر، مدل معمولا منجر به بیش‌برازش می‌شود و در نتیجه به خوبی با داده‌های آزمایشی مطابقت ندارد و باعث خطای پیش‌بینی بالاتر برای مجموعه آزمایشی می‌شود. هدف یافتن راه‌حل بهینه است و این یک موازنه بین سوگیری و واریانس است.

راه‌های مختلفی برای تنظیم سوگیری و واریانس وجود دارد. اکثر الگوریتم‌ها دارای پارامترهایی هستند که پیچیدگی مدل را تنظیم می‌کنند. این فرآیند اغلب به عنوان "**تنظیم ابرپارامترها**" شناخته می‌شود، که بخشی ضروری از مرحله ارزیابی مدل است.

سوگیری ناتوانی یک مدل یادگیری ماشین برای بدست آوردن رابطه واقعی بین متغیرهای داده است. این امر ناشی از فرضیات اشتباهی است که درون الگوریتم یادگیری است. مدل با سوگیری بالا توجه بسیار کمی به داده‌های آموزشی دارد و مدل را بیش از حد ساده می‌کند و همیشه منجر به خطای بالایی در آموزش و داده‌های آزمون می‌شود.

واریانس نشان می دهد که در صورت استفاده از داده های آموزشی متفاوت، برآورد تابع هدف چقدر تغییر می کند. به عبارت دیگر، واریانس بیان می‌کند که یک متغیر تصادفی چقدر با مقدار مورد انتظارش تفاوت دارد. مدل با واریانس بالا توجه زیادی به داده‌های آموزشی می‌کند و به داده‌هایی که قبلا ندیده تعمیم نمی‌یابد. در نتیجه، چنین مدل‌هایی روی داده‌های آموزشی بسیار خوب عمل می‌کنند، اما نرخ خطای بالایی در داده‌های آزمایشی دارند.

## روش‌های ارزیابی

در حالی‌که آموزش مدل یک گام کلیدی است، نحوه تعمیم‌دهی مدل برروی داده‌های دیده نشده جنبه‌ای به همان اندازه مهم است که باید پس از طراحی هر مدل یادگیری ماشین در نظر گرفته شود. باید این اطمینان حاصل شود که آیا مدل واقعا کارا است و می‌توان به نتایج پیش‌بینی‌های آن اعتماد کرد یا خیر.

یک الگوریتم طبقه‌بندی می‌تواند برروی یک مجموعه داده خاص با مجموعه‌ای منحصر به فرد از پارامترها آموزش داده شود که می‌تواند مرز تصمیم‌گیری را متناسب با داده‌ها ایجاد کند. نتیجه آن الگوریتم خاص نه‌تنها به پارامترهای ارائه شده برای آموزش مدل بستگی دارد، بلکه به نوع داده‌های آموزشی نیز بستگی دارد. اگر داده‌های آموزشی حاوی واریانس کم یا داده‌ها یکنواخت باشد، ممکن است مدل منجر به بیش‌برازش گردد و نتایج جانبدارانه‌ای را نسبت به داده‌های دیده نشده ایجاد کند. بنابراین، از رویکردهایی همانند اعتبارسنجی متقابل برای به‌حداقل رساندن بیش‌برازش استفاده می‌شود. اعتبارسنجی متقابل تکنیکی است که مجموعه آموزشی اصلی را به دو مجموعه داده آموزشی و آزمایشی (اعتبارسنجی) تقسیم می‌کند. رایج‌ترین روش اعتبارسنجی متقابل، اعتبارسنجی متقابل چند_بخشی است که مجموعه داده اصلی را به k بخش با اندازه یکسان تقسیم می‌کند. k یک عدد مشخص شده توسط کاربر است که معمولا ۵ یا ۱۰ انتخاب می‌شود. در این روش هر بار یکی از زیر مجموعه‌های k به‌عنوان مجموعه اعتبارسنجی (آزمون) مورد استفاده قرار می‌گیرد و k-1 زیر مجموعه‌ی دیگر برای تشکیل یک مجموعه آموزشی کنار هم قرار می‌گیرند. برای بدست آوردن کارآیی کل مدل، برآورد خطا در تمام آزمایش‌ها به‌طور میانگین محاسبه می‌شود.

**در تکنیک اعتبارسنجی متقابل چند-بخشی هر یک از داده‌ها دقیقا یک بار در یک مجموعه آزمایشی قرار و یک بار در یک مجموعه آموزشی قرار می‌گیرد. این امر به‌طور قابل توجهی بایاس و واریانس را کاهش می‌دهد، چراکه تضمین می‌کند هر نمونه‌ای از مجموعه داده اصلی این شانس را دارد که در مجموعه آموزشی و آزمایشی ظاهر شود. اگر داده های ورودی محدودی داشته باشیم، اعتبارسنجی متقابل چند-بخشی از بهترین روش‌ها برای ارزیابی کارآیی یک مدل است.**

## معیارهای ارزیابی کارآیی

برای محاسبه معیارهای ارزیابی کارآیی یک مدل طبقه‌بندی نیاز به چهار ترکیب از کلاس واقعی و کلاس پیش‌بینی با عناوین، مثبت راستین، مثبت کاذب، منفی راستین و منفی کاذب داریم که می‌توان آن‌ها را در یک ماتریس درهم‌ریختگی (Confusion Matrix) نشان داد (جدول ۵_۱). جایی که:

- **مثبت راستین($TP$):** به عنوان مثال، وقتی مقدار واقعی کلاس "**بله**" بود، مدل هم "بله" را پیش‌بینی کرد (یعنی پیش‌بینی درست)
- **مثبت کاذب($FP$):** به عنوان مثال، یعنی زمانی که مقدار واقعی کلاس "نه" بود، اما مدل "**بله**" را پیش‌بینی کرد (یعنی پیش‌بینی اشتباه)
- **منفی کاذب($FN$):** به عنوان مثال، زمانی که مقدار واقعی کلاس "**بله**" بود، اما مدل "نه" را پیش‌بینی کرد (یعنی پیش‌بینی اشتباه)
- **منفی راستین($TN$):** به عنوان مثال، یعنی زمانی که مقدار واقعی کلاس "**نه**" بود و مدل هم "**نه**" را پیش‌بینی کرد (یعنی پیش‌بینی درست).

جدول ۲ـ۱. ماتریس درهم‌ریختگی

| | | کلاس پیش‌بینی‌شده | |
|---|---|---|---|
| | | مثبت | منفی |
| کلاس واقعی | مثبت | مثبت راستین (TP) | منفی کاذب (FN) |
| | منفی | منفی کاذب (FP) | منفی راستین (TN) |

رایج‌ترین معیاری که از ماتریس درهم‌ریختگی بدست می‌آید **دقت (accuracy)** یا معکوس آن، **خطای پیش‌بینی (prediction error)** است:

$$\text{دقت} = \frac{TP + TN}{TP + FN + FP + TN}$$

$$\text{خطای پیش‌بینی} = 1 - \text{دقت}$$

دقت، نسبت تعداد پیش‌بینی‌های درست به تعداد کل نمونه‌های ورودی است.

زمانی که باید یک مدل را ارزیابی کنیم، در اغلب اوقات از نرخ خطا و دقت استفاده می‌کنیم، اما چیزی که عمدتا روی آن تمرکز می کنیم این است که مدل ما چقدر قابل اعتماد است، چگونه روی یک مجموعه داده متفاوت عمل می‌کند (قابلیت تعمیم) و چقدر انعطاف‌پذیری دارد. بدون شک دقت معیار بسیار مهمی است که باید در نظر گرفته شود، اما همیشه تصویر کاملی را از کارآیی مدل ارائه نمی‌دهد.

وقتی می‌گوییم مدل قابل اعتماد است، منظور ما این است که مدل از داده‌ها به‌درستی و مطابق خواسته یادگیری بدست آورده است. بنابراین، پیش‌بینی‌های انجام شده توسط آن به مقادیر واقعی نزدیک است. در برخی موارد، مدل ممکن است منجر به دقت بهتری شود، اما نمی‌تواند داده‌ها را به‌درستی درک کند و بنابراین زمانی که داده‌ها متفاوت هستند، عملکرد ضعیفی دارد. این بدان

معنی است که مدل به اندازه کافی قابل اعتماد و قوی نیست و از این رو استفاده از آن را محدود می‌کند.

به عنوان مثال، ما ۹۸۰ سیب و ۲۰ پرتقال داریم و یک مدل داریم که هر میوه را به عنوان یک سیب دسته‌بندی می‌کند. از این‌رو دقت مدل ۹۸٪=۹۸۰/۱۰۰۰ است و بر اساس معیار دقت، ما یک مدل بسیار دقیق داریم. با این حال، اگر ما از این مدل برای پیش‌بینی میوه‌ها در آینده استفاده کنیم، با شکست مواجه خواهیم شد. چرا که این مدل تنها می‌تواند یک کلاس را پیش‌بینی کند.

دریافت تصویری کامل از مدل، به عنوان مثال اینکه چگونه داده‌ها را درک می‌کند و چگونه می‌تواند پیش‌بینی کند، به درک عمیق ما از مدل کمک می‌کند و به بهبود آن کمک می‌کند. بنابراین، فرض کنید مدلی را ایجاد کرده‌اید که دقت ۹۰٪ را بدست می‌آورد، از این‌رو چگونه می‌خواهید آن را بهبود ببخشید؟ برای تصحیح یک اشتباه، ابتدا باید متوجه آن اشتباه شویم. به‌طور مشابه، برای بهبود مدل ما باید به نحوهٔ عملکرد مدل در سطح عمیق‌تری نگاه کنیم. با این حال، این کار تنها با نگاه کردن به معیار دقت بدست نمی‌آید و از این‌رو معیارهای دیگری نیز در نظر گرفته می‌شود. معیارهایی همانند، **صحت (precision)** ، **فراخوانی(recall)** و F1 نمونه‌هایی از این معیارها هستند.

فراخوانی، به توانایی یک مدل در پیش‌بینی موارد مثبت از کل موارد مثبت راستین اشاره می‌کند. در حالی‌که صحت، کسری از موارد مثبت راستین را در بین نمونه‌هایی که به عنوان مثبت پیش‌بینی شده‌اند، اندازه‌گیری می‌کند. صحت و فراخوانی به تنهایی ممکن است برای ارزیابی مدل مناسب نباشد، بنابراین از امتیازF1 استفاده می‌شود که هم صحت و هم فراخوانی را شامل می‌شود و نشان می‌دهد طبقه‌بند چقدر دقیق است. هرچه امتیاز F1 بیشتر باشد، کارآیی مدل ما بهتر است. نحوهٔ محاسبات این معیارها به‌صورت زیر است:

$$\text{فراخوانی} = \frac{TP}{FN + TP}$$

$$\text{صحت} = \frac{TP}{FP + TP}$$

$$\text{F1} = 2 \times \frac{\text{فراخوانی} * \text{صحت}}{\text{فراخوانی} + \text{صحت}}$$

# ابزارها و کتابخانه‌های پایتون

## نصب پایتون

در این بخش مراحل نصب پایتون را در سیستم عامل Windows معرفی می‌کنیم. از آن‌جایی که هیچ محیط پایتون داخلی در سیستم عامل ویندوز وجود ندارد، باید به‌طور مستقل نصب شود. بسته نصب را می‌توان از وب سایت رسمی پایتون (www.Python.org) بارگیری کرد. پس از باز کردن وب سایت رسمی، نوار ناوبری را که دارای دکمه بارگیری (download) است جستجو کنید. وب‌سایت، پیوندی را به‌طور پیش فرض توصیه می‌کند، چرا که می تواند سیستم عامل شما را شناسایی کرده و آخرین نسخه Python 3.x را توصیه کند. پس از ورود به صفحه بارگیری نسخه مربوطه، مقدمه‌ای اساسی در مورد محیطی که می‌خواهید بارگیری کنید وجود دارد. چندین نسخه مختلف عمدتا برای سیستم عامل‌های مختلف طراحی شده‌اند. بسته به ۳۲ یا ٦٤ بیتی بودن سیستم، می‌توانید فایل‌های مختلفی را برای بارگیری انتخاب کنید. در صفحه جدیدی که باز می شود، می‌توانیم نسخه‌های دیگری را نیز پیدا کنیم، از جمله آخرین نسخه آزمایشی و نسخه مورد نیاز. اگر می‌خواهید نسخه ٦٤ بیتی ۳/۹/٦ را نصب کنید، روی پیوند ارائه شده در صفحه کنونی کلیک کنید.

پس از بارگیری پایتون، نوبت به نصب آن می‌رسد. نصب بسته ویندوز بسیار آسان است. درست مانند نصب سایر برنامه‌های ویندوز، ما فقط باید گزینه مناسب را انتخاب کرده و روی دکمه "بعدی" کلیک کنیم تا نصب کامل شود. هنگامی که گزینه‌ها در هنگام نصب ظاهر می‌شوند، برای رفتن به مرحله بعدی عجله نکنید. چرا که برای راحتی در آینده، باید یک دکمه را انتخاب کنید. پس از علامت‌گذاری دکمه "Add Python 3.9.6 to PATH" به متغیر محیط ، می‌توان در آینده دستورات پایتون را مستقیما و به‌راحتی در خط فرمان Windows اجرا کرد. پس از انتخاب "Add Python 3.9.6 to PATH"، نصب دلخواه را انتخاب کنید. البته امکان انتخاب مکان نصب نیز وجود دارد، که به‌طور پیش فرض در دایرکتوری کاربری در درایو C نصب شده است. با این حال، بهتر است بدانید دایرکتوری کاربر چیست تا بتوانید در مواقع ضروری فایل‌های Python.exe نصب شده را پیدا کنید. دستورالعمل‌ها را ادامه دهید تا پایتون با موفقیت در سیستم نصب شود.

## شروع کار با پایتون

راه‌اندازی پایتون به دو صورت امکان‌پذیر است:

۱) **با استفاده از IDLE خود پایتون.** اگر می‌خواهید پایتون را اجرا کنید، می‌توانید روی دکمه "شروع" در دسکتاب ویندوز کلیک و در کادر "جستجو" عبارت "IDLE" را تایپ کنید تا

به‌طور سریع وارد "read-evaluate-print-loop" شوید. پس از اجرای برنامه، با تصویری همانند زیر روبه‌رو می‌شوید:

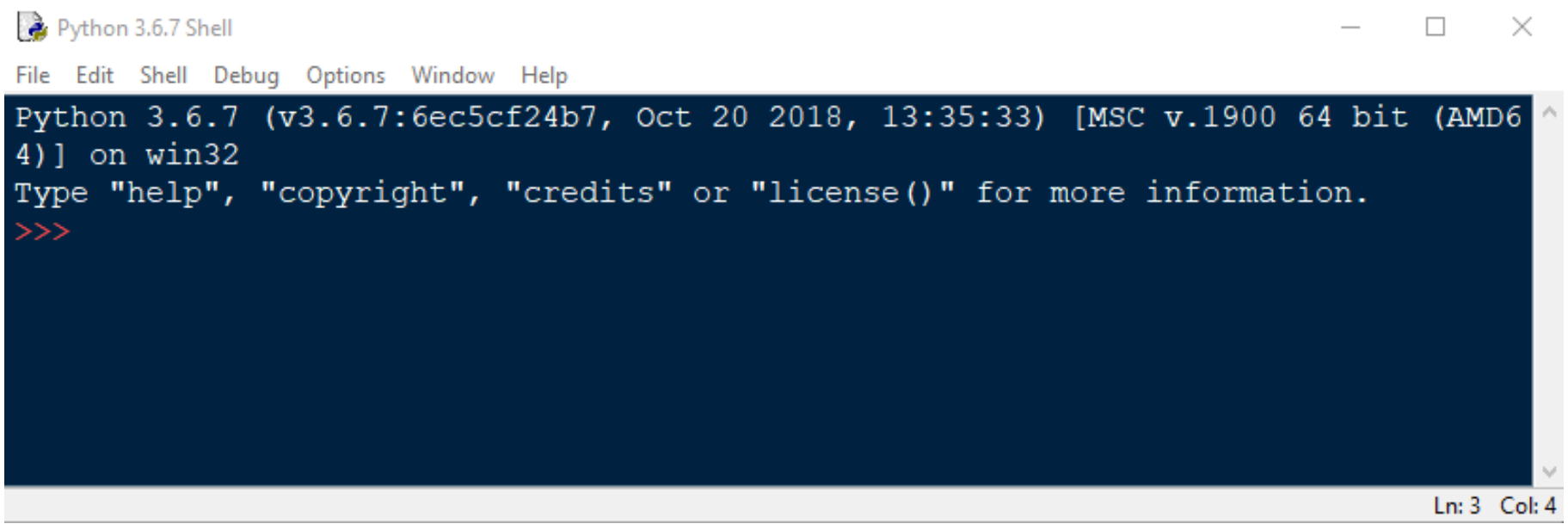

IDLE یک IDE (Integrated Development Environment) ساده خود پایتون است که یک ویرایشگر رابط گرافیکی را در اختیار کاربران قرار می‌دهد. عملکرد آن ساده به نظر می‌رسد و برای مبتدیان یادگیری زبان پایتون مناسب است. توسط IDLE یک محیط REPL ارائه می‌شود، یعنی ورودی کاربر () را می‌خواند، ارزیابی و محاسبه می‌کند ()، سپس نتیجه را چاپ می‌کند () و یک پیغام "حلقه" (منتظر ورودی بعدی) ظاهر می‌شود.

۲) **با استفاده از Windows Prompt.** راه دیگر برای راه‌اندازی پایتون، اجرای برنامه‌های پایتون از طریق خط فرمان ویندوز است. برای این کار کلیدهای "**Win+R**" را فشار دهید تا کادر اعلان باز شود و سپس در کادر باز شده، "**cmd**" را وارد کنید. اگر در هنگام نصب پایتون "Add Python 3.x to PATH" را علامت زده باشید، پایتون نصب‌شده به متغیر محیط ویندوز اضافه شده است. حال با وارد کردن کلمهٔ "**python**" پس از ظاهر شدن "**>>>**" پایتون با موفقیت اجرا می‌شود و با تصویری همانند زیر روبه‌رو می‌شوید:

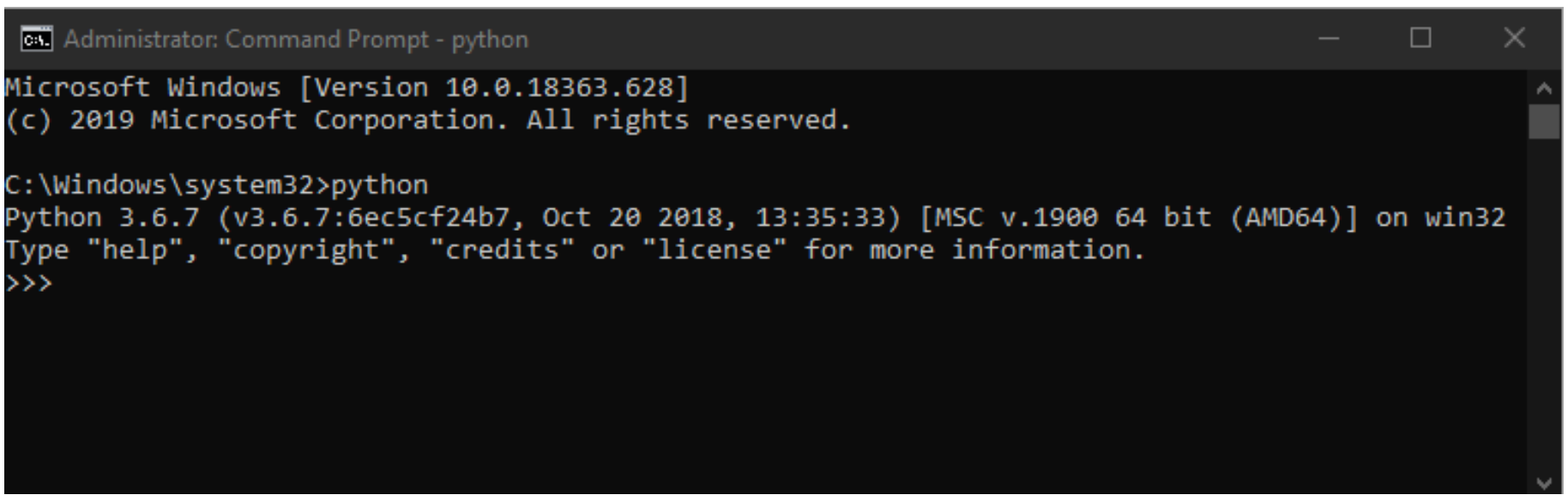

اعلان "<<<" بیانگر این است که نصب با پایتون موفقیت‌آمیز بوده و پایتون شروع به‌کار کرده است.

## نصب کتابخانه‌ها

برای مدیریت کتابخانه‌های پایتون باید از Pip استفاده کنید. Pip یک ابزار ضروری است که به شما امکان می‌دهد بسته‌های مورد نیاز خود را بارگیری، بروزرسانی و حذف کنید. علاوه‌براین با استفاده از آن می‌توان وابستگی‌های مناسب و سازگاری بین نسخه‌ها را بررسی کنید.

نصب یک کتابخانه با استفاده از Pip در خط فرمان ویندوز صورت می‌گیرد. برای مثال فرض کنید می‌خواهیم کتابخانه NumPy را نصب کنیم. مراحل زیر نحوه نصب این کتابخانه را نشان می‌دهد:

- ابتدا کلیدهای "Win+R" را فشار دهید تا کادر اعلان باز شود و سپس در کادر باز شده، "cmd" را وارد کنید. سپس دستور زیر را در خط فرمان وارد کنید:

```
> pip install numpy
```

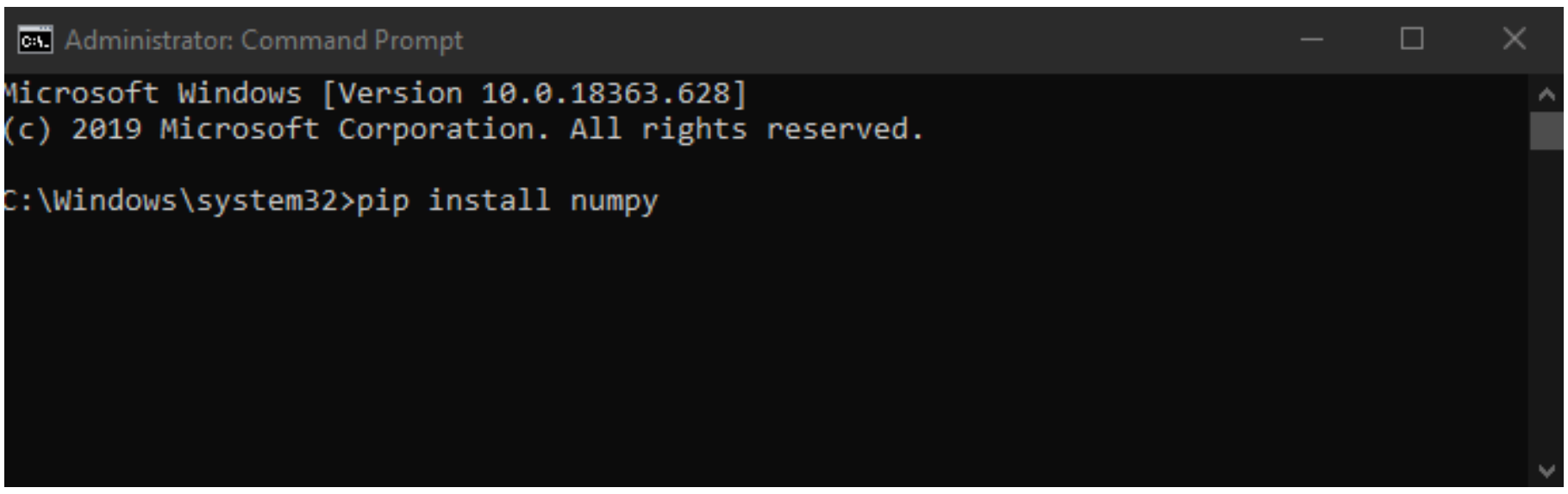

- برای اطمینان از نصب کتابخانه، از خط فرمان پایتون را اجرا کرده و دستور زیر را بنویسید:

```
>>> import numpy
```

- اگر کتابخانه به‌درستی نصب شده باشد پیغامی مشاهده نمی‌شود. در صورتی کتابخانه در رایانه شما نصب نشده باشد با اجرای دستور فوق، این پیغام را مشاهده خواهید کرد:

```
Traceback (most recent call last):
  File "<stdin>", line 1, in <module>
ImportError: No module named numpy
```

## JupyterNotebook

Jupyter Notebook یک ابزار فوق‌العاده قدرتمند برای توسعه و ارائه پروژه‌های یادگیری ماشین به‌صورت تعاملی است که می‌تواند علاوه بر اجرای کد، شامل متن، تصویر، صدا و یا ویدیو باشد. یک Notebook، کد و خروجی آن را با مصورسازی، متنِ روایی، معادلات ریاضی و سایر رسانه‌ها در قالب یک سند واحد ترکیب می‌کند. به عبارت دیگر، یک Notebook، یک سند واحد است که در آن می‌توانید کد را اجرا کنید، خروجی را نمایش دهید و همچنین توضیحات،

فرمول‌ها، نمودارها را اضافه، تا کار خود را شفاف‌تر، قابل فهم، تکرارپذیر و قابل اشتراک‌گذاری کنید.

برای نصب Jupyter Notebook، لازم است پایتون را از قبل نصب کرده باشید. حتی اگر قصد داشته باشید از جوپیتر برای سایر زبان‌های برنامه‌نویسی استفاده کنید، پایتون ستون اصلی جوپیتر است. برای نصب جوپیتر کافی است در خط فرمان ویندوز دستور زیر را بنویسید:

```
> pip install jupyter
```

برای اجرای Jupyter خط فرمان را باز کرده و دستور زیر را در آن بنویسید:

```
> jupyter notebook
```

پس از اجرای دستور فوق، مرورگر وب پیش‌فرض شما با Jupyter راه‌اندازی می‌شود. هنگام راه‌اندازی Jupyter Notebook به دایرکتوری خط فرمان توجه فرمایید، چرا که این دایرکتوری به فهرست اصلی تبدیل می‌شود که بلافاصله در جوپیتر نوت‌بوک ظاهر می‌شود و تنها به پرونده‌ها و زیردایرکتوری‌هایی موجود در آن دسترسی خواهید داشت. با اجرای دستور jupyter notebook با صفحه‌ای همانند زیر روبه‌رو می‌شوید:

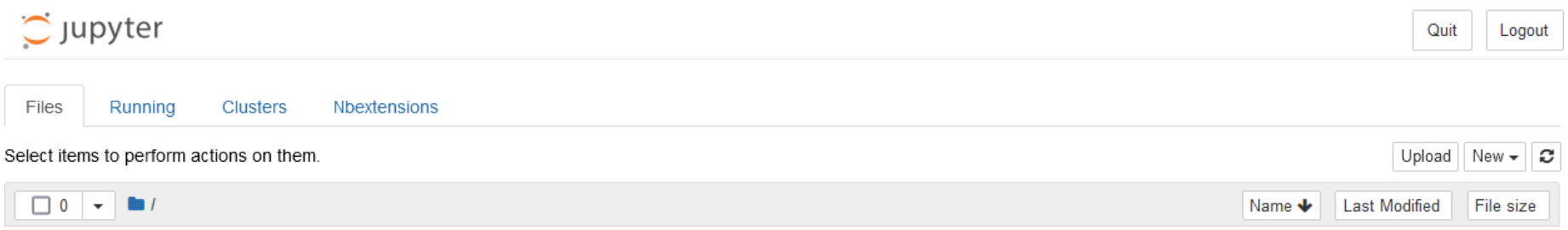

با این حال، این صفحه هنوز یک notebook نیست و تنها میزکار جوپیتر است که برای مدیریت نوت‌بوک‌های جوپیتر شما طراحی شده است و آن را به‌عنوان راه‌اندازی برای پیگردی (exploring)، ویرایش و ایجاد notebookهای خود در نظر بگیرید. notebookها و میزکار جوپیتر مبتنی‌بر مرورگر است و جوپیتر یک سرور محلی پایتون راه‌اندازی می‌کند تا این برنامه‌ها را به مرورگر وب شما ارتباط دهد.

برای ایجاد یک notebook جدید به به دایرکتوری که قصد دارید اولین notebook خود را در ایجاد کنید بروید و بر روی دکمه کشویی "New" که در قسمت بالای میزکار در سمت راست است کلیک کرده و گزینه "Python 3"را انتخاب کنید:

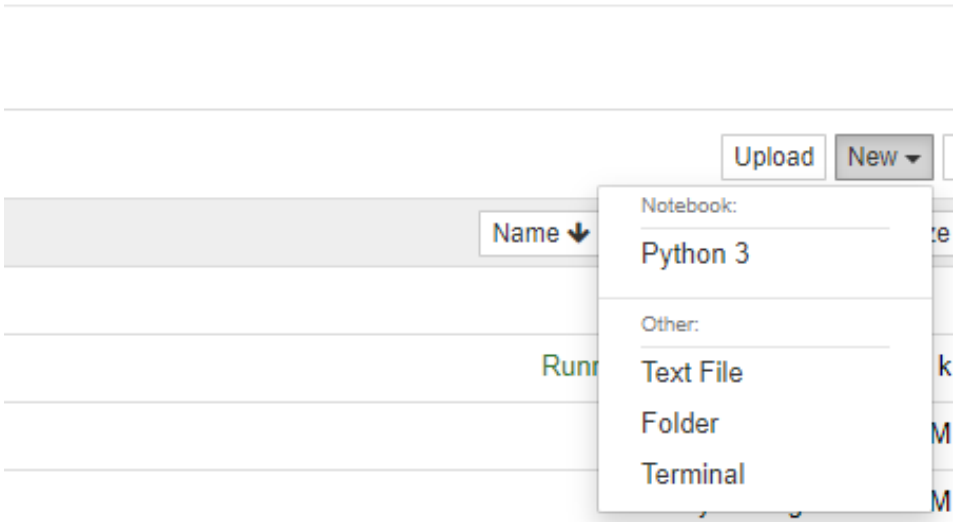

پس از آن، اولین نوت‌بوک شما در یک برگه جدید (new tab) همانند تصویر زیر باز می‌شود:

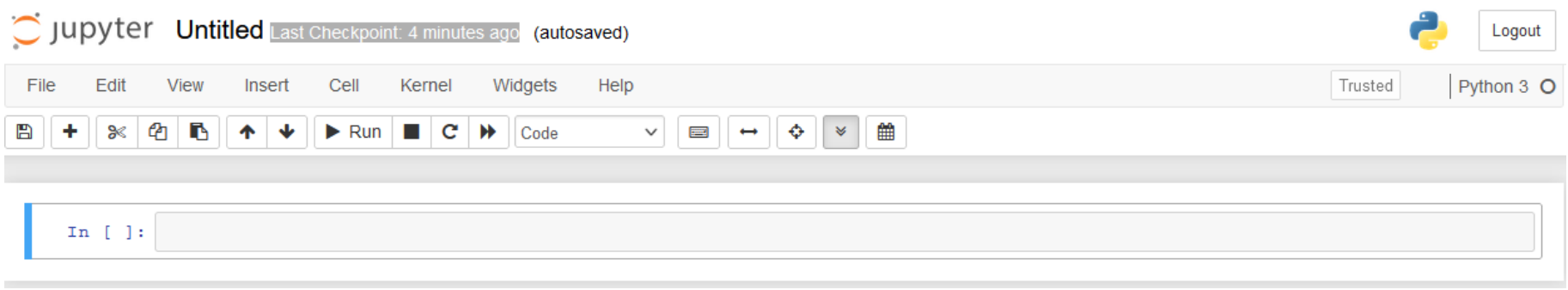

اگر به میزکار Jupyter بازگردید، فایل جدید Untitled.ipynb را مشاهده خواهید کرد و باید متن سبز رنگی را مشاهده کنید که به شما می‌گوید نوت‌بوک شما در حال اجرا است.

بیاید نحوه اجرای یک سلول را با یک مثال کلاسیک آزمایش کنیم: print('Hello World!') را در یک سلول تایپ کنید و برروی دکمه Run در نوار ابزار بالا کلیک کنید یا دکمه‌های Ctrl+Enter را فشار دهید. نتیجه آن به این صورت خواهد بود:

```
In [1]: print('Hello World!')

        Hello World!
```

# Colab

Colaboratory یا به اختصار Colab یک محصول تحقیقاتی گوگل (سرویس ابری) است که به توسعه‌دهندگان اجازه می‌دهد کدهای پایتون را از طریق مرورگر خود بنویسند و اجرا کنند. Google Colab یک ابزار عالی برای کارهای یادگیری عمیق است و به توسعه مدل‌ها با استفاده از چندین کتابخانه مانند Keras، Pytorch، OpenCv، Tensorflow و غیره کمک می‌کند. Colab یک نوت‌بوک مبتنی‌بر Jupyter است که نیازی به نصب ندارد و دارای نسخه رایگان عالی است که دسترسی رایگان به منابع محاسباتی Google مانند GPU و TPU را می‌دهد.

## چرا Colab؟

Colab برای همه چیز ایده‌آل است، از بهبود مهارت‌های کدنویسی پایتون گرفته تا کار با کتابخانه‌های یادگیری عمیق، مانند PyTorch، Keras، TensorFlow و OpenCV. می‌توانید نوت‌بوک‌ها را در Colab ایجاد، بارگذاری، ذخیره و به اشتراک بگذارید، Google Drive خود را نصب کنید و از هر چیزی که در آنجا ذخیره کرده‌اید استفاده کنید، نوت‌بوک‌ها را مستقیما از GitHub بارگذاری کنید، فایل‌های Kaggle را بارگذاری کنید، نوت‌بوک‌های خود را باگیری کنید و تقریبا هر کار دیگری را که ممکن است بخواهید انجام دهید را انجام دهید.

از دیگر ویژگی‌های عالی Google Colab، ویژگی همکاری (collaboration) است. اگر با چند برنامه‌نویس روی یک پروژه کار می‌کنید، استفاده از نوت‌بوک Google Colab عالی است. درست همانند همکاری در یک سند Google Docs، می‌توانید با استفاده از یک نوت‌بوک Colab با چندین برنامه‌نویس کدنویسی کنید. علاوه بر این، شما همچنین می‌توانید کارهای تکمیل شده خود را با توسعه‌دهندگان دیگر به اشتراک بگذارید.

به‌طور خلاصه می‌توان دلایل مختلف استفاده از Colab را به‌صورت زیر فهرست کرد:

- کتابخانه‌های از پیش نصب‌شده
- ذخیره‌شده در ابر
- همکاری
- استفاده از GPU و TPU رایگان

با این حال، دو سناریو وجود دارد که شما باید از Jupyter Notebook در ماشین خود استفاده کنید:

۱. اگر به حریم خصوصی اهمیت می‌دهید و می‌خواهید کد خود را مخفی نگه دارید، از Google Colab دوری کنید.
۲. اگر یک ماشین فوق‌العاده قدرتمند با دسترسی به GPU و TPU دارید.

## راه‌اندازی Google Colab

فرآیند راه‌اندازی Colab نسبتا آسان است و می‌تواند با مراحل زیر در هر نوع دستگاهی تکمیل شود:

۱. **از صفحه Google Colab دیدن کنید:**

http://colab.research.google.com

بارگذاری تارنمای فوق شما را به صفحه خوش‌آمدگویی Google Colaboratory هدایت می‌کند

۲. **روی دکمه ورود به سیستم (Sign in) در بالا سمت راست کلیک کنید:**

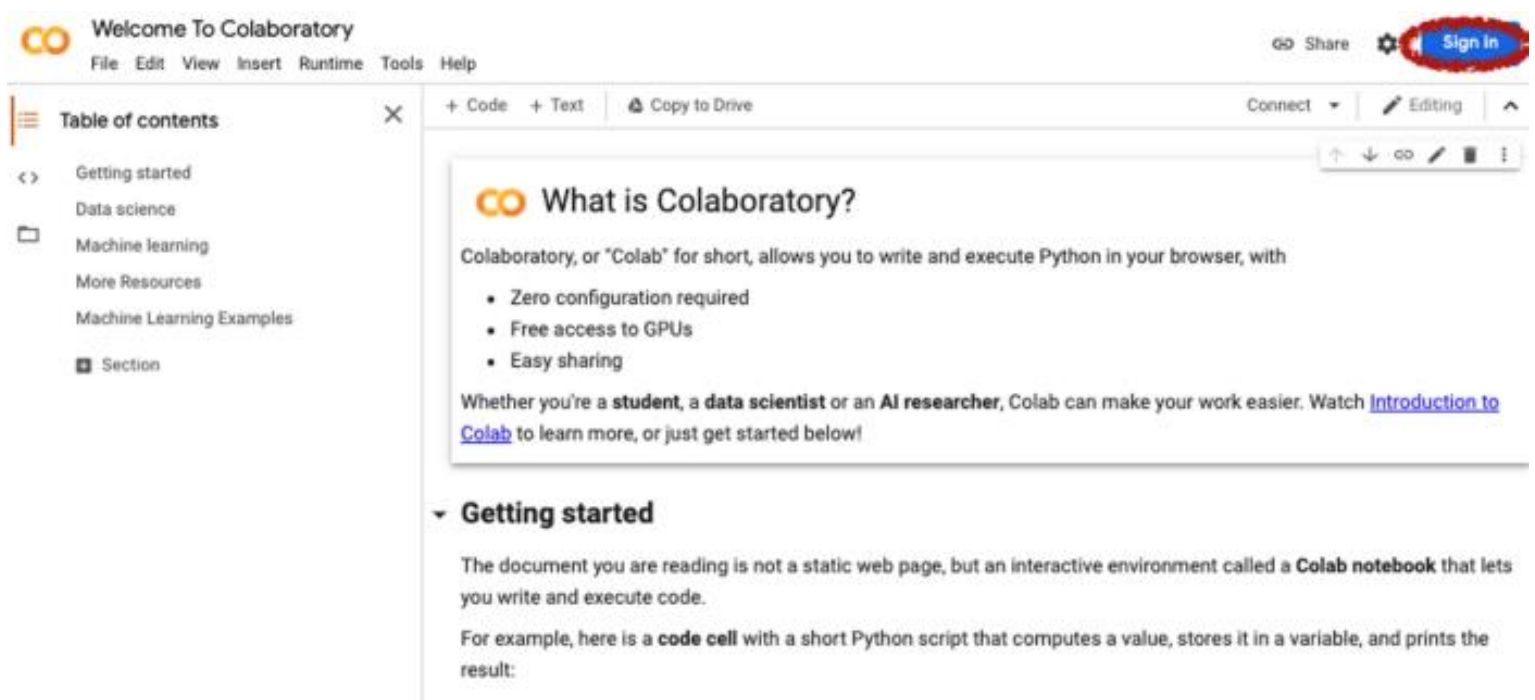

۳. با حساب **GMail** خود وارد شوید. اگر حساب **GMail** ندارید یکی بسازید:

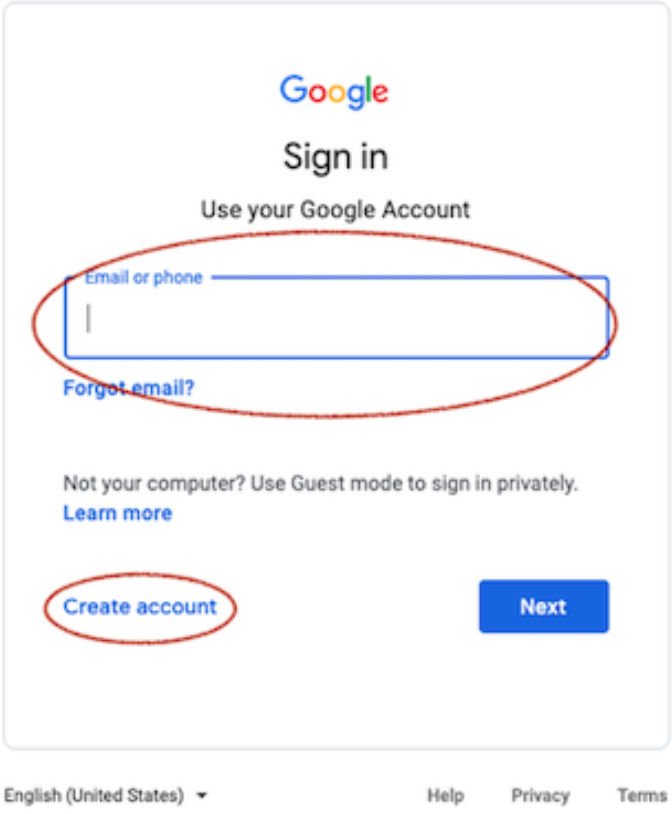

۴. به محض تکمیل فرآیند ورود به سیستم، آماده استفاده از **Google Colab** هستید.

۵. با کلیک بر روی **File> New notebook** به‌راحتی می توانید یک نوت‌بوک **Colab** جدید در این صفحه ایجاد کنید.

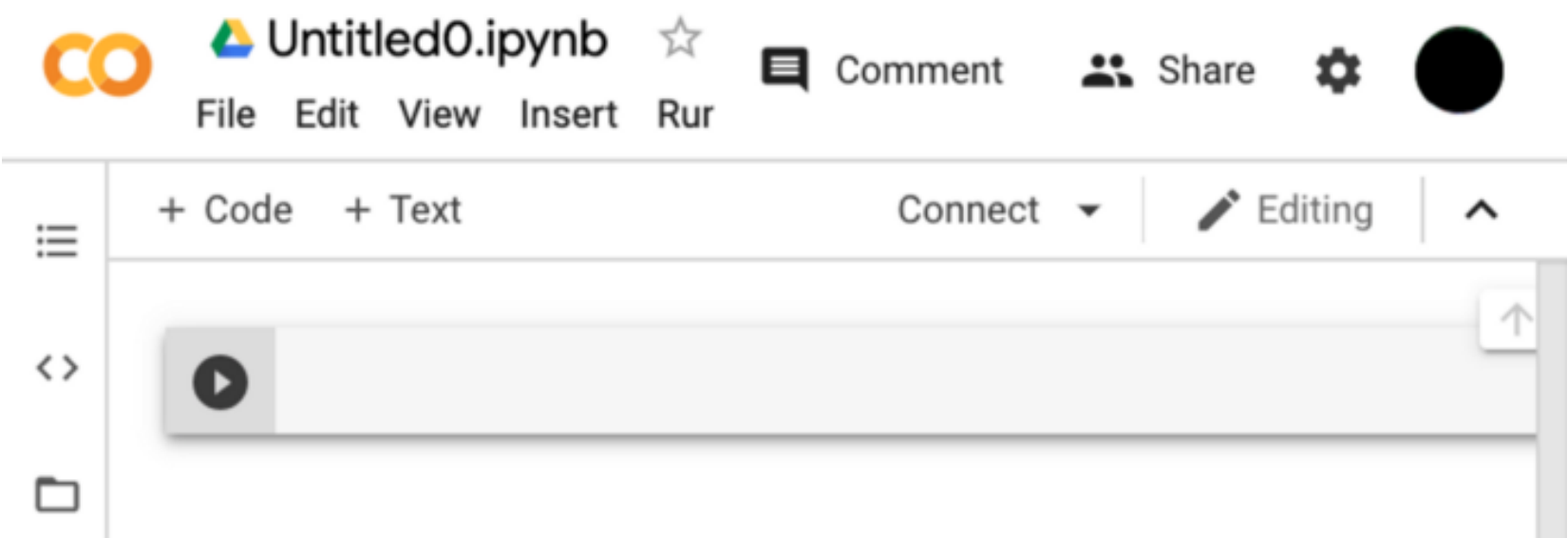

# چارچوب‌های یادگیری عمیق

توسعه یک شبکه عصبی عمیق و آماده‌سازی آن برای حل مشکلات، یک کار بسیار دشوار است. چراکه نیاز است تا قطعاتِ بسیار زیادی برای ایجاد و تنظیم یک جریان سیستماتیک در راستایِ دستیابی به اهدافی که با یادگیری عمیق قصد داریم بدست آوریم، کنار هم قرار گیرند. از این‌رو برای فعال‌کردنِ راه‌حل‌هایِ آسان‌تر، سریع‌تر و باکیفیت‌تر برای آزمایش‌ها و تحقیقات، محققین و یا دانشمندان داده، نیاز به یک چارچوب دارند. این چارچوب‌ها به محققان و توسعه‌دهندگان کمک می‌کند تا به جای سرمایه‌گذاریِ بیشترِ وقتِ خود بر روی عملیات‌های اساسی، برروی وظایفی که مهم‌تر هستند تمرکز کنند. چارچوب‌ها و پلتفرم‌های یادگیری عمیق، انتزاعی منصفانه بر روی وظایف پیچیده با توابع ساده ارائه می‌کنند که می‌توانند به عنوان ابزاری برای حل مشکلات بزرگ‌تر توسط محققان و توسعه‌دهندگان استفاده شوند.

## پای‌تورچ (PyTorch)

PyTorch یک محیط کاری یادگیری ماشین مبتنی‌بر Torch است که برای طراحی شبکه عصبی ایده‌آل است. PyTorch توسط آزمایشگاه تحقیقاتی هوش مصنوعی فیس‌بوک توسعه یافته و در ژانویه ۲۰۱۶ به عنوان یک کتابخانه رایگان و منبع‌باز منتشر شد و عمدتا در بینایی رایانه، یادگیری عمیق و برنامه‌های پردازش زبان طبیعی استفاده می‌شود و از توسعه نرم‌افزار مبتنی‌بر ابر پشتیبانی می‌کند. پیاده‌سازی یک شبکه عصبی در PyTorch نسبت به سایر محیط‌ها ساده‌تر و شهودی است. با پشتیبانی از CPU و GPU، شبکه‌های عصبی عمیق پیچیده را می‌توان با مجموعه داده‌های بزرگ آموزش داد.

### مزایا و معایب

**مزایا**

- یادگیری آسان
- انعطاف‌پذیر و سریع
- اشکال‌زدایی آسان

**معایب**

- عدم وجود ابزار مصورسازی مانند tensor board

## تنسورفلو (TensorFlow)

TensorFlow یکی از محبوب‌ترین محیط‌های کاری یادگیری ماشین و یادگیری عمیق است که توسط توسعه‌دهندگان و محققان استفاده می‌شود. TensorFlow در ابتدا در سال ۲۰۰۷ توسط تیم Google Brain راه‌اندازی شد و می‌تواند بر روی CPU و تسریع‌کننده‌های تخصصی هوش مصنوعی، از جمله GPU و TPU اجرا شود. TensorFlow در لینوکس ۶۴ بیتی، macOS، ویندوز و پلتفرم‌های محاسباتی موبایل، از جمله اندروید و iOS در دسترس است. مدل‌های آموزش دیده در TensorFlow را می‌توان بر روی دسکتاپ، مرورگرها و حتی میکروکنترلرها مستقر کرد. این پشتیبانی گسترده،

TensorFlow را منحصر به فرد و آماده تولید می‌کند. چه در حال کار با مسائل بینایی رایانه، پردازش زبان طبیعی یا مدل‌های سری زمانی باشید، TensorFlow یک پلتفرم یادگیری ماشین بالغ و قوی با قابلیت‌های زیاد است.

### مزایا و معایب

**مزایا**

- پشتیبانی عالی از گراف‌های محاسباتی، هم برای محاسبات و هم برای مصورسازی
- می‌توان TensorFlow را بر روی دسکتاپ، مرورگرها و حتی میکروکنترلرها مستقر کرد.

**معایب**

- منحنی یادگیری شیب دار به دلیل API های سطح پایین (یادگیری دشوار)
- درک برخی از پیام‌های خطا در TensorFlow می‌تواند بسیار دشوار باشد.

## کراس (Keras)

Keras یک رابط برنامه‌نویسی است که دانشمندان داده را قادر می‌سازد به‌راحتی به پلتفرم یادگیری عمیق TensorFlow دسترسی داشته باشند و از آن استفاده کنند. این یک رابط برنامه‌نویسی برنامه کاربردی (API) و محیط کاری یادگیری عمیق منبع‌باز است که در پایتون نوشته شده است که برروی TensorFlow اجرا می‌شود و اکنون در آن پلتفرم ادغام شده است. Keras قبلا از چندین پشتگاه (back end) پشتیبانی می‌کرد اما با شروع نسخه ۲/۴/۰ در ژوئن ۲۰۲۰ به‌طور انحصاری با TensorFlow مرتبط شده است. Keras به‌عنوان یک API سطح بالا، برای انجام آزمایش‌های آسان و سریع طراحی شده است که نسبت به سایر گزینه‌های یادگیری عمیق نیاز به کدنویسی کم‌تری دارد. هدف تسریع اجرای مدل‌های یادگیری ماشین، به ویژه، شبکه‌های عصبی عمیق، از طریق یک فرآیند توسعه با "سرعت تکرار بالا" است. مدل‌های Keras می‌توانند بر روی CPU یا GPU اجرا شوند و در چندین پلتفرم از جمله مرورگرهای وب و دستگاه‌های تلفن همراه Android و iOS مستقر شوند. Keras کندتر از TensorFlow و PyTorch است اما معماری ساده‌ای دارد و خواناتر، مختصرتر، کاربر پسند و قابل‌توسعه است. Keras بیشتر برای مجموعه داده‌های کوچک مناسب است و به دلیل طراحی ساده و قابل درک آن برای مبتدیان توصیه می‌شود.

### مزایا و معایب

**مزایا**

- API سطح بالا عالی
- یادگیری آسان
- تولید آسان مدل ها
- کاربرپسند
- نیازی به داشتن پیشینه قوی در یادگیری عمیق نیست.

**معایب**

- برای مجموعه داده های کوچک مناسب است
- گاهی اوقات در GPU کند است.

اگر مبتدی هستید یا مدل‌های ساده‌ای را برای مسئله امتحان می‌کنید، Keras بهترین گزینه برای شما است. چراکه شروع با آن کار آسان‌تر است، آنقدر آسان که برای آموزش یک شبکه عصبی عمیق نیازی به دانستن چیزی در مورد یادگیری عمیق ندارید!! PyTorch اگرچه از این نظر نیز نمره خوبی می‌گیرد (در مقایسه با تنسورفلو خالص) اما Keras بهتر است.

## خلاصه فصل

- داده‌ها به قطعاتِ متمایزِ اطلاعاتی اطلاق می‌شوند که معمولا به گونه‌ای قالب‌بندی و ذخیره می‌شوند که با هدف خاصی مطابقت داشته باشد.
- هر نقطه داده اغلب با یک بردار ویژگی نشان داده می‌شود، هر ورودی در بردار نشان‌دهنده یک ویژگی است.
- داده‌ها را می‌توان به عنوان قابل خواندن توسط ماشین، قابل خواندن توسط انسان یا هر دو دسته‌بندی کرد.
- یادگیری بانظارت یکی از پرکاربردترین شاخه‌های یادگیری ماشین است که از داده‌های آموزشی برچسب‌گذاری شده برای کمک به مدل‌ها در پیش‌بینی دقیق استفاده می‌کند.
- در طبقه‌بندی کلاس‌ها از قبل مشخص هستند و اغلب با عنوان هدف، برچسب یا دسته نامیده می‌شوند.

- رگرسیون یک فرآیند آماری است که رابطه معناداری بین متغیرهای وابسته و مستقل پیدا می‌کند و به عنوان یک الگوریتم، یک عدد پیوسته را پیش‌بینی می‌کند.
- یادگیری بدون‌نظارت در یادگیری ماشین زمانی است که به هیچ وجه دسته‌بندی یا برچسب‌گذاری داده‌ها وجود ندارد.

## آزمونک

۱. رویکردهای متفاوت یادگیری ماشین را نام ببرید، تفاوت‌ها، مزایا و معایب هر یک شرح دهید؟
۲. منظور از تعمیم‌دهی در یادگیری ماشین چیست؟
۳. تفاوت پارامتر و ابرپارامتر در چیست؟
۴. چرا برای آموزش یک الگوریتم یادگیری ماشین، داده‌ها تقسیم‌بندی می‌شوند؟
۵. آیا مدلی که در مجموعه‌یِ آموزشی به دقتی برابر ۹۸٪ دست یافته است ولی در مجموعه داده آزمون دقتی برابر ۷۹٪ دارد، می‌تواند مدل قابل قبولی باشد یا خیر؟ دلیل بیاورید.
۶. فرض کنید مدلی در یک مجموعه داده با دو کلاس متفاوت و به شدت نامتوازن دقتی برابر ۹۹٪ بدست آورده است، آیا تنها با براساس معیار دقت می‌توان گفت این مدل کارایی بسیار بالایی دارد؟ علت را شرح دهید.
۷. از مجموعه داده اعتبارسنجی به چه دلیل استفاده می‌شود؟
۸. بیش‌برازش به چه علت اتفاق می‌افتد؟
۹. واریانس بالا و سوگیری بالا چه چیزی را نشان می‌دهد؟

# ۳ شبکه‌های عصبی پیش‌خور

## اهداف یادگیری:

- آشنایی با پرسپترون
- آشنایی با شبکه عصبی پیش‌خور
- بهینه‌سازها
- توابع زیان
- پیاده‌سازی شبکه عصبی در **keras**

# مقدمه

در این فصل، به معرفی ساختار شبکه‌های عصبی می‌پردازیم، نحوه‌ی عملکرد یک نورون را به تفصیل شرح می‌دهیم و سپس فرآیند آموزش در شبکه‌های عصبی و مفاهیمی که در این زمینه وجود دارند را تشریح خواهیم کرد. این مفاهیم به عنوان پایه‌ای برای فصل‌های بعدی هستند.

## شبکه‌های عصبی مصنوعی (Artificial neural networks)

**شبکه‌های عصبی مصنوعی** یا به‌طور خلاصه **ANN** دسته‌ای از مدل‌های یادگیری ماشین هستند که به‌طور کلی از مطالعات مربوط به سیستم عصبی مرکزی پستانداران الهام گرفته شده‌اند. به عبارت دیگر، آن‌ها یک مدل محاسباتی هستند که نحوه‌ی عملکرد سلول‌های عصبی در مغز انسان را تقلید می‌کند. هر شبکه عصبی مصنوعی از چندین "**نورون**" مرتبط تشکیل شده است که در "**لایه‌ها**" سازماندهی شده‌اند. نورون‌های هر یک از لایه‌ها، پیام‌ها را به نورون‌های لایه بعدی می‌فرستند.

شبکه عصبی مصنوعی تلاشی برای شبیه‌سازی شبکه‌ای از نورون‌هایی است که مغز انسان را تشکیل می‌دهند تا رایانه بتواند توانایی یادگیری بدست آورد و به شیوه‌ای انسانی تصمیم بگیرد.

یک شبکه عصبی مصنوعی دارای یک لایه ورودی، یک لایه خروجی و یک یا چند لایه پنهان می‌باشد که بهم متصل هستند. لایه اول از نورون‌های ورودی تشکیل شده است. این نورون‌ها داده‌ها را به لایه‌های عمیق‌تر می‌فرستند. هر لایه بعد از لایه ورودی، به‌جای ورودی خام، خروجی لایه قبلی را به عنوان ورودی دریافت می‌کند. در نهایت، آخرین لایه خروجی مدل را تولید می‌کند. نمونه‌های آموزشی به‌طور مستقیم مشخص می‌کنند که برای هر ورودی $x$ چه خروجی باید تولید شود. لایه خروجی سعی در محاسبه مقداری دارد که نزدیک به خروجی مشخص برای نمونه‌های آموزشی مشابه باشد. با این حال، رفتار لایه‌های داخلی مستقیما تحت تاثیر نمونه‌های آموزشی نیستند و این الگوریتم آموزشی است که با تصمیم‌گیری‌های خود در جهت تولید خروجی موردنظر، چگونگی عملکرد این لایه‌ها را تعیین می‌کند. در نتیجه، عملکرد لایه‌های داخلی براساس خروجی مطلوبی که تحت نمونه‌های آموزشی بدست آورده است، به‌صورت واضح مشخص نیست و همانند یک جعبه سیاه عمل می‌کند، از این‌رو به این لایه‌ها، **لایه پنهان** گفته می‌شود.

وظیفه لایه‌های پنهان، تبدیل ورودی به چیزی است که لایه خروجی می‌تواند از آن استفاده کند. با افزایش تعداد لایه‌های پنهان، به سمت یک شبکه عمیق می‌رویم که توانایی حل مسائل پیچیده‌تر را نسبت به همتایان کم عمق خود دارا می‌باشد.

دانشمندان علوم اعصاب شناختی، از زمانی که دانشمندان رایانه برای اولین بار اقدام به ساخت شبکه عصبی مصنوعی اولیه کردند، اطلاعات بسیار زیادی در مورد مغز انسان آموخته‌اند. یکی از چیزهایی که آن‌ها آموختند این است که بخش‌های مختلف مغز مسئول پردازش جنبه‌های مختلف اطلاعات هستند و این بخش‌ها به صورت سلسله‌مراتبی مرتب شده‌اند. بنابراین، ورودی وارد مغز می‌شود و هر سطح از نورون‌ها بینش را ارائه می‌دهد و سپس اطلاعات به سطح بالاتر بعدی منتقل می‌شود. این دقیقا مکانیزمی است که ANNها سعی در تکرار آن دارند.

شبکه‌های عصبی مصنوعی به دلیل تطبیق‌پذیر بودن قابل توجه هستند، به این معنی که با یادگیری از داده‌ها، خود را اصلاح می‌کنند و در اجراهای بعدی اطلاعات بیشتری کسب می‌کنند.

برای اینکه شبکه‌های عصبی مصنوعی بتواند توانایی یادگیری بدست آورند، باید حجم عظیمی از داده‌ها را در اختیار داشته باشند که **مجموعه آموزشی** نامیده می‌شود. هنگامی که می‌خواهید به شبکه‌های عصبی مصنوعی بیاموزید که چگونه یک گربه را از سگ تشخیص دهد، مجموعه آموزشی هزاران تصویر با برچسب سگ ارائه می‌کند تا شبکه شروع به یادگیری کند. هنگامی که با حجم قابل توجهی از داده‌ها آموزش داده شد، سعی می‌کند داده‌های آینده را بر اساس آنچه که فکر می‌کند می‌بیند (یا می‌شنود، بسته به مجموعه داده‌ها) در دسته‌های مختلف طبقه‌بندی کند. در طول دوره آموزش، خروجی ماشین با توضیحات ارائه شده توسط انسان (برچسب‌ها) از آنچه باید مشاهده شود مقایسه می‌شود. اگر آن‌ها یکسان باشند، مدل به خوبی کارآیی خوبی دارد. اگر نادرست باشد، از **پس‌انتشار** (**backpropagation**) برای تنظیم یادگیری خود استفاده می‌کند.

## پرسپترون (perceptron)

**نورون** عنصر اساسی در هر شبکه عصبی مصنوعی است. ساده‌ترین نوع مدل‌سازی یک نورون را **پرسپترون** گویند که می‌تواند دارای تعداد زیادی ورودی تنها با یک خروجی باشد. در شکل ۱ـ۳ شمایی از پرسپترون رسم شده است. پرسپترون از یادگیری بانظارت برای طبقه‌بندی یا پیش‌بینی خروجی استفاده می‌کند. یک پرسپترون تک لایه با ترسیم یک **مرز تصمیم‌گیری** (**decision boundary**) با استفاده از یک خط جداساز، داده‌ها را طبقه‌بندی می‌کند.

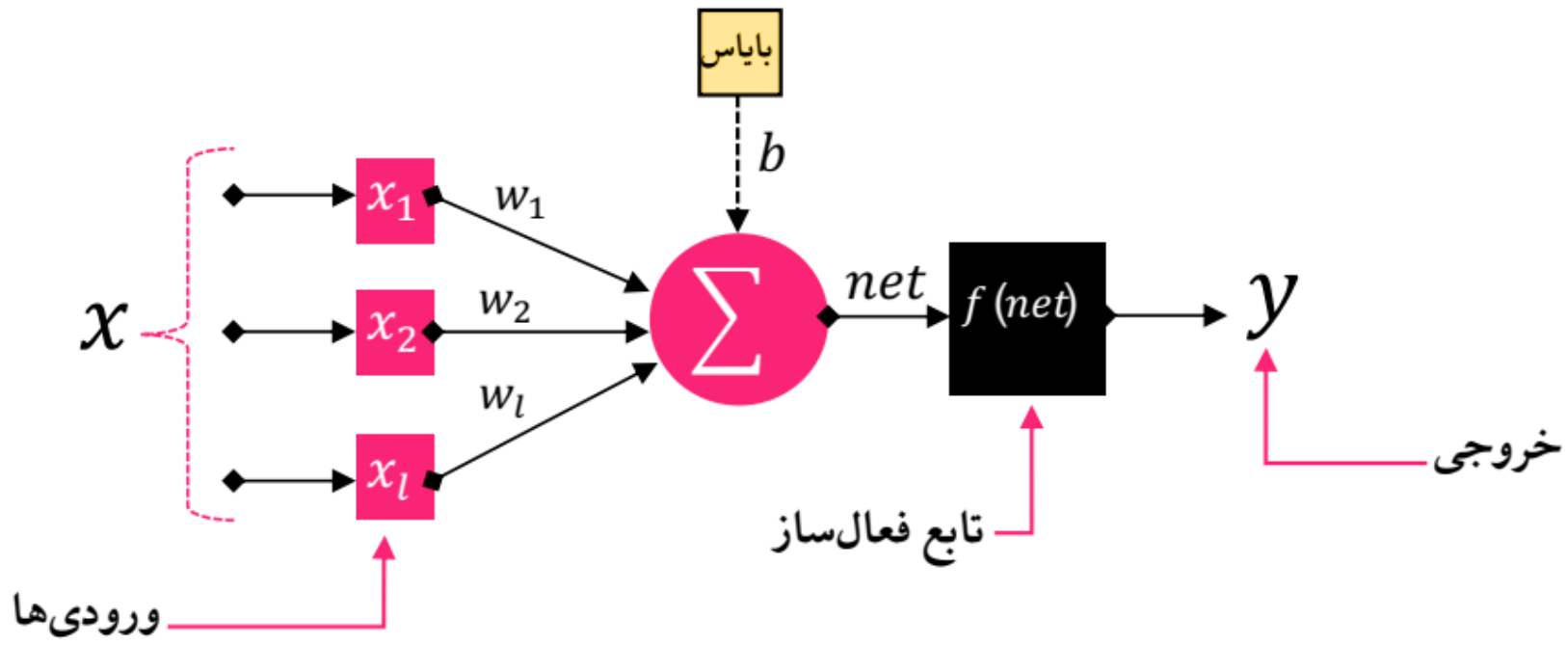

**شکل ۳ـ۱. پرسپترون**

بیایید نگاهی به نحوه عملکرد پرسپترون بیندازیم. پرسپترون با گرفتن برخی ورودی‌های عددی همراه با آنچه به عنوان **وزن‌ها** (**weights**) و **سوگیری** (**bias**) شناخته می‌شود، کار می‌کند. سپس این ورودی‌ها را با وزن‌های مربوط ضرب می‌کند (که به عنوان **مجموع وزنی** شناخته می‌شود). سپس این حاصل‌ضرب همراه با سوگیری بهم اضافه می‌شوند. **تابع فعال‌سازی** ( **Activation Function**) مجموع وزنی و بایاس را به عنوان ورودی می‌گیرد و خروجی نهایی را برمی‌گرداند. **گیج کننده بود!!!!**... بیایید پرسپترون را تجزیه کنیم، تا نحوه‌یِ کار آن را بهترکنیم. یک پرسپترون (شکل ۳ـ۱) از چهار بخش اصلی تشکیل شده است: **مقادیر ورودی**، **وزن‌ها و بایاس**، **مجموع وزنی** و **تابع فعال‌سازی**. فرض کنید یک نورون و سه ورودی $x_1$، $x_2$، $x_3$ داریم که به ترتیب در وزن‌های $w_1$، $w_2$، $w_3$ ضرب می‌شوند:

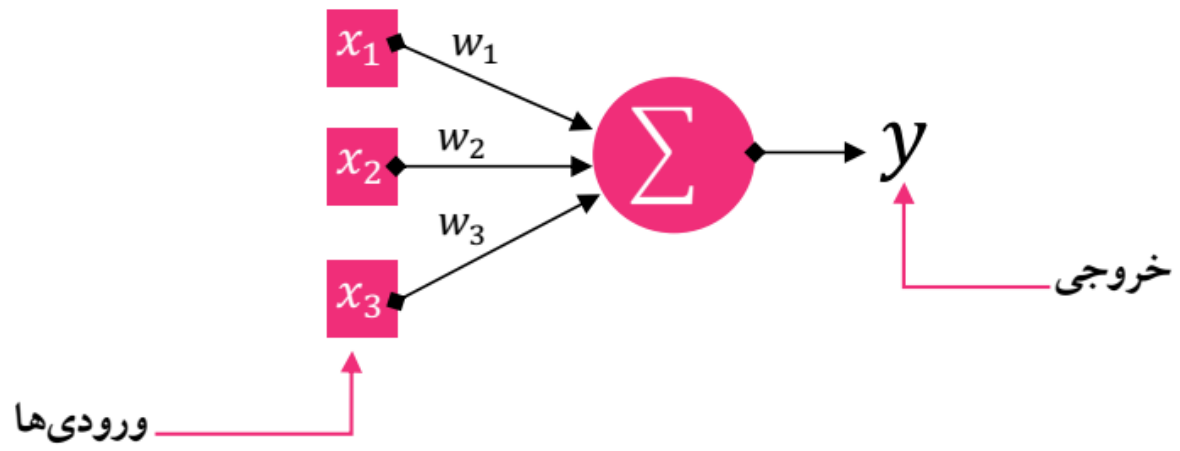

ایده ساده است، با توجه به مقدار عددی ورودی‌ها و وزن‌ها، تابعی در داخل نورون وجود دارد که یک خروجی تولید می‌کند. **حال سوال این است که این تابع چیست؟** این تابع این‌گونه عمل می‌کند:

$$y = x_1w_1 + x_2w_2 + x_3w_3$$

این تابع را **مجموع وزنی** می‌نامند، چراکه مجموع وزن‌ها و ورودی‌ها است. تا این‌جا همه چیز خوب به نظر می‌رسد، با این حال، اگر بخواهیم خروجی‌ها در محدوده خاصی قرار گیرند، مثلاً

۰ تا ۱ چه کاری باید کرد؟! ما می‌توانیم این کار را با استفاده از چیزی به نام **تابع فعال‌سازی** انجام دهیم. یک تابع فعال‌سازی، تابعی است که ورودی داده شده (در این مورد، ورودی مجموع وزنی خواهد بود) را به یک خروجی مشخص، بر اساس مجموعه‌ای از قوانین تبدیل می‌کند:

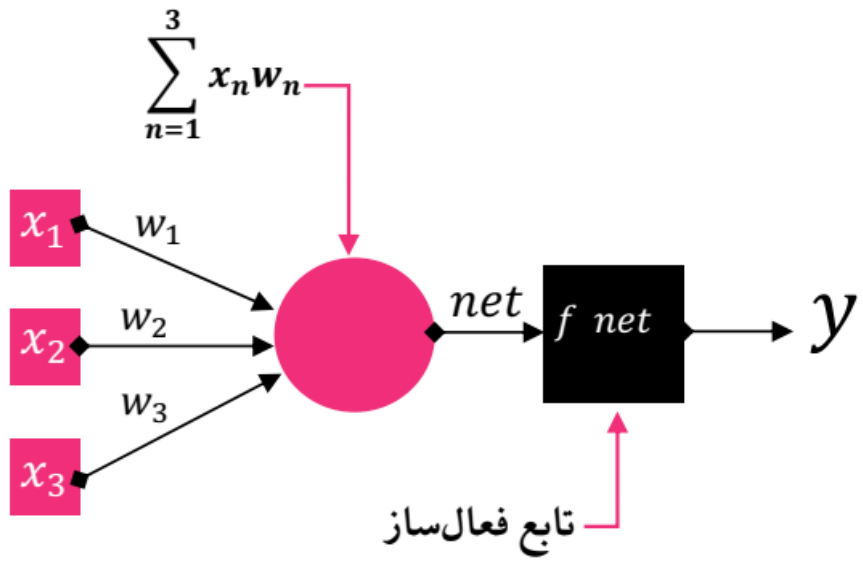

اکنون تقریبا همه‌یِ چیزهایی که برای ساختن پرسپترون نیاز است را در اختیار داریم. آخرین چیزی که سوگیری است. سوگیری، یک پارامتر اضافی در شبکه عصبی است که برای تنظیم خروجی به همراه مجموع وزنی ورودی‌های نورون استفاده می‌شود. علاوه بر این، مقدار سوگیری به این امکان را می‌دهد تا تابع فعال‌سازی به راست یا چپ تغییر پیدا کند. یک راه ساده تر برای درک سوگیری از طریق ثابت $c$ یک تابع خطی است:

$$y = mx + c$$

این به شما امکان می‌دهد خط را به پایین و بالا ببرید تا پیش‌بینی را با داده‌ها بهتر تطبیق دهید. اگر ثابت $c$ وجود نداشته باشد، خط از مبدأ (۰، ۰) عبور می‌کند و شما تطبیق ضعیف‌تری خواهید داشت. از این‌رو، سوگیری‌ها اجازه می‌دهند تا تغییرات بیشتر و بیشتری از وزن‌ها یاد گرفته شوند. به‌طور خلاصه، تغییرات بیشتر به این معنی است که سوگیری‌ها بازنمایی غنی‌تری از فضای ورودی را به وزن‌های آموخته شده مدل اضافه می‌کنند. بنابراین، معادله نهایی نورون به این صورت محاسبه می‌شود:

$$\text{خروجی} = \sum(\text{ورودی} * \text{وزن}) + \text{سوگیری}$$

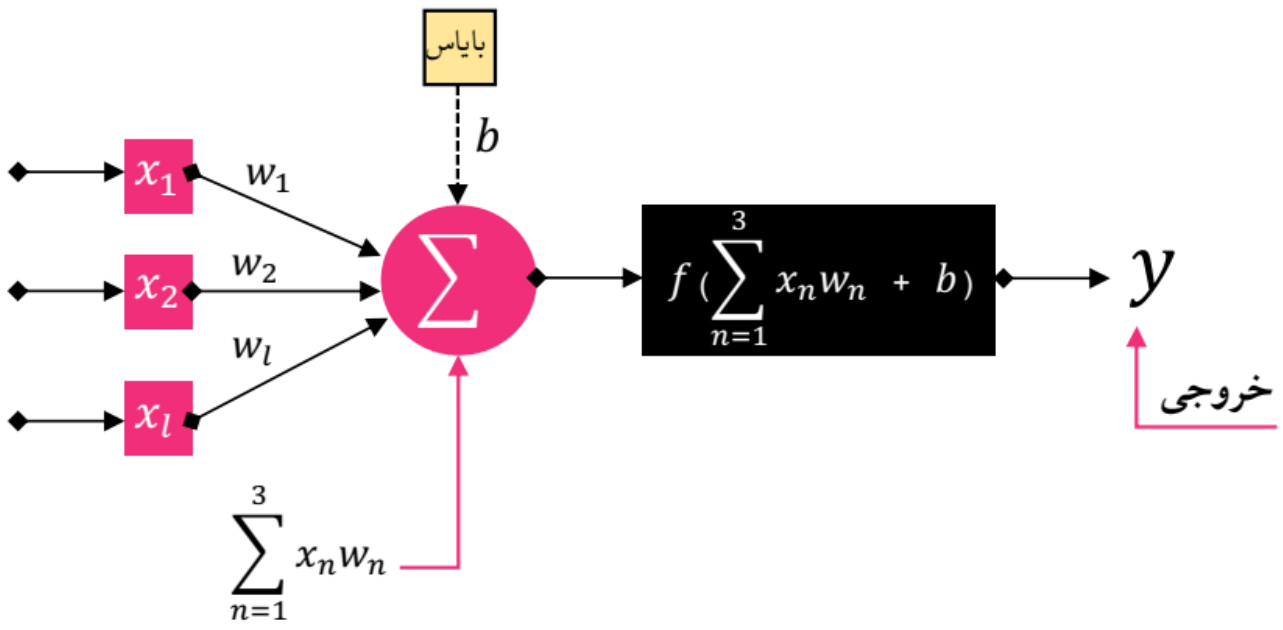

همان‌طور که پیش‌تر بیان شد، از پرسپترون برای طبقه‌بندی دودویی استفاده می‌شود. بیایید یک پرسپترون ساده را در نظر بگیریم و با یک مثال ساده نحوه‌ی کار آن را در طبقه‌بندی دودویی داده‌ها بهتر درک کنیم. در این پرسپترون دو ورودی $x_1$ و $x_2$ داریم که به ترتیب با وزن های $w_1$ و $w_2$ ضرب می‌شود و همچنین دارای یک بایاس است:

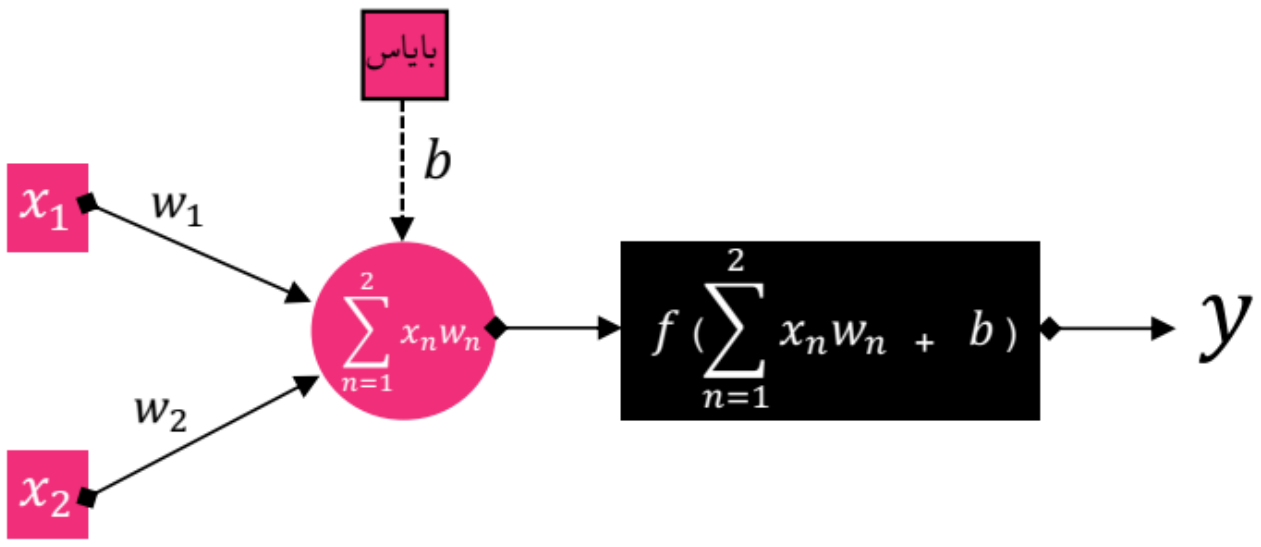

بیایید همچنین یک نمودار با دو دسته مختلف داده ایجاد کنیم که با اشکال دایره و مستطیل نشان داده شده‌اند:

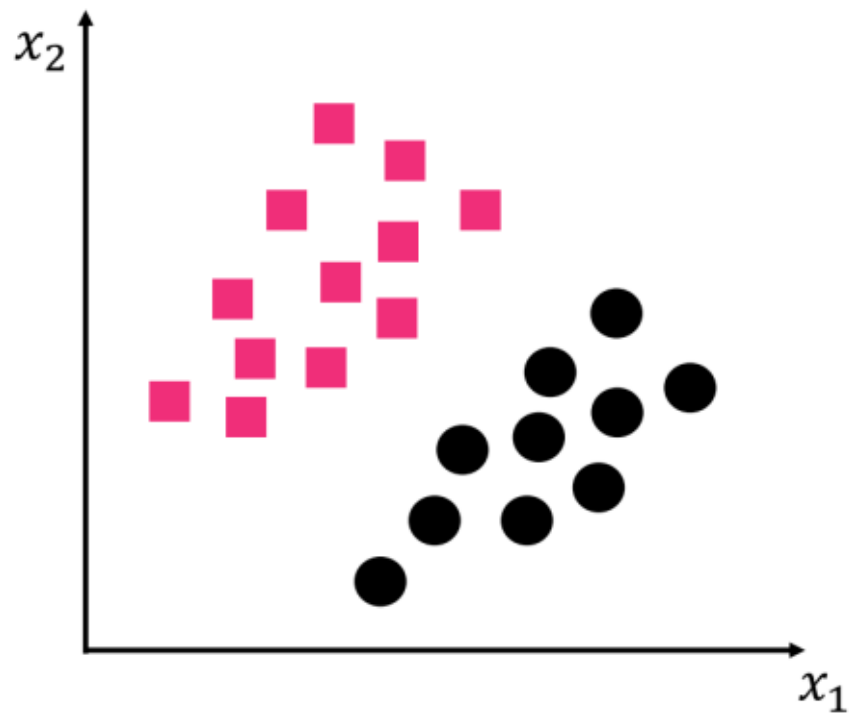

فرض کنید هدف ما این بود که این داده‌ها را جدا کنیم تا بین دایره و مستطیل تمایز وجود داشته باشد. یک پرسپترون می‌تواند یک مرز تصمیم برای یک طبقه‌بند دودویی ایجاد کند، جایی که مرز تصمیم مناطقی از فضا روی یک نمودار است که نقاط داده‌های مختلف را از هم جدا می‌کند. برای درک بهتر این موضوع، بیایید کمی با تابع بازی کنیم. می‌توانیم بگوییم:

$w_1 = 0.5$
$w_2 = 0.5$
$b = 0$

بر این اساس، تابع پرسپترون به این صورت خواهد بود:

$$0.5x_1 + 0.5x_2 = 0$$

و نمودار آن به صورت زیر خواهد بود:

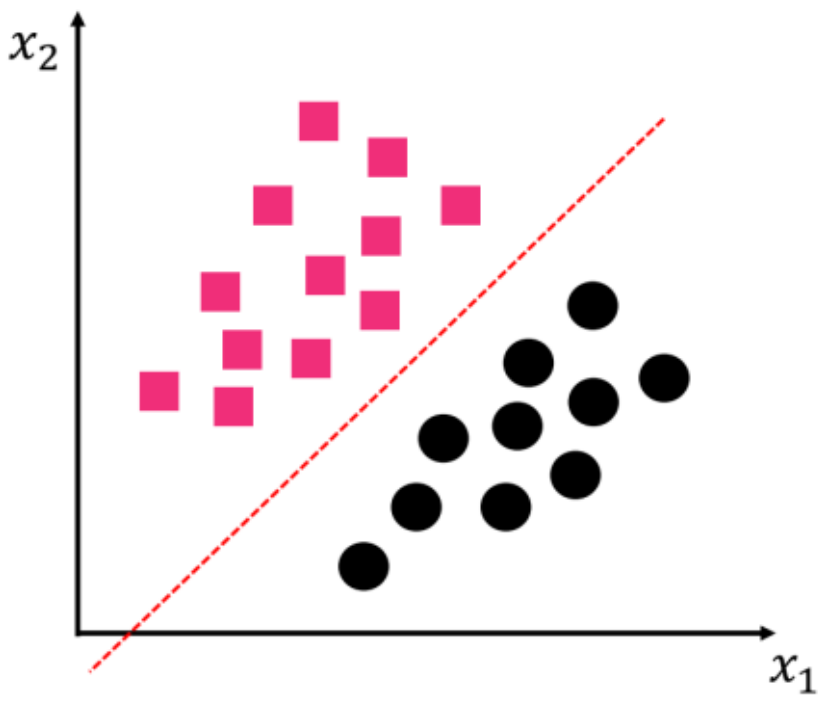

فرض کنید تابع فعال‌سازی، در این مورد، یک **تابع پله‌ای (step function)** ساده است که ۰ یا ۱ را خروجی می‌دهد. سپس تابع پرسپترون اشکال مستطیل را ۱ و اشکال دایره را با ۰ نشان می‌دهد. به عبارت دیگر:

$$\begin{cases} 1, & if\ 0.5x_1 + 0.5x_2 \geq 0 \\ 0, & if\ 0.5x_1 + 0.5x_2 < 0 \end{cases}$$

بنابراین، تابع $0.5x_1 + 0.5x_2 = 0$ یک مرز تصمیم ایجاد می‌کند که اشکال مستطیل و دایره را از هم جدا می‌کند.

## الگوریتم یادگیری پرسپترون

یادگیری پرسپترون یک عملیات نسبتا ساده است. هدف ما بدست آوردن مجموعه‌ای از وزن‌های $w$ است که به طور دقیق هر نمونه را در مجموعه آموزشی ما طبقه‌بندی می‌کند. به منظور آموزش پرسپترون، ما به طور مکرر شبکه را با داده‌های آموزشی خود چندین مرتبه تغذیه می‌کنیم. هر بار که شبکه مجموعه کاملی از داده‌های آموزشی را دید، می‌گوییم یک **دوره (epoch)** گذشته است. دوره پارامتری است که توسط کاربر، قبل از آموزش تعیین می‌شود.

شبه کد الگوریتم یادگیری پرسپترون (الگوریتم ۱.۳) را می‌توان به صورت خلاصه کرد:

"**یادگیری**" واقعی در مراحل (۲.ب) و (۲.ج) صورت می گیرد. ابتدا بردار ویژگی $x_j$ را از شبکه عبور می‌دهیم، حاصل‌ضرب داخلی با وزن‌های $w$ را می‌گیریم و خروجی $y_j$ را بدست می‌آوریم. سپس، این مقدار از تابع پله عبور داده می‌شود که اگر $x > 0$ باشد ۱ و در غیر این صورت ۰ برگردانده می‌شود. اکنون باید بردار وزن خود را بروز کنیم تا در جهتی قدم برداریم که نزدیک‌تر به طبقه‌بندی درست داده‌ها است. این عمل با **بروز رسانی بردار وزن توسط قانون دلتا** در مرحله (۲.ج) مدیریت می‌شود.

عبارت $(d_j - y_j)$ تعیین می‌کند که آیا طبقه‌بندی خروجی درست است یا نه. اگر طبقه‌بندی درست باشد، این اختلاف صفر خواهد بود. در غیر این صورت، تفاوت یا مثبت یا منفی خواهد بود و به ما جهتی می‌دهد که وزن‌های در آن بروز می‌شوند (در نهایت ما را به طبقه‌بندی درست نزدیک می‌کند). سپس $(d_j - y_j)$ را در $x_j$ ضرب می‌کنیم که ما را به طبقه‌بندی درست نزدیک می‌کند. مقدار $\alpha$ **نرخ یادگیری** (**learning rate**) ما است و میزان بزرگی (یا کوچکی) یک گام را کنترل می‌کند. *بسیار مهم است که این مقدار بدرستی تنظیم شود.* هرچند مقدار بزرگ $\alpha$ باعث می‌شود که گامی در جهت درست برداریم، با این حال، می‌تواند براحتی ما را از بهینه محلی یا سراسری عبور دهد. درمقابل، مقدار کمی $\alpha$ به ما اجازه می‌دهد گام‌های کوچک را در جهت درست برداریم و تضمین می‌کند که از بهینه محلی یا سراسری تجاوز نمی‌کنیم. با این حال، این گام‌های کوچک ممکن است زمان زیادی طول بکشد تا یادگیری ما همگرا شود. در نهایت، بردار وزن قبلی را در زمان $t$، $w_j(t)$ اضافه می‌کنیم که فرآیند گام برداشتن به سمت طبقه‌بندی درست را کامل می‌کند. اگر این روش آموزشی را کمی گیج کننده می‌بینید، نگران نباشید.

فرآیند یادگیری پرسپترون تا زمانی که تمام نمونه‌های آموزشی بدرستی طبقه‌بندی شوند یا به تعداد از پیش تعیین‌شده (توسط کاربر) دوره‌ها برسد، اجازه داده می‌شود تا ادامه یابد. اگر $\alpha$ به اندازه کافی کوچک باشد و داده‌های آموزشی به صورت خطی قابل تفکیک باشند، خاتمه تضمین می‌شود. به عبارت دیگر، با داشتن فرضیات مناسب می‌توان نشان داد یادگیری در پرسپترون با تکرار الگوریتم آن، به وزن‌های درست همگرا خواهد شد. یعنی یادگیری شبکه منجر به تخمین وزن‌هایی خواهد شد که شبکه را قادر می‌سازد تا مقادیر درست را در خروجی تولید کند.

**الگوریتم ۱.۳ الگوریتم یادگیری پرسپترون**

۱. بردار وزن $w$ خود را با مقادیر تصادفی کوچک مقداردهی اولیه کنید.
۲. تا زمانی که پرسپترون همگرا شود:
   أ. یک حلقه برروی هر بردار ویژگی $x_j$ و برچسب کلاس واقعی $d_j$ در مجموعه آموزشی بزنید.
   ب. $x$ را بگیرید و آن را از طریق شبکه عبور دهید و مقدار خروجی را محاسبه کنید:

$$y_j = f\big(w(t).x_j\big)$$

   ج. وزن‌های $w$ را به‌روزرسانی کنید:

$$w: w_i(t+1) = w_i(t) + \alpha\big(d_j - y_j\big)x_{j,i}$$

حال پرسش اینجاست، اگر داده‌های ما به صورت خطی قابل تفکیک نباشند یا انتخاب ضعیفی در $\alpha$ داشته باشیم، چه اتفاقی می‌افتد؟ آیا آموزش به صورت بی‌نهایت ادامه خواهد داشت؟ در این مورد، خیر. معمولا ما بعد از اینکه تعداد دوره‌های معینی انجام شد، متوقف می‌شویم یا اگر

تعداد طبقه‌بندی‌های اشتباه در تعداد زیادی از دوره‌ها تغییر نکرده باشد (که نشان می‌دهد داده‌ها به صورت خطی قابل تفکیک نیستند).

## پیاده‌سازی پرسپترون در پایتون

اکنون که الگوریتم پرسپترون را مطالعه کردیم، بیایید الگوریتم آن را در پایتون پیاده‌سازی کنیم (این پیاده‌سازی تنها برای این است که با عملکرد پرسپترون و روند آموزشی که در کتابخانه‌ها وجود دارد آشنایی پیدا کنید. از این‌رو اگر این موارد برای شما کمی مشکل به نظر می‌رسد، نگران نباشید، چراکه در ادامه تنها از کتابخانه‌ها و چارچوب‌ها استفاده می‌شود و نیازی به کد زدن این موارد نیست). در ابتدا کد زیر را وارد کنید:

```
# import the necessary packages
import numpy as np

class Perceptron:
  def __init__(self, N, alpha=0.1):
    # initialize the weight matrix and store the learning rate
    self.W = np.random.randn(N + 1) / np.sqrt(N)
    self.alpha = alpha
```

خط ۵ سازنده کلاس Perceptron ما را تعریف می‌کند، که یک پارامتر مورد نیاز و سپس یک پارامتر اختیاری را می پذیرد:

- N: تعداد ستون‌ها در بردارهای ویژگی ورودی ما است.
- alpha: نرخ یادگیری ما برای الگوریتم پرسپترون است. این مقدار را به‌طور پیش‌فرض برروی 0.1 قرار می‌دهیم.

در خط ۷ ماتریس وزن $W$ با مقادیر تصادفی از توزیع نرمال (گاوسی) با میانگین صفر و واریانس واحد نمونه‌برداری شده است. ماتریس وزن دارای N +1 ورودی است، یکی برای هر یک از N ورودی در بردار ویژگی، به علاوه یک ورودی برای سوگیری. W را بر ریشه مربع تعداد ورودی‌ها تقسیم می‌کنیم، تکنیک رایجی که برای مقیاس‌بندی ماتریس وزن که منجر به همگرایی سریع‌تر می‌شود.

سپس، بیایید تابع پله را تعریف کنیم:

```
  def step(self, x):
    return 1 if x > 0 else 0
```

برای آموزش پرسپترون، تابعی به نام fit تعریف می‌کنیم. اگر تجربه قبلی با یادگیری ماشین و کتابخانه scikit-learn داشته باشید، می‌دانید که نامگذاری این تابع برای آموزش با این نام معمول است:

```
def fit(self, X, y, epochs=10):
  X = np.c_[X, np.ones((X.shape[0]))]
```

متد fit به دو پارامتر الزامی و به دنبال آن یک پارامتر اختیاری نیاز دارد: مقدار X داده‌های آموزشی ما و متغیر y برچسب‌های کلاس خروجی هدف ما هستند (یعنی آنچه شبکه ما باید پیش‌بینی کند). در نهایت، پارامتر دوره را داریم که تعداد دوره‌هایی که Perceptron آموزش خواهد یافت. خط آخر کد، سوگیری را با درج ستونی از یک‌ها در داده‌های آموزشی اعمال می‌کند که به ما امکان می‌دهد بایاس را به عنوان یک پارامتر قابل آموزش مستقیما در داخل ماتریس وزن در نظر بگیریم. حال، بیایید روند آموزش واقعی را مرور کنیم:

```
  # loop over the desired number of epochs
  for epoch in np.arange(0, epochs):
    # loop over each individual data point
    for (x, target) in zip(X, y):
      # take the dot product between the input features
      # and the weight matrix, then pass this value
      # through the step function to obtain the prediction
      p = self.step(np.dot(x, self.W))
      # only perform a weight update if our prediction
      # does not match the target
      if p != target:
        # determine the error
        error = p - target
        # update the weight matrix
        self.W += -self.alpha * error * x
```

در ابتدا از یک حلقه استفاده می‌کنیم و آن را به تعداد دوره اجرا می‌کنیم. برای هر دوره، همچنین بر روی هر نقطه داده جداگانه $x$ و برچسب کلاس هدف خروجی از حلقه استفاده می‌کنیم. سپس، حاصل‌ضرب داخلی بین ویژگی‌های ورودی $x$ و ماتریس وزن $W$ گرفته می‌شود تا خروجی را از تابع پله عبور دهد و پیش‌بینی توسط پرسپترون بدست آید. تنها در صورتی بروزرسانی وزن را انجام می‌دهیم که پیش‌بینی ما با هدف مطابقت نداشته باشد. اگر این‌طور باشد، خطا را با محاسبه علامت (مثبت یا منفی) تعیین می‌کنیم.

بروز رسانی ماتریس وزن در خط آخر کد انجام می‌شود، جایی که ما یک گام به سمت طبقه‌بندی صحیح برمیداریم. در طی یک تعداد از دوره‌ها، پرسپترون ما قادر است الگوهایی را در داده‌های زیربنایی بیاموزد و مقادیر ماتریس وزن را طوری تغییر دهد که نمونه‌های ورودی خود را بدرستی طبقه‌بندی کنیم.

آخرین تابعی که باید تعریف کنیم، predict است و همانطور که از نام آن پیداست، برای پیش‌بینی برچسب‌های کلاس برای مجموعه داده‌های ورودی استفاده می‌شود:

```
def predict(self, X, addBias=True):
  X = np.atleast_2d(X)
  if addBias:
```

```
        X = np.c_[X, np.ones((X.shape[0]))]
        return self.step(np.dot(X, self.W))
```

متد predict ما به مجموعه‌ای از داده‌های ورودی X نیاز دارد که باید طبقه‌بندی شوند. همچنین یک بررسی در کد انجام می‌شود تا ببیند آیا یک ستون بایاس باید اضافه شود یا خیر.

اکنون که Perceptron خود را پیاده‌سازی کرده‌ایم، بیایید سعی کنیم آن را در مجموعه داده‌های بیتی (AND، OR و XOR) اعمال کنیم و ببینیم که چگونه کار می‌کند.

ابتدا در مجموعه داده OR آن را آزمایش می‌کنیم. برای این کار کد زیر را وارد کنید:

```
# construct the OR dataset
X = np.array([[0, 0], [0, 1], [1, 0], [1, 1]])
y = np.array([[0], [1], [1], [1]])
# define our perceptron and train it
print("[INFO] training perceptron...")
p = Perceptron(X.shape[1], alpha=0.1)
p.fit(X, y, epochs=20)
```

در خطوط ۲ و ۳ مجموعه داده OR را تعریف می‌کنیم. خطوط ۶ و ۷ پرسپترون ما را با نرخ یادگیری $\alpha = 0.1$ در ۲۰ دوره آموزش می‌دهند.

بعد از آموزش Perceptron خود، باید آن را روی داده‌ها ارزیابی کنیم تا تایید کنیم که در واقع تابع OR را یاد گرفته است:

```
# now that our perceptron is trained we can evaluate it
print("[INFO] testing perceptron...")
# now that our network is trained, loop over the data points
for (x, target) in zip(X, y):
  # make a prediction on the data point and display the result
  pred = p.predict(x)
  print("[INFO] data={}, ground-truth={}, pred={}".format(
    x, target[0], pred))
```

کد نهایی به‌صورت زیر است:

```
import numpy as np
class Perceptron:
  def __init__(self, N, alpha=0.1):
    self.W = np.random.randn(N + 1) / np.sqrt(N)
    self.alpha = alpha

  def step(self, x):
    return 1 if x > 0 else 0

  def fit(self, X, y, epochs=10):
    X = np.c_[X, np.ones((X.shape[0]))]
    for epoch in np.arange(0, epochs):
      for (x, target) in zip(X, y):
        p = self.step(np.dot(x, self.W))
        if p != target:
          error = p - target
```

```
            self.W += -self.alpha * error * x
    def predict(self, X, addBias=True):
      X = np.atleast_2d(X)
      if addBias:
        X = np.c_[X, np.ones((X.shape[0]))]
      return self.step(np.dot(X, self.W))

X = np.array([[0, 0], [0, 1], [1, 0], [1, 1]])
y = np.array([[0], [1], [1], [1]])
# define our perceptron and train it
print("[INFO] training perceptron...")
p = Perceptron(X.shape[1], alpha=0.1)
p.fit(X, y, epochs=20)

print("[INFO] testing perceptron...")

for (x, target) in zip(X, y):
  pred = p.predict(x)
  print("[INFO] data={}, ground-truth={}, pred={}".format(
    x, target[0], pred))
```

بعد از اجرای کد بالا، خروجی به صورت زیر نمایش داده می‌شود:

```
[INFO] training perceptron...
[INFO] testing perceptron...
[INFO] data=[0 0], ground-truth=0, pred=0
[INFO] data=[0 1], ground-truth=1, pred=1
[INFO] data=[1 0], ground-truth=1, pred=1
[INFO] data=[1 1], ground-truth=1, pred=1
```

همان‌طور که مشاهده می‌شود، پرسپترون ما توانست عملیات OR را یاد بگیرد. عملگر OR برای $x_0 = 0$ و $x_1 = 0$ صفر است و همه ترکیبات دیگر یک هستند.

حالا بیایید به سراغ تابع AND برویم، کد زیر را وارد کنید:

```
X = np.array([[0, 0], [0, 1], [1, 0], [1, 1]])
y = np.array([[0], [0], [0], [1]])
# define our perceptron and train it
print("[INFO] training perceptron...")
p = Perceptron(X.shape[1], alpha=0.1)
p.fit(X, y, epochs=20)
# now that our perceptron is trained we can evaluate it
print("[INFO] testing perceptron...")
# now that our network is trained, loop over the data points
for (x, target) in zip(X, y):
  # make a prediction on the data point and display the result
  # to our console
  pred = p.predict(x)
  print("[INFO] data={}, ground-truth={}, pred={}".format(
    x, target[0], pred))
```

توجه کنید که در اینجا تنها خطوط کدی که تغییر کرده‌اند، خطوط ۱ و ۲ هستند که در آن مجموعه داده AND را به جای مجموعه داده OR تعریف کرده‌ایم.

بعد از اجرای کد پیشین، خروجی به صورت زیر نمایش داده می‌شود:

```
[INFO] training perceptron...
[INFO] testing perceptron...
[INFO] data=[0 0], ground-truth=0, pred=0
[INFO] data=[0 1], ground-truth=0, pred=0
[INFO] data=[1 0], ground-truth=0, pred=0
[INFO] data=[1 1], ground-truth=1, pred=1
```

مشاهده شد که دوباره Perceptron ما توانست بدرستی تابع AND را مدل کند. عملگر AND فقط زمانی درست است که هم $x_0 = 1$ و $x_1 = 1$ باشد و برای همه ترکیب‌های دیگر AND صفر است.

در نهایت، اجازه دهید تا نگاهی به تابع غیرخطی XOR با پرسپترون بیندازیم. کد زیر را وارد کنید:

```
X = np.array([[0, 0], [0, 1], [1, 0], [1, 1]])
y = np.array([[0], [1], [1], [0]])
# define our perceptron and train it
print("[INFO] training perceptron...")
p = Perceptron(X.shape[1], alpha=0.1)
p.fit(X, y, epochs=20)
# now that our perceptron is trained we can evaluate it
print("[INFO] testing perceptron...")
# now that our network is trained, loop over the data points
for (x, target) in zip(X, y):
  # make a prediction on the data point and display the result
  pred = p.predict(x)
  print("[INFO] data={}, ground-truth={}, pred={}".format(
    x, target[0], pred))
```

با اجرای کد بالا خروجی به صورت بدست آمد:

```
[INFO] training perceptron...
[INFO] testing perceptron...
[INFO] data=[0 0], ground-truth=0, pred=1
[INFO] data=[0 1], ground-truth=1, pred=1
[INFO] data=[1 0], ground-truth=1, pred=0
[INFO] data=[1 1], ground-truth=0, pred=0
```

بیاید دوباره کد بالا را اجرا کنیم. خروجی این بار به صورت زیر بدست آمد:

```
[INFO] training perceptron...
[INFO] testing perceptron...
[INFO] data=[0 0], ground-truth=0, pred=0
[INFO] data=[0 1], ground-truth=1, pred=0
[INFO] data=[1 0], ground-truth=1, pred=0
[INFO] data=[1 1], ground-truth=0, pred=1
```

مهم نیست چند بار این آزمایش را با نرخ‌های یادگیری متفاوت یا روش‌های مقداردهی اولیه متفاوت اجرا کنید، چراکه هرگز نمی‌توانید تابع XOR را با پرسپترون تک لایه مدل‌سازی کنید.

در عوض، آنچه ما نیاز داریم، تعداد لایه‌های بیشتر با توابع فعال‌سازی غیرخطی است.

پرسپترون تنها یک طبقه‌بند خطی است و هرگز نمی‌تواند داده‌هایی را که به صورت خطی قابل تفکیک نیستند را از یکدیگر جدا کند. همچنین، این الگوریتم فقط برای مسائل طبقه‌بندی دودیی استفاده می‌شود.

## پرسپترون چند لایه (شبکه عصبی پیش‌خور)

همان‌طور که بیان شد، محدودیت اصلی شبکه‌های عصبی پرسپترون، عدم توانایی در طبقه‌بندی داده‌هایی است که جدایی‌پذیر خطی نیستند. استفاده از یک لایه پنهان در ساختار شبکه‌ها گریزی است بر این محدودیت. به عبارت دیگر، در راستای حل این محدودیت، می‌توان از لایه پنهان بین لایه ورودی و خروجی استفاده کرد. نمونه‌ای از این شبکه‌ها که اساس یادگیری عمیق نیز می‌باشد، شبکه‌های **پرسپترون چندلایه** (**multilayer perceptron**) یا به اختصار **MLP** هستند که همچنین از آن‌ها با عنوان **شبکه‌های عصبی پیش‌خور** ( **feed forward neural network**) نام برده می‌شود. این شبکه‌ها از پرکاربردترین‌ها شبکه‌ها در یادگیری عمیق به دلیل سازگاری آن با انواع مسائل هستند. چرا که برای ورودی آن هیچ محدودیتی وجود ندارد که داده‌ها تصویر، متن و یا ویدیو باشد.

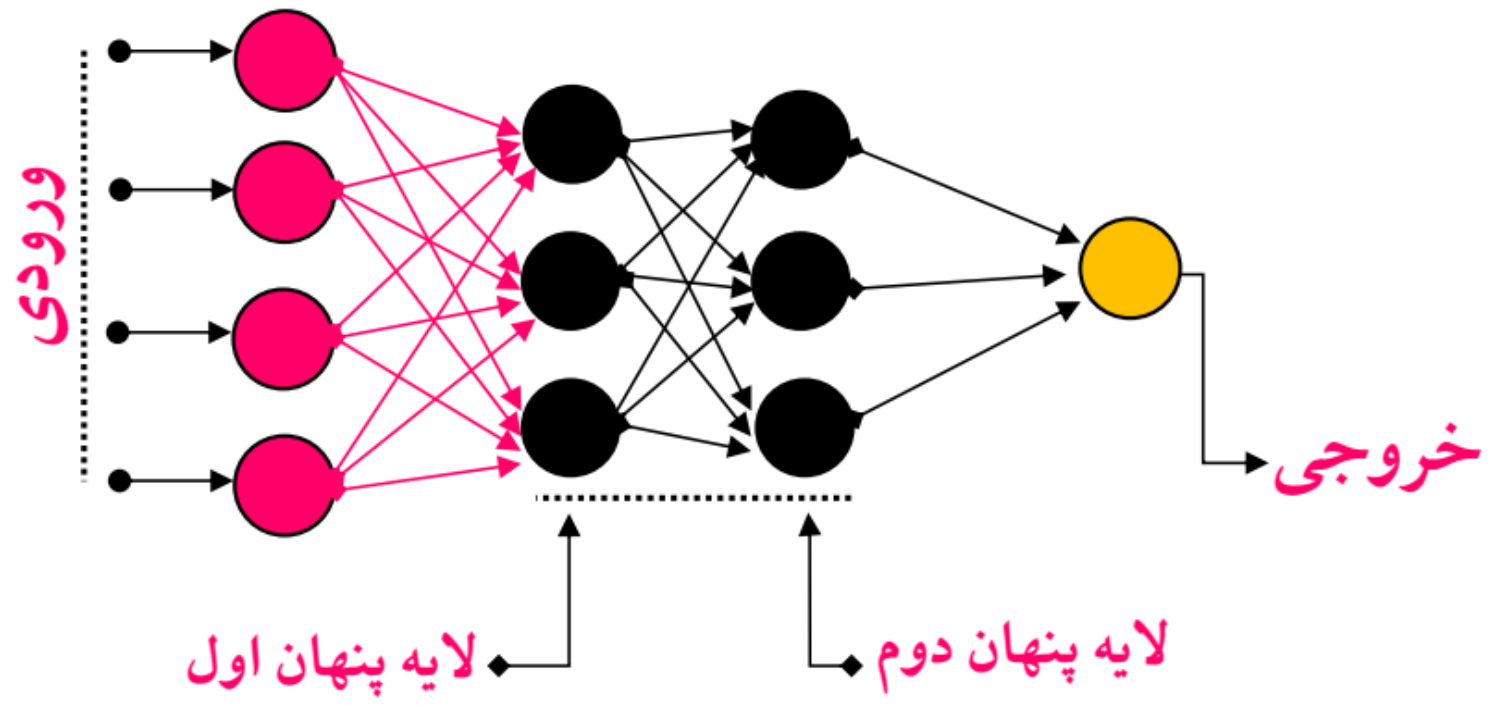

شکل ۳ـ۲. یک شبکه عصبی با دو لایه پنهان

در یک MLP، داده‌ها در جهت روبه‌جلو از لایه ورودی به خروجی جریان می‌یابند. در این نوع از شبکه‌ها با رفتن از هر لایه به لایه دیگر، جمع وزن‌دار مجموعه نورون‌های لایه قبل محاسبه و با اعمال یک تابع فعال‌ساز غیرخطی به لایه دیگر منتقل می‌شوند. دلیل نام‌گذاری آن به **پیش‌خور** (**Feed forward**) عدم وجود اتصال بازخوردی است که از طریق آن خروجی‌های مدل دوباره به عنوان ورودی خود مدل محسوب شوند (مقادیر فقط از ورودی به لایه‌های پنهان و سپس به خروجی می‌روند و هیچ مقداری به لایه‌های قبلی بازگردانده نمی‌شود. در مقابل، یک

شبکه بازگشتی اجازه می‌دهد مقادیر به عقب برگردانده شوند). به عبارت دیگر، در یک شبکه پیش‌خور، فعال‌سازی‌ها در شبکه، همیشه از طریق دنباله‌ای از لایه‌ها به جلو جریان می‌یابند. این شبکه همچنین یک شبکه **کاملاً متصل** (**fully connected**) است، چراکه هر یک از نورون‌های شبکه به گونه‌ای بهم متصل شده‌اند که ورودی‌ها را از تمام نورون‌های لایه قبلی دریافت می‌کند و فعال‌سازی خروجی خود را به تمام نورون‌های لایه بعدی منتقل می‌کند. در شکل ۳ــ۲ شمایی از یک شبکه عصبی با ۲ لایه پنهان قابل مشاهده است.

هنگامی که این شبکه در حال پردازش مجموعه‌ای از ورودی‌های خارجی است، ورودی‌ها از طریق نورون‌های حسگر در لایه ورودی به شبکه ارائه می‌شوند. این باعث می‌شود که نورون‌های لایه بعدی سیگنال‌های فعال‌سازی را در پاسخ به این ورودی‌ها تولید کنند. و این فعال‌سازی‌ها از طریق شبکه جریان می‌یابند تا به لایه خروجی برسند. فعال‌سازی نورون‌های این لایه، پاسخ شبکه به ورودی‌ها و خروجی نهایی است. لایه‌های داخلی شبکه که نه لایه ورودی هستند و نه لایه خروجی، **لایه‌های پنهان** نامیده می‌شوند.

**عمق** یک شبکه عصبی برابر است با تعداد لایه‌های پنهان به اضافه لایه خروجی است. بنابراین، شبکه در شکل ۳ــ۲ دارای سه لایه است. تعداد لایه‌های مورد نیاز برای در نظر گرفتن عمق یک شبکه یک سوال باز است. با این حال، ثابت شده است که یک شبکه با سه لایه نورون (یعنی دو لایه پنهان و یک لایه خروجی) می‌تواند هر تابعی را به دقت دلخواه تقریب بزند. بنابراین، در اینجا حداقل تعداد لایه‌های پنهان لازم برای عمیق در نظر گرفتن یک شبکه را به عنوان دو تعریف می‌کنیم. تحت این تعریف، شبکه در شکل ۳ــ۲ به عنوان یک شبکه عمیق توصیف می‌شود. با این حال، بیشتر شبکه‌های عمیق بیش از دو لایه پنهان دارند. امروزه برخی از شبکه‌های عمیق ده‌ها یا حتی صدها لایه دارند.

هر شبکه عصبی پیش‌خور از ورودی‌ها، تعداد دلخواه لایه‌ها پنهان و لایه‌ای که خروجی‌ها را محاسبه می‌کند به نام لایه خروجی تشکیل می‌شود. این رویکرد مبتنی‌بر لایه، جایی است که نام یادگیری عمیق از آن گرفته شده است، چراکه، عمق یک شبکه عصبی پیش‌خور، تعداد لایه‌هایی را که یک شبکه عصبی پیش‌خور از آن تشکیل شده است را توصیف می‌کند.

نورون‌ها در MLP با الگوریتم یادگیری پس‌انتشار آموزش داده می‌شوند. MLPها به عنوان **تقریب‌گرهای جامع** (**universal approximators**) عمل می‌کنند. به عبارت دیگر، آن‌ها می‌توانند هر تابع پیوسته‌ای را تقریب بزنند و می‌توانند مسائلی که به صورت خطی قابل تفکیک نیستند را حل کنند.

نشان داده شده است که شبکه‌هایِ عصبِ پیش‌خور تنها با یک لایه‌یِ پنهان، می‌توانند برای تقریب هر تابع پیوسته مورد استفاده قرار گیرند.

### مسائل مرتبط با طراحی و آموزش شبکه‌های عصبی

هدف از فرآیند یادگیری در شبکه‌های عصبی، یافتن مجموعه‌ای از مقادیر وزنی است که باعث می‌شود خروجی شبکه عصبی تا حد امکان با مقادیر هدف واقعی مطابقت داشته باشد. مسائل مختلفی در طراحی و آموزش شبکه پرسپترون چندلایه وجود دارد:

- **انتخاب تعداد لایه‌های پنهان برای استفاده در شبکه.**
- **تصمیم‌گیری برای استفاده از چند نورون در هر لایه پنهان.** تعداد نورون‌ها در لایه(های) پنهان یکی از تصمیم‌گیری‌های مهم در طراحی یک شبکه عصبی است. اگر به تعداد کافی نورون استفاده نشود، شبکه قادر به مدل‌سازی داده‌های پیچیده نخواهد بود و تطابق حاصل ضعیف خواهد بود. اگر تعداد زیادی نورون استفاده شود، زمان آموزش ممکن است بیش از حد طولانی شود و بدتر از آن، شبکه ممکن است منجر به بیش‌برازش شود. هنگامی که بیش‌برازش اتفاق می‌افتد، شبکه شروع به مدل‌سازی نویز تصادفی در داده‌ها می‌کند. نتیجه این است که مدل به خوبی با داده‌های آموزشی مطابقت دارد، اما به‌طور ضعیفی به داده‌های جدید و دیده نشده تعمیم می‌یابد. برای آزمایش این مورد باید از اعتبارسنجی استفاده شود.
- **یافتن یک راه‌حل بهینه سراسری که از کمینه‌های محلی اجتناب کند.** یک شبکه عصبی معمولی ممکن است صدها وزن داشته باشد که مقادیر آن‌ها باید برای تولید یک راه حل بهینه پیدا شود. اگر شبکه‌های عصبی مدل‌های خطی (مانند رگرسیون خطی) باشند، یافتن مجموعه بهینه وزن‌ها کار سختی نیست. اما خروجی یک شبکه عصبی به عنوان تابعی از ورودی‌ها اغلب بسیار غیرخطی است. این امر فرآیند بهینه‌سازی را بسیار پیچیده می‌کند. اگر خطا را به عنوان تابعی از وزن‌ها ترسیم کنید، احتمالا یک سطح ناهموار با کمینه‌های محلی زیادی مانند زیر مشاهده خواهید کرد:

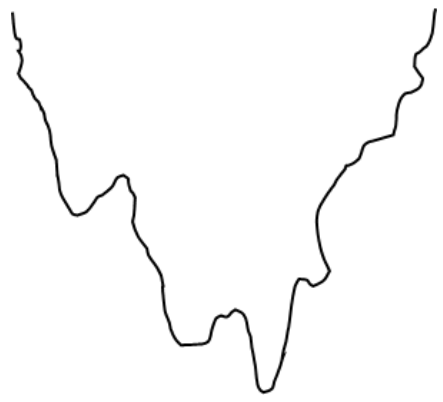

این تصویر بسیار ساده شده است زیرا تنها مقدار یک وزن را نشان می‌دهد (در محور افقی).

- **همگرایی به یک راه‌حل بهینه در یک دوره زمانی معقول.**
- **اعتبارسنجی شبکه عصبی برای آزمایش بیش‌برازش.**

## تابع فعال‌سازی

تابع فعال‌سازی از اهمیت زیادی در یادگیری عمیق برخوردار است. هدف توابع فعال‌سازی دریافت عددی از ورودی و محاسبه یک سری عملیات ریاضی است تا خروجی بین بازه ۰ تا ۱ یا ۱ ـ تا ۱ را تولید کند. تابع فعال‌سازی در هر نورون مصنوعی اگر سیگنال‌های دریافتی به حد آستانه رسیده باشند، سیگنال‌های خروجی را برای سطح بعدی ارسال می‌کنند. به بیان خیلی ساده تابع فعال‌ساز تصمیم می‌گیرد که یک نرون باید فعال شود یا خیر.

در یادگیری عمیق، یک شبکه عصبی بدون تابع فعال‌سازی فقط یک مدل رگرسیون خطی ساده است!؟ چراکه این توابع در واقع محاسبات غیرخطی را در ورودی یک شبکه عصبی انجام می‌دهند و آن را قادر به یادگیری و انجام وظایف پیچیده‌تر می‌کنند. بنابراین، مطالعه انواع مختلف و تجزیه و تحلیل مزایا و معایب هر تابع فعال‌سازی، برای انتخاب نوع مناسب تابع فعال‌سازی که بتواند غیرخطی بودن و دقت را در یک مدل شبکه عصبی خاص ارائه دهد، بسیار ضروری است. همچنین، به دلیل مشکل **محو گرادیان (Vanishing gradient)** که بعدا در مورد آن صحبت خواهیم کرد، تنظیم تابع فعال‌سازی مناسب برای شبکه بسیار مهم است.

تابع فعال‌سازی مورد استفاده در شبکه‌های عمیق، نمی‌تواند از هر تابعی باشد، بلکه باید ویژگی‌های خاصی را در خود به همراه داشته باشد. یکی از ویژگی‌های مهم یک تابع فعال‌سازی این است که باید **مشتق‌پذیر** باشد. شبکه از خطاهایی که در لایه خروجی محاسبه می‌شود، یاد می‌گیرد. یک تابع فعال‌سازی مشتق‌پذیر برای انجام بهینه‌سازی پس‌انتشار در حالی که **انتشار عقب‌گرد (propagating backwards)** را برای محاسبه گرادیان‌های خطا با توجه به وزن‌ها و سپس بهینه‌سازی وزن‌ها با استفاده از گرادیان کاهشی انجام می‌دهد (یا هر تکنیک بهینه‌سازی دیگری برای کاهش خطا)، مورد نیاز است.

نقش تابع فعال‌سازی استخراج خروجی از مجموعه مقادیر ورودی است که به یک نورون (یا یک لایه) داده می‌شود.

### چرا توابع فعال‌سازی غیرخطی ضروری هستند؟

تابع فعال‌سازی یک مرحله اضافی را در هر لایه در طول **انتشار جلورو (forward propagation)** معرفی می‌کنند، اما محاسبه آن ارزشش را دارد. فرض کنید یک شبکه عصبی داریم که بدون توابع فعال‌سازی کار می‌کند. در آن صورت، هر نورون تنها با استفاده از وزن‌ها و بایاس‌ها، یک تبدیل خطی روی ورودی‌ها انجام می‌دهد. از این‌رو دیگر اهمیتی ندارد که از چند لایه پنهان در شبکه عصبی خود استفاده کنید چراکه همه لایه‌ها به یک شکل رفتار می‌کنند به این علت که ترکیب دو تابع خطی خود یک تابع خطی است و شبکه قدرتی بیشتر از یک رگرسیون خطی را نخواهد داشت.

فرض کنید یک بردار ورودی x و سه لایه پنهان دارید که با ماتریس‌های وزن $W_1$، $W_2$ و $W_3$ نشان داده شده‌اند. بدون هیچ تابع فعال‌سازی، شبکه عصبی شما خروجی y=x $W_1W_2W_3$ را دارد که برابر است با $y=xW$ به طوری که $W = W_1W_2W_3$ و این چیزی نیست جز ضرب ماتریس. حال، با معرفی یک تابع فعال‌سازی غیرخطی پس از هر تبدیل خطی، دیگر این اتفاق نمی‌افتد:

$$y = f_1\left(W_1 f_2(W_2 f_3(W_3 x))\right)$$

اکنون هر لایه می‌تواند بر روی نتایج لایه غیرخطی قبلی ایجاد شود که اساسا منجر به یک تابع غیرخطی پیچیده می‌شود.

یک تابع فعال‌سازی به شبکه عصبی مصنوعی اضافه می‌شود تا به شبکه کمک کند الگوهای پیچیده در داده‌ها را یاد بگیرد.

## ویژگی‌های مطلوب یک تابع فعال‌سازی

هم‌چنان که پیش‌تر بیان گردید، تابع فعال‌سازی در شبکه‌های عصبی باید ویژگی‌های خاصی را در خود به همراه داشته باشد. ویژگی‌های مطلوبی که یک تابع فعال‌سازی باید داشته باشد را می‌توان به صورت زیر خلاصه کرد:

- **غیرخطی** (**Nonlinear**): همان‌طور که پیش‌تر بیان شد، اگر تابع فعال‌سازی خطی باشد، یک پرسپترون با چندین لایه پنهان را می‌توان براحتی به یک پرسپترون تک لایه فشرده کرد، چراکه ترکیب خطی از بردار ورودی را می‌توان به سادگی به عنوان یک ترکیب خطی منفرد از بردار ورودی بیان کرد. در این صورت عمق شبکه تاثیری نخواهد داشت. بنابراین، غیرخطی بودن در تابع فعال‌سازی زمانی که مرز تصمیم ماهیت غیرخطی داشته باشد، ضروری است. از آنجایی که یک شبکه عصبی مصنوعی الگوها یا مرز را از داده‌ها یاد می‌گیرد، غیرخطی بودن در تابع فعال‌سازی ضروری است تا شبکه عصبی مصنوعی بتواند هر مرز خطی یا غیرخطی را براحتی یاد بگیرد.
- **صفرـمرکز** (**Zero-Centered**): خروجی تابع فعال‌سازی باید صفرـمرکز باشد تا گرادیان‌ها به جهت خاصی تغییر نکنند. زمانی به تابعی صفرـمرکز گویند که محدوده آن دارای مقادیر مثبت و منفی باشد. اگر تابع فعال‌سازی شبکه صفرـمرکز نباشد، همیشه مثبت یا همیشه منفی است. بنابراین، خروجی یک لایه همیشه به مقادیر مثبت یا منفی منتقل می‌شود. در نتیجه، بردار وزن نیاز به بروز رسانی بیشتری دارد تا بدرستی آموزش داده شود. بنابراین، اگر تابع فعال‌سازی در صفرـمرکز نباشد، تعداد دوره‌های مورد نیاز برای آموزش شبکه افزایش می‌یابد. به همین دلیل است که ویژگی صفرـمرکز مهم است، اگرچه ضروری نیست.

- **هزینه محاسباتی (Computational Expense):** توابع فعال‌سازی بعد از هر لایه اعمال می‌شوند و باید میلیون‌ها بار در شبکه‌های عمیق محاسبه شوند. بنابراین، محاسبه آن‌ها باید از نظر محاسباتی ارزان باشد.
- **مشتق‌پذیر (Differentiable):** شبکه‌های عصبی با استفاده از فرآیند گرادیان کاهشی آموزش داده می‌شوند، بنابراین لازم است تابع فعال‌سازی با توجه به ورودی مشتق‌پذیر باشد. این یک نیاز ضروری برای عملکرد یک تابع فعال‌سازی است.
- **پیوسته (Continuous):** یک تابع نمی‌تواند مشتق‌پذیر شود مگر اینکه پیوسته باشد.
- **کران‌دار (Bounded):** داده‌های ورودی از طریق یک سری پرسپترون که هر کدام حاوی یک تابع فعال‌سازی هستند، منتقل می‌شود. در نتیجه، اگر تابع در یک کران محدود نباشد، مقدار خروجی ممکن است منفجر (explode) شود. برای کنترل این انفجارِ مقادیر، ماهیت کران‌دار تابع فعال‌سازی مهم است اما ضروری نیست.

## مشکلاتی که توابع فعال‌سازی با آن مواجه هستند

**مشکل محو گرادیان (Vanishing Gradient problem)** و مشکل **نورون مرده (dead neuron)** دو مشکل عمده‌ای هستند که توابع فعال‌سازی پرکاربرد با آن مواجه هستند:

- **مشکل محو گرادیان:** شبکه‌های عصبی با استفاده از گرادیان کاهشی و الگوریتم پس‌انتشار آموزش داده می‌شوند. هنگام استفاده از الگوریتم پس‌انتشار، در فاز عقب‌گرد محاسبه گرادیان کوچک و کوچک‌تر می‌شود. این اتفاق به این دلیل بوجود می‌آید که گرادیان کاهشی در هر تکرار مشتقات جزئی را با طی کردن از لایه پایانی به سمت لایه ابتدایی با استفاده از قانون زنجیره‌ای می‌یابد. در شبکه‌ای با داشتن $n$ لایه پنهان، مشتقات این $n$ لایه در یکدیگر ضرب می‌شود. حال اگر این مشتقات کوچک باشند، با رفتن به لایه‌های اولیه به صورت نمایی کاهش پیدا و (یا در بدترین حالت صفر می‌شوند و یادگیری شبکه متوقف می‌شود) همین امر سبب پدیده **محو گرادیان** می‌شود. از آنجایی که این گرادیان‌های کوچک در تکرار الگوریتم بروزرسانی نمی‌شوند و این لایه‌های اولیه اغلب در شناخت داده‌ها موثر هستند، منجر به عدم دقت کافی شبکه می‌شوند و این لایه‌ها نمی‌توانند به درستی یاد بگیرند. به عبارت دیگر، گرادیان آن‌ها به دلیل عمق شبکه و فعال‌سازی که مقدار را به صفر می‌برد، از بین می‌روند. *از این‌رو، ما می‌خواهیم تابع فعال‌سازی گرادیان را به سمت صفر تغییر ندهد.* در مقابل محو گرادیان، **انفجار گرادیان (Exploding Gradients)** وجود دارد. اگر مقادیر گرادیان‌ها بزرگ باشند، منجر به بروز رسانی‌های بسیار بزرگ وزن مدل شبکه عصبی در طول آموزش می‌شوند. با رشد نمایی

از طریق انتقال به لایه‌ها، این گرادیان‌های بزرگ در نهایت سبب سرریز شده و وزن‌ها دیگر توانایی بروزرسانی نخواهند داشت و شبکه‌ای ناپایداری را پدید می‌آورند.

- **نورون مرده:** وقتی یک تابع فعال‌سازی بخش بزرگی از ورودی را به صفر یا تقریبا صفر وادار می‌کند، آن نورون‌های متناظر برای کمک به خروجی نهایی غیرفعال (مرده) هستند. در حین بروز رسانی وزن‌ها، این احتمال وجود دارد که وزن‌ها به گونه‌ای بروز شوند که مجموع وزنی بخش بزرگی از شبکه به اجبار صفر شود. یک شبکه به سختی از چنین وضعیتی بهبود می‌یابد و بخش بزرگی از ورودی نمی‌تواند به شبکه کمک کند. این منجر به یک مشکل می‌شود زیرا ممکن است بخش بزرگی از ورودی در طول اجرای شبکه به طور کامل غیرفعال شود. این نورون‌هایی که به‌شدت غیرفعال می‌شوند، "نورون‌های مرده" نامیده می‌شوند و این مشکل به عنوان مشکل **نورون مرده** نامیده می‌شود. به‌طور خلاصه، نورون مرده در اصطلاح شبکه عصبی مصنوعی به نورونی گفته می‌شود که در حین آموزش فعال نمی‌شود. این امر باعث می‌شود که نورون نتواند وزن خود را بروز کند زیرا مشتقات آن وزن‌های مربوط بسیار کوچک یا صفر خواهد بود. خطاها از طریق یک نورون مرده نیز منتشر نمی‌شوند، بنابراین روی دیگر نورون‌های شبکه تأثیر می‌گذارند.

## توابع فعال‌سازی پرکاربرد

در این بخش به پرکاربردترین توابع فعال‌سازی و ویژگی‌های آن‌ها می‌پردازیم.

### Sigmoid

تابع فعال‌ساز Sigmoid به‌صورت زیر تعریف می‌شود:

$$\sigma(x) = \frac{1}{1+e^{-x}}$$

که در آن $x$ ورودی تابع فعال‌ساز است. حال بیایید این را در پایتون کدنویسی کنیم:

```
def sigmoid(x):
  return 1/(1+np.exp(-x))
```

تابع Sigmoid پیوسته و در محدوده (۰،۱) کران‌دار و مشتق‌پذیر است اما صفر–مرکز نیست. این تابع هر مقدار حقیقی (real) را به عنوان ورودی می‌گیرد و مقادیری را در محدوده ۰ تا ۱ در خروجی می‌دهد. هرچه ورودی بزرگ‌تر باشد (مثبت بیشتر)، مقدار خروجی به ۱/۰ نزدیکتر خواهد شد، در حالی که هرچه ورودی کوچکتر (منفی‌تر) باشد، خروجی به ۰/۰ نزدیکتر خواهد بود.

مشتق تابع Sigmoid $\sigma(x)$، تابع Sigmoid $\sigma(x)$ ضرب در $1-\sigma(x)$ است:

$$\acute{\sigma}(x) = \sigma(x).(1-\sigma(x))$$

که در پایتون می‌توانیم آن را به‌صورت زیر کدنویسی کنیم:

```
def der_sigmoid(x):
  return sigmoid(x) * (1- sigmoid(x))
```

حال بیاید تابع Sigmoid و مشتق آن را مصورسازی کنیم. برای این کار کد زیر را وارد کنید:

```
import numpy as np
import matplotlib.pyplot as plt

# Sigmoid Activation Function
def sigmoid(x):
  return 1/(1+np.exp(-x))

# Derivative of Sigmoid
def der_sigmoid(x):
  return sigmoid(x) * (1- sigmoid(x))

# Generating data to plot
x_data = np.linspace(-10,10,100)
y_data = sigmoid(x_data)
dy_data = der_sigmoid(x_data)

# Plotting
plt.plot(x_data, y_data, x_data, dy_data)
plt.title('Sigmoid Activation Function & Derivative')
plt.legend(['sigmoid','der_sigmoid'])
plt.grid()
plt.show()
```

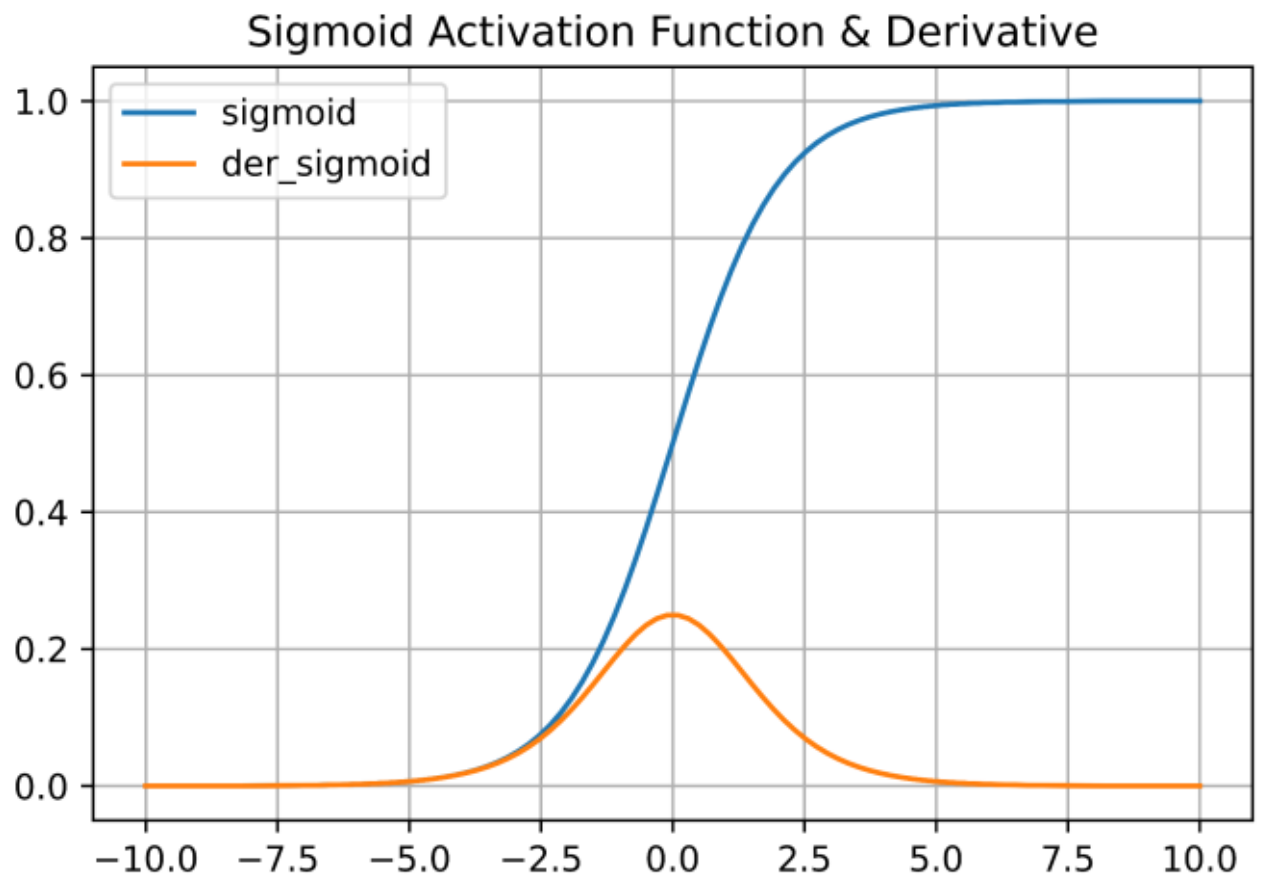

تابع Sigmoid از برخی اشکالات عمده رنج می‌برد. تابع Sigmoid در محدوده (۰،۱) محدود می‌شود. از این رو همیشه یک مقدار غیرمنفی به عنوان خروجی تولید می‌کند. بنابراین یک تابع فعال‌سازی صفر_مرکز نیست. تابع Sigmoid دامنه بزرگی از ورودی را به محدوده کوچکی (۰،۱) نگاشت می‌کند. بنابراین، یک تغییر بزرگ در مقدار ورودی منجر به تغییر

کوچک در مقدار خروجی می‌شود. این منجر به مقادیر گرادیان کوچک نیز می‌شود. به دلیل مقادیر کم گرادیان، با مشکل محو گرادیان مواجه می‌شود.

در کاربردهای عملی، تابع فعال‌ساز Sigmoid علیرغم محبوبیت آن در گذشته، به دلیل دو مشکل مهم کم‌تر مورد استفاده قرار می‌گیرد:

۱. **مشکل محو گرادیان دارد.** تابع Sigmoid منجر به مشکل محو گرادیان در الگوریتم پس‌انتشار می‌شود. در این حالت هیچ سیگنالی از طریق نورون‌ها منتقل نمی‌شود و بنابراین نورون در مرحله آموزش چیزی یاد نخواهد گرفت.
۲. **صفر-مرکز نیست.** خروجی‌های تابع Sigmoid صفر-مرکز نیستند. از این‌رو در الگوریتم پس‌انتشار، گرادیان‌هایی ایجاد می‌شود که یا همه مثبت و یا همه منفی هستند که برای بروز رسانی گرادیان وزن‌ها مناسب نیست.

از تابع فعال‌ساز Sigmoid به طور کلی برای مسائل طبقه‌بندی دودویی و طبقه‌بندی چندبرچسبی در لایه خروجی استفاده می‌شود چراکه باید احتمال را به عنوان خروجی پیش‌بینی کنیم. از آنجایی که احتمال هر چیزی فقط بین محدوده ۰ و ۱ است، Sigmoid به دلیل دامنه آن انتخاب مناسبی است.

**مزایا**

- محدوده خروجی آن از ۰ تا ۱ است، از این‌رو می‌تواند احتمالات را ایجاد کند. این باعث می‌شود که Sigmoid برای نورون‌های خروجی شبکه‌های عصبی با هدف طبقه‌بندی مفید باشند.
- در همه‌جا مشتق‌پذیر است.
- ماهیت آن غیرخطی است.

**معایب**

- از مشکل اشباع (saturation problem) رنج می‌برد. یک نورون در صورتی اشباع شده در نظر گرفته می‌شود که به حداکثر یا حداقل مقدار خود برسد، به طوری که مشتق آن برابر با ۰ باشد. در این صورت، وزن‌ها بروز نمی‌شوند که باعث یادگیری ضعیف برای شبکه‌های عمیق می‌شود.
- این یک تابع صفرمرکز نیست. بنابراین، گرادیان تمام وزن‌های متصل به یک نورون مثبت یا منفی است. در طول فرآیند بروزرسانی، این وزن‌ها تنها مجاز به حرکت در یک جهت، یعنی مثبت یا منفی در یک زمان هستند. این امر بهینه‌سازی تابع زیان را سخت‌تر می‌کند.
- هزینه محاسباتی زیادی دارد. این تابع یک عملیات نمایی انجام می‌دهد که در نتیجه زمان محاسبات بیشتری را می‌گیرد.

## tanh

تابع فعال‌سازی تانژانت هذلولوی‌گون یا tanh، ارتباط نزدیکی با تابع فعال‌سازی Sigmoid دارد و شکل ریاضی آن به صورت زیر است:

$$f(x) = \tanh(x) = \frac{\sinh(x)}{\cosh(x)} = \frac{e^{x} - e^{-x}}{e^{x} + e^{-x}} = 2\sigma(2x) - 1.$$

در پایتون می‌توانیم آن را به‌صورت زیر کدنویسی کنیم:

```
def htan(x):
  return (np.exp(x) - np.exp(-x))/(np.exp(x) + np.exp(-x))
```

همان طور که در معادله بالا مشاهده می‌شود، tanh به سادگی یک نسخه مقیاس‌شده از فعال‌ساز Sigmoid است. با این حال، صفرـمرکز است. از این‌رو، برخی از مشکلاتی را که فعال‌ساز Sigmoid دارد را از خود نشان نمی‌دهد. این تابع پیوسته، مشتق‌پذیر و کران‌دار در محدوده‌ای از (۱-، ۱) است. بنابراین، منفی، مثبت و صفر را به عنوان خروجی تولید می‌کند و ورودی‌های شدیدا منفی به tanh به خروجی‌های منفی نگاشت می‌شوند. بنابراین تابع فعال‌سازی tanh در صفرـمرکز می‌باشد و مشکل عدم صفرـمرکز بودن تابع Sigmoid را حل می‌کند.

مشتق تابع tanh به‌صورت زیر محاسبه می‌شود:

$$f(x) = 1 - f(x)^2$$

که در پایتون می‌توانیم آن را به‌صورت زیر بنویسیم:

```
def der_htan(x):
  return 1 - htan(x) * htan(x)
```

بیاید تابع tanh و مشتق آن را مصورسازی کنیم. برای این کار کد زیر را وارد کنید:

```
import numpy as np
import matplotlib.pyplot as plt

# Hyperbolic Tangent (htan) Activation Function
def htan(x):
  return (np.exp(x) - np.exp(-x))/(np.exp(x) + np.exp(-x))

# htan derivative
def der_htan(x):
  return 1 - htan(x) * htan(x)

# Generating data for Graph
x_data = np.linspace(-6,6,100)
y_data = htan(x_data)
dy_data = der_htan(x_data)

# Graph
plt.plot(x_data, y_data, x_data, dy_data)
plt.title('htan Activation Function & Derivative')
plt.legend(['htan','der_htan'])
plt.grid()
```

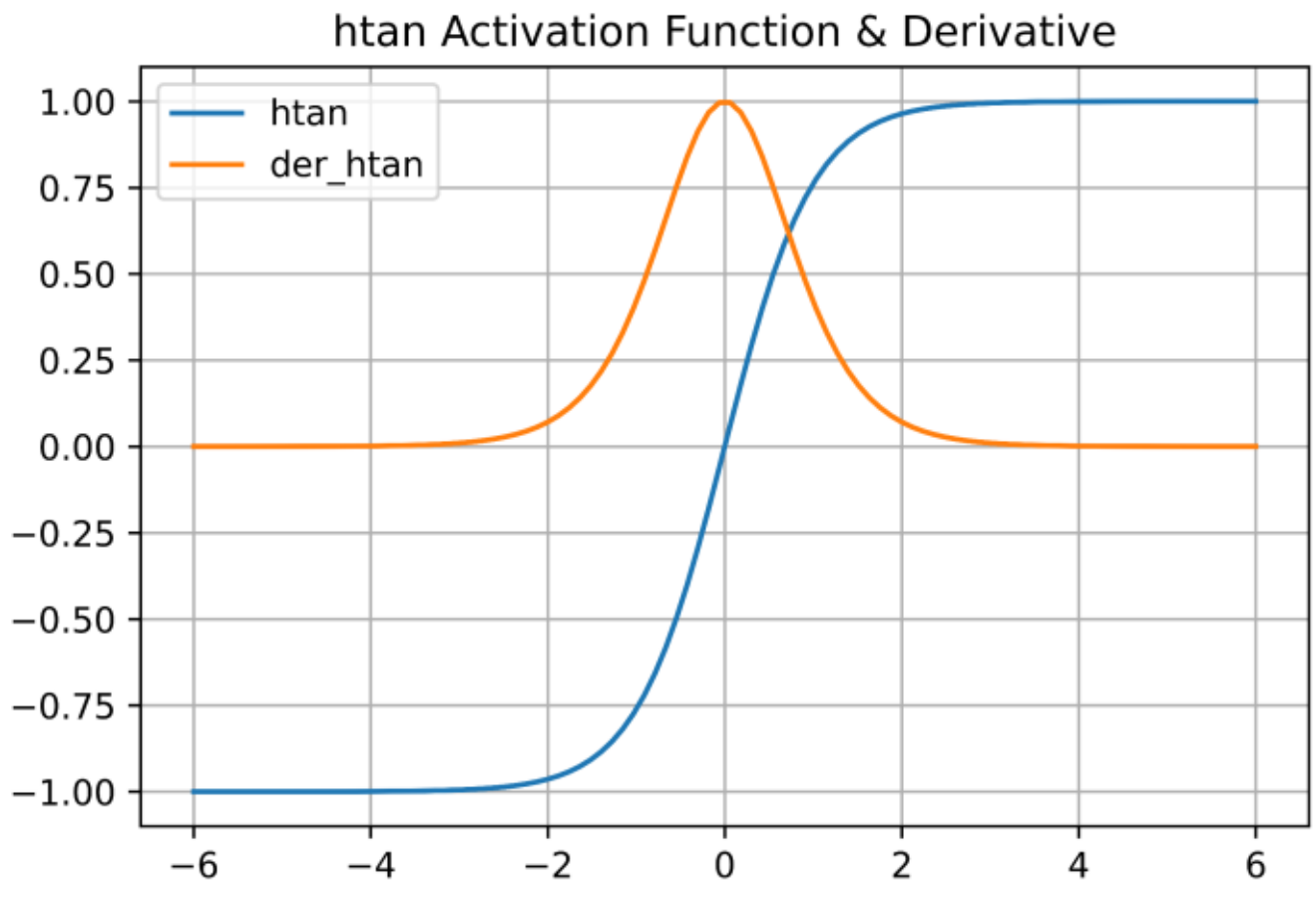

**مزایا**

- برخلاف Sigmoid، یک تابع صفر-مرکز است تا بهینه‌سازی تابع زیان آسان‌تر شود.
- خروجی نورون را در محدوده‌ای بین ۱ – و ۱ نرمال می‌کند.

**معایب**

- از نظر محاسباتی گران است.
- مستعد محو گرادیان است.

## ReLU

توابع فعال‌سازی sigmoid و tanh را نمی‌توان در شبکه‌هایی با لایه‌های زیاد به دلیل مشکل محو گرادیان استفاده کرد. تابع فعال‌سازی ReLU که محبوب‌ترین تابع فعال‌سازی در یادگیری عمیق است (در لایه پنهان)، بر مشکل محو گرادیان غلبه می‌کند و به شبکه اجازه می‌دهد سریع‌تر یاد بگیرد و عملکرد بهتری داشته باشد. تابع ReLU به صورت زیر تعریف می‌شود:

$$f(x) = \begin{cases} 0 \ for \ x \leq 0 \\ x \ for \ x > 0 \end{cases}$$

که در پایتون می‌توانیم آن را به‌صورت زیر بنویسیم:

```
def ReLU(x):
  data = [max(0,value) for value in x]
  return np.array(data, dtype=float)
```

مشتق تابع ReLU به‌صورت زیر محاسبه می‌شود:

$$\acute{f}(x) = \begin{cases} 0\ for\ x \leq 0 \\ 1\ for\ x > 0 \end{cases}$$

در پایتون می‌توانیم آن را به‌صورت زیر کدنویسی کنیم:

```
def der_ReLU(x):
  data = [1 if value>0 else 0 for value in x]
  return np.array(data, dtype=float)
```

برای مصورسازی تابع ReLU و مشتق آن کد زیر را وارد کنید:

```
import numpy as np
import matplotlib.pyplot as plt

# Rectified Linear Unit (ReLU)
def ReLU(x):
  data = [max(0,value) for value in x]
  return np.array(data, dtype=float)

# Derivative for ReLU
def der_ReLU(x):
  data = [1 if value>0 else 0 for value in x]
  return np.array(data, dtype=float)

# Generating data for Graph
x_data = np.linspace(-10,10,100)
y_data = ReLU(x_data)
dy_data = der_ReLU(x_data)

# Graph
plt.plot(x_data, y_data, x_data, dy_data)
plt.title('ReLU Activation Function & Derivative')
plt.legend(['ReLU','der_ReLU'])
plt.grid()
plt.show()
```

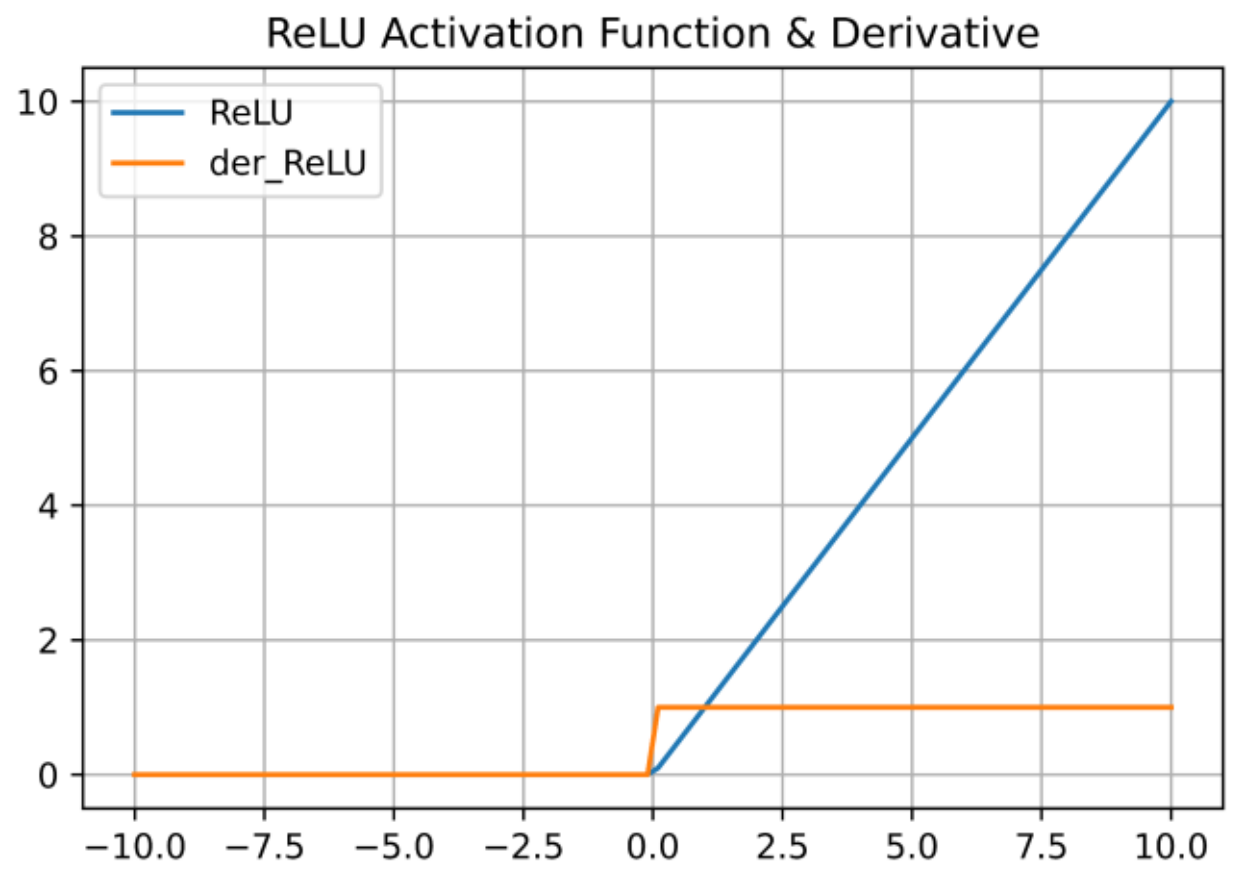

**مزایا**

- به میزان زیادی همگرایی گرادیان کاهشی تصادفی را در مقایسه با توابع Sigmoid تسریع می‌کند.
- می‌تواند با مشکل محو گرادیان مقابله کند.
- تابع محاسباتی ارزانی است.
- پرکاربردترین تابع فعال‌سازی است.
- خیلی سریع همگرا می‌شود.

**معایب**

- صفر-مرکز نیست.
- مشکل نورون مرده دارد. سمت منفی نمودار مقدار گرادیان را صفر می‌کند. به همین دلیل، در طول فرآیند پس‌انتشار، وزن‌ها و سوگیری‌ها برای برخی نورون‌ها بروز نمی‌شوند. این می‌تواند نورون‌های مرده‌ای ایجاد کند که هرگز فعال نمی‌شوند. به عبارت دیگر، تمام مقادیر ورودی منفی بلافاصله صفر می‌شوند، از این رو توانایی مدل را برای آموزش درست از داده‌ها کاهش می‌دهد.

هر زمان که ReLU ورودی منفی را دریافت کند، خروجی صفر می‌شود. بنابراین، از طریق پس‌انتشار چیزی یاد نمی‌گیرد (زیرا نمی‌توانید در آن انتشار عقبگرد انجام دهد). به عبارت دیگر، اگر مشتق صفر باشد، کل فعال‌سازی صفر می‌شود، بنابراین هیچ مشارکتی از آن نورون در شبکه وجود ندارد.

## Softmax

از تابع sofmax به عنوان خروجی در مسائل طبقه‌بندی چند کلاسی برای یافتن احتمالات برای کلاس‌های مختلف استفاده می‌شود (برخلاف Sigmoid که برای طبقه‌بندی دودویی ترجیح داده می‌شود). تابع Softmax احتمالات هر کلاس هدف را برروی تمام کلاس‌های هدف ممکن محاسبه می‌کند (که به تعیین کلاس هدف کمک می کند):

$$Softmax\ (z_i) = \frac{e^{z_i}}{\sum_{k=1}^{k} e^{z_k}}\ for\ j = 1, \dots, k$$

Softmax احتمال را برای یک نقطه داده متعلق به هر کلاس، جداگانه برمی‌گرداند. توجه داشته باشید که مجموع همه مقادیر ۱ است.

در پایتون می‌توانیم آن را به‌صورت زیر کدنویسی کنیم:

```
def softmax(x):
    return np.exp(x) / np.sum(np.exp(x), axis=0)
```

برای مصورسازی تابع Softmax کد زیر را وارد کنید:

```
import numpy as np
import matplotlib.pyplot as plt

# Softmax Activation Function
def softmax(x):
    return np.exp(x) / np.sum(np.exp(x), axis=0)

# Generating data to plot
x_data = np.linspace(-10,10,100)
y_data = softmax(x_data)

# Plotting
plt.plot(x_data, y_data)
plt.title('Softmax Activation Function')
plt.legend(['Softmax'])
plt.grid()
plt.show()
```

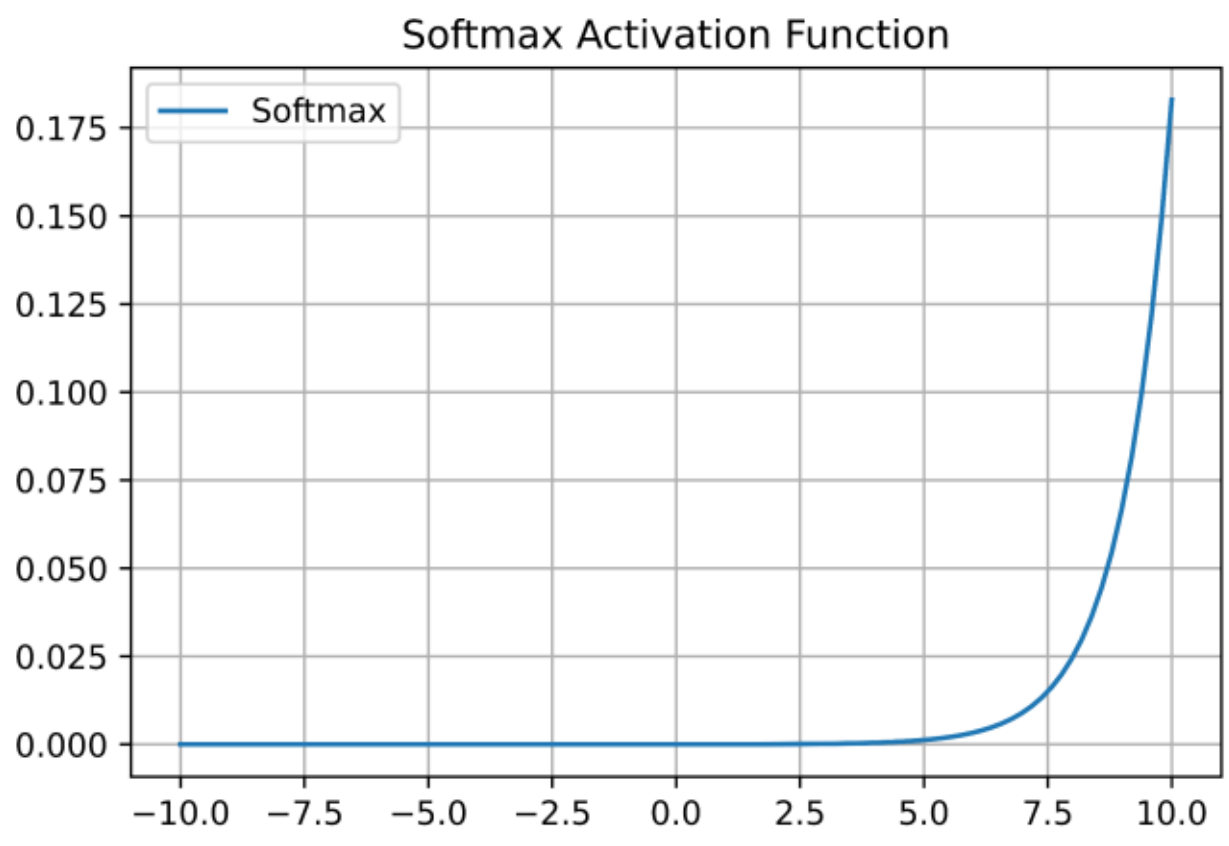

برای مسائل طبقه‌بندی چندکلاسه، لایه خروجی به اندازه کلاس هدف نورون دارد. به عنوان مثال، فرض کنید شما ٤ کلاس [A, B, C, D] دارید. از این‌رو ٤ نورون در لایه خروجی وجود خواهد داشت. فرض کنید خروجی تابع Softmax برای یک داده برابر با [۰/۱۹،۰/۴۱،۰/۱۴،۰/۲۶] شده است، با دیدن مقدار احتمال می‌توان گفت ورودی متعلق به کلاس C است.

## بهینه‌سازها و توابع زیان (LossFunction)

بهینه‌سازها الگوریتم‌هایی هستند که برای کمینه‌کردن تابع زیان/هزینه (**تابع زیان** خطای نمونه‌ی آموزشی واحد است، در حالی‌که **تابع هزینه (cost function)** خطای کل مجموعه داده آموزشی است) از طریق بروزرسانی وزن‌های شبکه استفاده می‌شوند. الگوریتم‌های بهینه‌سازی نقش بسیار

مهمی را در فرآیند آموزش شبکه بر عهده داشته و تاثیر مستقیمی بر زمان صرف شده آموزش دارند. به بیانی دیگر، الگوریتم‌های بهینه‌سازی قلب یادگیری عمیق هستند که مسئول کار پیچیده مدل‌های یادگیری عمیق برای یادگیری از داده‌ها هستند.

برای اینکه بفهمیم بهینه‌سازی چیست؟ ابتدا باید هدف را شناسایی کنیم. هدف به ویژگی‌های خاصی از سیستم به نام متغیر بستگی دارد. *هدف ما یافتن مقادیر متغیرهایی است که هدف را بهینه می‌کند.* اغلب متغیرها به نوعی محدود هستند. از منظر ریاضی، بهینه‌سازی فرآیندی است برای به **بیشینه‌سازی** (**maximizing**) یا **کمینه‌سازی** (**minimizing**) یک تابع هدف $f(x)$ با جستجوی متغیرهای مناسب $x$ با توجه به محدودیت‌های $c_i$، که می‌تواند به صورت فشرده به صورت زیر نوشته شود:

$$min_{x \epsilon R^n} f(x) subject\ to \begin{cases} c_i(x) = 0, & i \epsilon \mathcal{E} \\ c_i(x) \geq 0, & i \epsilon \mathcal{I} \end{cases}$$

که در آن $\mathcal{E}$ و $\mathcal{I}$ به ترتیب مجموعه‌ای از شاخص‌ها برای محدودیت‌های برابری و نابرابری هستند. این عبارت ریاضی قطعا در نگاه اول دلهره آور است!!، شاید به این دلیل که برای توصیف کلی بهینه‌سازی است. اما نگران نباشید، در ادامه مطالب همه چیز مشخص خواهد شد.

چندین روش مختلف برای بهینه‌سازی در یادگیری عمیق وجود دارد. به عنوان مثال، ساده‌ترین الگوریتم بهینه‌سازی مورد استفاده در یادگیری عمیق، **گرادیان کاهشی Gradient Descent** است. گرادیان کاهشی یک الگوریتم بهینه‌سازی مرتبه اول تکراری است که برای یافتن کمینه (بیشینه) محلی یک تابع معین استفاده می‌شود. این بدان معنی است که هنگام انجام بروزرسانی پارامترها فقط اولین مشتق را در نظر می‌گیرد. در هر تکرار، ما پارامترها را در جهت مخالف گرادیان تابع هدف $J(\theta)$ بروز می‌کنیم. اندازه گامی که در هر تکرار برای رسیدن به کمینه محلی برمی‌داریم با نرخ یادگیری $\alpha$ تعیین می‌شود. بنابراین جهت گرادیان به سمت پایین را دنبال می‌کنیم تا کمینه محلی برسیم.

روش دیگر بهینه‌سازی نیوتن است که با استفاده از مشتق مرتبه دوم با یافتن ریشه‌های یک تابع در بهبود بهینه‌سازی کمک می‌کند. البته این روش در مقایسه با روش‌های گرادیان کاهشی که مبتنی‌بر مشتق مرتبه اول هستند، پیچیدگی محاسباتی را به میزان قابل توجهی افزایش می‌دهد. از همین رو، گرادیان کاهشی در آموزش شبکه‌های عصبی بیشتر ترجیح داده می‌شود. الگوریتم‌های متفاوتی مبتنی‌بر گردایان کاهشی وجود دارند. در ادامه این بخش، پس تشریح کامل آن، به معرفی و بررسی سایر نسخه‌های این الگوریتم می‌پردازیم.

**تعریف ۱.۳ بهینه‌سازها**

درفرآیند آموزش شبکه های عصبی، بهینه‌سازها به دنبال مجموعه‌ای از پارامترها که وزن ها نیز نامیده می‌شوند، هستند تا مقدار زیان تابع زیان تا حد امکان کوچک شود.

## گرادیان کاهشی (نزول گرادیان)

الگوریتم گرادیان کاهشی، یک روش بهینه‌سازی مبتنی‌بر تکرار است که تلاش می‌کند با تغییر وزن‌های داخلی شبکه عصبی، مقدار تابع زیان را کمینه کند. در این روش، وزن‌های شبکه به‌صورت تدریجی بروز می‌شوند و در هر تکرار، الگوریتم تلاش می‌کند با ترفندی مقدار تابع زیان را تضعیف کند. این عمل تا جایی که تکرار آن منجر به تغییری در تابع زیان نشود، انجام می‌شود.

سه روش کلی برای استفاده از گرادیان کاهشی وجود دارد: **گرادیان کاهشی با یک نمونه**، **گرادیان کاهشی کامل** و **گرادیان کاهشی ریز دسته‌ای**. هنگامی که بروزرسانی تنها با یک نمونه انجام شود به آن روش **گرادیان کاهشی تصادفی** (**Stochastic gradient descent**) گفته می‌شود.

در این روش، با ورود هر نمونه به شبکه بروزرسانی اعمال و وزن‌های جدید بدست می‌آیند. نقطه ضعف این روش، گیر افتادن در کمینه محلی است. علاوه براین، نتایج ناپایداری را به سبب پاسخ به هربار ورود نمونه به شبکه دارد.

گرادیان کاهشی تصادفی، زمان بروزرسانی کاهش می‌دهد و مقداری از افزونگی محاسباتی را حذف می‌کند که به طور قابل توجهی محاسبه را تسریع می‌کند.

در روش گرادیان کاهشی کامل، شبکه با تمام نمونه‌های آموزشی تغذیه می‌شود. شبکه پس از محاسبه خطا برای تمام نمونه‌ها یک بار بروزرسانی را انجام می‌دهد. اگرچه این روش به مدل در فرار از کمینه محلی کمک کرده و همگرایی پایدارتری در مقایسه با گرادیان کاهشی یک نمونه‌ای ارائه می‌دهد، با این همه، زمان آموزش طولانی‌تر را در پی دارد. همچنین تغذیه تمام نمونه‌های آموزشی به شبکه به‌دلیل کمبود حافظه همیشه امکان‌پذیر نیست. در خلا بین این دو روش، **گرادیان کاهشی ریزدسته‌ای** (**mini-batch gradient descent**) استفاده می‌شود. در این روش شبکه با گروهی از نمونه‌های آموزشی تغذیه می‌شود تا از مزیت هر دو روش قبلی استفاده کند. بروزرسانی پارامترها با استفاده از گروهی از نمونه‌ها مزیتی مهم دارد: *با استفاده از این روش مدل نسبت به نمونه‌های نویزدار مقاوم‌تر بوده و واریانس کم‌تری در بروزرسانی پارامترها دارد*. این عمل همگرایی پایدارتری را ارائه می‌دهد. با این همه، این روش‌ها نیاز به انتخاب **نرخ یادگیری** ($\alpha$) دارند که انتخاب آن همیشه آسان نیست. علاوه براین، نرخ یادگیری یکسان در تمام مراحل آموزشی و برای همه پارامترهای مختلف نمی‌تواند بهینه باشد. اگر $\alpha$ خیلی بزرگ انتخاب شود، آموزش ممکن است نوسان کند، همگرا نشود یا از کمینه‌های محلی مربوط بگذرد. در مقابل، اگر نرخ یادگیری خیلی کوچک انتخاب شود، به طور قابل توجهی فرآیند همگرایی را به تاخیر می‌اندازد. از این‌رو، یک تکنیک رایج برای دورزدن این مساله استفاده از **نرخ واپاشی**

(**rate decay**) یادگیری است. به عنوان مثال، با استفاده از واپاشی گامی می‌توان نرخ یادگیری را هر چند دوره به میزانی کاهش داد. این امر این امکان را می‌دهد تا میزان یادگیری زیادی در ابتدای آموزش و نرخ یادگیری کم‌تری در پایان آموزش وجود داشته باشد. با این حال، این روش واپاشی نیز، به خودی خود یک ابرپارامتر است و بسته به کاربرد باید با دقت طراحی شود.

هدف بهینه‌سازهای نرخ یادگیری تطبیقی، حل مشکل یافتن نرخ یادگیری درست است. در این روش‌ها، نرخ یادگیری $\alpha$ یک متغیر سراسری نیست، اما در عوض هر پارامترِ قابل آموزش، نرخِ یادگیری جداگانه‌ای برای خود دارد. در حالی که این روش‌ها اغلب هنوز نیاز به تنظیم ابرپارامتر دارند، بحث اصلی این است که آن‌ها برای طیف وسیع‌تری از پیکربندی‌ها به خوبی کار می‌کنند؛ اغلب زمانی که تنها از ابرپارمترهای پیش‌فرضِ پیشنهادی استفاده می‌کنند.

## پیاده‌سازی گرادیان کاهشی در پایتون

همانطور که پیش‌تر بیان شد در بهینه‌سازی ما قصد داشتیم یک تابع را $min$ کنیم. حال بیاید ببینیم چگونه می‌توان تابع $min$ را حل کنیم؟ به لطف حساب دیفرانسیل و انتگرال، ابزاری به نام گرادیان داریم. توپی را در بالای تپه تصور کنید. می‌دانیم که تپه در نقاط مختلف دارای شیب‌های (گرادیان‌ها) مختلف است. به دلیل جاذبه، توپ به دنبال منحنی تپه به سمت پایین حرکت می‌کند. توپ به کدام سمت می‌رود؟ تندترین شیب. پس از مدتی، توپ به کمینه محلی می‌رسد که در آن زمین نسبت به اطراف خود صاف باشد. این ماهیت گرادیان کاهشی است. می‌توانیم مسیر صاف توپ را به مراحل کوچک تبدیل کنیم. در مرحله k_ام، دو کمیت خواهیم داشت: طول گام $\alpha_k$ و جهت $p_k$. برای مشاهده گردایان کاهشی در عمل، ابتدا چند کتابخانه را وارد می‌کنیم.

```
import numpy as np
import matplotlib.pyplot as plt
from matplotlib.ticker import MaxNLocator
from itertools import product
```

برای شروع، یک تابع هدف ساده $f(x) = x^2 - 2x - 3$ را تعریف می‌کنیم که $x$ اعداد حقیقی هستند. از آنجایی که گرادیان کاهشی از گرادیان استفاده می‌کند، گرادیان $f$ را نیز تعریف می‌کنیم، که فقط مشتق اول $f$ است، یعنی $\nabla f(x) = 2x - 2$:

```
def func(x):
    return x**2 - 2*x - 3

def fprime(x):
    return 2*x - 2
```

در مرحله بعد، توابع پایتون را برای رسم تابع هدف و مسیر یادگیری در طول فرآیند بهینه‌سازی تعریف می‌کنیم:

```
def plotFunc(x0):
    x = np.linspace(-5, 7, 100)
    plt.plot(x, func(x))
    plt.plot(x0, func(x0), 'ko')
    plt.xlabel('$x$')
    plt.ylabel('$f(x)$')
    plt.title('Objective Function')

def plotPath(xs, ys, x0):
    plotFunc(x0)
    plt.plot(xs, ys, linestyle='--', marker='o', color='#F12F79')
    plt.plot(xs[-1], ys[-1], 'ko')
```

از نمودار زیر، براحتی می‌توانیم ببینیم که $f$ دارای کمینه مقدار $x = 1$ است . فرض کنید از $x = -4$ شروع می‌کنیم (که با یک نقطه مشکی در زیر نشان داده شده است)، می‌خواهیم ببینیم که آیا گرادیان کاهشی می‌تواند کمینه محلی $x = 1$ را تعیین کند یا خیر.

```
x0 = -4
plotFunc(x0)
```

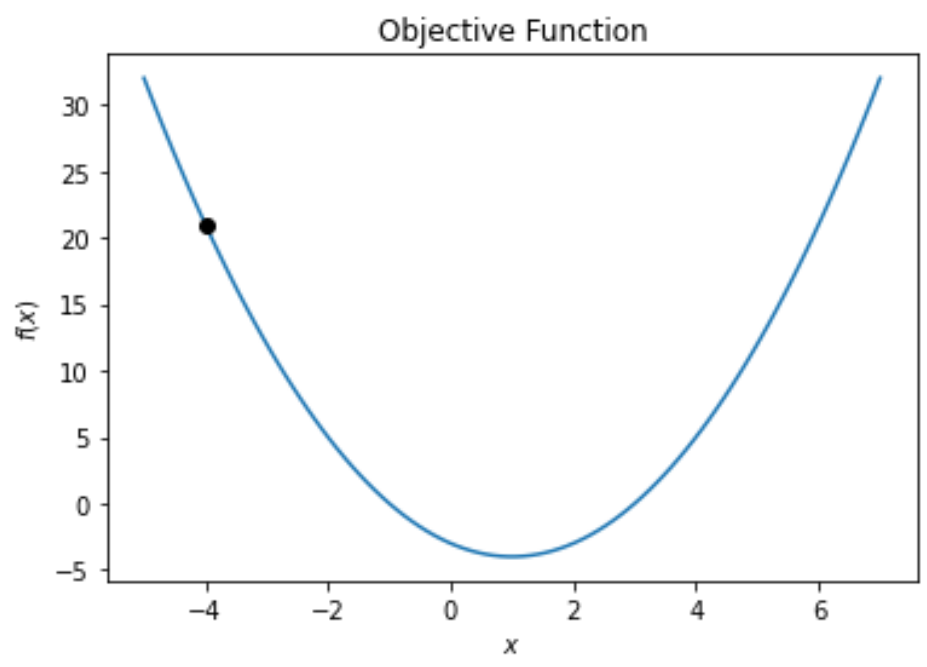

یک الگوریتم گرادیان کاهشی ساده را به این صورت تعریف می‌کنیم: برای هر نقطه $x_k$ در ابتدای مرحله $k$، طول گام $\alpha_k$ را ثابت نگه می داریم و جهت $p_k$ را منفی مقدار گرادیان قرار می دهیم. با استفاده از فرمول زیر این مراحل را انجام می‌دهیم:

$$x_{k+1} = x_k + \alpha_k p_k$$

جایی که گرادیان بالاتر از یک مقدار تلورانس معین است (در مورد ما $1 \times 10^{-5}$) و تعداد مراحل یک مقدار معین است (در مورد ما ۱۰۰۰).

```
def GradientDescentSimple(func, fprime, x0, alpha, tol=1e-5, max_iter=1000):
    # initialize x, f(x), and -f'(x)
    xk = x0
    fk = func(xk)
    pk = -fprime(xk)
    # initialize number of steps, save x and f(x)
    num_iter = 0
    curve_x = [xk]
    curve_y = [fk]
    # take steps
    while abs(pk) > tol and num_iter < max_iter:
```

```
        # calculate new x, f(x), and -f'(x)
        xk = xk + alpha * pk
        fk = func(xk)
        pk = -fprime(xk)
        # increase number of steps by 1, save new x and f(x)
        num_iter += 1
        curve_x.append(xk)
        curve_y.append(fk)
    # print results
    if num_iter == max_iter:
        print('Gradient descent does not converge.')
    else:
        print('Solution found:\n  y = {:.4f}\n  x = {:.4f}'.format(fk, xk))

    return curve_x, curve_y
```

از $x = -4$ شروع می‌کنیم و الگوریتم گرادیان کاهشی را روی $f$ با سناریوهای مختلف اجرا می کنیم:

$\alpha_k = 0.1$
$\alpha_k = 0.9$
$\alpha_k = 1 \times 10^{-4}$
$\alpha_k = 1.01$

**سناریو اول:** $\alpha_k = 0.1$

```
xs, ys = GradientDescentSimple(func, fprime, x0, alpha=0.1)
plotPath(xs, ys, x0)
```

```
Solution found:
  y = -4.0000
  x = 1.0000
```

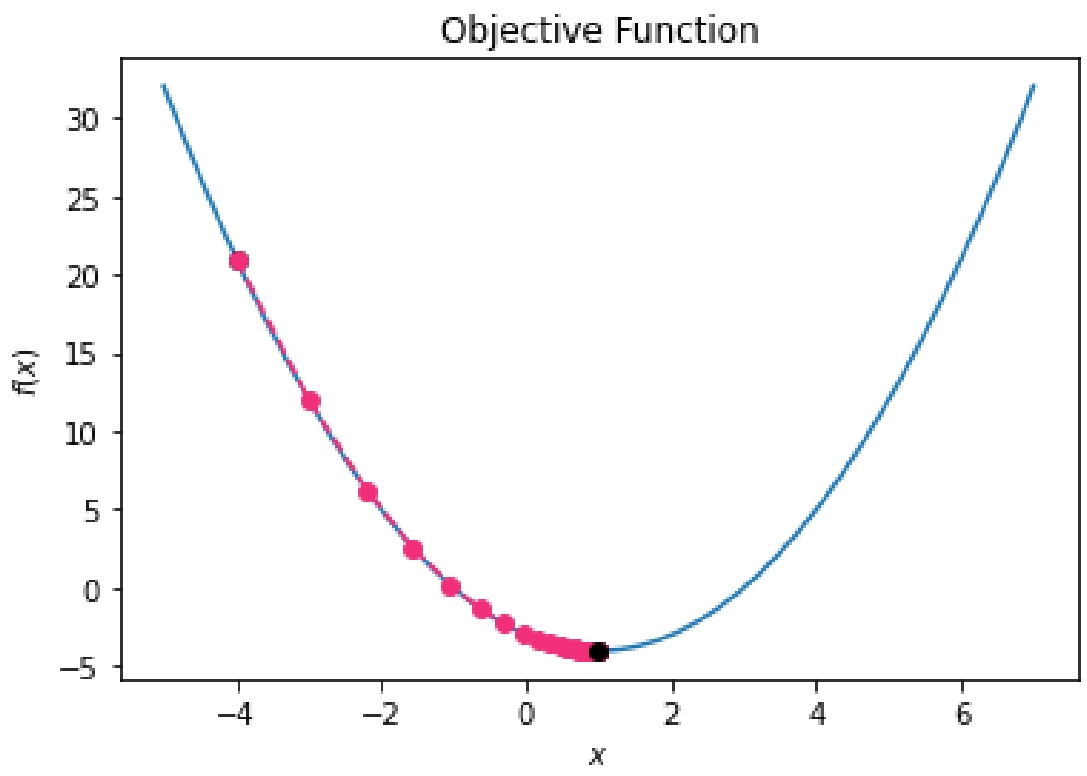

**سناریو دوم:** $\alpha_k = 0.9$

```
xs, ys = GradientDescentSimple(func, fprime, x0, alpha=0.9)
plotPath(xs, ys, x0)
```

```
Solution found:
  y = -4.0000
  x = 1.0000
```

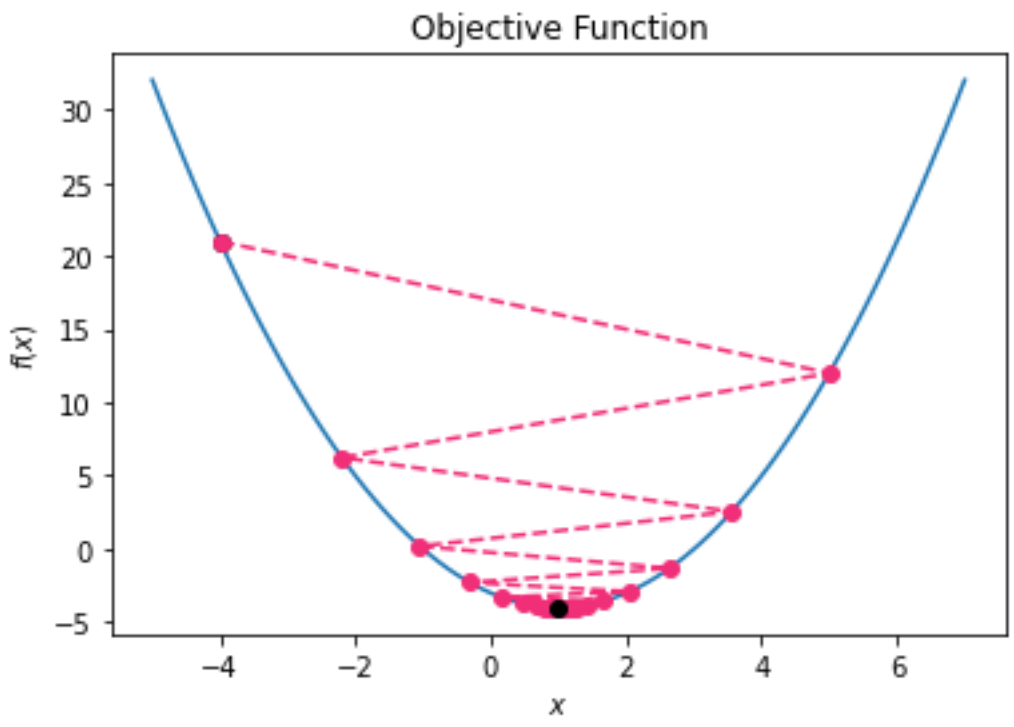

سناریو سوم: $\alpha_k = 1 \times 10^{-4}$

```
xs, ys = GradientDescentSimple(func, fprime, x0, alpha=1e-4)
plotPath(xs, ys, x0)
```

```
Gradient descent does not converge.
```

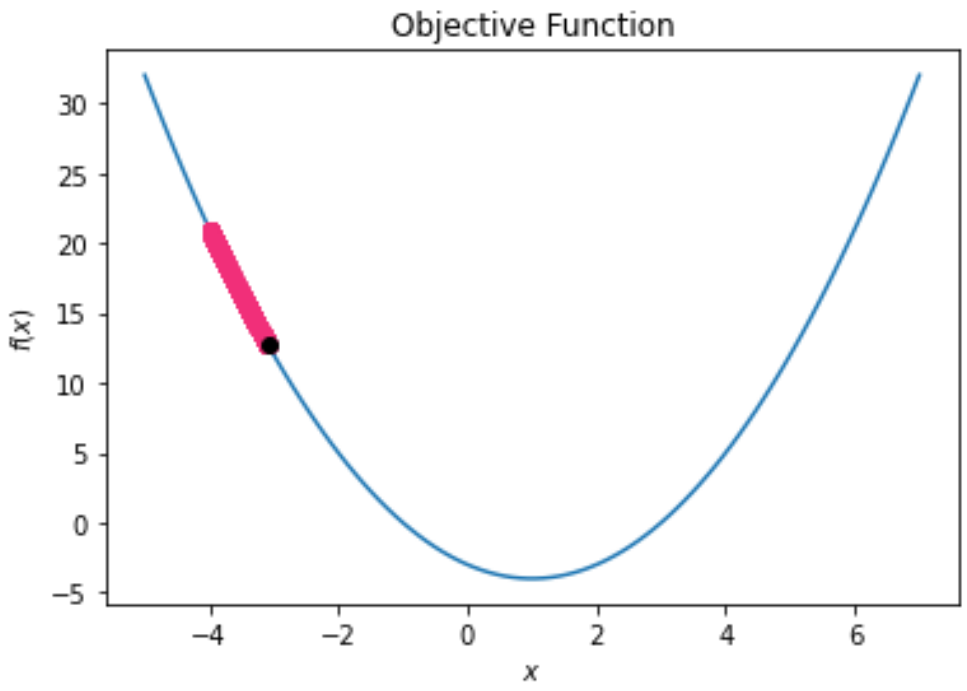

سناریو چهارم: $\alpha_k = 0.1$

```
xs, ys = GradientDescentSimple(func, fprime, x0, alpha=1.01, max_iter=8)
plotPath(xs, ys, x0)
```

```
Gradient descent does not converge.
```

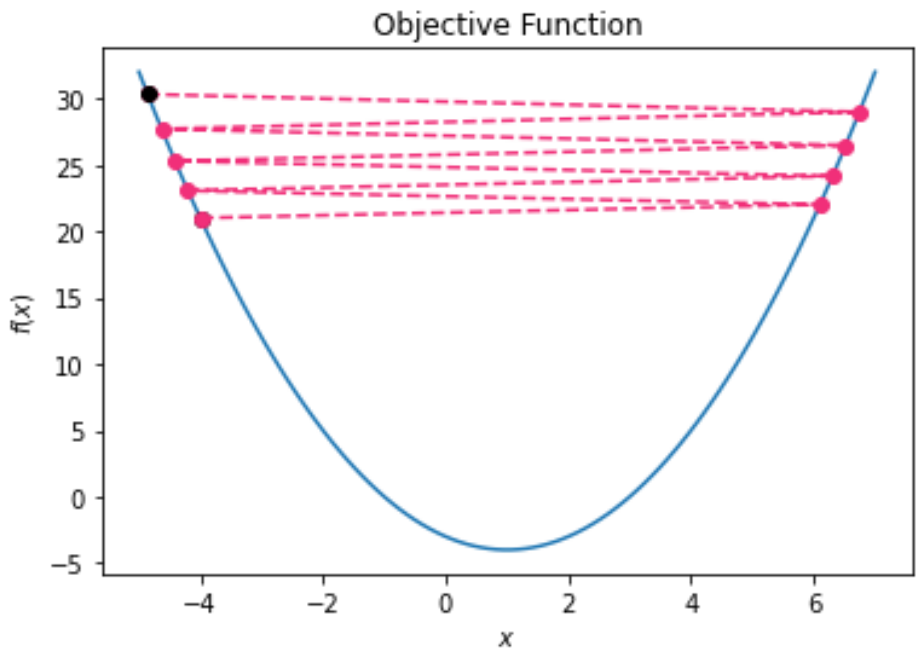

چیزی که از مصورسازی‌های بالا در سناریوهای مختلف بدست آوردیم، به صورت زیر خلاصه می‌شود:

سناریوی اول براحتی همگرا شد. حتی اگر طول گام ثابت باشد، جهت به سمت صفر کاهش می‌یابد و از این رو منجر به همگرایی می‌شود.

سناریوی دوم نیز با وجود اینکه مسیر یادگیری بدلیل طول گام بزرگ در اطراف راه حل در نوسان است، همگرا می‌شود.

سناریوی سوم به سمت راه حل حرکت می‌کند. با این حال، طول گام بقدری کوچک است که تعداد تکرارها به حداکثر می‌رسد و نمی‌تواند جواب را بیابد. در این مورد، افزایش max_iter مشکل را حل می‌کند.

سناریوی چهارم بدلیل طول گام بزرگ متفاوت است. در اینجا، $\text{max_iter} = 8$ را تنظیم کرده‌ایم تا مصورسازی را بهتر کنیم.

چیزی که می‌توان فهمید این است که راه حل $x = 1$ را می‌توان با گرادیان کاهشی با طول گام مناسب بدست آورد.

شاید تعجب کنید که چرا از راه حل تحلیلی دقیق استفاده نمی‌کنیم: مشتق $f$ را بگیرید، سپس $x$ را طوری حل کنید که مشتق صفر شود. برای مثال قبلی، متوجه می‌شویم که $x$ای که $f$ را به کمینه می‌کند، $\nabla f(x) = 2x - 2$ را برآورده کند، یعنی $x = 1$. بله، این یک راه است. اما زمانی که با یک مسئله بهینه‌سازی مواجه می‌شوید که در آن مشتق $f$ سخت است یا حل آن غیرممکن است، دیگر این تکنیک توصیه‌شده نمی‌شود.

*توجه داشته باشید که پیاده‌سازی ساده‌یِ بالا تنها برای درک بهتر نحوه کار گرادیان کاهشی همراه با مصورسازی بوده است. در عمل نیازی به پیاده‌سازی نیست و چارچوب‌های یادگیری عمیق، پیاده‌سازی کارآمد این الگوریتم‌ها را در خود جای داده‌اند.*

## گرادیان کاهشی مبتنی بر تکانه (Momentum)

روش گرادیان کاهشی استاندارد با استفاده از نرخ یادگیری $\alpha$ یک گام کوچک در جهت مخالف گرادیان برمی‌دارد. این پارامتر در طول کل فرآیند یادگیری دارای یک مقدار ثابت است. معادله ریاضی آن به صورت زیر می‌باشد:

$$\theta_{t+1} = \theta_t - \alpha . \nabla_t(\theta_t)$$

که در آن $\theta_t$ پارامترها/وزن‌های مدل در دوره $t$، $\nabla_t(\theta_t)$ گرادیان برای هر وزن $\theta_t$ در دوره $t$ و $\alpha$ نرخ یادگیری است. گرادیان کاهشی استاندارد، مستقل از مراحل قبلی، با اندازه گام ثابت در سراشیبی حرکت می‌کند. اگر تابع هدف شما همانند یک دره طولانی با بهینه‌های محلی و دیوارهای شیب‌دار در دو طرف باشد (شکل ۳ـ۳ را ببینید)، بروزرسانی وزن‌ها بسیار کند

خواهد بود و منجر به تعداد زیادی مرحله می‌شود. برای حل این موضوع، تکانه به الگوریتم گرادیان کاهشی اضافه می‌شود. ایده اصلی تکانه، اضافه کردن حافظه کوتاه‌مدت به گرادیان کاهشی است. به عبارت دیگر، به جای استفاده از گرادیان مرحله فعلی برای هدایت جستجو، **تکانه**، گرادیان‌های گام‌های گذشته را نیز برای تعیین جهت انباشته می‌کند. این مکانیسم را می‌توان به صورت زیر اجرا کرد:

$$\theta_{t+1} = \theta_t - v_t$$

$$v_t = \gamma . v_{t-1} - \alpha . \nabla_t(\theta_t)$$

در این معادلات، $\gamma$ عبارت تکانه است که تاثیر گرادیان‌های قبلی را بر بروزرسانی فعلی تعیین می‌کند. ایده کلی این است که این عبارت را تا حد امکان نزدیک به ۱ قرار دهیم و برای نرخ یادگیری مقداری تا حد امکان بالاتر انتخاب کرده، در حالی که همگرایی پایدار را حفظ می‌کنیم.

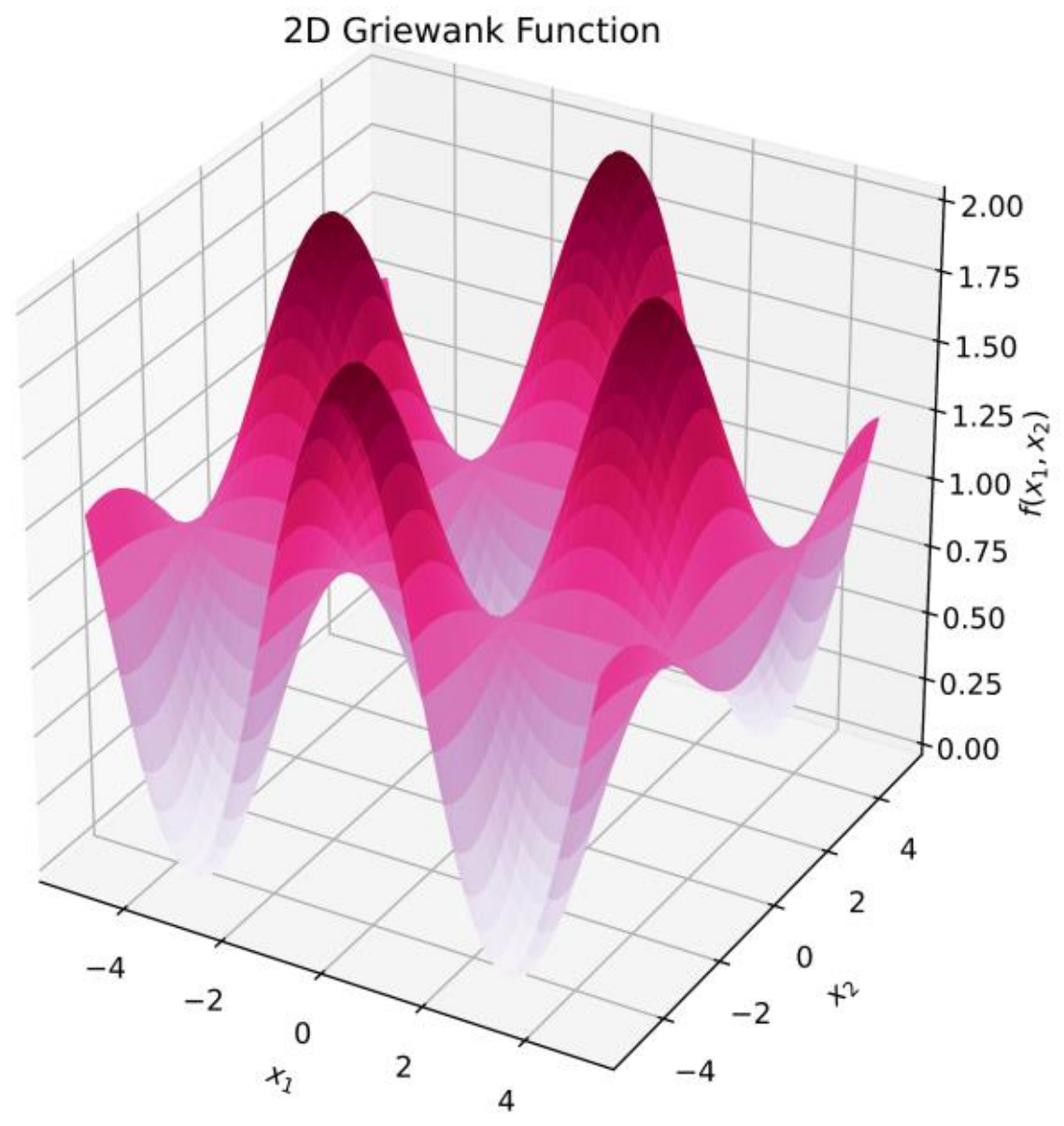

**شکل ۳ـ۳. مصورسازی تابع Griewank**

تکانه با اضافه کردن تاریخچه به معادله بروزرسانی پارامتر بر اساس گرادیانی که در بروزرسانی‌های قبلی با آن مواجه شده است، تغییری در گرادیان کاهشی بوجود می‌آورد (به نوعی یک حافظه به آن اضافه می‌کند) که به همگرایی سریع‌تر منجر می‌شود.

## گرادیان کاهشی تسریع‌شده نستروف (NAG)

قیاس توپی که از دره غلت می‌زند و از طریق تکانه سرعت می‌گیرد، ویژگی خوبی برای پیاده‌سازی به نظر می‌رسد. با این حال، یک توپ وقتی در دره باشد متوقف نمی‌شود، بیشتر و بیشتر خواهد چرخید تا زمانی که هیچ سرعتی باقی نماند. از طرفی، اگر در طرف دیگر دره یک سربالایی وجود داشته باشد، این امر باعث می‌شود توپ در نهایت به عقب برگردد، اما قبل از توقف توپ در نقطه‌ی کمینه دره، نوسان‌های متعددی خواهد داشت. نستروف (۱۹۸۳)، ایده‌ای را برای جلوگیری از بالا رفتن بیش از حد توپ در سربالایی مطرح کرد. برای محاسبه اندازه‌گام در دوره فعلی، گرادیان را در مکان تقریبی که توپ با استفاده از تکانه استاندارد به پایان می‌رسد، محاسبه می‌کند. سپس، از این گرادیان برای ایجاد یک اصلاح استفاده می‌شود که به ویژه در مواردی که شیب سربالایی در محل تقریبی وجود دارد، مناسب است. این کار سبب می‌شود تا از اندازه گام‌های خیلی بزرگ که به طرف دیگر دره می‌رود، جلوگیری کند. نستروف این روش را به صورت زیر پیاده‌سازی می‌کند:

$$\theta_{t+1} = \theta_t - v_t$$

$$v_t = \gamma . v_{t-1} - \alpha . \nabla_t(\theta_t - \gamma . v_{t-1})$$

این روش **گرادیان کاهشی تسریع‌شده نستروف** (**Nesterov Accelerated Gradient Descent**) یا به اختصار **NAG** نامیده می‌شود. نقطه ضعف این روش محاسبه $\nabla_t(\theta_t - \gamma . v_{t-1})$ است که می‌تواند برای برخی از شبکه‌ها از نظر محاسباتی زمان‌بر باشد. سوتسکور[1] و همکاران (۲۰۱۳) پیشنهاد می‌کنند که محاسبه $v_t$ به روش زیر اصلاح شود:

$$v_t = \gamma . m_t + \alpha . \nabla_t(\theta_t)$$

$$m_t = \gamma . m_{t-1} + \alpha . \nabla_t(\theta_t)$$

### الگوریتم‌های گرادیان کاهشی در keras

هنگام استفاده از Keras، برای انتخاب بهینه‌ساز، می‌توان با نمونه‌سازی مستقیم کلاس SGD و استفاده از آن در زمان کامپایل مدل، بهینه‌ساز گرادیان کاهشی تصادفی (SGD) را سفارشی (customize) کرد:

```
from tensorflow.keras.optimizers import SGD
...
sgd = SGD(learning_rate=0.0001, momentum=0.8, nesterov=True)
model.compile(optimizer=sgd, loss = ..., metrics= ...)
```

[1] Sutskever

کلاس SGD پارامتر learning_rate (نرخ یادگیری با تنظیم پیش‌فرض روی ۰/۰۱)، تکانه (momentum)، نستروف (nesterov) با یک مقدار بولین و یک پارامتر واپاشی (decay) اختیاری را می‌پذیرد.

## الگوریتم‌های بهینه‌سازی نرخ یادگیری تطبیقی

تمام نسخه‌های گرادیان کاهشی که تاکنون معرفی شدند، برای محاسبه بروزرسانی هر پارامتر منفرد، از نرخ یادگیری یکسانی استفاده می‌کنند. اما نکته اینجاست که، اطلاعات گرادیان در یک بروزرسانی منفرد، می‌تواند برای یک پارامتر بسیار مهم‌تر از یک پارامتر دیگر باشد. از این‌رو، در چنین شرایطی پارامتری با بروزرسانی‌های مهم‌تر، باید گام بزرگتری در جهت گرادیان بردارد تا پارامتر با بروزرسانی کم‌تر مهم. به عبارت دیگر، برای سرعت بخشیدن به فرآیند یادگیری، هر پارامتری که باید بروز شود، باید نرخ یادگیری خاص خود را در هر دوره داشته باشد.

از این‌رو، الگوریتم‌های متفاوتی برای حل این مشکل در جهت تطبیق نرخ یادگیری در مراحل مختلف الگوریتم ارائه شده است تا همگرایی سریع‌تر شبکه را بوجود آورند. در ادامه به بررسی برخی از این الگوریتم‌ها می‌پردازیم.

### Adagrad

یافتن نرخ یادگیری بهینه برای یک الگوریتم یادگیری عمیق می‌تواند یک مشکل پیچیده باشد. اگر نرخ یادگیری خیلی بالا تنظیم شود، پارامتر ممکن است با نوسانات زیادی حرکت کند تا به سطح قابل قبولی از زیان برسد. از طرف دیگر، تنظیم نرخ یادگیری بسیار پایین منجر به پیشرفت بسیار کند خواهد شد. Adagrad برای این منظور ارائه شد، جایی که انتخاب دستی نرخ‌های یادگیری مختلف برای هر بعد از مسئله به دلیل حجم ابعاد غیرعملی است. Adagrad به‌طور تطبیقی پارامتر نرخ یادگیری را برای هر بعد مقیاس‌بندی می‌کند تا اطمینان حاصل شود که فرآیند آموزش نه خیلی کند است و نه خیلی فرار و نادقیق. برای انجام این کار، الگوریتم AdaGrad به صورت پویا دانش هندسه داده‌های مشاهده‌شده در تکرارهای گذشته را ترکیب می‌کند. سپس، این دانش را برای **تنظیم نرخ یادگیری پایین‌تر برای ویژگی‌های متداول‌تر** و **نرخ‌های یادگیری بالاتر برای ویژگی‌های نسبتا نادر** اعمال می‌کند. در نتیجه، ویژگی‌های نادر برجسته می‌شوند و یادگیرنده را قادر می‌سازد تا ویژگی‌های نادر و در عین حال بسیار پیش‌گویانه را شناسایی کند.

در این روش نرخ یادگیری جداگانه برای هر پارامتر مدل، بر اساس تاریخچه کامل گرادیان‌های مجذور پارامتر محاسبه می‌شود. گرادیان مجذور مشابه با اهمیت گرادیان است. آنها برای هر پارامتر حساب می‌شوند و نرخ یادگیری فعلی با تقسیم گرادیان فعلی بر مجذور گرادیان به اضافه یک مقدار کوچک $\varepsilon$ (برای جلوگیری از تقسیم بر صفر) محاسبه می‌شود. این بدان معناست که، هر چه گرادیان‌های بدست‌آمده قبلی بزرگ‌تر باشند، گرادیان فعلی اهمیت کم‌تری دارد و در نتیجه

نرخ یادگیری و بدین ترتیب اندازه گام در دوره فعلی کوچکتر خواهد بود. معادله ریاضی این روش به‌صورت زیر است:

$$\theta_{t+1} = \theta_t - \alpha.\frac{\nabla_t(\theta_t)}{\sqrt{G_t} + \varepsilon}$$

$$G_t = G_{t-1} + \nabla_t(\theta_t)^2$$

### الگوریتم Adagrad در keras

با استفاده از قطعه کد زیر می‌توان از بهینه‌ساز Adagrad در Keras استفاده کرد:

```
from tensorflow.keras.optimizers import Adagrad
...
adagrad = Adagrad(learning_rate=0.0001, epsilon=1e-6)
model.compile(optimizer=adagrad, loss = ..., metrics= ...)
```

### AdaDelta

ایده پشت الگوریتم Adagrad خوب است، اما الگوریتم دارای برخی نقاط ضعف است. پس از مدتی، انباشت گرادیان‌های مربع به مقدار زیادی می‌رسد که نرخ یادگیری استفاده شده در بروزرسانی‌ها بسیار کم می‌شود و از این‌رو تقریبا نمی‌توان هیچ پیشرفتی ایجاد کرد. الگوریتم AdaDelta، سعی می کند به این مشکل رسیدگی کند، به‌طوری که پنجره گرادیان‌های گذشته انباشته بجای گرفتن کل تاریخ، به اندازه ثابت محدود شود. برای این‌کار، بجای جمع تمام گرادیان‌های مجذور از ابتدای آموزش، فرض کنید مجموع رو به زوالی از این گرادیان‌ها را حفظ می‌کنیم. ما می‌توانیم این را به عنوان یک لیست در حال اجرا از جدیدترین گرادیان‌ها برای هر وزن در نظر بگیریم. هر بار که وزن‌ها را بروزرسانی می‌کنیم، گرادیان جدید را در انتهای فهرست قرار می‌دهیم و قدیمی‌ترین را از شروع حذف می‌کنیم. برای یافتن مقداری که برای تقسیم گرادیان جدید استفاده می‌کنیم، همه مقادیر موجود در لیست را جمع می‌کنیم، اما ابتدا همه آن‌ها را بر اساس موقعیت آن‌ها در لیست در یک عدد ضرب می‌کنیم. مقادیر اخیر در یک مقدار بزرگ ضرب می‌شوند، در حالی که قدیمی‌ترین‌ها در یک مقدار بسیار کوچک ضرب می‌شوند. به این ترتیب مجموع در حال اجرا ما به شدت توسط گرادیان‌های اخیر تعیین می‌شود، اگرچه به میزان کم‌تری تحت تاثیر گرادیان‌های قدیمی‌تر است. به‌طور خلاصه، برای پیاده‌سازی کارآمد، بجای ذخیره کردن گرادیان‌های قبلی مربوط، از میانگین رو به زوال نمایی از همه گرادیان‌های مجذور گذشته استفاده می‌شود.

این الگوریتم به‌طور تطبیقی میزان بروزرسانی وزن‌ها را در هر مرحله با استفاده از مجموع وزنی هر مرحله تغییر می‌دهد. از آنجایی که Adadelta میزان یادگیری وزن‌ها را به صورت

جداگانه تنظیم می‌کند، هر وزنی که برای مدتی در شیب تند قرار داشته باشد، سرعت خود را کاهش می‌دهد تا به صورت خارج نشود، اما زمانی که آن وزن در قسمت صاف‌تری قرار می‌گیرد، می‌تواند گام‌های بزرگ‌تری بردارد.

جدای از این، نویسندگان مقاله روشی را برای حذف نیاز به نرخ یادگیری از الگوریتم معرفی می‌کنند. آنها خاطرنشان می‌کنند که واحدهای نرخ یادگیری و میانگین زوال تمام گرادیان‌های مجذور گذشته با هم مطابقت ندارند. آنها این مشکل را با جایگزین کردن نرخ یادگیری با میانگین رو به زوال دیگری، از پارامتر مربع بروزرسانی $\Delta\theta_t^2$ حل می‌کنند. این شبیه به اهمیت تغییرات قبلی پارامتر است. با تقسیم ریشه دوم آن بر جذر اهمیت گرادیان‌های قبلی، به مقداری می‌رسیم که کم و بیش نسبت اهمیت گرادیان‌های قبلی است که برای بروزرسانی پارامترها استفاده شده است، یا به عبارت دیگر تاثیر گرادیان‌های قبلی بر مقدار فعلی پارامتر. نماد ریاضی آن به شرح زیر است:

$$\theta_{t+1} = \theta_t - \nabla_t(\theta_t).\frac{\sqrt{E(\Delta\theta_t^2)_t + \varepsilon}}{\sqrt{E(\nabla^2)_t + \varepsilon}}$$

$$E(\Delta\theta_t^2)_t = \gamma.E(\Delta\theta^2)_{t-1} + (1-\gamma)\,\Delta\theta_t^2$$

$$E(\nabla^2)_t = \gamma.E(\nabla^2)_{t-1} + (1-\gamma)\nabla_t(\theta_t)^2$$

### الگوریتم Adadelta در keras

با استفاده از قطعه کد زیر می‌توان از بهینه‌ساز Adadelta در Keras استفاده کرد:

```
from tensorflow.keras.optimizers import Adadelta
...
adadelta= Adadelta(learning_rate=0.0001, epsilon=1e-6)
model.compile(optimizer= adadelta, loss = ..., metrics= ...)
```

### RMSprop

الگوریتمی که بسیار شبیه به Adadelta است، اما از ریاضیات کمی متفاوت استفاده می‌کند، RMSprop نامیده می‌شود. این نام از آن‌جایی ناشی می‌شود که از یک عملیات میانگین مربعات ریشه (root mean-squared)، که اغلب به اختصار RMS نامیده می‌شود، برای تعیین تطبیقی که به گرادیان‌ها اضافه می‌شود (یا منتشر می شود) استفاده می‌کند. نماد ریاضی آن به شرح زیر است:

$$\theta_{t+1} = \theta_t - \alpha.\frac{\nabla_t(\theta_t)}{\sqrt{E(\nabla^2)_t + \varepsilon}}$$

$$E(\nabla^2)_t = \gamma.E(\nabla^2)_{t-1} + (1-\gamma)\nabla_t(\theta_t)^2$$

### الگوریتم RMSprop در keras

با استفاده از قطعه کد زیر می‌توان از بهینه‌ساز RMSprop در Keras استفاده کرد:

```
from tensorflow.keras.optimizers import RMSprop
...
rms_prop = RMSprop(learning_rate=0.0001, epsilon=1e-6)
model.compile(optimizer= rms_prop, loss = ..., metrics= ...)
```

### Adam

الگوریتم های قبلی ایده ذخیره لیستی از گرادیان های مجذور با هر وزن را به اشتراک می‌گذارند. سپس، با جمع کردن مقادیر موجود در این لیست، شاید پس از مقیاس‌گذاری آن‌ها، یک ضریب مقیاس ایجاد می‌کنند. گرادیان در هر مرحله‌یِ بروزرسانی بر این مجموع تقسیم می‌شود . Adagrad هنگام ایجاد ضریب مقیاس‌بندی به همه عناصر موجود در لیست وزن یکسانی می‌دهد، در حالی که Adadelta و RMSprop عناصر قدیمی‌تر را کم‌اهمیت تلقی می‌کنند و بنابراین سهم کم‌تری در کل دارند.

مربع کردن گرادیان قبل از قرار دادن آن در لیست از نظر ریاضی مفید است، اما وقتی یک عدد را مربع می‌کنیم، نتیجه همیشه مثبت است. این بدان معناست که ما مسیر مثبت یا منفی بودن آن گرادیان در لیست خود را از دست می‌دهیم که اطلاعات مفیدی است. بنابراین، برای جلوگیری از از دست دادن این اطلاعات، می‌توانیم لیست دومی از گرادیان‌ها را بدون مجذور کردن آنها نگه داریم. سپس می‌توانیم از هر دو لیست برای استخراج ضریب مقیاس خود استفاده کنیم. این رویکرد الگوریتمی است به نام تخمین لحظه تطبیقی ( adaptive moment estimation) یا به طور معمول Adam.

### الگوریتم Adam در keras

با استفاده از قطعه کد زیر می‌توان از بهینه‌ساز Adam در Keras استفاده کرد:

```
from tensorflow.keras.optimizers import Adam
...
adam= Adam(learning_rate=0.0001, beta_1=0.9, beta_2=0.999, epsilon=1e-6)
model.compile(optimizer= adam, loss = ..., metrics= ...)
```

## تابع زیان

در مرحله‌یِ آموزش شبکه‌های عصبی از یک **امتیاز** (**score**) برای نشان دادن وضعیت فعلی استفاده می‌شود. بر اساس این امتیاز، پارامترهای وزن بهینه جستجو می‌شوند. به عبارت دیگر، یک شبکه عصبی با استفاده از امتیاز به عنوان راهنما، پارامترهای بهینه را جستجو می‌کند. این

امتیاز از طریق یک **تابع زیان (تابع ضرر)**، براساس اندازه‌گیری میزان خطا بین مقادیر پیش‌بینی شده و واقعی محاسبه می‌شود. فرمول ساده‌یِ زیر، تابع زیان را به عنوان یک معادله نشان می‌دهد:

$$Loss = y - \hat{y}$$

که در آن $y$ و $\hat{y}$ به ترتیب به عنوان مقدار واقعی و مقدار پیش‌بینی‌شده هستند.

**تعریف ۲.۳ تابع زیان**

کمیتی است که به صورت دوره‌ای در طول آموزش ارزیابی می‌شود و میزان پیشرفت یادگیری را بیان می‌کند.

یک شبکه عصبی از طریق یک فرآیند بهینه‌سازی آموزش داده می‌شود و از یک تابع زیان برای محاسبه خطا بین مقدار پیش‌بینی‌شده مدل و خروجی مورد انتظار (خروجی واقعی) استفاده می‌کند. برای اهداف مختلف آموزش، فرآیند بهینه‌سازی ممکن است تابع زیان را کمینه یا بیشینه کند، به این معنی که باید راه‌حل مناسبی مانند مجموعه‌ای از پارامترها را ارزیابی کند تا به ترتیب به کم‌ترین یا بالاترین نمره برسد.

با استفاده از تابع زیان می‌توان نحوه‌یِ مدل‌سازی الگوریتم را برروری داده‌ها ارزیابی کرد. انتخاب تابع زیان به نوع مساله بستگی دارد و برای مسائل مختلف طبقه‌بندی و رگرسیون این تابع زیان متفاوت خواهد بود. در یک مساله طبقه‌بندی، قصد داریم تا یک توزیع احتمالاتی برای مجموعه کلاس‌ها پیش‌بینی کنیم. حال آنکه، در مسائل رگرسیون، قصد داریم یک مقدار خاص را بیابیم.

## توابع زیان برای رگرسیون

همچنان که پیش‌تر بیان شد، مدل‌های رگرسیون با پیش‌بینی یک مقدار پیوسته به‌عنوان مثال قیمت خودرو، پیش‌بینی وام و غیره سروکار دارند. در این بخش، پرکاربردترین توابع زیان مربوط به رگرسیون فهرست شده‌اند.

### Mean Square Error

از معروف‌ترین توابع زیان در رگرسیون می‌باشد، و میانگین اختلاف مربعات بین مقادیر واقعی و پیش‌بینی شده را توسط معادله زیر محاسبه می‌کند:

$$MSE(y, \hat{y}) = \frac{1}{n}\sum_{i=1}^{n}(y_i - \hat{y}_i)^2$$

که در آن $n$ تعداد نمونه‌های آموزشی است.

با استفاده از قطعه کد زیر می‌توان از تابع زیان Mean Square Error در Keras استفاده کرد:

```
from tensorflow import keras
...
loss_fn = keras.losses.MeanSquaredError()
model.compile(loss=loss_fn, optimizer=..., metrics= ...)
```

### Mean Absolute Error

این تابع میانگین اختلاف قدرمطلق بین مقادیر واقعی و پیش‌بینی شده را توسط معادله زیر محاسبه می‌کند:

$$MAE(y,\hat{y}) = \frac{1}{n}\sum_{i=1}^{n}|y_i - \hat{y}_i|$$

با استفاده از قطعه کد زیر می‌توان از تابع زیان Mean Absolute Error در Keras استفاده کرد:

```
from tensorflow import keras
...
loss_fn = keras.losses.MeanAbsoluteError()
model.compile(loss=loss_fn, optimizer=..., metrics= ...)
```

## توابع زیان برای طبقه‌بندی

در این بخش توابع زیان مربوط به طبقه‌بندی فهرست شده‌اند.

### Cross Entropy

این تابع فاصله بین دو توزیع احتمال را محاسبه می‌کند و به صورت زیر تعریف می‌شود:

$$Cross\ Entropy(y,\hat{y}) = \frac{1}{n}\sum_{i=1}^{n} y_i \log(\hat{y})$$

با استفاده از قطعه کد زیر می‌توان از تابع زیان Cross Entropy در Keras استفاده کرد:

```
from tensorflow import keras
...
loss_fn = keras.losses.CategoricalCrossentropy()
model.compile(loss=loss_fn, optimizer=..., metrics= ...)
```

### Binary Cross Entropy

برای طبقه‌بندهایی دودویی از Binary Cross Entropy استفاده می‌شود که به صورت زیر تعریف می‌شود:

$$Binary\ Cross\ Entropy(y,\hat{y}) = -\frac{1}{n}\sum_{i=1}^{n}(y_i \log(\hat{y}) + (1-y_i)(1-\log(\hat{y})))$$

با استفاده از قطعه کد زیر می‌توان از تابع زیان Binary Cross Entropy در Keras استفاده کرد:

```
from tensorflow import keras
...
loss_fn = keras.losses.binary_crossentropy
model.compile(loss=loss_fn, optimizer=..., metrics= ...)
```

## پس‌انتشار

یادگیری در یک شبکه عصبی با چندین لایه به‌طور کلی شبیه نحوه یادگیری یک پرسپترون است، با تطبیق وزن‌ها. با این حال، اینکه چگونه هر وزن باید تغییر کند کمی دشوارتر است. در مورد یک پرسپترون، برای بدست آوردن خروجی یک گره، تنها یک ضرب داخلی بین داده‌های ورودی و وزن‌ها مورد نیاز است. از آنجایی که در یک MLP چندین لایه وجود دارد، خروجی یک گره در آخرین لایه با گرفتن ضرب داخلی وزن‌ها و خروجی گره‌ها در لایه قبلی تعیین می‌شود. این مقادیر اخیر هر کدام به یک روش محاسبه می‌شوند. این بدان معنی است که خطای بدست آمده در خروجی لایه نهایی می تواند ناشی از وزن‌های لایه آخر و همچنین وزن‌های لایه(های) قبلی باشد. بنابراین سوال این است که مسبب این خطای بدست آمده، وزن‌های کدام لایه است. گاهی اوقات به این مشکل **تخصیص اعتبار** (**credit assignment**) نیز می‌گویند. روشی که می‌تواند این مشکل را حل کند، **پس‌انتشار** نام دارد.

الگوریتم پس انتشار احتمالا اساسی‌ترین بلوک سازنده در یک شبکه عصبی است. پس‌انتشار اساسا تدبیر هوشمندانه‌ای برای محاسبه مؤثر گرادیان در شبکه‌های عصبی چندلایه است. این الگوریتم از قاعده زنجیره‌ای حساب دیفرانسیل استفاده می‌کند و گرادیان خطا را در مسیرهای مختلف از یک گره تا خروجی محاسبه می‌کند و از دو فاز اصلی به نام فاز جلورو (پیش‌رو) و فاز عقبگرد (پس‌رو) تشکیل می‌شود. در این الگوریتم، پس از هر گذرِ جلورو در یک شبکه، پس‌انتشار یک گذر عقبگرد انجام می‌دهد و در عین حال پارامترهای مدل (وزن ها و بایاس‌ها) را تنظیم می‌کند.

الگوریتم پس انتشار اجازه می‌دهد تا گرادیان یک لایه را در یک زمان به روشی کارآمد محاسبه کنید، از مقادیر محاسبه شده مجدد استفاده کنید و از آخرین لایه از طریق قانون زنجیره‌ای عقبگرد کنید.

به‌طور خلاصه و در یک سطح بسیار بالا، بر اساس آنچه تاکنون بیان شد، یک شبکه عصبی این مراحل را در طول آموزش چندین مرتبه تکرار و اجرا می‌کند (شکل ۳ـ ۴):

- **یک فاز جلورو برای تولید خروجی بر اساس پارامترهای فعلی و داده‌های ورودی**
- **محاسبه تابع زیان برای شکاف بین خروجی‌های فعلی و خروجی‌های هدف**
- **یک فاز عقبگرد برای محاسبه گرادیان زیان نسبت به پارامترها**

- **یک مرحله بهینه‌سازی که از گرادیان‌ها برای بروزرسانی پارامترها استفاده می‌کند تا زیان برای تکرار بعدی کاهش یابد.**

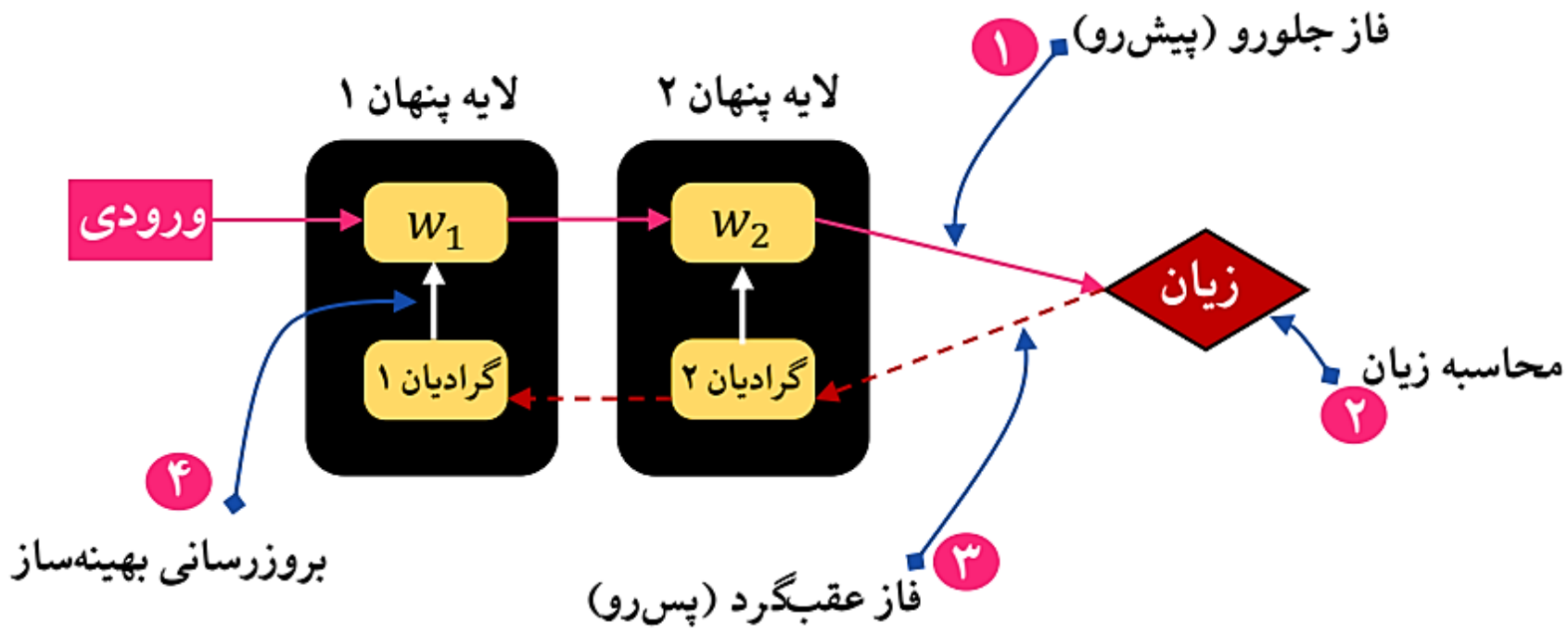

شکل ۳ــ۴. فرآیند آموزش یک شبکه عصبی

## وزن‌دهی اولیه‌ی پارامترها

یک عنصر بسیار مهم در طراحی شبکه‌های عصبی، مقداردهی اولیه وزن‌ها است. در شبکه‌های عصبی، مقداردهی اولیه‌یِ وزن‌ها باید با دقت بسیاری انتخاب شود. برای مثال، اگر چندین نورون در یک لایه‌یِ پنهان، وزن‌های مشابهی داشته باشند، گرادیان‌های یکسانی را دریافت خواهند کرد. در نتیجه، هنگام تصحیح گرادیان همه نورون‌ها به یک شکل بهسازی می‌شوند. به عبارت دیگر، برای همه آن‌ها نتایج یکسانی محاسبه می‌شود که منجر به هدر رفتن ظرفیت مدل می‌شود. بنابراین، شبکه معادل دنباله‌ای از لایه‌های تک نورونی است.

اگر پیکربندی نهایی را می‌دانستیم (یا حتی به‌طور تقریبی می‌دانستیم)، می‌توانستیم آن‌ها را طوری تنظیم کنیم که براحتی در چند تکرار به نقطه بهینه برسند. اما، متأسفانه ما نمی‌دانیم که بهینه محلی در کجا قرار دارد. از این‌رو، راهبردهای تجربی با هدف کمینه کردن زمان آموزش توسعه یافته و آزمایش شده‌اند. در حالت ایده‌آل، واریانس‌های فعال‌ساز باید تقریبا در سراسر شبکه و همچنین واریانس‌های وزن پس از هر مرحله پس‌انتشار ثابت باقی بمانند. این دو شرط به منظور بهبود فرآیند همگرایی و جلوگیری از مشکلات محو و انفجار گرادیان اساسی هستند.

به‌طور معمول، وزن‌هایِ شبکه‌های عصبی با استفاده از یک توزیع گاوسی با میانگین صفر و یک انحراف معیار کوچک مقداردهی اولیه می‌شوند. با این حال، مشکلی که وجود دارد این است که توزیع خروجی‌های یک نورونِ به‌طور تصادفی مقداردهی اولیه‌شده، دارای واریانسی است که با تعداد ورودی‌ها افزایش می‌یابد. برای نرمال کردن واریانس خروجی هر نورون به ۱، کافی است از یک توزیع نرمال استاندارد استفاده کنید و وزن را بر اساس جذر گنجایش ورودی $n_{in}$، که تعداد ورودی‌های آن است، مقیاس کنید:

$$w_0 \sim \frac{\mathcal{N}(0,1)}{\sqrt{n_{in}}}$$

این روش **مقیاس‌بندی واریانس (variance scaling)** نامیده می‌شود و می‌تواند علاوه بر استفاده از تعداد واحدهای ورودی (Fan-In)، براساس تعداد واحدهای خروجی (Fan-Out) یا میانگین آن‌ها اعمال شود. این ایده بسیار شهودی است، اگر تعداد اتصالات ورودی یا خروجی زیاد هستند، وزن‌ها باید کوچکتر باشند تا از خروجی‌های بزرگ جلوگیری شود.

با استفاده از قطعه کد زیر در Keras می‌توان از variance scaling استفاده کرد:

```
from keras import initializers
initializer = initializers.VarianceScaling(scale=1.0, mode='fan_in',
distribution='normal')
```

گلورت و بنجیو تجزیه و تحلیلی را برروی گرادیان‌های پس‌انتشار انجام دادند و یک مقداردهی اولیه (معروف به مقداردهی اولیه Xavier) را توصیه کردند:

$$w_0 \sim \sqrt{\frac{2}{n_{in} + n_{out}}}\mathcal{N}(0,1)$$

جایی که $n_{out}$ تعداد واحدهای خروجی را توصیف می‌کند. این یک نوع قوی‌تر از روش قبلی است، چراکه هم اتصالات ورودی و هم اتصالات خروجی (که به نوبه خود اتصالات ورودی هستند) را در نظر می‌گیرد. هدف، تلاش برای برآوردن دو الزام ارائه شده قبلی است. ثابت شده است که مقداردهی اولیه Xavier در بسیاری از معماری‌های عمیق بسیار موثر است و اغلب انتخاب پیش‌فرض است.

با استفاده از قطعه کد زیر در Keras می‌توان از Xavier استفاده کرد:

```
from keras import initializers
initializer = initializers.GlorotNormal()
```

هدف از مقداردهی اولیه وزن‌ها، جلوگیری از انفجار یا محو خروجی‌های فعال‌سازها در طول مسیر عبور از یک شبکه عصبی عمیق است. اگر یکی از این دو اتفاق بیفتد، گرادیان‌های خطا (زیان) یا خیلی بزرگ یا خیلی کوچک خواهند بود که به‌طور سودمند به عقب جریان پیدا نمی‌کنند. حتی اگر اصلا بتواند این کار را انجام دهد شبکه زمان بیشتری نیاز دارد تا همگرا شود.

همه این روش‌ها اصول مشترکی دارند و در بسیاری از موارد قابل تعویض هستند. همان‌طور که قبلا ذکر شد، Xavier یکی از قوی‌ترین هاست و در اکثر مسائل کارکرد خوبی دارد. با این حال، فقط اعتبارسنجی برروی مجموعه داده واقعی می‌تواند این را تأیید کند.

## پارامترها

در هر شبکه عصبی عمیق، دو نوع پارامتر مختلف وجود دارد: یک نوع پارامتر مدل و دیگری **اَبَرپارامتر** (**hyper parameter**). پارامترهای مدل آن دسته از پارامترهایی هستند که به‌طور خودکار توسط مدل از داده‌های آموزشی شناسایی می‌شوند. در مقابل، اَبَرپارامترها آن دسته از پارامترهایی هستند که قابل تنظیم هستند و باید برای بدست آوردن بهترین عملکرد مدل تنظیم شوند. به عبارت دیگر، این پارامترها در طول آموزش آموخته نمی‌شوند، اما در شروع فرآیند یادگیری توسط کاربر تنظیم می‌شوند. اَبَرپارامترها کل فرآیند یادگیری در شبکه عصبی را تحت تاثیر قرار می‌دهند. برخی از اَبَرپارامترها شامل تعداد لایه‌های پنهان است که ساختار شبکه را تعیین می‌کند. نرخ یادگیری یک اَبَرپارامتر دیگر است که به درک نحوه آموزش شبکه کمک می‌کند. انتخاب اَبَرپارامتر بهینه نقش مهمی در کل فرآیند آموزش شبکه دارد.

**تعریف ۳.۳ ابرپارامتر**

**ابرپارامترها، آرگومان‌های مدل هستند که مقدار آن‌ها قبل از شروع فرآیند یادگیری تنظیم می‌شود.**

ابرپارامترها را می‌توان به‌عنوان داشبوردی با کلیدها و شماره‌گیری‌هایی در نظر بگیرید که نحوه‌ی عملکرد الگوریتم را کنترل می‌کنند. تنظیم ابرپارامترهای مختلف اغلب به مدل‌هایی با عملکرد متفاوت منجر می‌شوند.

## تعمیم و منظم‌سازی

هدف اصلی از ساخت یک مدل یادگیری عمیق، یادگیری یک تابع یا یک وظیفه (task) می‌باشد. برای دستیابی به این هدف، به شبکه داده‌های آموزشی را تغذیه می‌کنیم. پس از آموزش یک شبکه‌ی عصبی با گرادیان کاهشی و پس‌انتشار، فرض می‌کنیم که این عملکرد خوب روی داده‌های دیده‌نشده باقی می‌ماند (یعنی داده‌هایی که در طول آموزش درگیر نشده‌اند). حال آن‌که، لزوما این بدان معنا نیست که مدل قادر به پیش‌بینی درست خروجی برای داده‌های دیده‌نشده هم باشد. بنابراین، دو مجموعه اضافی از داده‌ها برای بهینه‌سازی معرفی می‌شوند، مجموعه اعتبارسنجی و مجموعه آزمایشی. هر سه مجموعه داده از هم مستقل هستند، به طوری که هیچ نمونه‌ای در بین آن‌ها مشترک نیست.

مجموعه اعتبارسنجی در شبکه‌های عصبی، معمولا برای تنظیم دقیق ابرپارمترهای مدل مانند معماری شبکه یا نرخ یادگیری استفاده می‌شود. مجموعه آزمون فقط برای ارزیابی نهایی در راستایِ بررسی عملکرد شبکه در داده‌های دیده‌نشده استفاده می‌شود. اگر یک شبکه‌ی عصبی به خوبی **تعمیم** نیابد (قابلیت انتقال دانش به داده‌های غیرقابل مشاهده را تعمیم گویند)، یعنی زیانِ آموزشِ کم‌تری نسبت به زیانِ آزمون داشته باشد، به این حالت **بیش‌برازش** گفته می‌شود.

در حالی که سناریوی معکوس، زمانی که زیان آزمون نسبت به زیان آموزش بسیار کم‌تر باشد، **کم‌برازش** نامیده می‌شود (شکل ۳ــ۵).

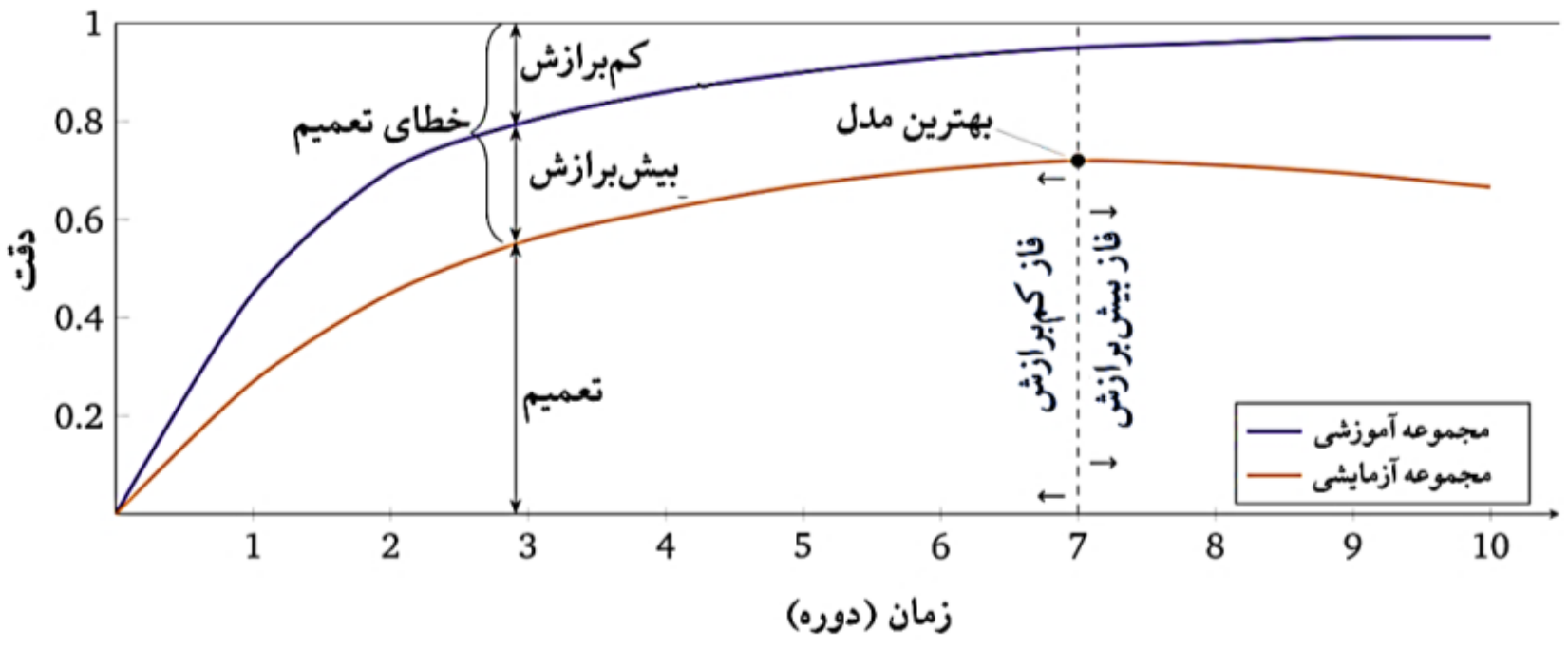

**شکل ۳ــ۵. رفتار تعمیم‌دهی در منحنی یادگیری با توجه به معیار دقت در داده‌های آموزشی و آزمون**

بیش‌برازش پدیده‌ای است که در نهایت همه شبکه‌های عصبی را تحت تاثیر قرار می‌دهد. این به دلیل این واقعیت است که آن‌ها فقط از مجموعه داده‌های آموزشی یاد می‌گیرند: *زیرمجموعه‌ای از تمام داده‌های ممکن.* این‌که آن‌ها در این زیرمجموعه چقدر خوب عمل می‌کنند، تعیین می‌کند که وزن‌های آنها چقدر پاداش یا جریمه بشود. به عبارت دیگر، حتی اگر هدف ما دانش تعمیم باشد، خود شبکه‌ها برای دستیابی به دقت بالا در مجموعه داده‌های خاصی طراحی شده‌اند. به این ترتیب، آن‌ها در نهایت با توجه به توانایی انجام این کار، شروع به حفظ مجموعه داده‌های خود خواهند کرد. این به‌خاطر سپردن باعث می‌شود شبکه منجر به بیش‌برازش شود. شبکه‌ای که بیش‌برازش را آغاز کرده است، ویژگی‌های منحصربه‌فرد و جزئیات خاص داده‌ای را که به طور انحصاری در مجموعه داده‌های آموزشی یافت می‌شود، به خاطر می‌سپارد و آن‌ها را به‌عنوان مفاهیمِ کلیِ مشترک در همه‌یِ ورودی‌های داده مشابه به اشتباه می‌گیرد. بنابراین چنین شبکه‌ای برای تجزیه و تحلیل ورودی‌ها جدید و ناآشنا (مجموعه داده آزمایشی)، زمان دشوارتری خواهد داشت. این به این دلیل است که بخشی از صفات متمایزکننده‌یِ شناسایی شده قبلی، در داده‌های جدید وجود ندارند. علاوه بر این، با افزایش دقت شبکه در طول آموزش، شکاف بین خطای ایجاد شده در طول آموزش و خطای تولید شده در طول آزمایش نیز افزایش می‌یابد. با این حال، گاهی اوقات هنگام تلاش برای حل وظایف بسیار پیچیده، استفاده از یک مدل پیچیده اجتناب‌ناپذیر است. افزودن لایه‌های بیشتر یا نورون‌ها بیشتر، سطح بیشتری از استخراج ویژگی را ممکن می‌سازد که می‌تواند منجر به یک سطح بالاتر از دقت تا یک نقطه خاص می‌شود.

به‌طور معمول، بیش‌برازش و کم‌برازش در شبکه‌های عصبی عمیق، مستقیما با ظرفیت مدل مرتبط است. به زبان ساده، ظرفیتِ مدلِ یک شبکه‌یِ عصبیِ عمیق، به‌طور مستقیم با تعداد پارامترهای داخل شبکه در ارتباط است. ظرفیت مدل تعیین می‌کند که یک شبکه عمیق تا چه

حد قادر به برازش با طیف گسترده‌ای از توابع است. اگر ظرفیت خیلی کم باشد، شبکه ممکن است نتواند مجموعه آموزشی را تطبیق دهد (کم‌برازش)، در حالی که ظرفیت مدل خیلی بزرگ ممکن است منجر به حفظ نمونه‌های آموزشی (بیش‌برازش) شود. *کم‌برازش معمولاً برای شبکه‌های عصبی عمیق، مشکل چندانی ندارد. چراکه این مشکل را می‌توان با استفاده از معماری شبکه‌یِ قوی‌تر یا عمیق‌تر با پارامترهای بیشتر برطرف کرد.* با این حال، برای اینکه بتوان از شبکه‌های عمیق برای داده‌های جدید و دیده‌نشده استفاده کرد، باید بیش‌برازش را کنترل کرد. فرآیند کاهش اثر بیش‌برازش یا جلوگیری از آن را **منظم‌سازی** (**regularization**) می‌گویند. منظم‌سازی روشی برای کنترل بیش‌برازش یا بهتر است بگوییم **بهبود خطای تعمیم** است.

مناسب بودن داده‌های آموزشی را نیز نباید نادیده گرفت. چراکه موفقیت در تعمیم یا حتی برازش کافی در داده‌های آموزشی به این امر بستگی دارد. در غیر این صورت، مدل ممکن است تمایل داشته باشد که بیش از حد با ویژگی‌هایِ خاصِ داده‌هایِ آموزشی سازگار شود. این امر از یک طرف به مقدار داده‌های موجود برای آموزش بستگی دارد، چراکه ممکن است یک مجموعه آموزشی کوچک برای تشخیص الگوها و ساختارهای کلی کافی نباشد و از طرف دیگر به کیفیت داده‌های آموزشی؛ به ویژه در مورد یادگیری بانظارت در موردِ صحتِ برچسب‌هایِ هدف که از قبل توسط انسان یا حتی متخصصان انسانی تنظیم شده است. علاوه براین، اطمینان از اینکه توزیع و ویژگی‌های داده‌های آموزشی با داده‌های آزمون مطابقت دارد یا به طور کلی با داده‌هایی که مدل آموخته‌شده برای استفاده در آینده برنامه‌ریزی شده است مطابقت داشته باشد، ضروری است.

## توقف زودهنگام (Early stopping)

در آموزش مدل‌های بزرگ، معمولا خطای آموزش و اعتبارسنجی در طول زمان کاهش می‌یابد، اما در یک نقطه، خطایِ اعتبارسنجی شروع به افزایش می‌کند. در این مرحله، مدل شروع به بیش‌برازش می‌کند و ویژگی‌های خاص مجموعه آموزشی را یاد می‌گیرد. برای توقف در این مرحله از روش توقف زودهنگام استفاده می‌شود. این روش، مدلی که کم‌ترین خطای اعتبارسنجی را دارد برمی‌گرداند. بنابراین، آموزش به یک مجموعه اعتبارسنجی نیاز دارد تا به صورت دوره‌ای خطای اعتبارسنجی را ارزیابی کند.

در این روش، آموزش پس از اولین افزایش خطای اعتبارسنجی متوقف نمی‌شود. بلکه، شبکه تا رسیدن به آستانه "**تعداد دوره‌های بدون پیشرفت**" آموزش بیشتری می‌بیند. سپس، از طریق ارزیابی دوره‌های بعدی، روندِ خطایِ اعتبارسنجی را برای آموزش بیشتر دریافت می‌کنیم. به عنوان مثال، اگر ۱۰ بار متوالی خطای اعتبارسنجی نسبت به بهترین خطای اعتبارسنجی هیچ پیشرفتی نداشته باشد، آموزش متوقف می‌شود و مدلی که بهترین خطای اعتبارسنجی را دارد برگردانده می‌شود.

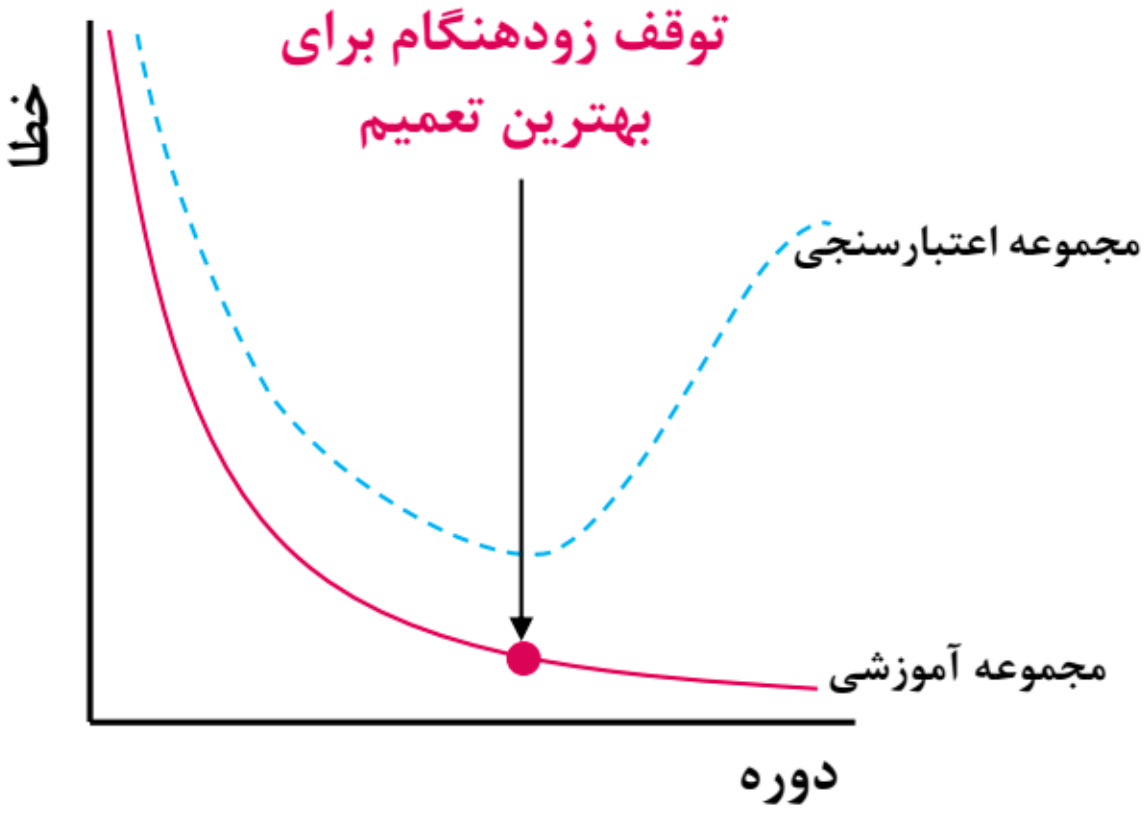

**شکل ۳ـ۶. توقف زودهنگام**

توقف زودهنگام، به نظارت عملکرد مدل در هر دوره براساس مجموعه اعتبارسنجی می‌پردازد و خاتمه آموزش را مشروط به عملکرد اعتبارسنجی می‌داند.

## حذف تصادفی (Dropout)

حذف تصادفی، یک روش منظم‌سازی برای شبکه‌های عصبی است. ایده کلیدی این است که نورون‌ها به‌صورت تصادفی در هر تکرار آموزش حذف شوند. اگر یک نورون حذف شود، همه گرادیان‌های وابسته صفر هستند و بنابراین وزن مربوط بروز نمی‌شود. فرآیند اجرای این روش به این صورت است که در هر دوره تکرار آموزش، با یک احتمال $p$ هر نرون باقی و با احتمال $(1-p)$ از شبکه حذف خواهد شد. این عمل باعث خواهد شد که در هر دوره تمام ویژگی‌ها یادگرفته نشوند و با هربار ورود یک داده ویژگی‌های متفاوتی از آن برای طبقه‌بندی استفاده شود. در شکل ۳ـ۴ یک شبکه عصبی معمولی و یک شبکه عصبی همراه با حذف تصادفی قابل مشاهده است.

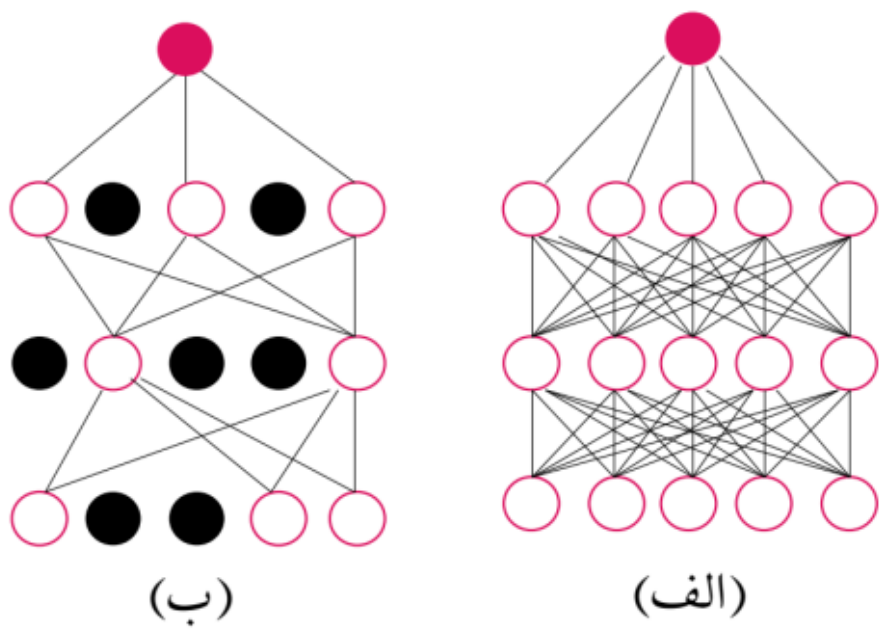

**شکل ۳ـ۶. یک شبکه عصبی (الف) قبل از حذف تصادفی و (ب) بعد از حذف تصادفی**

متداول‌ترین تفسیر از اثر حذف تصادفی این است که به طور ضمنی **گروهی (ensemble)** از مدل‌ها را آموزش می‌دهد. چراکه در هر تکرار نسخه متفاوتی از مدل را ایجاد می‌کند و هر وزن با مجموعه وزن‌های دیگری بروز می‌شود. یادگیریِ گروهی از چندین مدل، یک تکنیک رایج در یادگیری ماشین برای کاهش خطای تعمیم با این ایده است که یک پیش‌بینی نادرست از یک مدل واحد می‌تواند توسط مدل‌های دیگر جبران شود.

از سوی دیگر، حذف تصادفی را می‌توان اینگونه تفسیر کرد که شبکه عصبی مجبوربه بازنمایی اضافی از دانش بدست آمده از طریق یادگیری است. چراکه دانش خاصی در مورد کلاس‌ها یا ورودی‌های خاص لزوماً در برخی از تکرارها در دسترس نیست، به‌دلیل اینکه دانشی که در این نورون‌ها برای این ورودی‌ها رمزگذاری شده است، در حال حاضر حذف شده‌اند. از این‌رو، برای یک شبکه عصبی دشوار است تا برروی نمونه‌های آموزشی خاص، بیش‌برازش کند، چراکه برخی نورون‌های خاص همیشه قابل دستیابی نیستند.

با استفاده از قطعه کد زیر در Keras می‌توان به حذف تصادفی دسترسی پیدا کرد:

```
from keras.layers import Dropout
from keras.models import Sequential

model = Sequential()
...
model.add(Dropout(0.5))
```

## نرمال‌سازی دسته‌ای (BatchNormalization)

آموزش یک شبکه، وزن‌های هر لایه را تغییر می‌دهد. این تغییر باعث می‌شود که **توزیع (distribution)** ورودی در طول بروزرسانی لایه‌های قبلی تغییر کند، این اثر **تغییر متغیر داخلی (internal covariate shift)** نامیده می‌شود. این مشکل از آنجا ناشی می‌شود که پارامترها در طول فرآیند آموزش مدام تغییر می‌کنند، این تغییرات به نوبه خود مقادیر توابع فعال‌سازی را تغییر می‌دهد. تغییر مقادیر ورودی از لایه‌های اولیه به لایه‌های بعدی سبب همگرایی کندتر در طول فرآیند آموزش می‌شود، چرا که داده‌های آموزشی لایه‌های بعدی پایدار نیستند. به عبارت دیگر، شبکه‌های عمیق ترکیبی از چندین لایه با توابع مختلف بوده و هر لایه فقط یادگیری بازنمایی کلی از ابتدای آموزش را فرا نمی‌گیرد، بلکه باید با تغییر مداوم در توزیع‌های ورودی با توجه به لایه‌های قبلی تسلط پیدا کند. حال آن‌که بهینه‌ساز بر این فرض بروزرسانی پارامترها را انجام می‌دهد که در لایه‌های دیگر تغییر نکنند و تمام لایه‌ها را همزمان بروز می‌کند، این عمل سبب نتایج ناخواسته‌ای هنگام ترکیب توابع مختلف خواهد شد. در راستای مقابله با تغییر توزیع در طول یادگیری، **نرمال‌سازی دسته‌ای** معرفی شد. در این روش، نرمال‌سازی برروی داده‌های ورودی یک لایه را به گونه‌ای انجام می‌دهد، که دارای میانگین صفر و انحراف معیار یک شوند.

با قرار دادن نرمال‌سازی دسته‌ای بین لایه‌های پنهان و با ایجاد ویژگی واریانس مشترک، سبب کاهش تغییرات داخلی لایه‌های شبکه می‌شویم.

از طریق اعمال نرمال‌سازی دسته‌ای می‌توان میزان نرخ یادگیری را افزایش داد و این امر منجر به آموزش سریع‌تر می‌شود. علاوه بر این، دقت در مقایسه با همان شبکه بدون نرمال‌سازی دسته‌ای در حال افزایش است.

با استفاده از قطعه کد زیر در Keras می‌توان از نرمال‌سازی دسته‌ای استفاده کرد:

```
from keras.layers import BatchNormalization
from keras.models import Sequential

model = Sequential()
...
model.add(BatchNormalization())
```

## پیاده‌سازی شبکه عصبی در keras

در این بخش نحوه پیاده‌سازی شبکه عصبی پیش‌خور در Keras را خواهیم آموخت. برای اولین مثال، شبکه عصبی برای پیش‌بینی اینکه قیمت خانه‌ها بالاتر یا پایین‌تر از مقدار متوسط است را می‌سازیم.

مجموعه داده‌ای که استفاده خواهیم کرد از داده‌های[1] مسابقه Kaggle برای پیش‌بینی ارزش خانه زیلو اقتباس شده است. ما تعداد ویژگی‌های ورودی را کاهش داده‌ایم و کار را به پیش‌بینی اینکه قیمت خانه بالاتر یا کم‌تر از مقدار متوسط است تغییر داده‌ایم. برای دانلود[2] مجموعه داده اصلاح شده از پیوندی که در پانویس قرار دارد استفاده کنید.

قبل از کدنویسی برای ساخت هر مدل یادگیری ماشین، اولین کاری که باید انجام دهیم این است که داده‌های خود را در قالبی قرار دهیم که برای الگوریتم مناسب باشد. برای این مثال، کارهای زیر را انجام می‌دهیم:

- ابتدا فایل CSV را می‌خوانیم و آن‌ها را به آرایه تبدیل می‌کنیم. آرایه‌ها فرمت داده‌ای هستند که الگوریتم ما می‌تواند آن‌ها را پردازش کند.
- مجموعه داده خود را به ویژگی‌های ورودی که آن را $x$ و برچسب که آن را $y$ می‌نامیم تقسیم می‌کنیم.
- داده‌ها را نرمال‌سازی می‌کنیم.
- مجموعه داده را به مجموعه آموزشی، مجموعه اعتبارسنجی و مجموعه آزمون تقسیم می‌کنیم.

---

[1] https://www.kaggle.com/c/zillow-prize-1/data
[2] https://drive.google.com/file/d/1h6LPHNs4F_FnxwfdE_fCIsGeEh30tDBf/view?usp=sharing

بیایید شروع کنیم! ابتدا کتابخانه pandas را وارد می‌کنیم، کد زیر را در سلول notebook خود تایپ کرده و Alt+Enter را فشار دهید:

```
import pandas as pd
```

این فقط به این معنا است که اگر بخواهم به کد موجود در بسته "pandas" اشاره کنم، آن را با نام pd ارجاع خواهم داد. سپس با اجرای کد زیر فایل CSV خود را می‌خوانیم:

```
df = pd.read_csv('housepricedata.csv')
```

این خط کد به این معنا است که ما فایل `'housepricedata.csv'` را می‌خوانیم و آن را در متغیر `df` ذخیره می‌کنیم. اگر بخواهیم بفهمیم در `df` چه چیزی وجود دارد، کافی است `df` را در سلول notebook تایپ کرده و Alt+Enter را فشار دهید:

```
df
```

خروجی شما باید چیزی شبیه به این باشد:

| | LotArea | OverallQual | OverallCond | TotalBsmtSF | FullBath | HalfBath | BedroomAbvGr | TotRmsAbvGrd | Fireplaces | GarageArea | AboveMedianPrice |
|---|---|---|---|---|---|---|---|---|---|---|---|
| 0 | 8450 | 7 | 5 | 856 | 2 | 1 | 3 | 8 | 0 | 548 | 1 |
| 1 | 9600 | 6 | 8 | 1262 | 2 | 0 | 3 | 6 | 1 | 460 | 1 |
| 2 | 11250 | 7 | 5 | 920 | 2 | 1 | 3 | 6 | 1 | 608 | 1 |
| 3 | 9550 | 7 | 5 | 756 | 1 | 0 | 3 | 7 | 1 | 642 | 0 |
| 4 | 14260 | 8 | 5 | 1145 | 2 | 1 | 4 | 9 | 1 | 836 | 1 |
| ... | ... | ... | ... | ... | ... | ... | ... | ... | ... | ... | ... |
| 1455 | 7917 | 6 | 5 | 953 | 2 | 1 | 3 | 7 | 1 | 460 | 1 |
| 1456 | 13175 | 6 | 6 | 1542 | 2 | 0 | 3 | 7 | 2 | 500 | 1 |
| 1457 | 9042 | 7 | 9 | 1152 | 2 | 0 | 4 | 9 | 2 | 252 | 1 |
| 1458 | 9717 | 5 | 6 | 1078 | 1 | 0 | 2 | 5 | 0 | 240 | 0 |
| 1459 | 9937 | 5 | 6 | 1256 | 1 | 1 | 3 | 6 | 0 | 276 | 0 |

1460 rows × 11 columns

ویژگی‌های ورودی ما در ده ستون اول هستند. در آخرین ستون، ویژگی (برچسب) را داریم که می‌خواهیم پیش‌بینی کنیم: آیا قیمت خانه بالاتر از میانگین است یا خیر؟ (۱ برای بله و ۰ برای خیر). اکنون که دیدیم داده‌ها چگونه به نظر می‌رسند، می‌خواهیم آن‌ها را به آرایه‌هایی تبدیل کنیم تا ماشین آن‌ها را پردازش کند:

```
dataset = df.values
```

برای تبدیل دیتافریم (dataframe) خود به آرایه، فقط مقادیر `df` را (با `df.values`) در متغیر `dataset` ذخیره می‌کنیم. برای دیدن آنچه در داخل متغیر "`dataset`" وجود دارد، کافی است `dataset` را در سلول notebook خود تایپ کنید و سلول را اجرا کنید (Alt+Enter):

```
dataset
```

همان‌طور که می‌بینید، اکنون همه در یک آرایه ذخیره می‌شوند:

```
array([[ 8450,     7,     5, ...,     0,   548,     1],
       [ 9600,     6,     8, ...,     1,   460,     1],
       [11250,     7,     5, ...,     1,   608,     1],
       ...,
```

```
[ 9042,     7,      9, ...,      2,    252,       1],
[ 9717,     5,      6, ...,      0,    240,       0],
[ 9937,     5,      6, ...,      0,    276,       0]])
```

اکنون مجموعه داده خود را به ویژگی‌های ورودی (X) و ویژگی که می‌خواهیم پیش‌بینی کنیم (Y) تقسیم می‌کنیم. برای انجام این تقسیم، ما به سادگی ۱۰ ستون اول آرایه خود را به متغیری به نام X و آخرین ستون آرایه خود را به متغیری به نام Y اختصاص می‌دهیم. کد انجام اولین انتساب به این صورت است:

```
X = dataset[:,0:10]
```

این ممکن است کمی عجیب به نظر برسد، اما اجازه دهید تا آن‌را شرح دهم که چه چیزی در داخل [] قرار دارد. همه چیز قبل از کاما (,) به ردیف‌های آرایه و همه چیز بعد از کاما به ستون‌های آرایه اشاره دارد. از آنجایی که سطرها را از هم جدا نمی‌کنیم، ":" را قبل از کاما قرار می‌دهیم. این به این معنا است که تمام سطرهای مجموعه داده را برداریم و آن را در X قرار دهیم. می‌خواهیم ۱۰ ستون اول را استخراج کنیم، بنابراین "`0:10`" بعد از کاما به معنای گرفتن ستون‌های ۰ تا ۹ و قرار دادن آن در X است (ستون ۱۰ را شامل نمی‌شود). ستون‌های ما از شاخص ۰ شروع می‌شوند، بنابراین ۱۰ ستون اول ستون‌های ۰ تا ۹ هستند.

سپس آخرین ستون آرایه خود را به Y اختصاص می‌دهیم:

```
Y = dataset[:,10]
```

اکنون مجموعه داده خود را به ویژگی‌های ورودی (X) و برچسب، یعنی آنچه می‌خواهیم پیش‌بینی کنیم (Y) تقسیم کرده‌ایم. مرحله بعدی پردازش این است که اطمینان حاصل کنیم که مقیاس ویژگی‌های ورودی مشابه است. در حال حاضر، ویژگی‌هایی مانند مساحت زمین به صورت هزار، امتیاز برای کیفیت کلی از ۱ تا ۱۰ متغیر است و تعداد شومینه ها ۰، ۱ یا ۲ است. این امر شروع اولیه شبکه عصبی را دشوار می‌کند که باعث ایجاد برخی مشکلات عملی می‌شود. یکی از راه‌های مقیاس‌سازی داده‌ها استفاده از بسته scikit-learn است. ابتدا آن را وارد می‌کنیم:

```
from sklearn import preprocessing
```

کد بالا می‌گوید که من می‌خواهم از کد `preprocessing` در بسته `sklearn` استفاده کنم. سپس، از تابعی به نام مقیاس‌کننده min-max استفاده می‌کنیم که مجموعه داده را به‌گونه‌ای مقیاس‌بندی می‌کند که همه ویژگی‌های ورودی بین ۰ و ۱ قرار بگیرند:

```
min_max_scaler = preprocessing.MinMaxScaler()
X_scale = min_max_scaler.fit_transform(X)
```

توجه داشته باشید که ما ۰ و ۱ را برای کمک به آموزش شبکه عصبی خود انتخاب کردیم. اکنون مجموعه داده‌های مقیاس‌شده ما در آرایه `X_scale` ذخیره می‌شود. اگر می‌خواهید ببینید `X_scale` چه شکلی است، به سادگی سلول زیر را اجرا کنید:

```
X_scale
```

بعد از اجرای کد بالا، خروجی زیر را خواهید دید:

```
array([[0.0334198 , 0.66666667, 0.5       , ..., 0.5       , 0.        ,
        0.3864598 ],
       [0.03879502, 0.55555556, 0.875     , ..., 0.33333333, 0.33333333,
        0.32440056],
       [0.04650728, 0.66666667, 0.5       , ..., 0.33333333, 0.33333333,
        0.42877292],
       ...,
       [0.03618687, 0.66666667, 1.        , ..., 0.58333333, 0.66666667,
        0.17771509],
       [0.03934189, 0.44444444, 0.625     , ..., 0.25      , 0.        ,
        0.16925247],
       [0.04037019, 0.44444444, 0.625     , ..., 0.33333333, 0.        ,
        0.19464034]])
```

اکنون، به آخرین مرحله خود در پردازش داده‌ها رسیده‌ایم، یعنی تقسیم مجموعه داده به یک مجموعه آموزشی، یک مجموعه اعتبارسنجی و یک مجموعه آزمون (آزمایشی). برای این منظور از کد scikit-learn به نام "train_test_split" استفاده خواهیم کرد، که همان‌طور که از نام آن پیداست، مجموعه داده ما را به یک مجموعه آموزشی و یک مجموعه آزمایشی تقسیم می‌کند. ابتدا کد مورد نیاز خود را وارد می‌کنیم:

```
from sklearn.model_selection import train_test_split
```

سپس مجموعه داده خود را به صورت زیر تقسیم می‌کنیم:

```
X_train, X_val_and_test, Y_train, Y_val_and_test =
train_test_split(X_scale, Y, test_size=0.3)
```

قطعه کد بالا به scikit-learn می‌گوید که اندازه val_and_test ما ۳۰٪ از کل مجموعه داده خواهد بود. کد، داده‌های تقسیم شده را در چهار متغیر اول در سمت چپ علامت مساوی ذخیره می‌کند. متأسفانه، این تابع فقط به ما کمک می‌کند مجموعه داده خود را به دو قسمت تقسیم کنیم. از آنجایی که ما یک مجموعه اعتبارسنجی و مجموعه آزمون جداگانه می‌خواهیم، می‌توانیم از همان تابع برای انجام دوباره تقسیم در val_and_test استفاده کنیم:

```
X_val, X_test, Y_val, Y_test = train_test_split(X_val_and_test,
Y_val_and_test, test_size=0.5)
```

کد بالا اندازه val_and_test را به‌طور مساوی به مجموعه اعتبارسنجی و مجموعه آزمون تقسیم می‌کند. به طور خلاصه، ما اکنون در مجموع شش متغیر برای مجموعه داده‌های خود داریم که از آن‌ها استفاده خواهیم کرد:

- X_train
- X_val
- X_test
- Y_train
- Y_val
- Y_test

اگر می‌خواهید ببینید که شکل آرایه‌ها برای هر یک از آن‌ها چگونه است (یعنی چه ابعادی دارند)، به سادگی کد زیر را اجرا کنید:

```
print(X_train.shape, X_val.shape, X_test.shape, Y_train.shape,
Y_val.shape, Y_test.shape)
```

```
(1022, 10) (219, 10) (219, 10) (1022,) (219,) (219,)
```

همان‌طور که می‌بینید، مجموعه آموزشی دارای ۱۰۲۲ نقطه داده است درحالی که مجموعه اعتبارسنجی و آزمون هر کدام دارای ۲۱۹ نقطه داده هستند. متغیرهای X دارای ۱۰ ویژگی ورودی هستند، در حالی که متغیرهای Y فقط یک ویژگی برای پیش‌بینی دارند.

اکنون نوبت به ساخت و آموزش اولین شبکه عصبی ما رسیده است. اولین کاری که باید انجام دهیم این است که معماری را پیکره‌بندی کنیم. فرض کنید شبکه عصبی با معماری به شکل زیر می‌خواهیم:

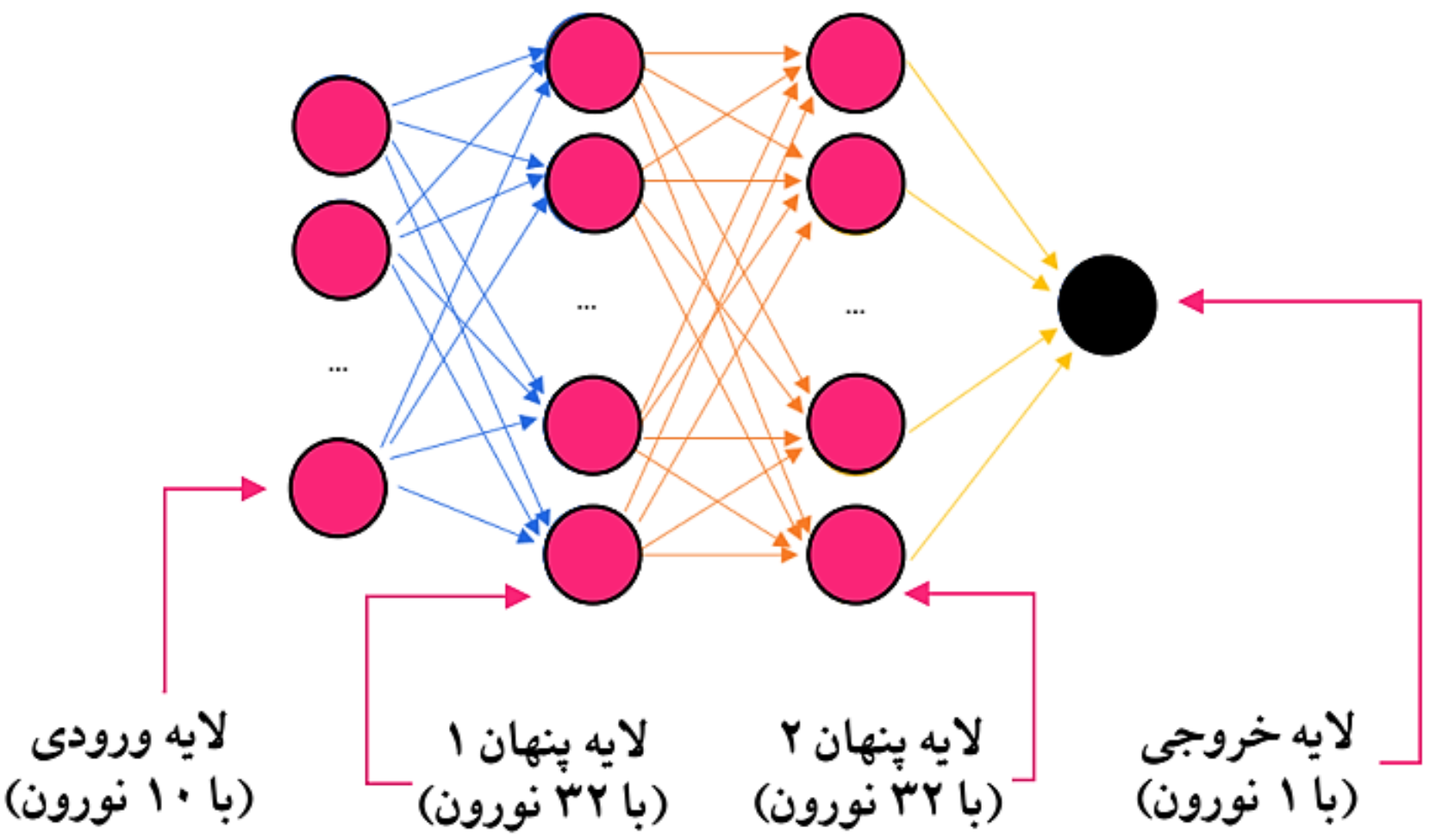

به عبارت دیگر، می‌خواهیم این لایه‌ها را داشته باشیم:

- لایه پنهان ۱: ۳۲ نورون با تابع فعال‌سازی ReLU
- لایه پنهان ۲: ۳۲ نورون با تابع فعال‌سازی ReLU
- لایه خروجی: ۱ نورون با تابع فعال‌سازی Sigmoid

حال باید این معماری را در Keras توصیف کنیم. ما از مدل ترتیبی (Sequential) استفاده خواهیم کرد، به این معنا که تنها باید لایه‌های بالا را به ترتیب توصیف کنیم. ابتدا، بیایید کد لازم را از Keras وارد کنیم:

```
from keras.models import Sequential
from keras.layers import Dense
```

سپس، مشخص می‌کنیم که مدل ترتیبی ما به این صورت است:

```
model = Sequential([
    Dense(32, activation='relu', input_shape=(10,)),
    Dense(32, activation='relu'),
    Dense(1, activation='sigmoid'),
])
```

دقیقا مانند شکل قبلی که معماری خود راترسیم کرده‌ایم، قطعه کد بالا همین معماری را تعریف کرده است. قطعه کد بالا را می‌توان اینگونه تفسیر کرد:

```
model = Sequential([...])
```

این کد می‌گوید که ما مدل خود را در متغیر model ذخیره می‌کنیم و آن را به صورت متوالی (لایه به لایه) در بین براکت‌ها توصیف می‌کنیم.

```
Dense(32, activation='relu', input_shape=(10,))
```

در اولین لایه، یک لایه کاملا متصل با ۳۲ نورون داریم، تابع فعال‌سازی ReLU و شکل (shape) ورودی ۱۰ است، چراکه ما ۱۰ ویژگی ورودی داریم. توجه داشته باشید که "**Dense**" به یک لایه کاملا متصل اشاره دارد.

```
Dense(32, activation='relu'),
```

لایه دوم ما نیز یک لایه کاملا متصل با ۳۲ نورون و تابع فعال‌سازی ReLU است. توجه داشته باشید که ما مجبور نیستیم شکل ورودی را توصیف کنیم، چراکه Keras می‌تواند از خروجی لایه اول ما این را نتیجه بگیرد.

```
Dense(1, activation='sigmoid'),
```

لایه سوم ما یا همان لایه خروجی یک لایه کاملا متصل با ۱ نورون و تابع فعال‌سازی sigmoid است. همین‌طور که دیدید توانستیم معماری مدل خود را به صورت کد بنویسیم.

اکنون که معماری خود را مشخص کرده‌ایم، باید بهترین پارامترها را برای آن پیدا کنیم. قبل از شروع آموزش، باید مدل را توسط موارد زیر پیکربندی کنیم:

- **به او بگویید از کدام الگوریتم می‌خواهید برای انجام بهینه‌سازی استفاده کنید.**
- **به او بگویید از چه تابع زیانی استفاده کند.**
- **به آن بگویید که چه معیارهای دیگری را می‌خواهید جدا از تابع زیان ردیابی کنید.**

برای پیکربندی مدل با این تنظیمات، باید تابع model.compile را فراخوانی کنیم، به این صورت:

```
model.compile(optimizer='sgd',
              loss='binary_crossentropy',
              metrics=['accuracy'])
```

تنظیمات زیر را بعد از `model.compile` داخل براکت‌ها قرار می‌دهیم:

```
optimizer='sgd',
```

'sgd' به گرادیان کاهشی تصادفی اشاره دارد (در اینجا، به گرادیان کاهشی ریزدسته‌ای اشاره دارد.

```
loss='binary_crossentropy',
```

برای خروجی‌هایی که مقادیر ۱ یا ۰ را می‌گیرند، از تابع زیان 'binary_crossentropy' (آنتروپی متقاطع دودویی) استفاده می‌شود.

```
metrics=['accuracy']
```

در نهایت، ما می‌خواهیم دقت را نیز همراه با تابع زیان ردیابی کنیم. حالا وقتی این سلول را اجرا کردیم، آماده آموزش هستیم!

آموزش شبکه در keras بسیار ساده است و از ما می‌خواهد تنها یک خط کد بنویسیم:

```
hist = model.fit(X_train, Y_train,
          batch_size=32, epochs=100,
          validation_data=(X_val, Y_val))
```

برای این کار از تابع "fit" استفاده می‌کنیم که برازش پارامترها به داده‌ها را انجام می‌دهد. باید مشخص کنیم که روی چه داده‌هایی آموزش می‌دهیم که توسط X_train و Y_train مشخص شده‌اند. سپس، اندازه ریزدسته (توسط پارامتر batch_size) خود را مشخص می‌کنیم و مدت زمانی که می‌خواهیم آن را آموزش دهیم (epochs) مشخص می‌کنیم. در نهایت، ما مشخص می‌کنیم که داده‌های اعتبارسنجی ما چیست تا مدل به ما بگوید در هر نقطه در مورد داده‌های اعتبارسنجی چگونه عمل می‌شود. این تابع یک تاریخچه را تولید می‌کند که آن را در متغیر hist ذخیره می‌کنیم. زمانی که به مصورسازی رسیدیم، از این متغیر استفاده خواهیم کرد. حالا، سلول را اجرا کنید و آموزش آن را تماشا کنید! خروجی شما باید به شکل زیر باشد:

```
Epoch 1/100
32/32 [==============================] - 0s 5ms/step - loss: 0.6990 - accuracy: 0.3542 -
val_loss: 0.6974 - val_accuracy: 0.3699
Epoch 2/100
32/32 [==============================] - 0s 2ms/step - loss: 0.6955 - accuracy: 0.4022 -
val_loss: 0.6943 - val_accuracy: 0.4110
Epoch 3/100
32/32 [==============================] - 0s 2ms/step - loss: 0.6926 - accuracy: 0.4706 -
val_loss: 0.6915 - val_accuracy: 0.4703
Epoch 4/100
32/32 [==============================] - 0s 2ms/step - loss: 0.6899 - accuracy: 0.5499 -
val_loss: 0.6889 - val_accuracy: 0.5616
Epoch 5/100
32/32 [==============================] - 0s 2ms/step - loss: 0.6874 - accuracy: 0.6468 -
val_loss: 0.6864 - val_accuracy: 0.6758
Epoch 6/100
32/32 [==============================] - 0s 2ms/step - loss: 0.6849 - accuracy: 0.7133 -
val_loss: 0.6842 - val_accuracy: 0.7123
Epoch 7/100
32/32 [==============================] - 0s 2ms/step - loss: 0.6828 - accuracy: 0.7524 -
val_loss: 0.6821 - val_accuracy: 0.7489
Epoch 8/100
32/32 [==============================] - 0s 2ms/step - loss: 0.6807 - accuracy: 0.7564 -
val_loss: 0.6801 - val_accuracy: 0.7717
Epoch 9/100
32/32 [==============================] - 0s 2ms/step - loss: 0.6787 - accuracy: 0.7779 -
val_loss: 0.6781 - val_accuracy: 0.8037
Epoch 10/100
32/32 [==============================] - 0s 2ms/step - loss: 0.6767 - accuracy: 0.8072 -
val_loss: 0.6761 - val_accuracy: 0.8128
Epoch 11/100
```

```
32/32 [==============================] - 0s 2ms/step - loss: 0.6746 - accuracy: 0.8317 -
val_loss: 0.6740 - val_accuracy: 0.8219
Epoch 12/100
32/32 [==============================] - 0s 2ms/step - loss: 0.6725 - accuracy: 0.8239 -
val_loss: 0.6717 - val_accuracy: 0.8265
:
:
:
Epoch 95/100
32/32 [==============================] - 0s 2ms/step - loss: 0.2931 - accuracy: 0.8865 -
val_loss: 0.3051 - val_accuracy: 0.9041
Epoch 96/100
32/32 [==============================] - 0s 2ms/step - loss: 0.2920 - accuracy: 0.8816 -
val_loss: 0.3043 - val_accuracy: 0.8995
Epoch 97/100
32/32 [==============================] - 0s 2ms/step - loss: 0.2911 - accuracy: 0.8855 -
val_loss: 0.3044 - val_accuracy: 0.9041
Epoch 98/100
32/32 [==============================] - 0s 2ms/step - loss: 0.2901 - accuracy: 0.8865 -
val_loss: 0.3030 - val_accuracy: 0.8995
Epoch 99/100
32/32 [==============================] - 0s 2ms/step - loss: 0.2896 - accuracy: 0.8806 -
val_loss: 0.3025 - val_accuracy: 0.8995
Epoch 100/100
32/32 [==============================] - 0s 2ms/step - loss: 0.2884 - accuracy: 0.8816 -
val_loss: 0.3017 - val_accuracy: 0.8995
```

اکنون می‌بینید که مدل در حال آموزش است! با مشاهده اعداد، باید بتوانید کاهش زیان و افزایش دقت را در طول زمان مشاهده کنید. در این مرحله، می‌توانید با ابرپارامترهای مختلف شبکه عصبی را آزمایش کنید. سلول‌ها را دوباره اجرا کنید تا ببینید وقتی که ابرپارامترهای خود را تغییر داده‌اید، آموزش شما چگونه تغییر می‌کند. هنگامی که از مدل نهایی خود راضی بودید، می‌توانید آن را در مجموعه آزمایشی ارزیابی کنید. برای یافتن دقت در مجموعه آزمایشی خود، این قطعه کد را اجرا می‌کنیم:

```
model.evaluate(X_test, Y_test)[1]
```

دلیل اینکه ما شاخص ۱ را بعد از تابع model.evaluate داریم این است که تابع زیان را به عنوان عنصر اول و دقت را به عنوان عنصر دوم برمی‌گرداند. از آنجایی که برای نمایش خروجی دقت کافیست، از این طریق می‌توان به آن دسترسی داشته باشید. به دلیل تصادفی بودن **نحوه تقسیم مجموعه داده‌ها** و همچنین **مقداردهی اولیه وزن‌ها**، هر بار که notebook خود را اجرا می‌کنیم، اعداد و نمودار کمی متفاوت خواهند بود. با این وجود، اگر از معماری که در بالا مشخص شده پیروی کرده باشید، باید دقت آزمون را بین ۸۰ تا ۹۵ درصد دریافت کنید! همانند خروجی زیر:

```
7/7 [==============================] - 0s 2ms/step - loss: 0.3281 - accuracy: 0.8584
0.8584474921226501
```

تبریک می‌گویم! شما توانستید اولین شبکه عصبی خود را طراحی کرده و آن را آموزش دهید. در بخش‌های قبلی در مورد بیش‌برازش و برخی تکنیک‌های منظم‌سازی صحبت کردیم. حالا چگونه بفهمیم که مدل ما در حال حاضر بیش‌برازش شده است؟ کاری که می‌توانیم انجام دهیم، این است که از زیان آموزشی و زیان اعتبارسنجی را برروی تعداد دوره‌های سپری شده ترسیم کنیم. برای مصورسازی این‌ها، از بسته matplotlib استفاده می‌کنیم. طبق معمول، باید کدی را که می‌خواهیم استفاده کنیم وارد کنیم:

```
import matplotlib.pyplot as plt
```

سپس، می‌خواهیم زیان آموزش و زیان اعتبارسنجی را مصورسازی کنیم. برای انجام این کار، این قطعه کد را اجرا کنید:

```
plt.plot(hist.history['loss'])
plt.plot(hist.history['val_loss'])
plt.title('Model loss')
plt.ylabel('Loss')
plt.xlabel('Epoch')
plt.legend(['Train', 'Val'], loc='upper right')
plt.show()
```

ما هر خط از قطعه کد بالا را توضیح خواهیم داد. دو خط اول می‌گوید که می‌خواهیم `loss` و `val_loss` را ترسیم کنیم. خط سوم عنوان این نمودار را مشخص می‌کند: `Model loss`. خط چهارم و پنجم به ما می‌گوید که محور y و x به ترتیب باید چه برچسبی داشته باشند. خط ششم شامل یک شرح برای نمودار ما است و مکان شرح در سمت راست بالا خواهد بود و خط هفتم به jupyter notebook می‌گوید که نمودار را نمایش دهد. خروجی شما باید چیزی شبیه به این باشد:

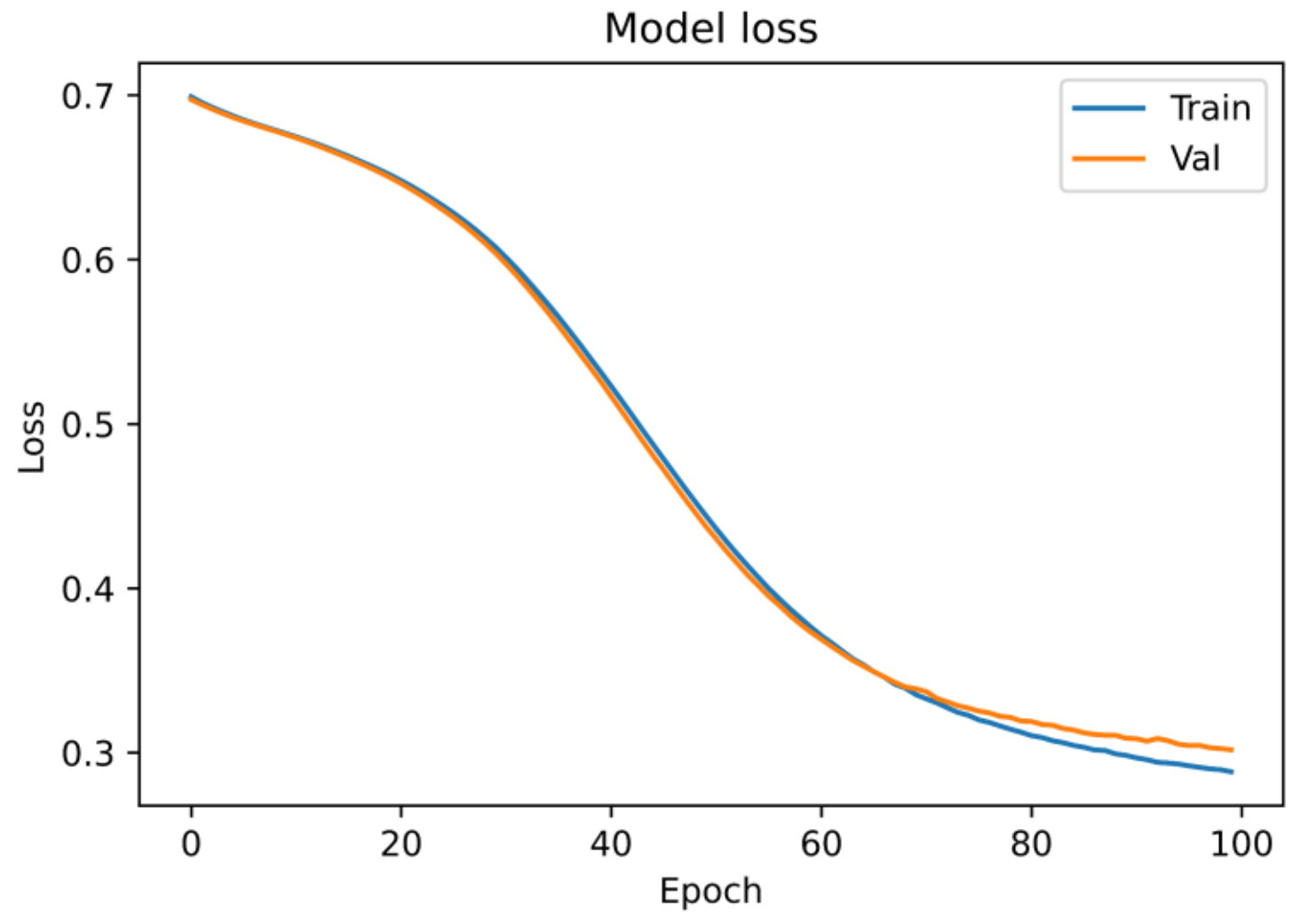

ما می‌توانیم همین کار را برای ترسیم دقت آموزشی و دقت اعتبارسنجی با کد زیر انجام دهیم:

```
plt.plot(hist.history['accuracy'])
plt.plot(hist.history['val_accuracy'])
plt.title('Model accuracy')
plt.ylabel('Accuracy')
plt.xlabel('Epoch')
plt.legend(['Train', 'Val'], loc='lower right')
plt.show()
```

شما باید نموداری که کمی شبیه به این در خروجی دریافت کنید:

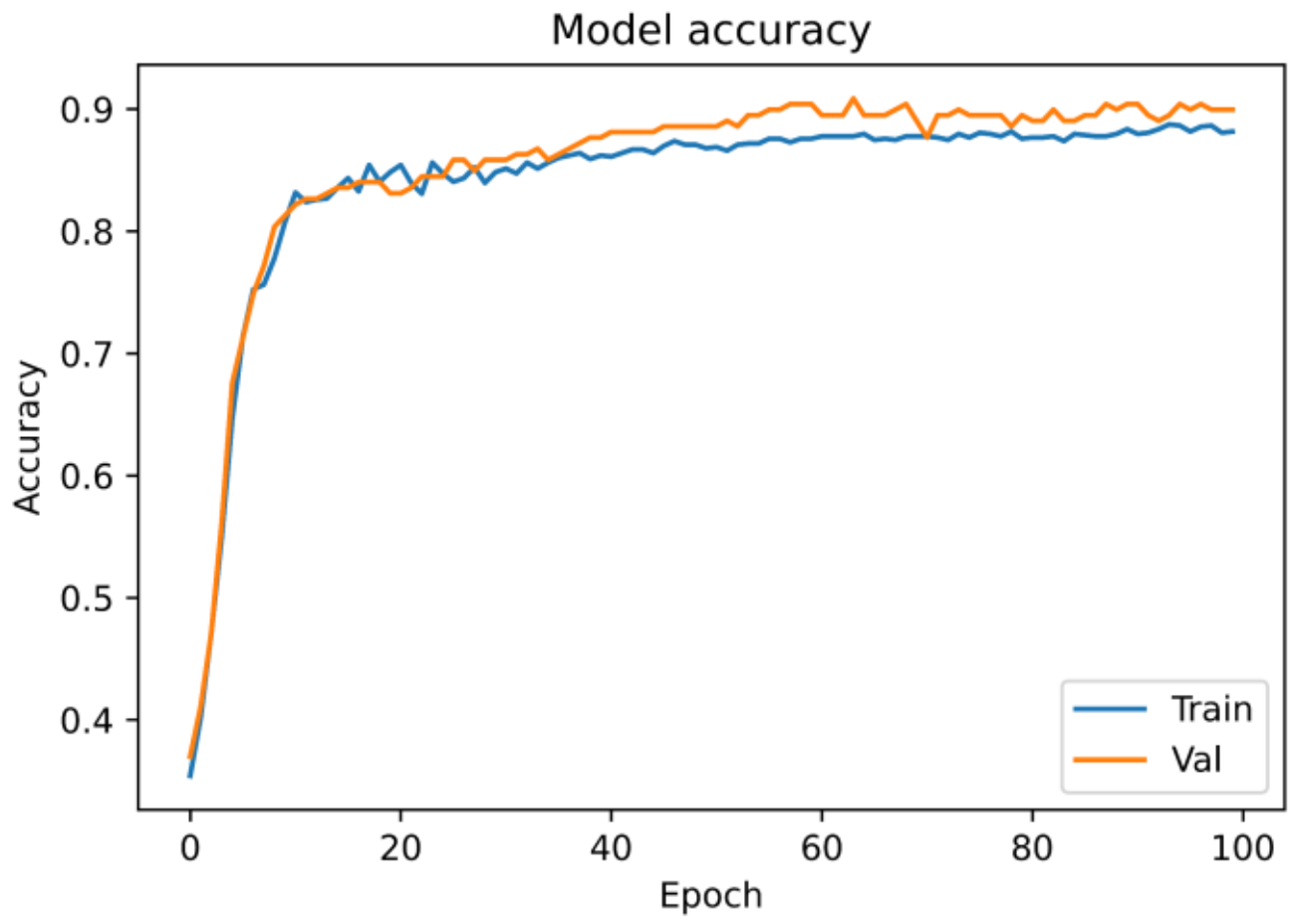

از آنجایی که پیشرفت‌های مدل ما در مجموعه آموزشی تا حدودی با بهبود مجموعه اعتبارسنجی مطابقت دارد، به نظر نمی‌رسد که بیش‌برازش مشکل بزرگی در مدل ما باشد (با این حال می‌توان آن را از طریق بهینه‌سازی ابرپارمترها بهبود بخشید).

به منظور معرفی منظم‌سازی به شبکه عصبی خود، بیایید با یک شبکه عصبی فرموله کنیم که به شدت در مجموعه آموزشی مطابقت داشته باشد. ما این را model_2 می‌نامیم.

```
model_2 = Sequential([
    Dense(1000, activation='relu', input_shape=(10,)),
    Dense(1000, activation='relu'),
    Dense(1000, activation='relu'),
    Dense(1000, activation='relu'),
    Dense(1, activation='sigmoid'),
])

model_2.compile(optimizer='adam',
              loss='binary_crossentropy',
              metrics=['accuracy'])

hist_2 = model_2.fit(X_train, Y_train,
          batch_size=32, epochs=100,
          validation_data=(X_val, Y_val))
```

در اینجا، ما یک مدل بسیار بزرگتر ساخته‌ایم و از بهینه ساز Adam استفاده می‌کنیم. Adam یکی از رایج‌ترین بهینه‌سازهایی است که در معماری‌های شبکه‌های عصبی استفاده می‌شود، به دلیل این‌که سریع‌تر به زیان کم‌تر می‌رسد. اگر این کد را اجرا کنیم و نمودارهای زیان را برای hist_2 با استفاده از کد زیر رسم کنیم (توجه داشته باشید که کد یکسان است با این تفاوت که به جای hist از hist_2 استفاده می‌کنیم):

```
plt.plot(hist_2.history['loss'])
plt.plot(hist_2.history['val_loss'])
plt.title('Model loss')
plt.ylabel('Loss')
```

```
plt.xlabel('Epoch')
plt.legend(['Train', 'Val'], loc='upper right')
plt.show()
```

ما یک نمودار به صورت زیر دریافت می‌کنیم:

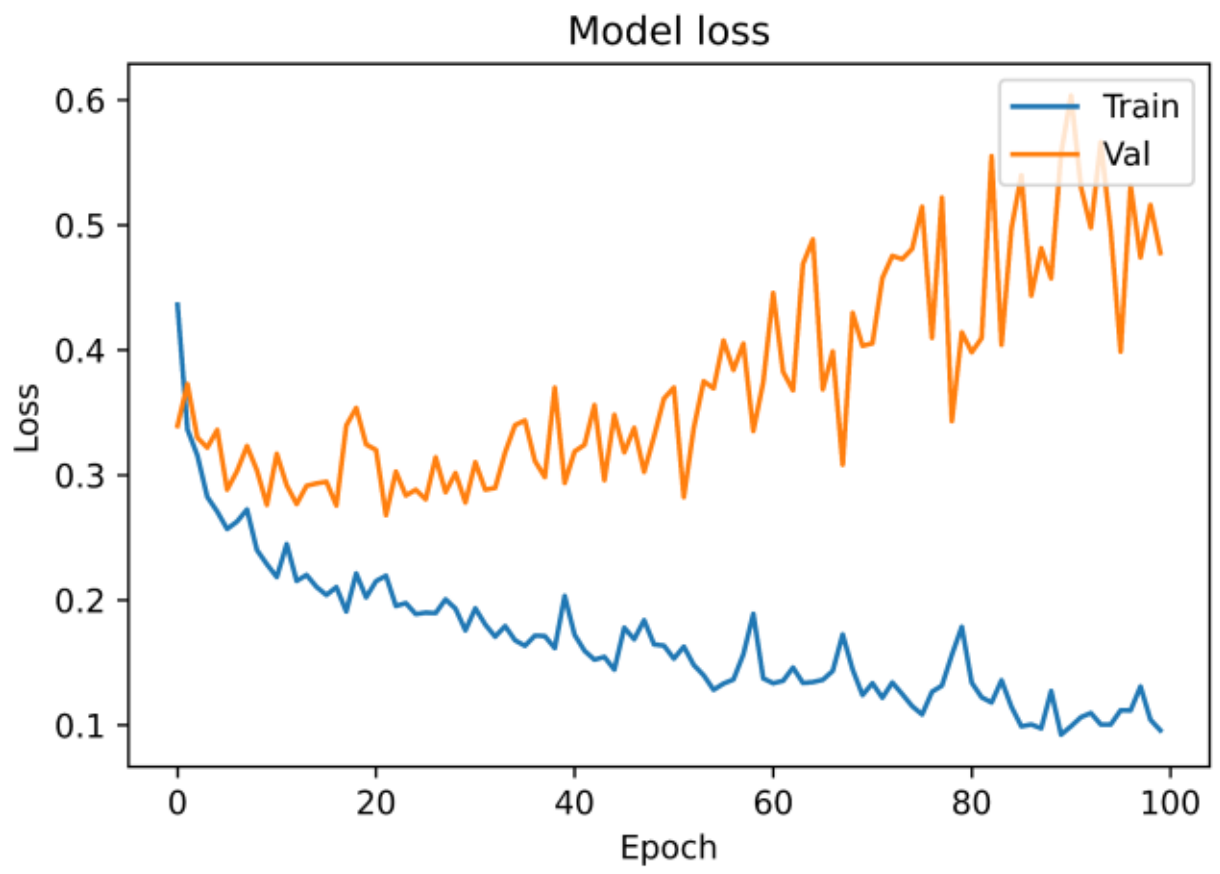

این نشانه بارز بیش‌برازش است. زیان آموزشی در حال کاهش است، اما زیان اعتبارسنجی بسیار بالاتر از زیان آموزشی و در حال افزایش است. اگر دقت را با استفاده از کد زیر ترسیم کنیم:

```
plt.plot(hist_2.history['accuracy'])
plt.plot(hist_2.history['val_accuracy'])
plt.title('Model accuracy')
plt.ylabel('Accuracy')
plt.xlabel('Epoch')
plt.legend(['Train', 'Val'], loc='lower right')
plt.show()
```

همچنین می‌توانیم تفاوت واضح‌تری بین دقت آموزشی و اعتبارسنجی را ببینیم:

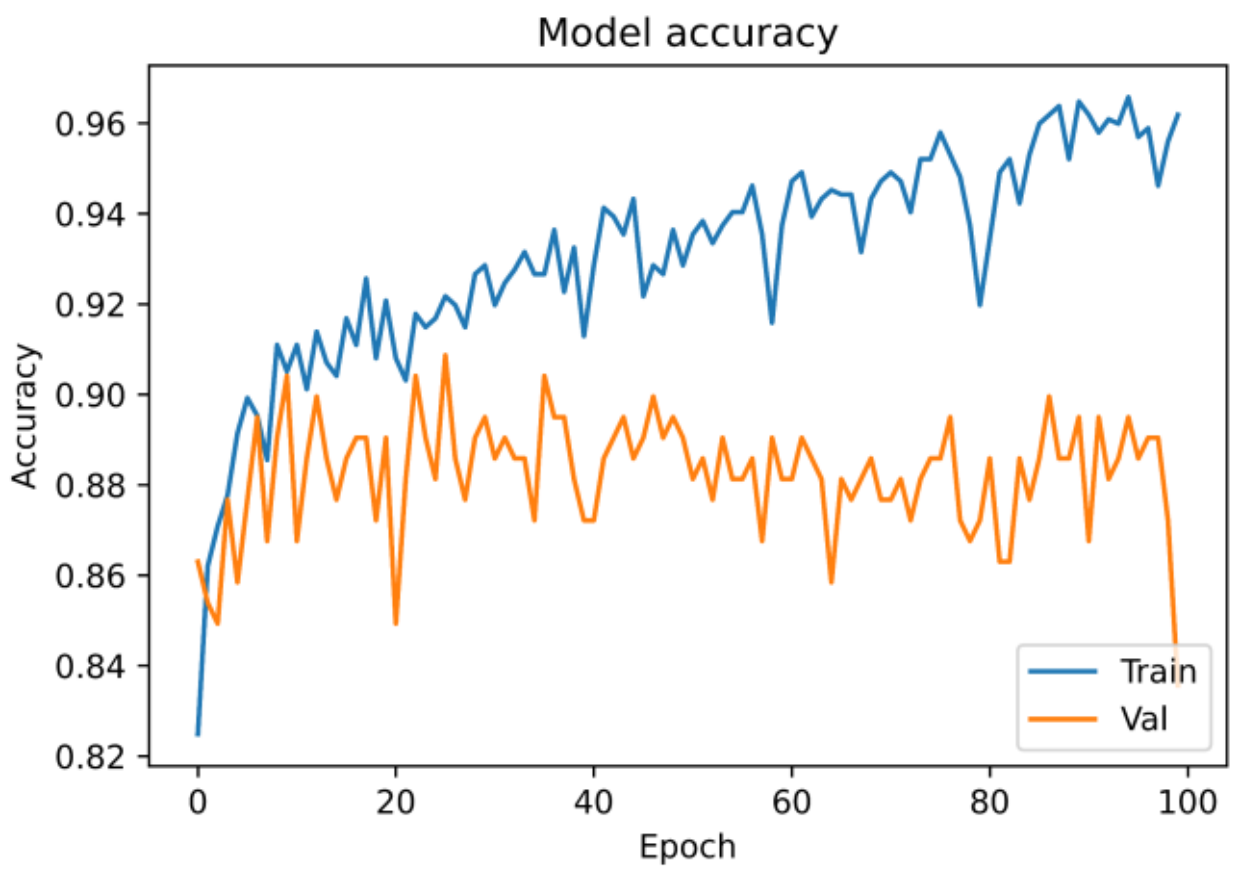

اکنون، بیایید برخی از استراتژی‌های خود را برای کاهش بیش‌برازش امتحان کنیم. در بخش‌های پیشین ما چندین روش را برای جلوگیری از بیش‌برازش معرفی کردیم. با این حال، در این بخش ما تنها از حذف تصادفی استفاده می‌کنیم. ابتدا، بیایید کدی را که برای حذف تصادفی نیاز داریم را وارد کنیم:

```
from keras.layers import Dropout
```

سپس مدل سوم خود را به صورت زیر مشخص می‌کنیم:

```
model_3 = Sequential([
    Dense(1000, activation='relu',  input_shape=(10,)),
    Dropout(0.5),
    Dense(1000, activation='relu'),
    Dropout(0.5),
    Dense(1000, activation='relu'),
    Dropout(0.5),
    Dense(1000, activation='relu'),
    Dropout(0.5),
    Dense(1, activation='sigmoid'),
])
```

آیا می‌توانید تفاوت بین مدل ۳ و مدل ۲ را تشخیص دهید؟ یک تفاوت اصلی وجود دارد:

برای اضافه کردن Dropout، یک لایه جدید مانند این اضافه کردیم:

```
Dropout(0.5),
```

این به این معنی است که نورون‌های لایه قبلی در حین آموزش `0.5` احتمال حذف دارند. بیایید آن را کامپایل کرده و با همان پارامترهای مدل ۲ خود اجرا کنیم.

```
model_3.compile(optimizer='adam',
              loss='binary_crossentropy',
              metrics=['accuracy'])

hist_3 = model_3.fit(X_train, Y_train,
          batch_size=32, epochs=100,
          validation_data=(X_val, Y_val))
```

اکنون، بیایید نمودارهای زیان و دقت را رسم کنیم.

```
plt.plot(hist_3.history['loss'])
plt.plot(hist_3.history['val_loss'])
plt.title('Model loss')
plt.ylabel('Loss')
plt.xlabel('Epoch')
plt.legend(['Train', 'Val'], loc='upper right')
plt.show()
```

خروجی به شکل زیر دریافت خواهیم کرد:

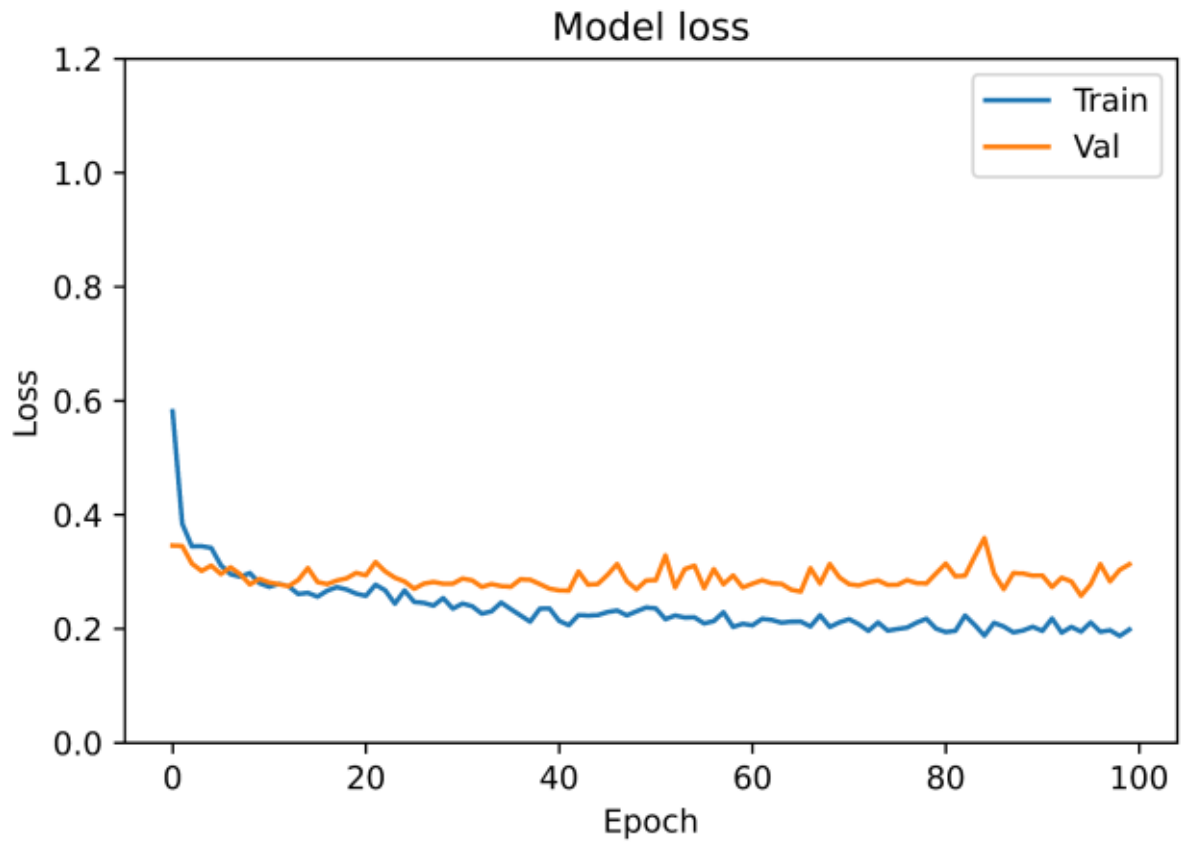

همان‌طور که مشاهده می‌شود، زیان اعتبارسنجی نسبت مدل ۲ بیشتر با از زیان آموزش ما مطابقت دارد (با این حال این مدل همچنان مطلوب نیست و مدل بیش‌برازش شده است). بیایید دقت را با قطعه کد مشابه ترسیم کنیم:

```
plt.plot(hist_3.history['accuracy'])
plt.plot(hist_3.history['val_accuracy'])
plt.title('Model accuracy')
plt.ylabel('Accuracy')
plt.xlabel('Epoch')
plt.legend(['Train', 'Val'], loc='lower right')
plt.show()
```

و خروجی به صورت زیر دریافت خواهیم کرد:

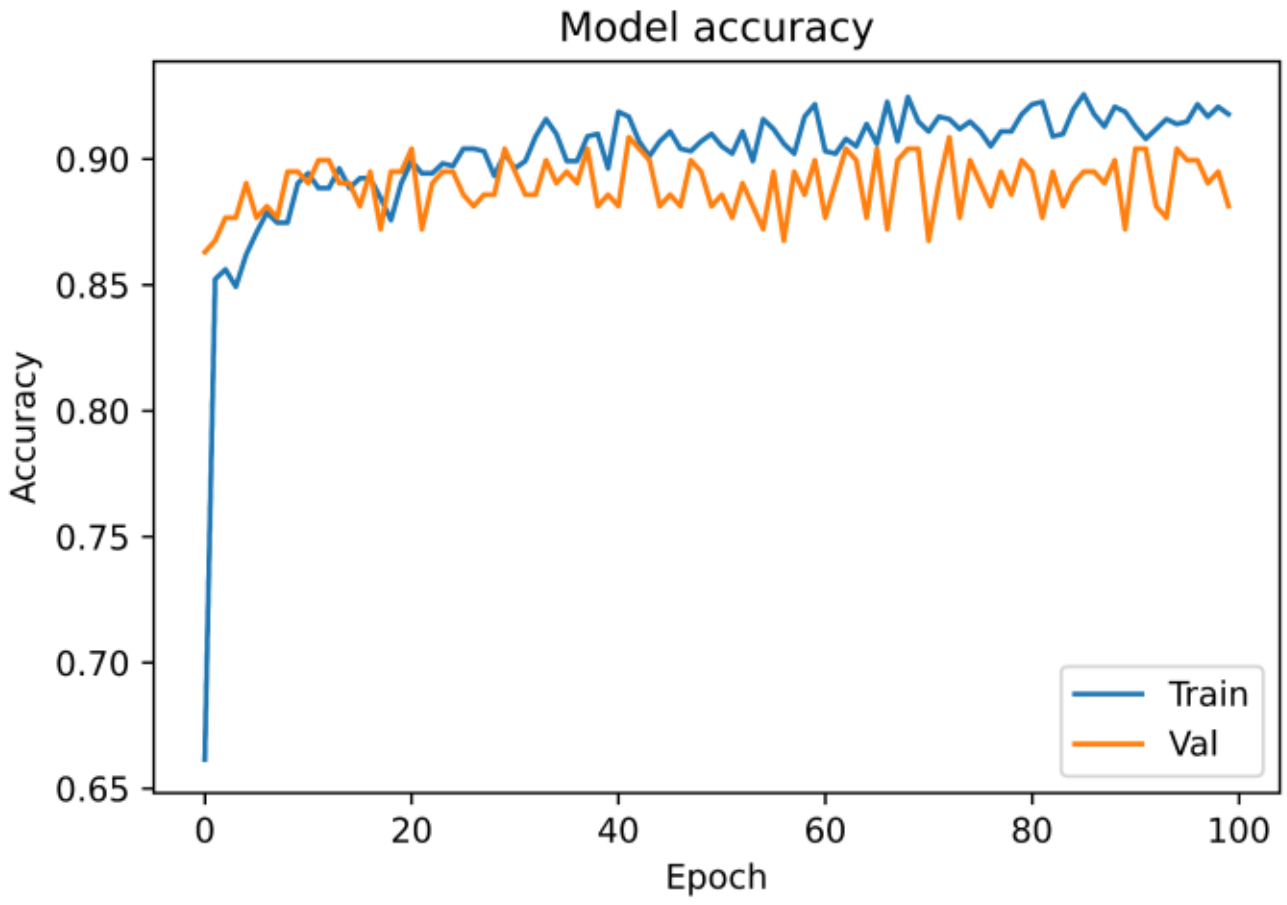

در مقایسه با مدل ۲، ما بیش‌برازش را به میزان قابل توجهی کاهش داده‌ایم! به این ترتیب است که ما تکنیک‌های منظم‌سازی را برای کاهش بیش‌برازش در مجموعه آموزشی اعمال می‌کنیم. می‌توانید به عنوان تمرین، ابرپارمترها را تغییر داده و نتایج را مقایسه کنید.

## خلاصه فصل

- شبکه‌های عصبی مصنوعی مدل محاسباتی هستند که نحوه‌یِ عملکرد سلول‌های عصبی در مغز انسان را تقلید می‌کند.
- هر شبکه عصبی مصنوعی دارای یک لایه ورودی، یک لایه خروجی و یک یا چند لایه پنهان می‌باشد.
- ساده‌ترین نوع مدل‌سازی یک نورون را پرسپترون گویند که می‌تواند دارای تعداد زیادی ورودی تنها با یک خروجی باشد.
- محدودیت اصلی شبکه‌های عصبی پرسپترون، عدم توانایی در طبقه‌بندی داده‌هایی است که جدایی‌پذیر خطی نیستند.
- هدف از فرآیند یادگیری در شبکه‌های عصبی، یافتن مجموعه‌ای از مقادیر وزنی است که باعث می‌شود خروجی شبکه عصبی تا حد امکان با مقادیر هدف واقعی مطابقت داشته باشد.
- تابع فعال‌ساز تصمیم می‌گیرد که یک نرون باید فعال شود یا خیر.

## آزمونک

۱. انتخاب نرخ یادگیری خیلی کوچک و یا خیلی بزرگ چه تاثیری در فرآیند یادگیری دارد؟
۲. پدیده محو گرادیان را شرح دهید؟
۳. ویژگی‌های مطلوب یک تابع فعال‌سازی چیست؟
۴. در لایه خروجی مسائل طبقه‌بندی دودویی و چندکلاسه از کدام تابع فعال‌سازی استفاده می‌شود؟
۵. بهینه‌سازها چه نقشی در فرآیند یادگیری شبکه‌های عصبی دارند؟

## تمرین

یک شبکه عصبی با دو لایه پنهان بسازید تا به طبقه‌بندی مجموعه داده Iris بپردازد. نمودارهای دقت و زیان را برای مجموعه آموزشی و اعتبارسنجی در حین آموزش مصورسازی کنید.

راهنمایی برای مقیاس‌بندی داده‌ها

```
scaler = StandardScaler()
X_scaled = scaler.fit_transform(X)
```

راهنمایی برای وارد کردن داده‌ها

```
from sklearn.datasets import load_iris
iris = load_iris()
X = iris['data']
y = iris['target']
```

# ۴ شبکه‌های عصبی کانولوشنی

## اهداف یادگیری:

- شبکه کانولوشنی چیست؟
- آشنایی با انواع لایه‌های آن
- طبقه‌بندی تصویر با شبکه عصبی کانولوشنی

## مقدمه

در این فصل به معرفی مفاهیم شبکه‌های عصبی کانولوشنی می‌پردازیم. این مفاهیم، شامل اجزای اصلی شبکه هستند که معماری یک شبکه عصبی کانولوشنی را تشکیل می‌دهند. شبکه‌های عصبی کانولوشنی برای داده‌های بدون ساختار همانند تصاویر عملکرد بسیار خوبی دارند. پس از آشنایی کامل با معماری شبکه‌های عصبی کانولوشنی، در انتهای فصل به پیاده‌سازی یک مثال عملی با استفاده از شبکه عصبی کانولوشنی در keras می‌پردازیم.

## شبکه عصبی کانولوشنی (CNN)

در یک شبکه عصبی پیش‌خور کاملا متصل، تمام گره‌های یک لایه به تمام گره‌های لایه بعدی متصل می‌شوند. هر اتصال دارای وزن $w_{i,j}$ است که باید توسط الگوریتم یادگیری آموخته شود. فرض کنید ورودی ما یک تصویر ٦٤ پیکسل در ٦٤ پیکسل در مقیاس خاکستری است. از آنجایی که هر پیکسل خاکستری را می توان با ۱ مقدار نشان داد، می‌گوییم اندازه کانال ۱ است. چنین تصویری را می‌توان با ٤٠٩٦ = ۱ × ٦٤ × ٦٤ نشان داد (کانال × ستون × سطر). از این رو، لایه ورودی یک شبکه پیش‌خور که چنین تصویری را پردازش می‌کند دارای ٤٠٩٦ گره است. فرض کنید لایه بعدی ۵۰۰ گره دارد. از آنجایی که تمام گره‌های لایه‌های بعدی کاملا بهم متصل هستند، بین ورودی و اولین لایه پنهان، ۲۰٤۸۰۰۰ = ۵۰۰ × ٤٠٩٦ وزن خواهیم داشت. حال آنکه، برای مسائل پیچیده، ما معمولا به چندین لایه پنهان در شبکه پیش‌خور خود نیاز داریم، زیرا یک شبکه پیش‌خور ساده ممکن است نتواند مدل نگاشت ورودی‌ها به خروجی‌ها را در داده‌های آموزشی بیاموزد. داشتن چندین لایه پنهان مشکل داشتن وزن‌های زیاد در شبکه پیش‌خور ما را تشدید می‌کند و با افزایش ابعاد فضای جستجو، فرآیند یادگیری را دشوارتر می‌کند. همچنین باعث می‌شود آموزش، زمان و منابع بیشتری را صرف کند. علاوه براین، تعداد زیاد پارامترها میل شبکه را به بیش‌برازش افزایش می‌دهد. از این‌رو، به منظور پرداختن به این مسائل، شبکه‌های عصبی کانولوشنی (CNN) به عنوان توسعه‌ِی بسیار محبوب شبکه‌های عصبی استاندارد معرفی شدند. شبکه‌ی عصبی کانولوشنی دسته‌ای از شبکه‌های عصبی پیش‌خور هستند که از لایه‌های پیچش برای تجزیه و تحلیل ورودی‌هایی با توپولوژی‌هایِ مشبکی، همانند تصاویر و ویدیوها استفاده می‌کنند. نام این شبکه‌ها بر اساس تابع ریاضی به نام **کانولوشن** یا **پیچش** است که در ساختار خود به کار می‌برند. به‌طور خلاصه، شبکه‌های کانولوشنی، شبکه‌های عصبی هستند که از کانولوشن به جای ضرب ماتریس، حداقل در یکی از لایه‌های خود استفاده می‌کنند.

چیزی که در مورد شبکه کانولوشنی خاص است، نحوه ساختاربندی اتصالات بین نورون‌ها و معماری لایه پنهان منحصر به فردی است که از مکانیسم پردازش داده‌های بصری خودمان در داخل قشر بینایی ما الهام گرفته شده است و برخلاف شبکه‌های عصبی پیش‌خور، لایه‌ها در CNN در ۳ بعد سازماندهی شده اند: عرض، ارتفاع و عمق.

یکی از مهم‌ترین ویژگی‌های شبکه کانولوشنی را که باید بدون توجه به اینکه چند لایه در معماری آن وجود دارد را به خاطر بسپارید، این است که کل معماری یک CNN از دو بخش اصلی تشکیل شده است:

- **استخراج ویژگی (Feature Extraction):** در این لایه، شبکه یک سری **کانولوشن (convolution)** و عملگر **ادغام (pooling)**، انجام می‌دهد. اگر بخواهیم تصویر یک گربه را در تصویر شناسایی کنیم، این قسمتی است که در آن ویژگی‌های خاصی همانند گوش، پنجه، رنگ خز گربه تشخیص داده می‌شود. به‌طور خلاصه، وظیفه این لایه، تشخیص ویژگی‌های مهم در پیکسل‌های تصویر است. لایه‌های نزدیک‌تر به ورودی یاد می‌گیرند که ویژگی‌های ساده مانند لبه‌ها و گرادیان‌های رنگ را شناسایی کنند، در حالی که لایه‌های عمیق‌تر ویژگی‌های ساده را با ویژگی‌های پیچیده‌تر ترکیب می‌کنند.
- **طبقه‌بندی:** در اینجا، لایه کاملا متصل به عنوان یک طبقه‌بند بعد از مرحله‌ی استخراج ویژگی‌ها عمل می‌کند. این لایه مشخص می‌کنند که کدام ویژگی بیشترین ارتباط را با یک کلاس خاص دارد، از این‌رو کلاس تصویر را بر اساس ویژگی‌های استخراج شده در مراحل قبلی پیش‌بینی می‌کند. معماری کلی آن در شکل ۴ـ۱ قابل مشاهده است.

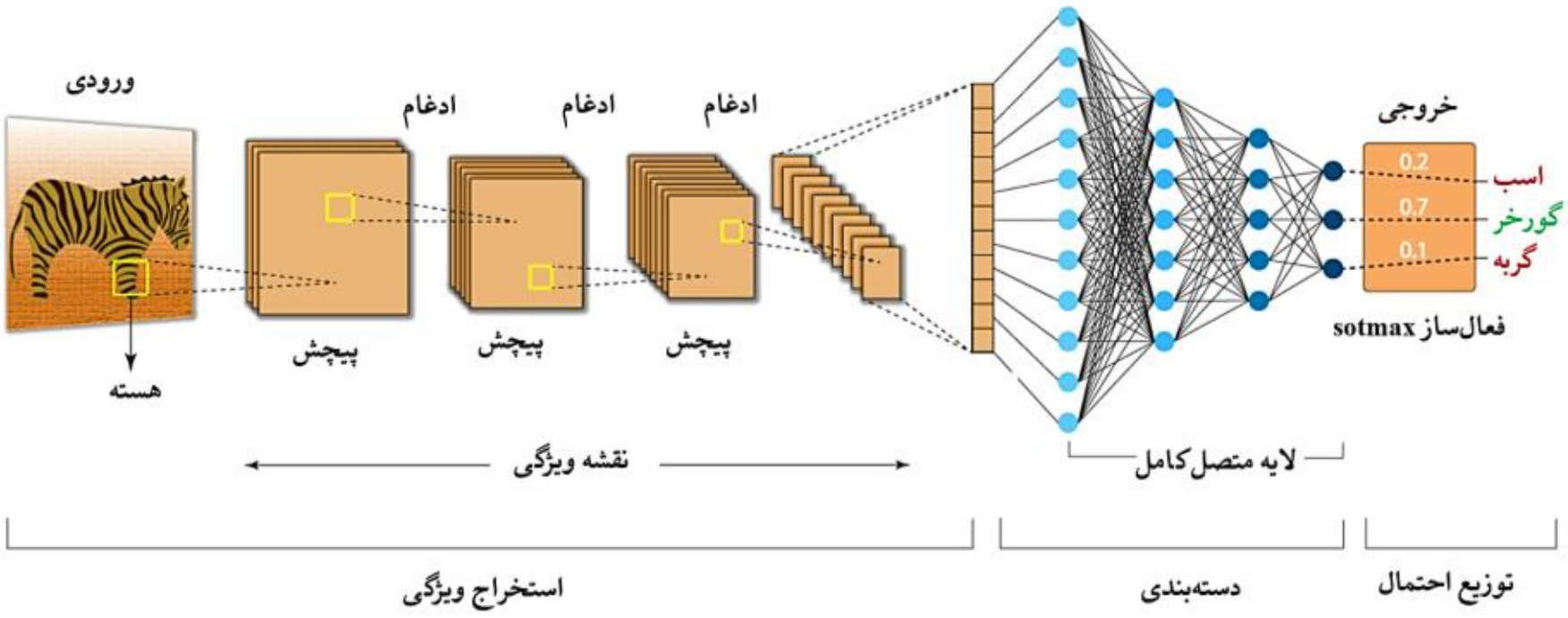

**شکل ٤ـ۱. شمایی کلی از ساختار یک شبکه کانولوشنی**

شبکه‌های عصبی کانولوشنی دارای سه ویژگی متمایز در مقایسه با سایر شبکه‌های عصبی هستند:

۱. **میدان پذیرای محلی (local receptive fields):** هر نورون در یک CNN مسئول یک منطقه تعریف شده از داده‌های ورودی است و این به نورون‌ها امکان می‌دهد تا الگوهایی

مانند خطوط، لبه‌ها و جزئیات کوچکی که تصویر را می‌سازند، بیاموزند. این ناحیه تعریف شده از فضا که یک نورون یا واحد در داده‌های ورودی در معرض آن قرار می‌گیرد، میدان پذیرای محلی نامیده می‌شود. میدان پذیرا با اندازه فیلتر یک لایه در یک شبکه عصبی کانولوشنی تعریف می‌شود.

۲. **اشتراک‌گذاری پارامتر (parameter sharing) و اتصال‌محلی (Local connectivity):** هر لایه کانولوشنی شامل چندین فیلتر می‌باشد و این یک ابرپارامتر از پیش تعریف شده است. هر یک از این فیلترها دارای یک عرض و ارتفاع تنظیم شده است که مربوط به میدان پذیرای محلی یک نورون است. فیلترهایی که بر روی داده‌های ورودی عمل می‌کنند، **نقشه ویژگی** (**feature map**) را در خروجی لایه کانولوشنی ایجاد می‌کنند. **اشتراک‌گذاری پارامتر** به اشتراک‌گذاری وزن‌ها توسط همه نورون‌ها در یک نقشه ویژگی است. از طرف دیگر، اتصال محلی مفهومی است که هر نورون فقط به زیرمجموعه‌ای از نورون‌ها متصل است، برخلاف یک شبکه عصبی پیش‌خور که در آن همه نورون‌ها کاملا بهم متصل هستند. این ویژگی‌ها به کاهش تعداد پارامترها در کل سیستم کمک می‌کند و محاسبات را کارآمدتر می‌کند.

۳. **زیرنمونه‌گیری (sub-sampling) یا ادغام (pooling)**: زیرنمونه‌گیری یا ادغام اغلب بلافاصله پس از یک لایه کانولوشنی در CNN می‌آید. نقش آن پایین آوردن خروجی یک لایه کانولوشنی در امتداد ابعاد فضایی ارتفاع و عرض است. عملکرد اصلی ادغام، کاهش تعداد پارامترهایی است که باید توسط شبکه یاد گرفته شود. این ویژگی، همچنین سبب کاهش اثر بیش‌برازش می‌شود و در نتیجه افزایش عملکرد و دقت کلی شبکه را به همراه دارد.

شبکه‌های کانولوشنی، نقش مهمی را در تاریخچه یادگیری عمیق به‌همراه داشته‌اند. آن‌ها نمونه‌ای مهم و موفق در فهم ما از مطالعه مغز در کاربردهای یادگیری ماشین هستند. شبکه‌های عصبی کانولوشنی جزو اولین شبکه‌های عصبی بودند که در حل و انجام کاربردهای تجاری مهم مورد استفاده قرار گرفته و حتی تا امروز در صدر برنامه‌های کاربردی تجاری یادگیری عمیق قرار دارند.

شبکه‌های کانولوشنی در یافتن الگوهای تکراری و استخراج ویژگی‌های محلی قدرت بسیار زیادی دارند.

## عملگر کانولوشن

شبکه‌های کانولوشنی به دسته‌ای از شبکه‌های عصبی تعلق دارند که تصویر را به عنوان ورودی می‌گیرند، آن را در معرض ترکیبی از وزن‌ها و سوگیری‌ها قرار می‌دهند، ویژگی‌ها را استخراج

می‌کنند و نتایج را بدست می‌آورند. آن‌ها تمایل به کاهش ابعاد تصویر ورودی با استفاده از یک **هسته** دارند که **استخراج ویژگی‌ها** را در مقایسه با شبکه عصبی پیش‌خور آسان‌تر می‌کند. اساس یک شبکه‌ی کانولوشنی عملگر کانولوشن است.

کانولوشن دوبعدی اساسا یک عملیات نسبتا ساده است. شما با یک هسته شروع می‌کنید، که یک ماتریس کوچک از وزن‌ها است. این هسته روی داده‌های ورودی دوبعدی می‌لغزد، **ضرب درایه‌ای** را با بخشی از ورودی که در حال حاضر روی آن است انجام می‌دهد و سپس نتایج را در یک پیکسل خروجی خلاصه می‌کند (شکل ٤ــ۲). هسته این فرآیند را برای هر مکانی که روی آن می‌لغزد تکرار می‌کند و یک ماتریس دو بعدی از ویژگی‌ها را به ماتریس دو بعدی دیگر از ویژگی‌ها تبدیل می‌کند. ویژگی‌های خروجی اساساً، مجموع وزن‌دار ویژگی‌های ورودی هستند که تقریبا در همان مکان پیکسل خروجی در لایه ورودی قرار دارند.

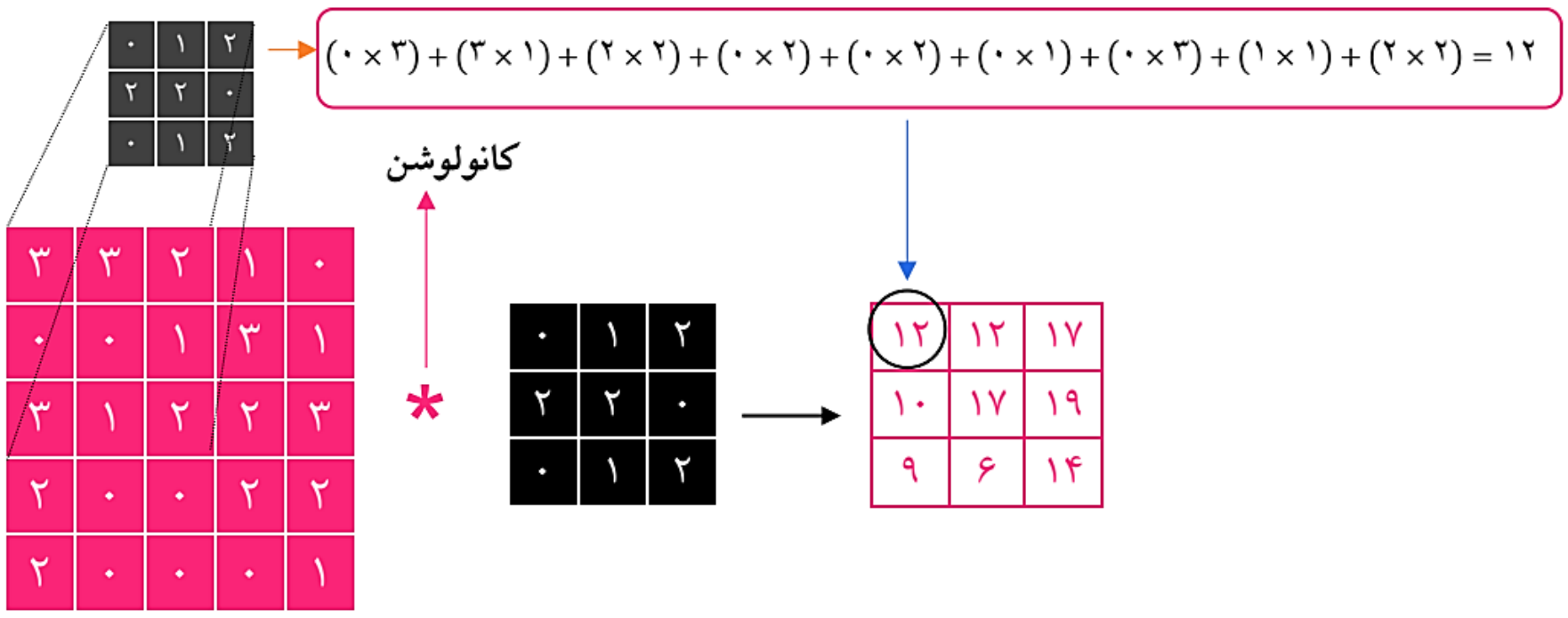

**شکل ٤ــ۲. عملگر کانولوشن**

در مثال بالا، ویژگی‌های ورودی **۲۵=۵×۵**، و ویژگی‌های خروجی **۹=۳×۳** است. اگر این یک لایه کامل متصل استاندارد بود، ماتریسِ وزنی به تعداد پارامتر **۲۲۵=۹×۲۹** خواهیم داشت که هر ویژگی خروجی مجموع وزنی هر ویژگی ورودی است. کانولوشن‌ها به ما این امکان را می‌دهند که این تبدیل را تنها با ۹ پارامتر انجام دهیم.

تأثیر کانولوشن، تأکید بر مرزهای اشکال مختلف است. هسته‌های متغیر را می‌توان به منظور برآوردن نیازهای دقیق تنظیم کرد. با این حال، به جای تلاش برای انجام آن به صورت دستی، یک شبکه کانولوشن عمیق این وظایف را به فرآیند یادگیری واگذار می‌کند.

کاربرد موازی هسته‌های (فیلترهای) مختلف، همپوشانی‌های پیچیده‌ای را به همراه دارد که می‌تواند استخراج ویژگی‌هایی را که واقعاً برای طبقه‌بندی مهم هستند، ساده کند. تفاوت اصلی بین یک لایه کاملا متصل و یک لایه کانولوشن، توانایی لایه دوم برای کار با یک هندسه موجود است که تمام عناصر مورد نیاز برای تشخیص یک شی از یک شی دیگر را رمزگذاری می‌کند. این عناصر را نمی‌توان فوراً تعمیم داد، اما به پردازش بعدی برای انجام یک ابهام‌زدایی نیاز دارد.

به عنوان مثال، چشم و بینی تقریبا شبیه به‌هم هستند. چگونه می‌توان تصویر را بدرستی تقسیم کرد؟ پاسخ با یک تحلیل مضاعف ارائه می‌شود: تفاوت‌های ظریفی وجود دارد که می‌توان آن‌ها را با فیلترهای ریزدانه کشف کرد و مهم‌تر از همه، هندسه سراسری اشیا، مبتنی بر روابط درونی است که تقریبا ثابت هستند. به عنوان مثال، چشم‌ها و بینی باید یک مثلث متساوی‌الساقین را تشکیل دهند، زیرا تقارن صورت دلالت بر فاصله یکسان بین هر چشم و بینی دارد. این را می‌توان از قبل، مانند بسیاری از تکنیک‌های پردازش تصویر انجام داد، یا به لطف قدرت یادگیری عمیق، می‌توان آن را به فرآیند آموزش واگذار کرد. از آنجایی که تابع هزینه و کلاس‌های خروجی به طور ضمنی تفاوت‌ها را کنترل می‌کنند، یک شبکه کانولوشن عمیق می‌تواند یاد بگیرد که برای رسیدن به یک هدف خاص چه چیزی مهم است و در عین حال تمام جزئیات بی‌فایده را کنار بگذارد.

## لایه کانولوشن

لایه کانولوشن مهم‌ترین بلوک سازنده یک CNN است. این لایه شامل مجموعه‌ای از **فیلترها** که همچنین به عنوان **هسته‌ها** (**kernels**) یا **آشکارسازهای ویژگی** (**feature detectors**) شناخته می‌شود، هستند که در آن هر فیلتر در تمام مناطق داده‌های ورودی اعمال می‌شود. به عبارت دیگر، وظیفه اصلی لایه کانولوشن، شناسایی ویژگی‌های یافت شده در مناطق محلی تصویر ورودی است که این ویژگی‌ها در کل مجموعه داده مشترک هستند. این شناسایی ویژگی‌ها از طریق اعمال **فیلترها** منجر به تولید **نقشه ویژگی** می‌شود. لایه کانولوشنی، یک فیلتر محلی را بر روی تصویر ورودی اعمال می‌کند. این امر منجر می‌شود طبقه‌بندی بهتری در پیکسل‌های همسایه‌ای که همبستگی بیشتری بین آن‌ها وجود دارد در همان تصویر صورت پذیرد. به عبارت دیگر، پیکسل‌های تصاویر ورودی می‌توانند با یک‌دیگر همبستگی داشته باشند. به عنوان مثال، در تصاویر صورت، بینی همیشه بین چشم‌ها و دهان قرار دارد. وقتی فیلتر را به زیرمجموعه‌ای از تصویر اعمال می‌کنیم، برخی از ویژگی‌های محلی را استخراج می‌کنیم. از این لایه به عنوان لایه **استخراج ویژگی** نیز یاد می‌شود. چراکه ویژگی‌های تصویر در این لایه استخراج می‌شوند.

هر لایه کانولوشن دارای مجموعه خاصی از ابرپارامترها است که هر یک از آن‌ها تعداد ارتباطات و اندازه‌یِ خروجیِ نقشه‌هایِ ویژگی را تعیین می‌کند:

- **اندازه هسته:** اندازه‌ی هسته‌ی K (گاهی اوقات **اندازه فیلتر** نیز نامیده می‌شود) **میدان پذیرا** را توصیف می‌کند که برای همه مکان‌های ورودی اعمال می‌شود. افزایش این پارامتر به لایه کانولوشن اجازه می‌دهد تا اطلاعات فضایی بیشتری را دریافت کند، در حالی که به‌طور هم‌زمان تعداد وزن‌های شبکه را افزایش می‌دهد.
- **تعداد هسته:** تعداد هسته‌ها مستقیما با تعداد پارامترهای قابل یادگیری و عمق D حجم خروجی یک لایه پیچش مطابقت دارد. همان‌طورکه هر هسته یک نقشه ویژگی

خروجی جداگانه تولید می‌کند، هسته‌های D یک نقشه‌یِ ویژگی خروجی با عمق D را تولید می‌کنند.

- **گام:** پیچش را می‌توان به عنوان جمع‌وزنی با "لغزاندن" یک هسته بر روی یک حجم ورودی درک کرد. با این حال، نیازی نیست که "لغزش" با یک فاصله یک پیکسل در یک زمان اتفاق بیفتد، چیزی که گام توصیف می‌کند. گام $S$ تعداد پیکسل‌هایی را که هسته بین هر محاسبه‌یِ ویژگیِ خروجی جابجا می‌شود را مشخص می‌کند. گام‌های بزرگ‌تر، نقشه‌هایِ ویژگیِ خروجیِ کوچکتری تولید می‌کنند، زیرا محاسبات کم‌تری انجام می‌شود. این مفهوم در شکل زیر نشان‌داده شده است:

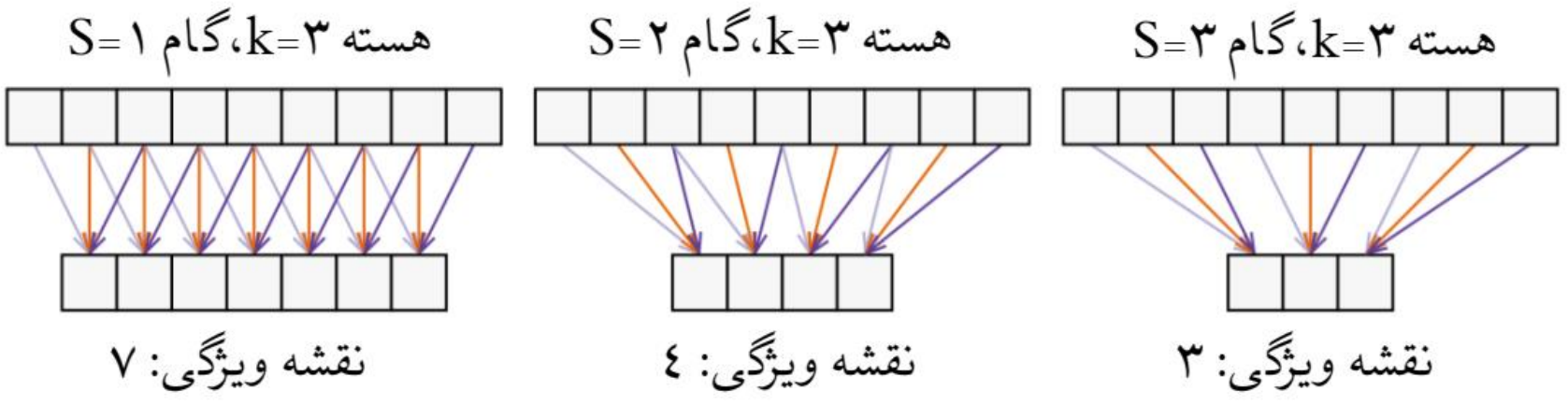

- **لایه‌گذاری‌ـ‌صفر:** به دلیل نحوه عملکرد عملیات پیچش، از لایه‌گذاری‌ـ‌صفر برای کنترل کاهش ابعاد پس از اعمال فیلترهای بزرگتر از ۱*۱ و جلوگیری از گم شدن اطلاعات در حاشیه استفاده می‌شود. به عبارت دیگر، از لایه‌گذاری‌ـ‌صفر اغلب استفاده می‌شود تا ابعاد فضایی لایه‌های ورودی و خروجی را یکسان نگه داشت. با اضافه کردن ورودیِ صفر در اطراف حاشیه، می‌توان کوچک‌شدن ابعاد فضایی هنگام انجام پیچش را دور زد. مقدار صفرهای اضافه شده در هر طرف برای هر بعد فضایی یک ابرپارمتر اضافی $P$ است. نمونه‌ای از لایه‌گذاری صفر در شکل زیر نشان داده شده است:

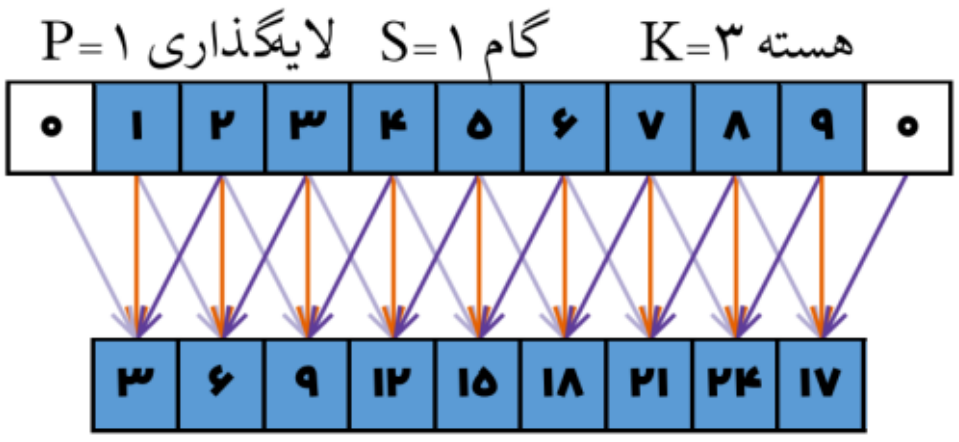

- **فراخش(Dilation):** فراخش یا ازهم‌گشودگی $d$ که اخیرا معرفی شده است، ابرپارمتر دیگری است که به لایه کانولوشن اجازه می‌دهد تا میدان پذیرای موثرتری نسبت به ورودی داشته باشد، در حالی که اندازه هسته را ثابت نگه می‌دارد. این امر با معرفی $d$ فاصله بین هر سلول از هسته بدست می‌آید. کانولوشن استاندارد، به سادگی از فراخش ۰ استفاده می‌کند. از این‌رو دارای یک هسته پیوسته است. با افزایش

فراخش این امکان برای یک لایه‌یِ کانولوشن وجود دارد که وسعت فضایی بیشتری از ورودی را بگیرد و در عین حال مصرف حافظه را ثابت نگه دارد. مفهوم کانولوشن‌های فراخش که گاهی اوقات **کانولوشن‌های آتروس** (**atrous convolutions**) نیز نامیده می‌شود، با فراخش‌های مختلف در شکل ۴ـ۳ نشان داده شده است.

با توجه به اندازه‌یِ حجم ورودی $W$، اندازه‌یِ هسته‌یِ $K$، گام $S$، فراخش $d$ و $P$ لایه‌گذاری، حجم خروجی حاصل به صورت زیر محاسبه می‌شود:

$$W_o = \left\lfloor \frac{W + 2P - K - (K-1)(d-1)}{S} \right\rfloor + 1.$$

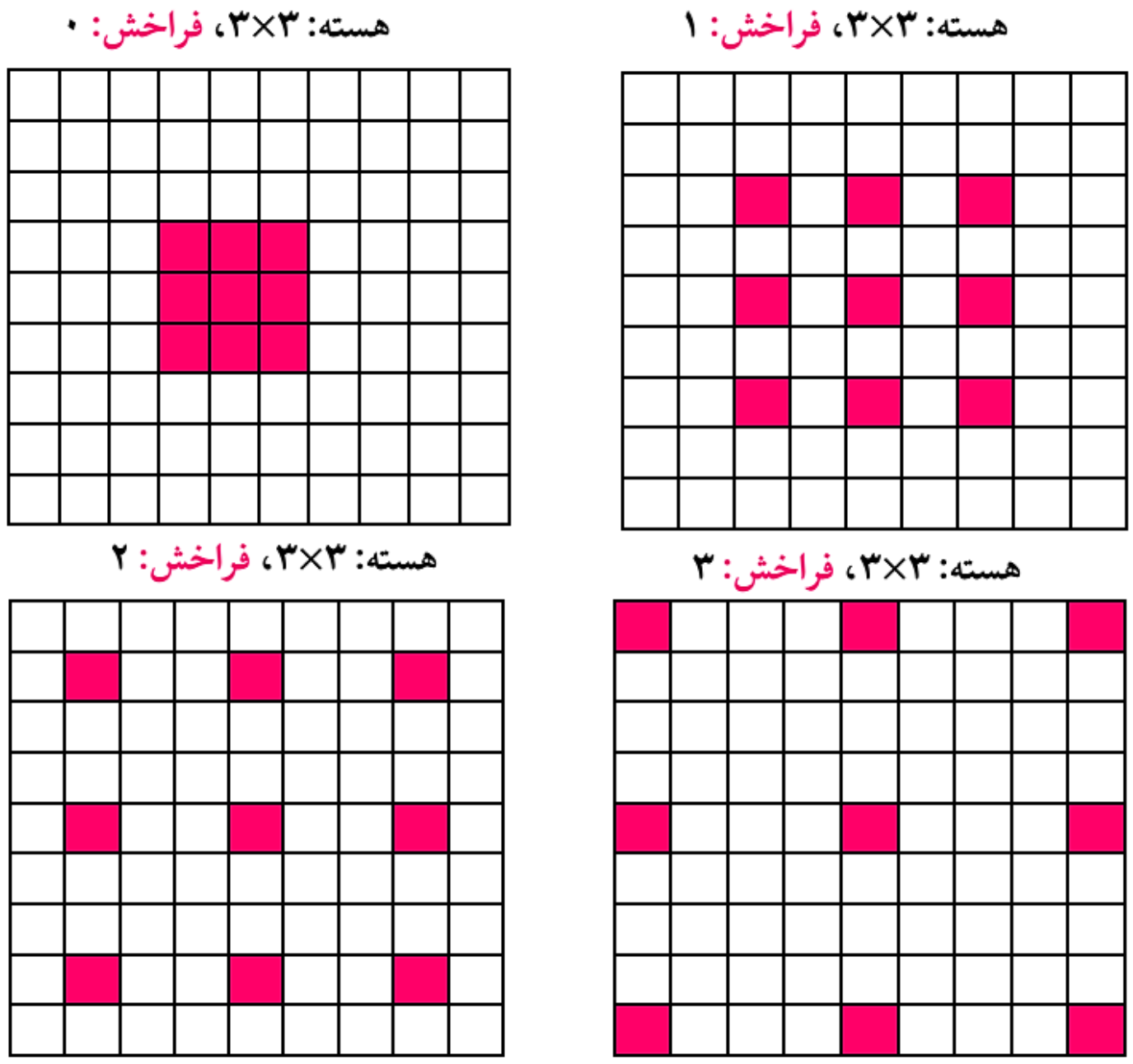

**شکل ٤ـ۳. فراخش روی ورودی دو بعدی با اندازه‌های مختلف.**

استفاده از کانولوشن دارای سه مزیت مهم است. اولا، شبکه‌های عصبی کانولوشنی معمولا دارای **ارتباط‌های خلوت** (**Sparse interactions**) هستند. شبکه‌های عصبی پیش‌خور از ماتریسی از پارامترها استفاده می‌کنند که ارتباط بین واحد ورودی و خروجی را توصیف می‌کند. این بدان معناست که هر واحد خروجی با هر واحد ورودی ارتباط دارد. با این حال، شبکه‌های عصبی کانولوشنی دارای ارتباط خلوت هستند که با کوچک‌تر کردن هسته از ورودی بدست می‌آید. به عنوان مثال، یک تصویر می‌تواند میلیون‌ها یا هزاران پیکسل داشته باشد، اما در حین پردازش آن با استفاده از هسته، می‌توانیم اطلاعات معنی‌داری که ده‌ها یا صدها پیکسل هستند

را شناسایی کنیم. این بدان معنی است که ما باید پارامترهای کم‌تری را ذخیره کنیم که نه تنها نیاز به حافظه را کاهش می‌دهد، بلکه کارایی آماری مدل را نیز بهبود می‌بخشد. ثانیاً، شبکه‌های عصبی پیچشی از **اشتراک‌گذاری پارامتر** استفاده می‌کنند. به این معنا که آن‌ها از پارامترهای مشابه برای چندین تابع دوباره استفاده می‌کنند. اشتراک‌گذاری پارامترها همچنین باعث آخرین مزیت اصلی یعنی **هم‌وردایی (Equivariance)** می‌شود. هم‌وردایی به این معنی است که اگر ورودی جابجا شود، خروجی نیز به همان صورت جابجا می‌شود. این ویژگی برای پردازش داده‌های دوبعدی ضروری است، چراکه اگر یک تصویر یا بخشی از یک تصویر به جای دیگری در تصویر منتقل شود، نمایش یکسانی خواهد داشت.

شبکه‌های عصبی پیش‌خور هر نورون ورودی را به تمام نورون‌های لایه بعدی متصل می‌کنند که به این فرآیند اتصال کامل گفته می‌شود. با این حال، این روش مستلزم محاسبه اضافی وزن‌ها است به‌طوری که بر سرعت آموزش مدل تاثیر زیادی می‌گذارد. به جای اتخاذ ارتباط کامل، CNN از اتصال جزئی استفاده می‌کند، یعنی هر نورون فقط به ناحیه‌ای از لایه ورودی متصل است که به عنوان میدان پذیرای محلی نورون پنهان شناخته می‌شود. بنابراین، یک CNN پارامترهای کم‌تری نسبت به شبکه‌های عصبی پیش‌خور دارد و در نتیجه سرعت آموزش سریع‌تری نیز دارد

در لایه‌های کانولوشن نقشه‌های ویژگی از داده‌های ورودی با استفاده از عملگر کانولوشن بدست می‌آید.

## لایه کانولوشن در keras

برای ایجاد یک لایه کانولوشن در Keras، ابتدا باید ماژول‌های مورد نیاز را به صورت زیر وارد کنید:

```
from keras.layers import Conv2D
```

سپس می‌توانید با استفاده از فرمت زیر یک لایه کانولوشن ایجاد کنید:

```
Conv2D (filters, kernel_size, strides, padding, activation='relu', input_shape)
```

شما باید آرگومان‌های زیر را وارد کنید:

- filters: تعداد فیلترها
- kernel_size: عددی که هم ارتفاع و هم عرض پنجره کانولوشن را مشخص می‌کند.
- strides: گام کانولوشنی. اگر چیزی را مشخص نکنید، به صورت پیش‌فرض روی یک تنظیم می‌شود.

- padding: valid یا same
- activation: معمولاً از تابع فعال‌ساز relu استفاده می‌شود.

هنگام استفاده از لایه کانولوشنی خود به عنوان اولین لایه در یک مدل، باید یک آرگومان input_shape اضافی را ارائه واردکنید. این یک تاپل است که ارتفاع، عرض و عمق (به ترتیب) ورودی را مشخص می‌کند.

مطمئن شوید که آرگومان input_shape در صورتی که لایه کانولوشنی اولین لایه در شبکه شما نیست، گنجانده نشده باشد.

## لایه ادغام

از مزایای لایه‌های کانولوشن این است که تعداد پارامترهای مورد نیاز را کاهش می‌دهد، عملکرد را بهبود می‌بخشد و بیش‌برازش را کاهش می‌دهد. پس از یک عملگر کانولوشن، اغلب عملیات دیگری انجام می‌شود: **ادغام**. لایه ادغام به کاهش میزان توان محاسباتی مورد نیاز برای پردازش داده‌ها کمک می‌کند و مسئول کاهش ابعاد است. با کمک کاهش ابعاد، میزان قدرت پردازش لازم برای پردازش مجموعه داده کاهش پیدا می‌کند. ادغام را می‌توان به دو نوع تقسیم کرد: **ادغام حداکثری (maximum pooling)** و **ادغام میانگین (average pooling)**. متداول‌ترین نوع اداغام، ادغام حداکثری و ایجاد شبکه‌های (۲×۲) در هر بخش و انتخاب نورون با حداکثر مقدار فعال‌سازی در هر شبکه و کنار گذاشتن بقیه است. بدیهی است که چنین عملیاتی ۷۵٪ از نورون‌ها را دور می‌اندازد و تنها نورون‌هایی را که بیشترین نقش را ایفا می‌کنند، حفظ می‌کند. در مقابل، در ادغام میانگین، میانگینِ مقدار هسته محاسبه می‌شود (شکل ۳ـ ٤).

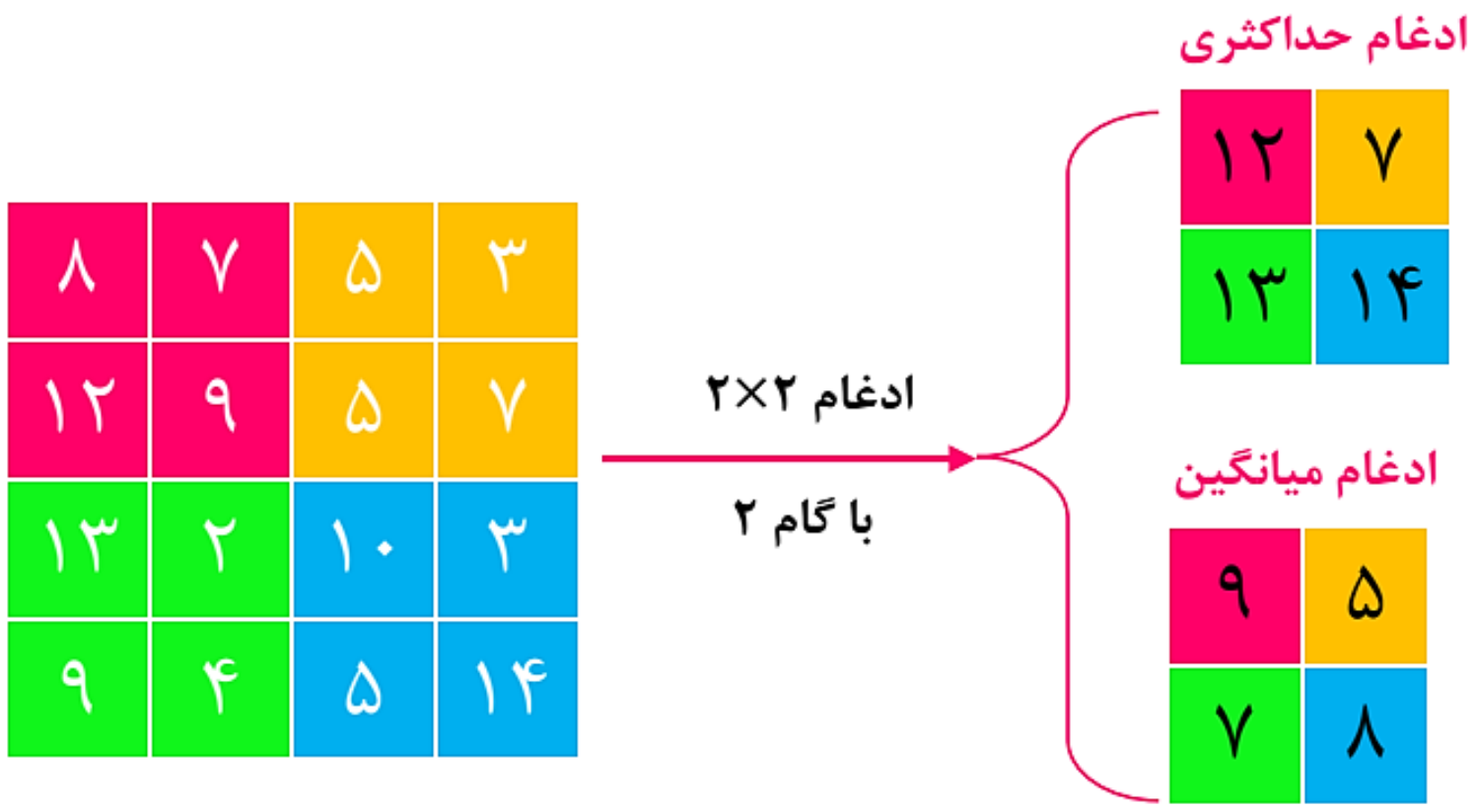

شکل ٤ـ۴. ادغام حداکثری و ادغام میانگین

برای هر لایه ادغام دو پارامترِ اندازه‌یِ **سلول** و **گام**، مشابه به پارامترهای گام و لایه‌گذاری در لایه‌های کانولوشن وجود دارد. یک انتخاب معمول انتخاب اندازه سلول ۲ وگام ۲ است. اگرچه انتخاب اندازه سلول ۳ و گام ۲ غیرمعمول نیست. البته باید توجه داشت که اگر اندازه سلول خیلی بزرگ باشد، ممکن است لایه ادغام اطلاعات زیادی را دور بیندازد و مفید نباشد.

لازم به ذکر است که مانند نحوه استفاده از توابع فعال‌سازی مختلف، می‌توانیم از عملگرهای ادغام متفاوت نیز استفاده کنیم. با این حال، استفاده از ادغام حداکثری یکی از رایج‌ترین عملگرها است، اما ادغام میانگین هم غیرمعمول نیست. در عمل، ادغام حداکثری اغلب بهتر عمل می‌کند، چراکه مرتبط‌ترین ساختارها را در تصویر حفظ می‌کند.

توجه داشته باشید که لایه‌های ادغام هیچ پارامتر جدیدی اضافه نمی‌کنند، چراکه آن‌ها به سادگی مقادیر (مانند حداکثر) را بدون نیاز به وزن یا بایاس اضافی استخراج می‌کنند.

## طبقه‌بندی تصویر با شبکه کانولوشنی در keras

برای انجام طبقه‌بندی تصویر، ابتدا به یک مجموعه داده و برچسب‌های موجود برای هر تصویر نیاز داریم. خوشبختانه، مجبور نیستیم به صورت دستی وب را برای تصاویر اسکرپ **(scrape)** کنیم و خودمان آن‌ها را برچسب‌گذاری کنیم، چراکه چند مجموعه داده استاندارد وجود دارد که می‌توانیم از آن‌ها استفاده کنیم. برای این مثال، از مجموعه داده **CIFAR-10** استفاده خواهیم کرد. جزئیات مجموعه داده به شرح زیر است:

- **ابعاد تصاویر:** تصاویر کوچک ۳۲ × ۳۲ پیکسل
- **برچسب‌ها:** ۱۰ برچسب شامل: هواپیما، خودرو، پرنده، گربه، گوزن، سگ، قورباغه، اسب، کشتی و کامیون
- **اندازه مجموعه داده:** ۶۰۰۰۰ تصویر که به ۵۰۰۰۰ داده برای آموزش و ۱۰۰۰۰ داده برای آزمایش تقسیم شده‌اند.

اولین کاری که باید انجام دهیم این است که مجموعه داده تصویر را وارد کنیم. این کار را با Keras انجام می‌دهیم؛ با اجرای کد زیر در jupyter notebook:

```
from keras.datasets import cifar10
(x_train, y_train), (x_test, y_test) = cifar10.load_data()
```

```
Downloading data from https://www.cs.toronto.edu/~kriz/cifar-10-python.tar.gz
170500096/170498071 [==============================] - 2s 0us/step
170508288/170498071 [==============================] - 2s 0us/step
```

اکنون داده‌هایی که نیاز داریم در آرایه‌های مربوط (x_train, y_train) و (x_test, y_test) ذخیره شده‌اند. اجازه دهید کمی مجموعه داده را مورد بررسی قرار دهیم. بیایید ببینیم شکل آرایه‌یِ ویژگی‌هایِ ورودی ما چگونه است:

```
print('x_train shape:', x_train.shape)
```

```
x_train shape: (50000, 32, 32, 3)
```

شکل آرایه به ما می‌گوید که x_train مجموعه داده شامل موارد زیر است:

- ۵۰۰۰۰ عکس
- ارتفاع ۳۲ پیکسل
- عرض ۳۲ پیکسل
- ۳ پیکسل در عمق (مرتبط با قرمز، سبز و آبی)

بیایید ببینیم شکل آرایه برچسب چگونه است:

```
print('y_train shape:', y_train.shape)
```

```
y_train shape: (50000, 1)
```

این بدان معنا است که برای هر یک از ۵۰۰۰۰ تصویر یک عدد (مرتبط با برچسب) وجود دارد.

اکنون، بیایید سعی کنیم نمونه‌ای از یک تصویر و برچسب آن را برای فهم بهتر ببینیم:

```
print(x_train[0])
```

```
[[[ 59  62  63]
  [ 43  46  45]
  [ 50  48  43]
  ...
  [158 132 108]
  [152 125 102]
  [148 124 103]]

 [[ 16  20  20]
  [  0   0   0]
  [ 18   8   0]
  ...
  [123  88  55]
  [119  83  50]
  [122  87  57]]

 [[ 25  24  21]
  [ 16   7   0]
  [ 49  27   8]
  ...
  [118  84  50]
  [120  84  50]
  [109  73  42]]

 ...

 [[208 170  96]
  [201 153  34]
  [198 161  26]
  ...
  [216 184 140]
  [151 118  84]
  [123  92  72]]]
```

در حالی‌که رایانه تصویر را اینگونه می‌بیند، اما برای ما چندان مفید نیست. بنابراین بیایید این تصویر از x_train[0] را با استفاده از بسته matplotlib مصورسازی کنیم:

```
import matplotlib.pyplot as plt
img = plt.imshow(x_train[0])
```

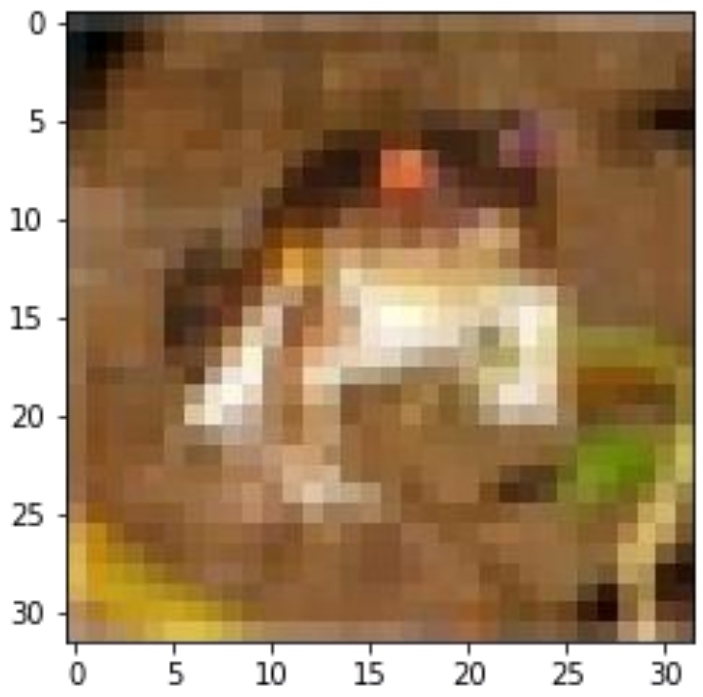

plt.imshow تابعی است که مقادیر پیکسل شماره‌گذاری شده را در `x_train[0]` به تصویر واقعی نشان می‌دهد. تصویر نمایش داده شده در بالا بسیار پیکسلی است، این به دلیل این است که اندازه تصویر ۳۲ × ۳۲ پیکسل است که بسیار کوچک می‌باشد. حال بیایید ببینیم که برچسب این تصویر در مجموعه داده ما چیست:

```
print('The label is:', y_train[0])
```

```
The label is: [6]
```

می‌بینیم که برچسب عدد "٦" است. تبدیل اعداد به برچسب بر اساس حروف الفبای انگلیسی به صورت زیر مرتب شده است:

| برچسب | شماره |
|---|---|
| هواپیما | ۰ |
| خودرو | ۱ |
| پرنده | ۲ |
| گربه | ۳ |
| گوزن | ۴ |
| سگ | ۵ |
| قورباغه | ۶ |
| اسب | ۷ |
| کشتی | ۸ |
| کامیون | ۹ |

بنابراین، از جدول می‌بینیم که تصویر بالا به عنوان تصویر یک قورباغه برچسب‌گذاری شده است (برچسب ٦). بیایید نمونه دیگری از یک تصویر را با تغییر شاخص به ۱ (تصویر دوم در مجموعه داده ما) به جای ۰ (تصویر اول در مجموعه داده ما) مشاهده کنیم:

```
img = plt.imshow(x_train[1])
```

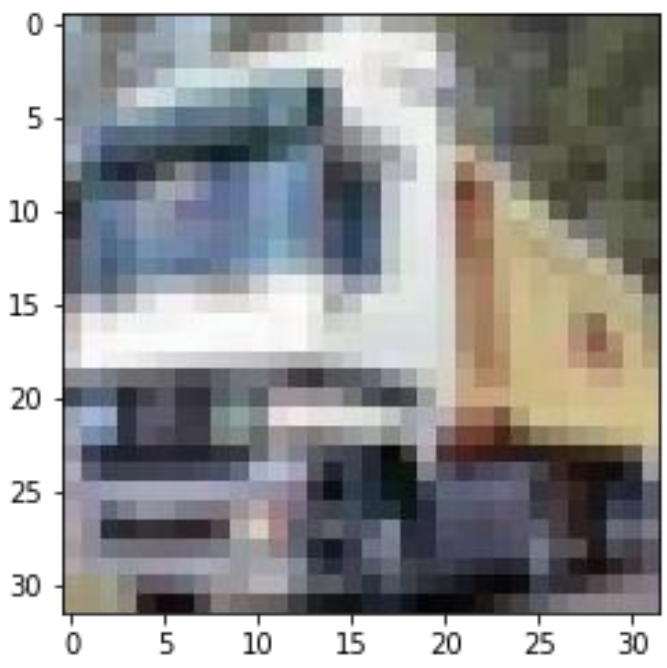

بیایید برچسب آن را نیز نمایش دهیم:

```
print('The label is:', y_train[1])
```

```
The label is: [9]
```

با استفاده از جدول پیشین، می‌بینیم که این تصویر به عنوان کامیون برچسب‌گذاری شده است.

اکنون که مجموعه داده خود را بررسی کرده‌ایم، باید آن را پردازش کنیم. اولین مشاهده‌ای که انجام می‌دهیم این است که برچسب‌های ما به عنوان شماره کلاس خیلی مفید نیستند. این به این دلیل است که کلاس‌ها نظم و ترتیبی ندارند. برای روشن شدن این موضوع مثالی می‌زنیم. اگر شبکه عصبی ما نتواند تصمیمی بگیرد که آیا تصویر یک خودرو (برچسب: ۱) است یا یک کامیون (برچسب: ۹)، چه اتفاقی می‌افتد. آیا باید میانگین در نظر بگیریم و آن را به عنوان یک سگ پیش‌بینی کنیم (برچسب: ۵)؟ قطعا چنین چیزی هیچ معنایی دارد.

در فصل پیشین، اولین شبکه عصبی خود را برای پیش‌بینی قیمت خانه با Keras ساختیم، ممکن است تعجب کنید که چرا توانستیم از برچسب‌های [۰] و [۱] در آنجا استفاده کنیم. این به این دلیل است که فقط دو کلاس وجود دارد و ما می‌توانیم خروجی شبکه عصبی را به عنوان یک احتمال تفسیر کنیم. یعنی اگر خروجی شبکه عصبی ۰/۶ باشد، به این معنی است که معتقد است با احتمال ۶۰ درصد بالاتر از میانگین قیمت خانه است. با این حال، این در پیکربندی چند کلاسه همانند این مثال کار نمی‌کند، جایی که تصویر می‌تواند به یکی از ۱۰ کلاس مختلف تعلق داشته باشد.

آنچه ما واقعا می‌خواهیم احتمال هر یک از ۱۰ کلاس مختلف است. برای آن، ما به ۱۰ نورون خروجی در شبکه عصبی خود نیاز داریم. از آنجایی که ما ۱۰ نورون خروجی داریم، برچسب‌های ما نیز باید با آن مطابقت داشته باشند. برای انجام این کار، برچسب را به مجموعه‌ای از ۱۰ عدد تبدیل می‌کنیم که هر عدد نشان می‌دهد آیا تصویر متعلق به آن کلاس است یا خیر. بنابراین اگر یک تصویر متعلق به کلاس اول باشد، اولین عدد این مجموعه ۱ و

تمام اعداد دیگر در این مجموعه ۰ خواهند بود. به این کدگذاری **one-hot** می‌گویند و جدول تبدیل برای این مثال به این صورت است:

| کدگذاری one-hot | برچسب | شماره |
|---|---|---|
| [1000000000] | هواپیما | ۰ |
| [0100000000] | خودرو | ۱ |
| [0010000000] | پرنده | ۲ |
| [0001000000] | گربه | ۳ |
| [0000100000] | گوزن | ۴ |
| [0000010000] | سگ | ۵ |
| [0000001000] | قورباغه | ۶ |
| [0000000100] | اسب | ۷ |
| [0000000010] | کشتی | ۸ |
| [0000000001] | کامیون | ۹ |

برای انجام این تبدیل ، دوباره از Keras استفاده می‌کنیم:

```
from keras.utils import np_utils
y_train_one_hot = keras.utils.np_utils.to_categorical(y_train, 10)
y_test_one_hot = keras.utils.np_utils.to_categorical(y_test, 10)
```

خط `y_train_one_hot = keras.utils.np_utils.to_categorical(y_train, 10)` به این معنی است که آرایه اولیه را فقط با عدد y_train می‌گیریم و آن را به کدگذاری one_hot، y_train_one_hot تبدیل می‌کنیم. عدد ۱۰ به عنوان پارامتر مورد نیاز است زیرا باید به تابع بگویید چند کلاس وجود دارد.

حال، فرض کنید می‌خواهیم ببینیم که برچسب تصویر دوم ما (کامیون با برچسب: ۹) در این کدگذاری چگونه به نظر می‌رسد:

```
print('The one hot label is:', y_train_one_hot[1])
```

```
The one hot label is: [0. 0. 0. 0. 0. 0. 0. 0. 0. 1.]
```

اکنون که برچسب‌های خود (y) را پردازش کرده‌ایم، ممکن است بخواهیم تصویر خود (x) را نیز پردازش کنیم. گام متداولی که ما انجام می‌دهیم این است که اجازه دهیم مقادیر بین ۰ و ۱ باشد که به آموزش شبکه عصبی ما کمک می‌کند. از آنجایی که مقادیر پیکسل ما از قبل مقادیری بین ۰ تا ۲۵۵ می‌گیرند، به سادگی باید آن‌ها را بر ۲۵۵ تقسیم کنیم:

```
x_train = x_train.astype('float32')
x_test = x_test.astype('float32')
x_train = x_train / 255
x_test = x_test / 255
```

در عمل، کاری که ما انجام می‌دهیم این است که نوع را به "float32" تبدیل می‌کنیم، که یک نوع داده است که می‌تواند مقادیر را با اعشار ذخیره کند. سپس، هر سلول را بر ۲۵۵ تقسیم می‌کنیم. در صورت تمایل، می‌توانید با اجرای سلول به مقادیر آرایه‌یِ اولین تصویر آموزشی نگاه کنید:

```
x_train[0]
```

```
array([[[0.23137255, 0.24313726, 0.24705882],
        [0.16862746, 0.18039216, 0.1764706 ],
        [0.19607843, 0.1882353 , 0.16862746],
        ...,
        [0.61960787, 0.5176471 , 0.42352942],
        [0.59607846, 0.49019608, 0.4       ],
        [0.5803922 , 0.4862745 , 0.40392157]],

       [[0.0627451 , 0.07843138, 0.07843138],
        [0.        , 0.        , 0.        ],
        [0.07058824, 0.03137255, 0.        ],
        ...,
        [0.48235294, 0.34509805, 0.21568628],
        [0.46666667, 0.3254902 , 0.19607843],
        [0.47843137, 0.34117648, 0.22352941]],

       [[0.09803922, 0.09411765, 0.08235294],
        [0.0627451 , 0.02745098, 0.        ],
        [0.19215687, 0.10588235, 0.03137255],
        ...,
        [0.4627451 , 0.32941177, 0.19607843],
        [0.47058824, 0.32941177, 0.19607843],
        [0.42745098, 0.28627452, 0.16470589]],

       ...,

       [[0.8156863 , 0.6666667 , 0.3764706 ],
        [0.7882353 , 0.6       , 0.13333334],
        [0.7764706 , 0.6313726 , 0.10196079],
        ...,
        [0.627451  , 0.52156866, 0.27450982],
        [0.21960784, 0.12156863, 0.02745098],
        [0.20784314, 0.13333334, 0.07843138]],

       [[0.7058824 , 0.54509807, 0.3764706 ],
        [0.6784314 , 0.48235294, 0.16470589],
        [0.7294118 , 0.5647059 , 0.11764706],
        ...,
        [0.72156864, 0.5803922 , 0.36862746],
        [0.38039216, 0.24313726, 0.13333334],
        [0.3254902 , 0.20784314, 0.13333334]],

       [[0.69411767, 0.5647059 , 0.45490196],
        [0.65882355, 0.5058824 , 0.36862746],
        [0.7019608 , 0.5568628 , 0.34117648],
        ...,
        [0.84705883, 0.72156864, 0.54901963],
        [0.5921569 , 0.4627451 , 0.32941177],
        [0.48235294, 0.36078432, 0.28235295]]], dtype=float32)
```

تاکنون ما فقط یک مجموعه آموزشی و یک مجموعه آزمایشی داریم. برخلاف مثال فصل پیش، مجموعه اعتبارسنجی خود را از قبل تقسیم نمی‌کنیم، چراکه میانبری برای این کار وجود دارد که بعدا معرفی خواهیم کرد.

مشابه با مثال قبلی، ابتدا باید معماری مدلی خود را تعریف کنیم. معماری CNN که ما خواهیم ساخت به شکل زیر است:

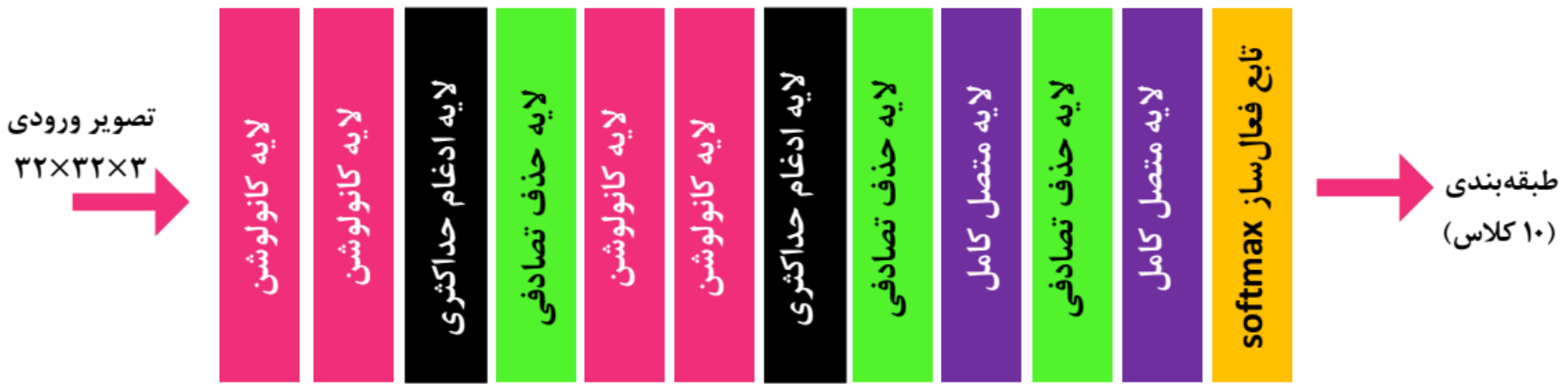

و مقادیرِ پارامترهای معماری بالا به صورت زیر خلاصه شده‌اند:

- Conv Layer (32 Filter size 3×3)
- Conv Layer (32 Filter size 3×3)
- Max Pool Layer (Filter size 2×2)
- Dropout Layer (Prob of dropout 0.25)
- Conv Layer (64 Filter size 3×3)
- Conv Layer (64 Filter size 3×3)
- Max Pool Layer (Filter size 2×2)
- Dropout Layer (Prob of dropout 0.25)
- FC Layer (512 neurons)
- Dropout Layer (Prob of dropout 0.5)
- FC Layer, Softmax (10 neurons)

این معماری دارای تعداد لایه‌های زیادی است (بیشتر از آنچه تاکنون دیده‌ایم)، اما همه از مفاهیمی ساخته شده‌اند که قبلا دیده‌ایم. با این حال، ساخت هر لایه فقط با یک خط کد انجام می‌شود و جای نگرانی نیست! به یاد بیاورید که تابع softmax به سادگی خروجی لایه قبلی را به توزیع‌های احتمالی تبدیل می‌کند، چیزی که ما برای مسئله طبقه‌بندی خود می‌خواهیم.

برای کدنویسی این مورد، از مدل ترتیبی Keras استفاده خواهیم کرد. با این حال، از آنجایی که ما لایه‌های زیادی در مدل خود داریم، روش جدیدی را برای تعیین دنباله معرفی می‌کنیم. ما کد را خط به خط مرور می‌کنیم تا بتوانید دقیقا آنچه را که انجام می‌دهیم دنبال کنید. ابتدا بخشی از کدهای مورد نیاز خود را وارد می‌کنیم:

```
from keras.models import Sequential
from keras.layers import Dense, Dropout, Flatten, Conv2D, MaxPooling2D
```

سپس، ما یک مدل ترتیبی خالی را ایجاد می‌کنیم:

```
model = Sequential()
```

ما هر بار یک لایه به این مدل خالی اضافه می‌کنیم. لایه اول (اگر از شکل قبلی با یاد داشته باشید) یک لایه کانولوشنی با اندازه فیلتر 3×3، اندازه گام ۱ و عمق ۳۲ است. لایه‌گذاری "same" و فعال‌ساز آن "relu" است (این دو پیکربندی برای همه لایه‌های CNN اعمال می‌شود). با این حال، اجازه دهید اولین لایه خود را با کد مشخص کنیم:

```
model.add(Conv2D(32, (3, 3), activation='relu', padding='same',
input_shape=(32,32,3)))
```

کاری که این تکه کد انجام می‌دهد، اضافه کردن این لایه به مدل ترتیبی خالی ما با استفاده از تابع `model.add()` است. اولین عدد یعنی ۳۲ به تعداد فیلترها اشاره دارد. جفت اعداد بعدی (۳،۳) به عرض و اندازه فیلتر اشاره دارد. سپس، فعال‌سازی را که `'relu'` و لایه‌گذاری را که `'same'` است مشخص می‌کنیم. توجه داشته باشید که ما گام را مشخص نکردیم. دلیلش این است که `stride=1` یک تنظیم پیش‌فرض است و تا زمانی که بخواهیم این تنظیم را تغییر ندهیم، نیازی به تعیین آن نداریم. اگر به خاطر داشته باشید، ما همچنین باید اندازه ورودی را برای لایه اول خود مشخص کنیم. لایه‌های بعدی این مشخصات را ندارند، چراکه می‌توانند اندازه ورودی را از اندازه خروجی لایه قبلی استنتاج کنند. لایه دوم ما در کد به این شکل است (نیازی به تعیین اندازه ورودی نداریم):

```
model.add(Conv2D(32, (3, 3), activation='relu', padding='same'))
```

لایه بعدی یک لایه ادغام حداکثری با اندازه ادغام ۲×۲ وگام ۲ است. پیش‌فرض برای گام ادغام حداکثری، اندازه ادغام است، بنابراین ما نیازی به تعیین گام نداریم:

```
model.add(MaxPooling2D(pool_size=(2, 2)))
```

در نهایت، یک لایه حذفی با احتمال ۰/۲۵، اضافه می‌کنیم تا از بیش‌برازش جلوگیری کنیم:

```
model.add(Dropout(0.25))
```

اکنون چهار لایه اول را با کد ایجاد کردیم. چهار لایه بعدی واقعا شبیه بهم به نظر می‌رسند:

```
model.add(Conv2D(64, (3, 3), activation='relu', padding='same'))
model.add(Conv2D(64, (3, 3), activation='relu', padding='same'))
model.add(MaxPooling2D(pool_size=(2, 2)))
model.add(Dropout(0.25))
```

در نهایت، ما باید لایه متصل کامل خود را کدنویسی کنیم، که مشابه کاری است که در مثال فصل پیش انجام دادیم. با این حال، در این مرحله، نورون‌های ما به‌جای یک ردیف، در قالب مکعب‌مانند قرار گرفته‌اند. برای اینکه این قالب مکعب‌مانند نورون‌ها را در یک ردیف قرار دهیم، ابتدا باید آن را صاف کنیم. این کار را با افزودن یک لایه Flatten انجام می‌دهیم:

```
model.add(Flatten())
```

اکنون، باید یک لایه متصل کامل با ۵۱۲ نورون و فعال‌ساز `relu` بسازیم:

```
model.add(Dense(512, activation='relu'))
```

سپس یک حذف تصادفی دیگر با احتمال ۰/۵ اضافه می‌کنیم:

```
model.add(Dropout(0.5))
```

و در آخر، ما یک لایه متصل کامل با ۱۰ نورون و فعال‌ساز softmax می‌سازیم:

```
model.add(Dense(10, activation='softmax'))
```

ساخت معماری ما اکنون به پایان رسید. حال، برای مشاهده خلاصه‌ای از معماری کامل، کد زیر را اجرا می‌کنیم:

```
model.summary()
```

```
Model: "sequential"
_________________________________________________________________
Layer (type)                 Output Shape              Param #
=================================================================
conv2d (Conv2D)              (None, 32, 32, 32)        896

conv2d_1 (Conv2D)            (None, 32, 32, 32)        9248

max_pooling2d (MaxPooling2D)  (None, 16, 16, 32)        0

max_pooling2d_1 (MaxPooling2D)    (None, 8, 8, 32)        0


dropout (Dropout)            (None, 8, 8, 32)          0

flatten (Flatten)            (None, 2048)              0

dense (Dense)                (None, 512)               1049088

dropout_1 (Dropout)          (None, 512)               0

dense_1 (Dense)              (None, 10)                5130

=================================================================
Total params: 1,064,362
Trainable params: 1,064,362
Non-trainable params: 0
_________________________________________________________________
```

سپس، مدل را با تنظیمات خود کامپایل می‌کنیم:

```
model.compile(loss='categorical_crossentropy',
              optimizer='adam',
              metrics=['accuracy'])
```

تابع زیانی که ما استفاده می‌کنیم آنتروپی متقاطع طبقه‌ای نامیده می‌شود. بهینه ساز ما در اینجا adam است. در نهایت، ما می‌خواهیم دقت مدل خود را ردیابی کنیم.

اکنون، زمان اجرای آموزش مدل است:

```
hist = model.fit(x_train, y_train_one_hot,
           batch_size=32, epochs=20,
           validation_split=0.2)
```

ما مدل خود را با اندازه دسته‌یِ ۳۲ و ۲۰ دوره آموزش می‌دهیم. با این حال، یک تفاوت در کد را متوجه شدید؟ ما از تنظیم validation_split=0.2 به جای validation_data استفاده می‌کنیم. با این میانبر، ما نیازی به تقسیم مجموعه داده‌های خود به یک مجموعه آموزشی و مجموعه

اعتبارسنجی در شروع نداریم. در عوض، ما به سادگی مشخص می‌کنیم که چه مقدار از مجموعه داده ما به عنوان یک مجموعه اعتبارسنجی استفاده می‌شود. در این مورد، ۲۰٪ از مجموعه داده ما به عنوان یک مجموعه اعتبارسنجی استفاده می‌شود. سلول را اجرا کنید و خواهید دید که مدل شروع به آموزش می‌کند:

```
Epoch 1/20
1250/1250 [==============================] - 27s 14ms/step - loss: 1.4824 - accuracy: 0.4623 - val_loss: 1.1698 -
val_accuracy: 0.5885
Epoch 2/20
1250/1250 [==============================] - 17s 14ms/step - loss: 1.1323 - accuracy: 0.6000 - val_loss: 1.0689 -
val_accuracy: 0.6276
Epoch 3/20
1250/1250 [==============================] - 17s 13ms/step - loss: 0.9810 - accuracy: 0.6534 - val_loss: 0.9494 -
val_accuracy: 0.6725
Epoch 4/20
1250/1250 [==============================] - 17s 14ms/step - loss: 0.8859 - accuracy: 0.6900 - val_loss: 0.9151 -
val_accuracy: 0.6828
Epoch 5/20
1250/1250 [==============================] - 18s 15ms/step - loss: 0.8015 - accuracy: 0.7171 - val_loss: 0.9117 -
val_accuracy: 0.6802
Epoch 6/20
1250/1250 [==============================] - 18s 14ms/step - loss: 0.7203 - accuracy: 0.7469 - val_loss: 0.8959 -
val_accuracy: 0.6874
Epoch 7/20
1250/1250 [==============================] - 17s 14ms/step - loss: 0.6486 - accuracy: 0.7709 - val_loss: 0.8811 -
val_accuracy: 0.7066
Epoch 8/20
1250/1250 [==============================] - 17s 14ms/step - loss: 0.5960 - accuracy: 0.7902 - val_loss: 0.9053 -
val_accuracy: 0.7055
Epoch 9/20
1250/1250 [==============================] - 17s 13ms/step - loss: 0.5312 - accuracy: 0.8129 - val_loss: 0.9100 -
val_accuracy: 0.6982
Epoch 10/20
1250/1250 [==============================] - 18s 14ms/step - loss: 0.4887 - accuracy: 0.8262 - val_loss: 0.9183 -
val_accuracy: 0.7077
Epoch 11/20
1250/1250 [==============================] - 18s 14ms/step - loss: 0.4483 - accuracy: 0.8415 - val_loss: 0.9454 -
val_accuracy: 0.7083
Epoch 12/20
1250/1250 [==============================] - 17s 13ms/step - loss: 0.4195 - accuracy: 0.8514 - val_loss: 1.0072 -
val_accuracy: 0.7062
Epoch 13/20
1250/1250 [==============================] - 18s 14ms/step - loss: 0.3889 - accuracy: 0.8625 - val_loss: 0.9655 -
val_accuracy: 0.7089
Epoch 14/20
1250/1250 [==============================] - 18s 14ms/step - loss: 0.3591 - accuracy: 0.8764 - val_loss: 1.0041 -
val_accuracy: 0.7042
Epoch 15/20
1250/1250 [==============================] - 17s 13ms/step - loss: 0.3372 - accuracy: 0.8814 - val_loss: 1.0220 -
val_accuracy: 0.7133
Epoch 16/20
1250/1250 [==============================] - 18s 14ms/step - loss: 0.3242 - accuracy: 0.8868 - val_loss: 1.0927 -
val_accuracy: 0.7041
Epoch 17/20
1250/1250 [==============================] - 23s 18ms/step - loss: 0.3053 - accuracy: 0.8925 - val_loss: 1.0579 -
val_accuracy: 0.7046
Epoch 18/20
1250/1250 [==============================] - 18s 14ms/step - loss: 0.2917 - accuracy: 0.8990 - val_loss: 1.0833 -
val_accuracy: 0.7061
Epoch 19/20
1250/1250 [==============================] - 17s 14ms/step - loss: 0.2825 - accuracy: 0.9014 - val_loss: 1.0957 -
val_accuracy: 0.7165
Epoch 20/20
1250/1250 [==============================] - 17s 14ms/step - loss: 0.2700 - accuracy: 0.9064 - val_loss: 1.0855 -
val_accuracy: 0.7131
```

پس از اتمام آموزش، می‌توانیم با استفاده از این کدی که در ساخت اولین شبکه عصبی خود دیده‌ایم، زیان آموزش و اعتبارسنجی را در طول تعداد دوره‌ها مصورسازی کنیم:

```
plt.plot(hist.history['loss'])
plt.plot(hist.history['val_loss'])
plt.title('Model loss')
plt.ylabel('Loss')
plt.xlabel('Epoch')
plt.legend(['Train', 'Val'], loc='upper right')
plt.show()
```

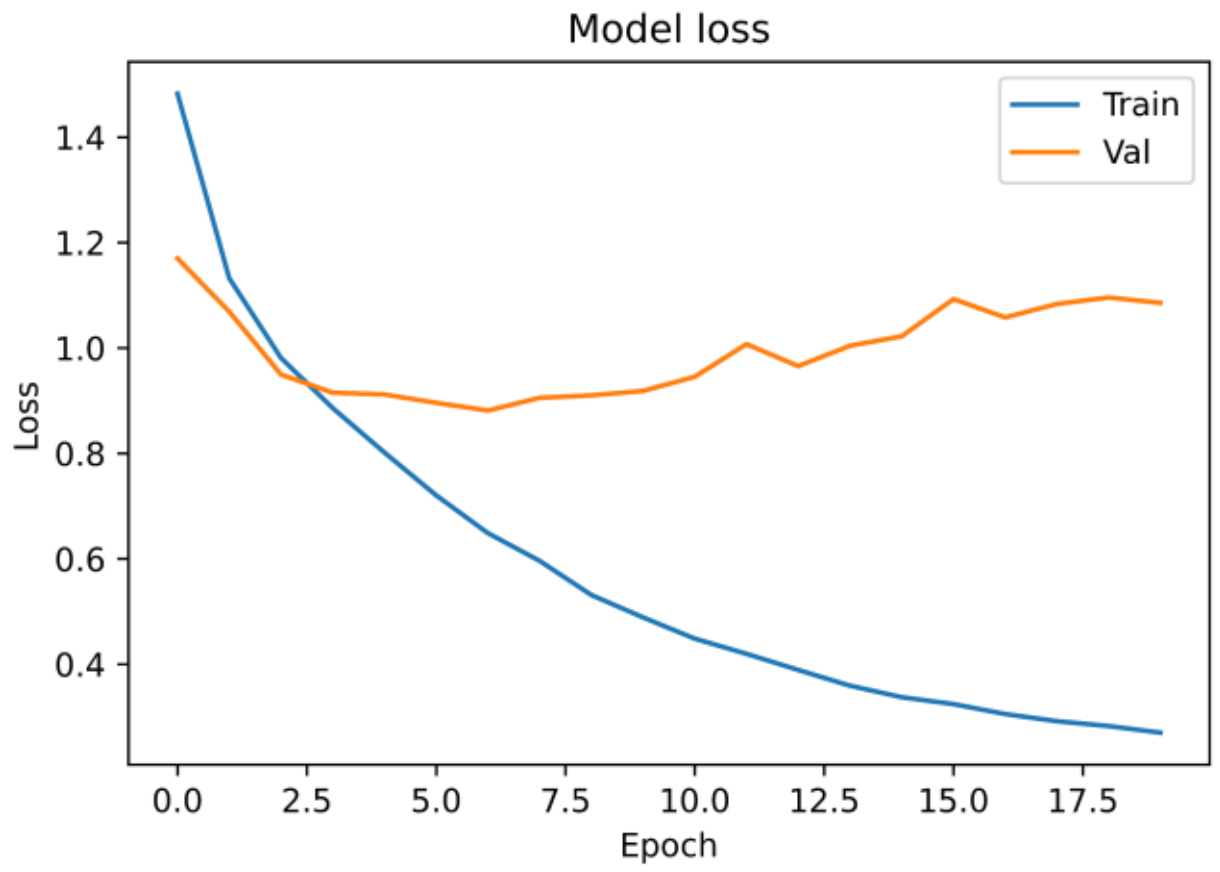

ما همچنین می‌توانیم دقت را مصورسازی کنیم:

```
plt.plot(hist.history['loss'])
plt.plot(hist.history['val_loss'])
plt.title('Model loss')
plt.ylabel('Loss')
plt.xlabel('Epoch')
plt.legend(['Train', 'Val'], loc='upper right')
plt.show()
```

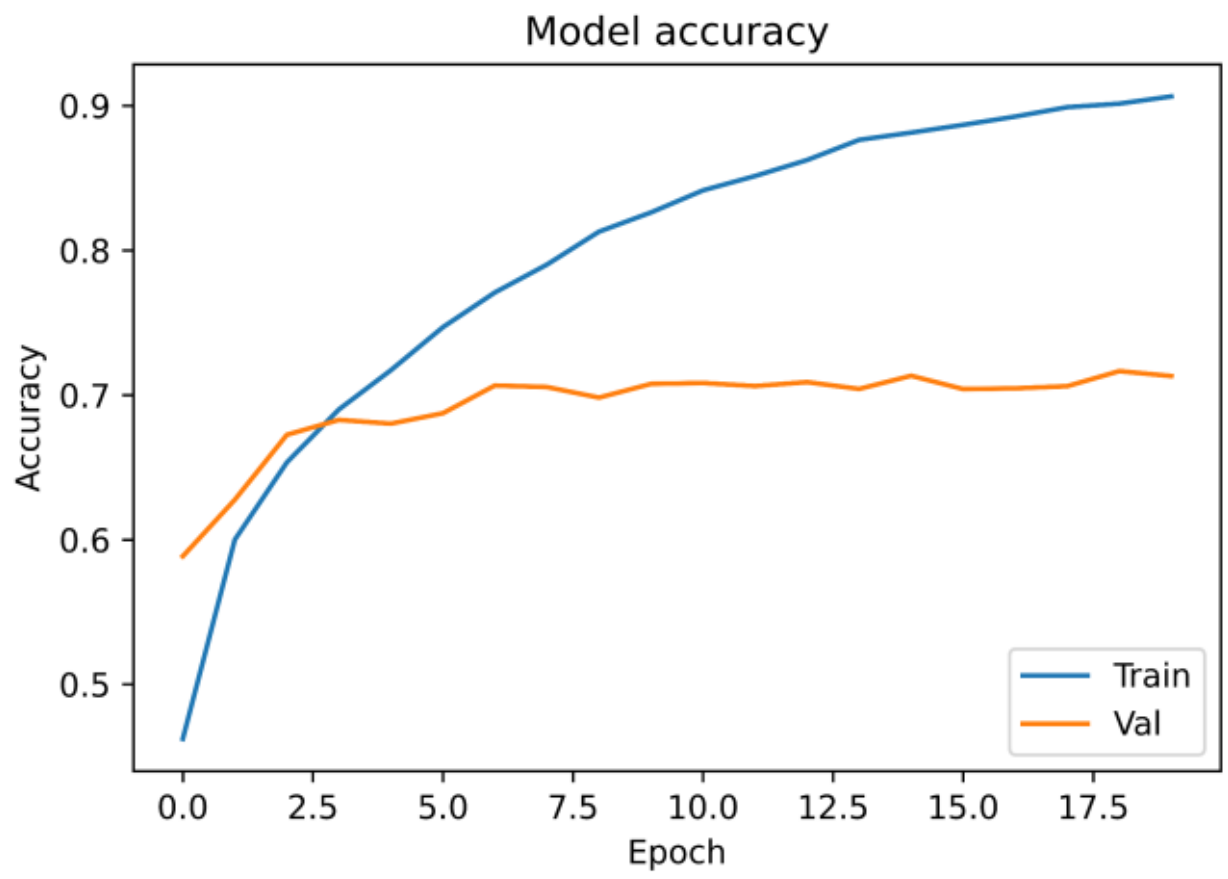

همان‌طور که مشاهده می‌شود، مدل دچار بیش‌برازش شده است. در این مرحله، به شما توصیه می‌کنم که به عقب برگردید و پارامترهای مختلف مانند تغییر معماری یا تغییر تعداد دوره‌ها را امتحان کنید تا ببینید آیا می‌توانید دقت val بهتری دریافت کنید. هنگامی که از مدل خود راضی بودید، می‌توانید آن را در مجموعه آزمایشی ارزیابی کنید:

```
model.evaluate(x_test, y_test_one_hot)[1]
```

```
313/313 [==============================] - 2s 5ms/step - loss: 1.1410 - accuracy: 0.6924
0.6923999786376953
```

همان‌طور که مشاهده می‌شود، مدل کارایی چندانی ندارد. با این حال از حدس زدن تصادفی بهتر عمل می‌کند.

در این مرحله، ممکن است بخواهید مدل آموزش دیده خود را ذخیره کنید (با فرض این‌که یک مدل باکارایی خوب با تنظیم دقیق ابرپارمترها ساخته‌اید). مدل در قالب فایلی به نام HDF5 (با پسوند h5) ذخیره می‌شود. ما مدل خود را با این خط کد ذخیره می‌کنیم:

```
model.save('my_cifar10_model.h5')
```

اگر می‌خواهید مدل ذخیره شده خود را در آینده بارگیری کنید، از این خط کد استفاده کنید:

```
from keras.models import load_model
model = load_model('my_cifar10_model.h5')
```

به‌طور خلاصه، ما اولین CNN خود را برای ایجاد یک طبقه‌بند تصویر ساخته‌ایم. برای انجام این کار، ما از مدل Keras Sequential برای مشخص کردن معماری استفاده کرده‌ایم و آن را بر روی مجموعه داده‌ای که قبلا پردازش کرده‌ایم آموزش داده‌ایم. ما همچنین مدل خود را ذخیره کرده‌ایم تا بتوانیم بعدا از آن برای انجام طبقه‌بندی تصاویر بدون نیاز به آموزش مجدد مدل استفاده کنیم.

اکنون که یک مدل داریم، بیایید آن را روی تصاویر خودمان امتحان کنیم. برای انجام این کار، یک تصویر (بر اساس یکی از ۱۰ کلاس مجموعه داده) از اینترنت دانلود کرده و در همان پوشه notebook خود قرار دهید. ما از تصویر گربه زیر استفاده می‌کنیم:

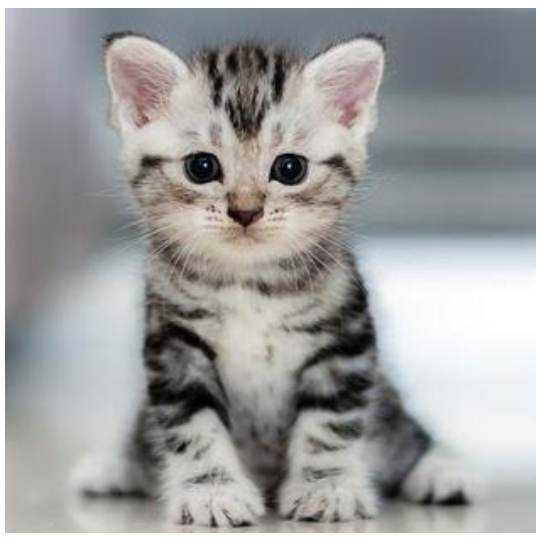

فایل تصویری من "cat.jpg" است. اکنون این فایل JPEG را به صورت آرایه‌ای از مقادیر پیکسل می‌خوانیم:

```
my_image = plt.imread("cat.jpg")
```

اولین کاری که باید انجام دهیم این است که اندازه تصویر گربه خود را تغییر دهیم تا بتوانیم آن را در مدل خود قرار دهیم (اندازه ورودی ۳ × ۳۲ × ۳۲). به جای اینکه خودمان یک تابع تغییر اندازه را کدنویسی کنیم، بیایید از بسته‌ای به نام "scikit-image" استفاده کنیم که به ما در انجام آن کمک می‌کند:

```
from skimage.transform import resize
my_image_resized = resize(my_image, (32,32,3))
```

ما می‌توانیم تصویر تغییر اندازه یافته خود را به صورت زیر مصورسازی کنیم:

```
img = plt.imshow(my_image_resized)
```

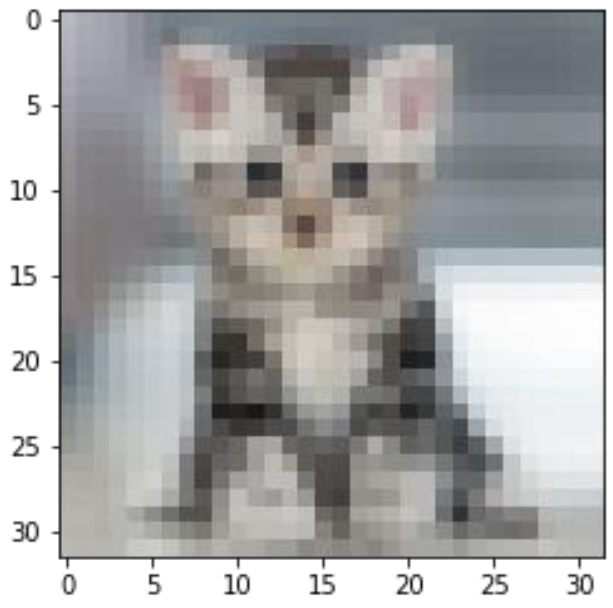

توجه داشته باشید که اندازه تصویر تغییر یافته دارای مقادیر پیکسلی است که قبلا بین ۰ و ۱ مقیاس‌بندی شده است، بنابراین نیازی نیست مراحل پیش‌پردازشی را که قبلا برای تصویر آموزشی خود انجام دادیم اعمال کنیم. اکنون، با استفاده از کد model.predict، می‌بینیم که وقتی تصویری از گربه داده می‌شود، مدل آموزش‌دیده ما چه خروجی خواهد داشت:

```
import numpy as np
probabilities = model.predict(np.array( [my_image_resized,] ))
```

خروجی‌های کد بالا، خروجی ۱۰ نورون مربوط به توزیع احتمال بر روی کلاس‌ها هستند. اگر سلول را اجرا کنیم، خواهیم داشت:

```
probabilities
```

```
array([[2.6720341e-02, 3.1344647e-05, 1.5095539e-01, 3.8518414e-01,
        3.3354717e-03, 3.2324010e-01, 5.1648129e-02, 5.7933435e-02,
        9.2716294e-04, 2.4454062e-05]], dtype=float32)
```

برای سهولت خواندن پیش‌بینی‌های مدل، قطعه کد زیر را اجرا کنید:

```
number_to_class = ['airplane', 'automobile', 'bird', 'cat', 'deer', 'dog', 'frog', 'horse', 'ship', 'truck']
index = np.argsort(probabilities[0,:])
print("Most likely class:", number_to_class[index[9]], "-- Probability:", probabilities[0,index[9]])
print("Second most likely class:", number_to_class[index[8]], "-- Probability:", probabilities[0,index[8]])
print("Third most likely class:", number_to_class[index[7]], "-- Probability:", probabilities[0,index[7]])
print("Fourth most likely class:", number_to_class[index[6]], "-- Probability:", probabilities[0,index[6]])
print("Fifth most likely class:", number_to_class[index[5]], "-- Probability:", probabilities[0,index[5]])
```

```
Most likely class: cat -- Probability: 0.31140402
Second most likely class: horse -- Probability: 0.296455
Third most likely class: dog -- Probability: 0.1401798
Fourth most likely class: truck -- Probability: 0.12088975
Fifth most likely class: frog -- Probability: 0.078746535
```

همانطور که می‌بینید، مدل به‌طور دقیق پیش‌بینی کرده است که تصویر ورودی در واقع تصویر یک گربه است. با این حال، این بهترین مدلی نیست که ما داریم و دقت آن بسیار پایین است، بنابراین نباید انتظار زیادی از آن نداشته باشید. این مثال، تنها اصول اساسی CNNها را در یک مجموعه داده بسیار ساده پوشش داده است. می‌توانید به عنوان تمرین، برای این مجموعه داده مدل‌های دیگری را بسازید و نتایج را مقایسه کنید.

## خلاصه فصل

- **شبکه‌های عصبی کانولوشنی، از کانولوشن به جای ضرب ماتریس، حداقل در یکی از لایه‌های خود استفاده می‌کنند.**
- **شناسایی ویژگی‌ها از طریق اعمال فیلترها منجر به تولید نقشه ویژگی می‌شود.**
- **از لایه کانولوشن به عنوان لایه استخراج ویژگی نیز یاد می‌شود. چراکه ویژگی‌های تصویر در این لایه استخراج می‌شوند.**

## آزمونک

۱. **معماری کلی یک شبکه کانولوشنی از چه قسمت‌های تشکیل شده است؟**
۲. **سه ویژگی متمایز شبکه‌های کانولوشنی را بیان کنید؟**
۳. **لایه کانولوشن از چه بخش‌هایی تشکیل شده است؟**
۴. **مزایای استفاده از کانولوشن چیست؟**
۵. **لایه ادغام در شبکه‌های کانولوشنی چه نقشی دارند؟**

# شبکه‌های عصبی بازگشتی

## اهداف یادگیری:

- شبکه عصبی بازگشتی چیست؟
- آشنایی با LSTM
- تولید متن و طبقه‌بندی متن با این شبکه ها

## مقدمه

معماری شبکه‌های عصبی که در فصل‌های پیشین مورد بحث قرار گرفتند، ورودی با اندازه ثابت را دریافت می‌کنند و خروجی با اندازه ثابت ارائه می‌کنند. این فصل با معرفی شبکه‌های عصبی بازگشتی (Recurrent Neural Networks) یا به اختصار RNN از این محدودیت دور می‌شویم. RNNها به ما کمک می‌کنند تا با توالی‌هایی با طول متغیر با تعریف یک ارتباط بازگشتی برروی این دنباله‌ها (sequences) مقابله کنیم. توانایی پردازش توالی‌های دلخواه ورودی باعث می‌شود RNNها برای کارهایی مانند مدل سازی زبان، تشخیص گفتار و یا غیره قابل استفاده باشند. در واقع، در تئوری، RNNها را می‌توان برای هر مشکلی اعمال کرد، زیرا ثابت شده است که آن‌ها **Turing-Complete** هستند[1]. این بدان معنی است که از نظر تئوری، آن‌ها می‌توانند هر برنامه‌ای را شبیه سازی کنند که یک کامپیوتر معمولی قادر به محاسبه آن نیست. به عنوان نمونه‌ای از این موضوع، Google DeepMind مدلی به نام **ماشین‌های تورینگ عصبی** را پیشنهاد کرده است که می‌تواند نحوه اجرای الگوریتم‌های ساده مانند مرتب‌سازی را بیاموزد.

## شبکه عصبی بازگشتی چیست؟

در فصل قبل، به تشریح معماری شبکه‌های عصبی کانولوشنی که پایه بسیاری از سیستم‌های بینایی کامپیوتری پیشرفته را تشکیل می‌دهند، پرداختیم. با این حال، ما دنیای اطراف خود را تنها با بینایی درک نمی‌کنیم. به عنوان مثال، صدا نیز نقش بسیار مهمی دارد. به‌طور مشخص‌تر، ما انسان‌ها عاشق برقراری ارتباط و بیان افکار و ایده‌های پیچیده از طریق توالی‌هایی نمادین و بازنمایی‌های انتزاعی هستیم. از این‌رو، منطقی است که ما می‌خواهیم ماشین‌ها بتوانند اطلاعات متوالی را درک کنند.

هنگامی که داده‌ها به‌گونه‌ای تنظیم شده باشند که هر قطعه به نوعی رابطه‌ای با قطعاتی که قبل و بعد از آن ایجاد می‌شوند، داشته باشد از آن‌ها به عنوان دنباله یاد می‌شود. **شبکه‌های عصبی بازگشتی** یا **مکرر**، نوعی شبکه‌ی عصبی مصنوعی هستند که برای تشخیص الگوها در داده‌های دنباله‌ای همانند متن، ژنوم، دست‌خط، کلمات گفتاری، داده‌های سری زمان، بازارهای سهام و غیره طراحی شده‌اند. ایده‌یِ پشت این شبکه‌های عصبی این است که به سلول‌ها اجازه می‌دهند از سلول‌های قبلی متصل به خود یاد بگیرند. می‌توان گفت که به نوعی، این سلول‌ها دارای "حافظه" هستند. از این‌رو، دانشِ پیچیده‌تری را از داده‌های ورودی می‌سازند.

[1] Alex Graves and Greg Wayne and Ivo Danihelka (2014). "Neural Turing Machines".

مدل‌هایی که در فصل‌های پیشین مورد مطالعه قرار گرفتند، یک ویژگی مشترک دارند. پس از تکمیل فرآیند آموزش، وزن‌ها ثابت می‌شوند و خروجی فقط به نمونه ورودی بستگی دارد. واضح است که این رفتار مورد انتظار یک طبقه‌بند است، اما سناریوهای زیادی وجود دارد که در آن پیش‌بینی باید تاریخچه مقادیر ورودی را در نظر بگیرد. سری زمانی یک نمونه کلاسیک در این خصوص است. بیایید فرض کنیم که باید دمای هفته آینده را پیش‌بینی کنیم. اگر سعی کنیم فقط آخرین مقدار $x(t)$ شناخته شده و یک MLP آموزش دیده برای پیش‌بینی $x(t+1)$ استفاده کنیم، نمی‌توان شرایط زمانی مانند فصل، تاریخچه فصل در طول سال‌ها و غیره را در نظر گرفت. رگرسیون قادر خواهد بود خروجی‌هایی را که کم‌ترین میانگین خطا را ایجاد می‌کند مرتبط کند، اما در موقعیت‌های واقعی، این کافی نیست. تنها راه معقول برای حل این مشکل این است که یک معماری جدید برای نورون مصنوعی تعریف کنیم تا یک حافظه برای آن فراهم کنیم. این مفهوم در شکل زیر نشان داده شده است:

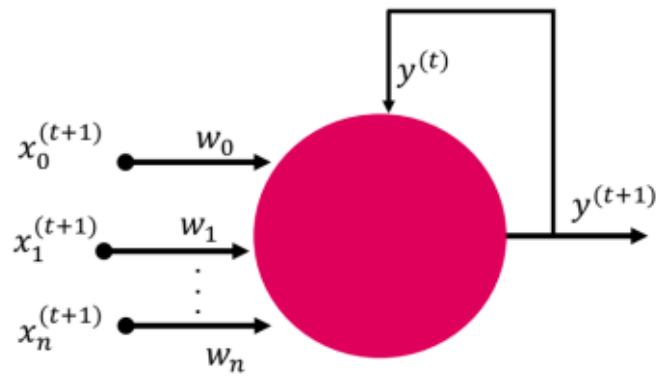

اکنون نورون دیگر یک واحد محاسباتی پیش‌خور خالص نیست، چراکه اتصال بازخوردی آن را مجبور می‌کند گذشته خود را به خاطر بسپارد و از آن برای پیش‌بینی مقادیر جدید استفاده کند.

شبکه‌های عصبی بازگشتی، نواقصِ شبکه‌هایِ عصبِی پیش‌خور را برطرف می‌کنند. چرا که شبکه‌های پیش‌خور تنها می‌توانند ورودی‌های با اندازه ثابت را بپذیرند و تنها خروجی‌هایی با اندازه ثابت تولید کنند و قادر به در نظر گرفتن ورودی‌های قبلی با همان ترتیب نیستند. با در نظر گرفتن ورودی‌های گذشته در توالی‌ها، شبکه‌ی عصبی بازگشتی قادر به گرفتن وابستگی‌های زمانی هستند که شبکه‌ی عصبی پیش‌خور قادر به آن نیست.

شبکه‌های عصبی بازگشتی، دنباله‌ای را به عنوان ورودی می‌گیرند و برای هر مرحله زمانی شبکه عصبی را ارزیابی می‌کنند. این شبکه‌ها را می‌توان به عنوان یک شبکه عصبی در نظر گرفت که دارای یک حلقه است که به آن اجازه می‌دهد حالت را حفظ کند. هنگامی که ارزیابی می‌شود، حلقه از طریق مراحلِ زمانیِ یک دنباله باز می‌شود. این حلقه‌ها یا پیوندهای مکرر دلیلی هستند که این شبکه‌ها را شبکه‌های بازگشتی می‌نامند. اینکه یک شبکه بازگشتی شامل حلقه است به این معنی است که خروجی یک نورون در یک نقطه زمانی ممکن است در نقطه زمانی دیگر به همان نورون بازگردانده شود. نتیجه این امر این است که شبکه نسبت به فعال‌سازی‌های گذشته (و بنابراین ورودی‌های گذشته که در این فعال‌سازی نقش داشته‌اند) حافظه دارد.

## ساختار شبکه عصبی بازگشتی

فرض کنید در یک شبکه عصبی سنتی، همانطور که در شکل زیر نشان داده شده است:

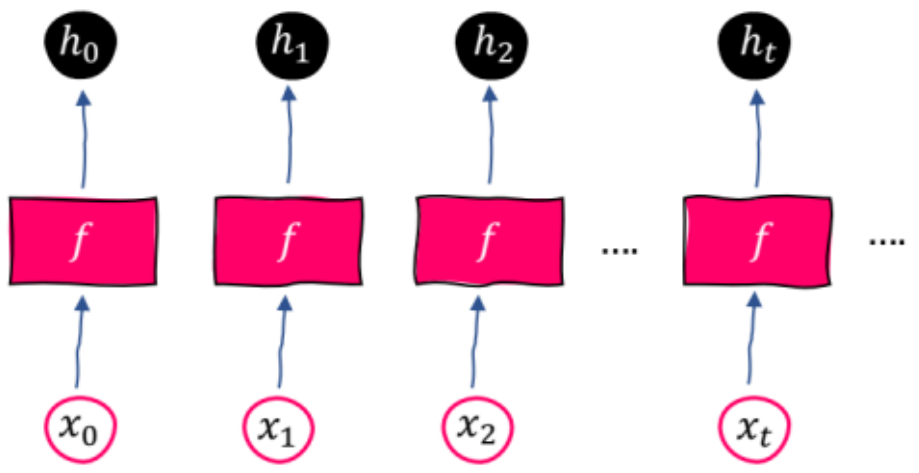

تعدادی ورودی $x_t$ داریم، که $t$ نشان دهنده یک گام زمانی یا یک ترتیب متوالی است. بیایید فرض کنیم ورودی‌هایِ $t$ مختلف، مستقل از یکدیگر باشند. خروجی شبکه در هر $t$ را می‌توان به صورت $h_t = f(x_t)$ نوشت.

در RNNها، حلقه بازخوردی، اطلاعاتِ وضعیتِ فعلی را به حالت بعدی منتقل می‌کند، همان‌طور که در نسخه بازشده شبکه، در شکل زیر نشان داده شده است:

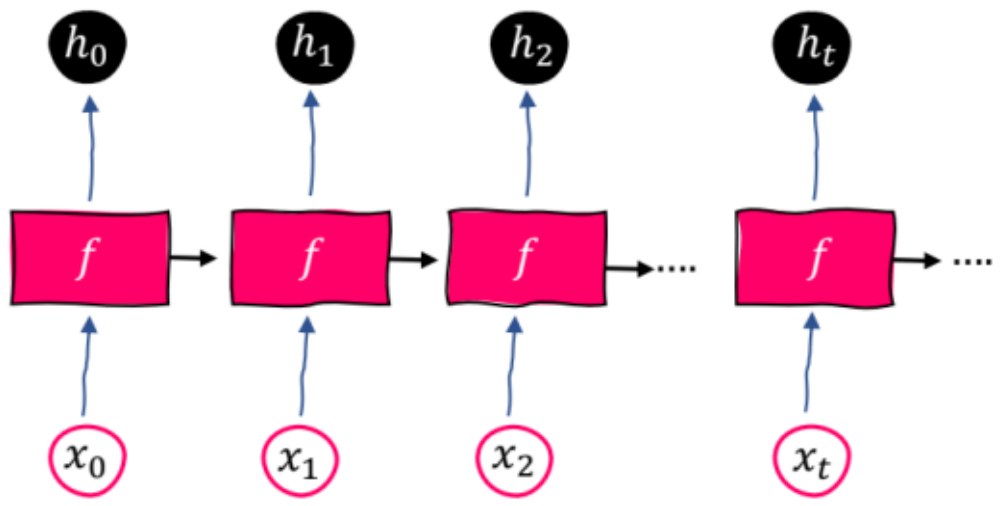

خروجی شبکه RNN در هر $t$ را می‌توان به صورت $h_t = f(h_{t-1}, x_t)$ نوشت. کار مشابه $f$ روی هر عنصر دنباله انجام می‌شود و خروجیِ $h_t$ وابسته به خروجیِ محاسبات قبلی است. از این‌رو، برخلاف شبکه‌های معمولی، که حالت فقط به ورودی جریان (و وزن شبکه) بستگی دارد، در اینجا $h_t$ تابعی از ورودی فعلی و همچنین وضعیت قبلی $h_{t-1}$ است. شما می توانید $h_{t-1}$ را به عنوان خلاصه‌ای از تمام ورودی‌های قبلی شبکه در نظر بگیرید.

به لطف این معماری زنجیره‌وار یا به عبارت دیگر، ذخیره‌ای اضافی (حافظه) از آنچه تاکنون محاسبه شده است، موفقیت زیادی در بکارگیری RNN در داده‌های سری زمانی و متوالی حاصل شده است.

RNNها نام خود را به این دلیل می‌گیرند چراکه تابع مشابهی را به‌طور مکرر روی یک دنباله اعمال می‌کنند.

RNN دارای سه مجموعه پارامتر (وزن) است:

- U ورودی $x_t$ را به حالت $h_t$ تبدیل می‌کند.
- W حالت قبلی $h_{t-1}$ را به حالت فعلی $h_t$ تبدیل می‌کند.
- V وضعیت داخلی تازه محاسبه شده $h_t$ را به خروجی $y$ نگاشت می‌کند.

U، V و W تبدیل خطی را روی ورودی‌هایِ مربوط اعمال می‌کنند. اساسی‌ترین مورد چنین تبدیلی، مجموع وزنی است که ما می‌شناسیم. اکنون می‌توانیم وضعیت داخلی و خروجی شبکه را به صورت زیر تعریف کنیم:

$$h_t = f(h_{t-1} * W + x_t * U)$$

$$y_t = h_t * V$$

در اینجا، f تابع فعال سازی غیرخطی است.

توجه داشته باشید که در یک RNN، هر حالت به تمام محاسبات قبلی از طریق این رابطه تکراری وابسته است. یک مفهوم مهم این است که RNNها در طول زمان دارای حافظه هستند، زیرا حالت‌های $h$ حاوی اطلاعاتی بر اساس مراحل قبلی هستند. در تئوری، RNNها می‌توانند اطلاعات را برای مدت زمان دلخواه خود به خاطر بسپارند، اما در عمل فقط می‌توانند به چند مرحله به عقب نگاه کنند.

**در RNN، هر حالت به تمام محاسبات قبلی توسط معادله بازگشتی وابسته است. پیامد مهم این امر سبب ایجاد حافظه در طول زمان می‌شود، چرا که حالت‌ها مبتنی بر مراحل قبلی هستند.**

## انواع معماری‌های RNN

از آنجا که RNNها به پردازش ورودی اندازه‌یِ ثابت، محدود نمی‌شوند، معماری‌های متفاوتی را شامل می‌شوند:

- **یک به یک:** همان‌طور که در شکل مقابل، قابل مشاهده است، در این معماری یک واحد ورودیِ RNN به یک واحد پنهان و یک واحد خروجی نگاشت می‌شود. این معماری یک پردازش بدون متوالی همانند، شبکه‌های عصبی پیش‌خور و شبکه‌ی عصبی کانولوشنی می‌باشد. نمونه‌ای از این پردازش طبقه‌بندی تصاویر می‌باشد.

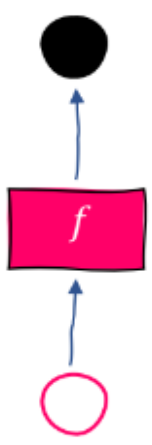

- **یک به چند:** همان‌طور که در شکل زیر، قابل مشاهده است، در این معماری یک واحد ورودی RNN به چند واحد پنهان و چند واحد خروجی نگاشت شده است. نمونه کاربردی

از این معماری شرح‌نویسی تصاویر می‌باشد. لایه ورودی یک تصویر را دریافت کرده و آن را به چندین کلمه نگاشت می‌کند.

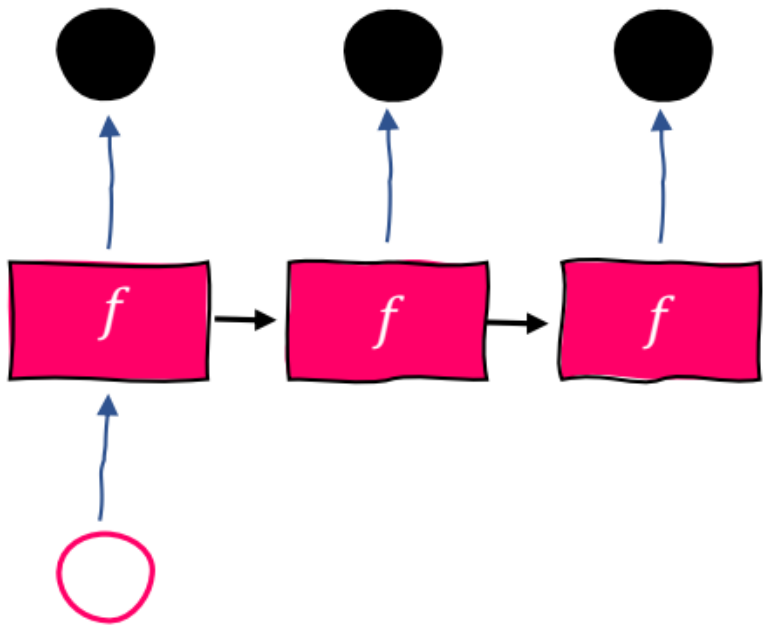

- **چند به یک:** همان‌طور که در شکل مقابل، قابل مشاهده است، در این معماری چند واحد ورودی RNN به چند واحد پنهان و یک واحد خروجی نگاشت شده است. یک نمونه کاربردی از این معماری طبقه‌بندی احساسات می‌باشد. لایه ورودی چندین نشانه از کلمات یک جمله را دریافت می‌کند و به صورت یک احساس مثبت یا منفی نگاشت می‌کند.

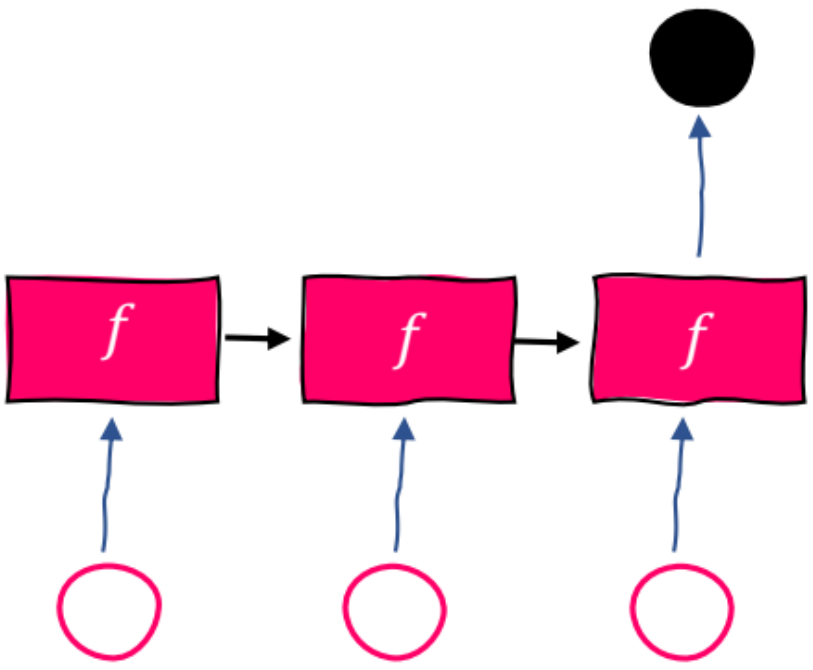

- **چند به چند:** همان‌طور که در شکل مقابل، قابل مشاهده است، در این معماری چند واحد ورودی RNN به چند واحد پنهان و چند واحد خروجی نگاشت شده است. نمونه کاربردی از این معماری ترجمه ماشینی است. لایه ورودی چندین نشانه از کلمات زبان مبدا را دریافت کرده و آن‌ها را به نشانه‌هایی از کلمات به زبان هدف نگاشت می‌کند.

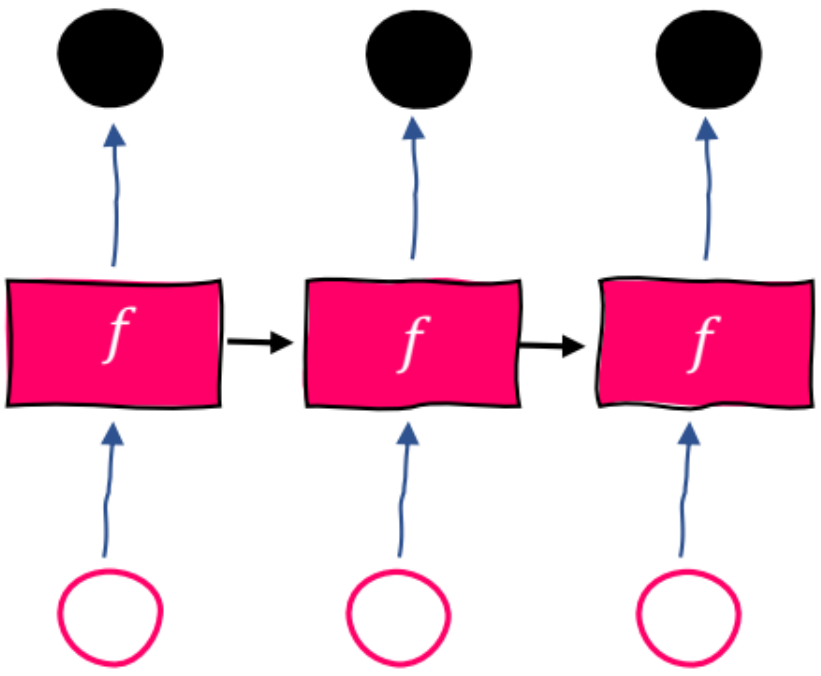

## تولید متن با RNN

RNNها معمولا به عنوان مدل‌های زبان (language models) در حوزه پردازش زبان طبیعی (natural language processing) استفاده می‌شوند. ما قصد داریم روی یک زبان نسبتا جالب

برای مدل‌سازیِ تولید متن کار کنیم، جایی که مدل‌های RNN برای یادگیری دنباله‌های متنی یک دامنه مشخص و سپس، تولیدِ توالیِ متنِ کاملا جدید و معقول در دامنه مورد نظر استفاده می‌شوند. مولدِ متنِ مبتنی بر RNN می‌تواند هر متن ورودی، مانند رمان‌هایی مانند هری‌پاتر، شعرهایی از شکسپیر و فیلم‌نامه‌هایِ فیلم‌هایی همانند جنگ ستارگان را بگیرد، و اشعار شکسپیر و فیلمنامه‌های فیلم جنگ ستارگان را تولید کند. اگر مدل به خوبی آموزش داده شده باشد، متن مصنوعی باید قابل قبول باشد و شبیه به متن اصلی خوانده شود. در این بخش از رمان "جنگ و صلح" از نویسنده روسی لئوتولستوی به عنوان مثال استفاده می‌کنیم. با این حال، می‌توانید از هر یک از کتاب‌های مورد علاقه خود برای ورودی‌های آموزشی استفاده کنید. پروژه گوتنبرگ (www.gutenberg.org) یک منبع عالی برای این کار است، با بیش از ۵۷۰۰۰ کتاب عالی رایگان که حق چاپ آن‌ها منقضی شده است.

ابتدا باید فایل txt جنگ و صلح را مستقیما از پیوند:

https://cs.stanford.edu/people/karpathy/char-rnn/warpeace_input.txt

دانلود می‌کنیم. از طرف دیگر، می‌توانیم آن را از پروژه گوتنبرگ

https://www.gutenberg.org/ebooks/2600

دانلود کنیم، اما باید برخی از پاک‌سازی‌ها را انجام دهیم. سپس فایل را می‌خوانیم، متن را به حروف کوچک تبدیل می‌کنیم و با چاپ ۱۰۰ کاراکتر اول، نگاهی گذرا به آن می‌اندازیم:

```
training_file = 'warpeace_input.txt'
raw_text = open(training_file, 'r').read()
raw_text = raw_text.lower()
raw_text[:100]
```

```
'ufeff"well, prince, so genoa and lucca are now just family estates
of thenbuonapartes. but i warn you, i'
```

اکنون باید تعداد کاراکترها را بشماریم:

```
n_chars = len(raw_text)
print('Total characters: {}'.format(n_chars))
```

```
Total characters: 3196213
```

سپس، می‌توانیم کاراکترهای یکتا و اندازه واژگان را بدست آوریم:

```
chars = sorted(list(set(raw_text)))
n_vocab = len(chars)
print('Total vocabulary (unique characters): {}'.format(n_vocab))
print(chars)
```

```
Total vocabulary (unique characters): 57
['\n', ' ', '!', '"', "'", '(', ')', '*', ',', '-', '.', '/', '0',
'1', '2', '3', '4', '5', '6', '7', '8', '9', ':', ';', '=', '?',
'a', 'b', 'c', 'd', 'e', 'f', 'g', 'h', 'i', 'j', 'k', 'l', 'm',
```

```
'n', 'o', 'p', 'q', 'r', 's', 't', 'u', 'v', 'w', 'x', 'y', 'z',
'à', 'ä', 'é', 'ê', '\ufeff']
```

اکنون، ما یک مجموعه داده آموزشی خام داریم که از بیش از ۳ میلیون کاراکتر و ۵۷ کاراکتر یکتا تشکیل شده است. اما چگونه می‌توانیم آن را به مدل RNN تغذیه کنیم؟ در معماری چند به چند، مدل توالی‌ها را می‌گیرد و توالی‌ها را همزمان تولید می‌کند. در مورد ما، می‌توانیم مدل را با دنباله‌های کاراکترهای با طول ثابت تغذیه کنیم. طول دنباله‌های خروجی با توالی‌های ورودی برابر است و یک کاراکتر از دنباله‌های ورودی آن‌ها جابه‌جا می‌شود. فرض کنید طول دنباله را برای کلمه learning روی ۵ قرار می‌دهیم. اکنون می‌توانیم با ورودی learn و خروجی earni یک نمونه آموزشی بسازیم. ما می‌توانیم این عمل را در شبکه به صورت زیر تجسم کنیم:

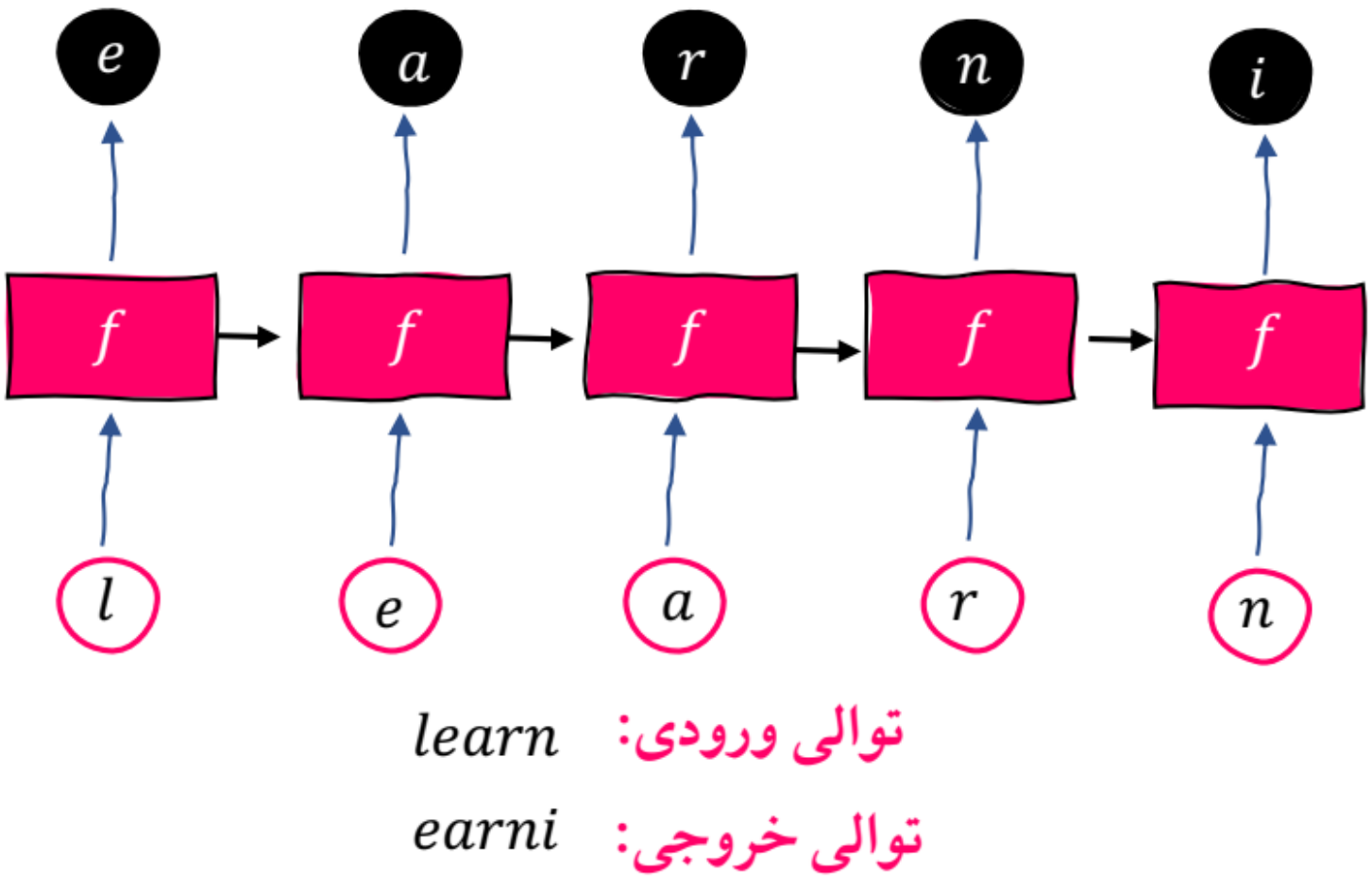

ما فقط یک نمونه آموزشی ساختیم. در مورد کل مجموعه آموزشی، می‌توانیم داده‌های متن خام را به دنباله‌هایی با طول مساوی، مثلا ۱۰۰ تقسیم کنیم. هر دنباله از کاراکترها ورودی یک نمونه آموزشی است .سپس، ما داده‌های متنی خام را به همان روش تجزیه و تحلیل می‌کنیم، اما این بار از کاراکتر دوم شروع می‌کنیم. هر یک از توالی‌های بدست آمدهِ خروجی، یک نمونه آموزشی است. برای مثال، با توجه به متن خام deep learning architectures و عدد ۵ به عنوان طول توالی، می‌توانیم پنج نمونه آموزشی را به صورت زیر بسازیم:

| ورودی | خروجی |
|---|---|
| deep_ | eep_l |
| learn | earni |
| ing_a | ng_ar |
| rchit | chite |
| ectur | cture |

در اینجا، _ نشانگرِ فضای خالی است.

از آنجایی که مدل‌های شبکه عصبی فقط داده‌های عددی را دریافت می‌کنند، دنباله‌های ورودی و خروجی کاراکترها با بردارهایِ کدگذاریِ one-hot نشان داده می‌شوند. ما با نگاشت ۵۷ کاراکتر به شاخص های ۰ تا ۵٦، یک دیکشنری ایجاد می‌کنیم:

```
index_to_char = dict((i, c) for i, c in enumerate(chars))
char_to_index = dict((c, i) for i, c in enumerate(chars))
print(char_to_index)
```

به عنوان مثال، کاراکتر *e* به برداری با طول ۵۷ تبدیل می‌شود که در شاخص ۳۰ بردارش مقدار ۱ قرار دارد و همه شاخص‌های دیگر آن ۰ وجود دارد (نحوه این کدگذاری را در فصل پیشین مشاهده کردید). با آماده شدن جدول جستجوی کاراکتر، می‌توانیم مجموعه داده آموزشی را به صورت زیر بسازیم:

```
import numpy as np
seq_length = 100
n_seq = int(n_chars / seq_length)
```

با تنظیم طول دنباله روی ۱۰۰، ما n_seq نمونه‌یِ آموزشی خواهیم داشت. سپس، ورودی و خروجی آموزشی را مقداردهی اولیه می‌کنیم:

```
X = np.zeros((n_seq, seq_length, n_vocab))
Y = np.zeros((n_seq, seq_length, n_vocab))
```

توجه داشته باشید که طول دنباله از نظر شکل اینگونه است:

**(تعداد نمونه، طول دنباله، ابعاد ویژگی)**

چنین فرمی مورد نیاز است، چراکه ما قصد داریم از Keras برای آموزش مدل RNN استفاده کنیم. سپس، هر یک از نمونه‌های n_seq را ایجاد می‌کنید:

```
for i in range(n_seq):
        x_sequence = raw_text[i * seq_length : (i + 1) * seq_length]
        x_sequence_ohe = np.zeros((seq_length, n_vocab))
        for j in range(seq_length):
                char = x_sequence[j]
                index = char_to_index[char]
                x_sequence_ohe[j][index] = 1.
        X[i] = x_sequence_ohe
        y_sequence = raw_text[i * seq_length + 1 : (i + 1) * seq_length + 1]
        y_sequence_ohe = np.zeros((seq_length, n_vocab))
        for j in range(seq_length):
                char = y_sequence[j]
                index = char_to_index[char]
                y_sequence_ohe[j][index] = 1.
        Y[i] = y_sequence_ohe
```

در صورت تمایل، می‌توانید با اجرای سلول‌های زیر شکل آرایه را ببینید:

```
X.shape
```

```
(31962, 100, 57)
```

```
Y.shape
```

```
(31962, 100, 57)
```

تا این‌جا، مجموعه داده آموزشی را آماده کردیم و اکنون زمان ساخت مدل RNN است. ابتدا تمام ماژول‌های لازم را وارد کنید:

```
from keras.models import Sequential
from keras.layers.core import Dense, Activation, Dropout
from keras.layers.recurrent import SimpleRNN
from keras.layers.wrappers import TimeDistributed
from keras import optimizers
from tensorflow import keras
```

اکنون، ابرپارامترها، از جمله اندازه دسته، تعداد نورون‌ها، تعداد لایه‌ها، احتمال حذف تصادفی و تعداد دوره‌ها را مشخص می‌کنیم:

```
batch_size = 100
n_layer = 2
hidden_units = 800
n_epoch = 300
dropout = 0.3
```

سپس، شبکه را ایجاد می‌کنیم:

```
model = Sequential()
model.add(SimpleRNN(hidden_units, activation='relu', input_shape=(None,
n_vocab), return_sequences=True))
model.add(Dropout(dropout))
for i in range(n_layer - 1):
    model.add(SimpleRNN(hidden_units, activation='relu',
return_sequences=True))
    model.add(Dropout(dropout))
model.add(TimeDistributed(Dense(n_vocab)))
model.add(Activation('softmax'))
```

در مورد مدلی که ساخته‌ایم، باید به چند نکته توجه داشت:

- `return_sequences=True`: خروجی لایه‌های بازگشتی، به یک دنباله تبدیل می‌شود و معماری چند به چند را ممکن می‌کند، همان‌طور که می‌خواستیم. در غیر این صورت، تبدیل به چند به یک می‌شود و آخرین عنصر به عنوان خروجی خواهد بود.
- `TimeDistributed`: از آنجایی که خروجی لایه‌های بازگشتی یک دنباله است، در حالی که لایه بعدی یک لایه متصل کامل است و ورودی متوالی نمی‌گیرد. از `TimeDistributed` برای دور زدن این مورد استفاده می‌شود.

برای بهینه‌ساز، ما RMSprop را با نرخ یادگیری ۰/۰۰۱ انتخاب می‌کنیم:

```
optimizer = keras.optimizers.RMSprop(learning_rate=0.001, rho=0.9,
epsilon=1e-08, decay=0.0)
```

با اضافه کردن زیان آنتروپی متقاطع چندکلاسه، ساخت مدل خود را به پایان رساندیم و در نهایت مدل را کامپایل می‌کنیم:

```
model.compile(loss= "categorical_crossentropy", optimizer=optimizer)
```

با استفاده از کد زیر می‌توانیم نگاهی به خلاصه‌ای از مدل بیندازیم:

```
model.summary()
```

```
_________________________________________________________________
 Layer (type)                Output Shape              Param #
=================================================================
 simple_rnn_4 (SimpleRNN)    (None, None, 800)         686400

 dropout_4 (Dropout)         (None, None, 800)         0

 simple_rnn_5 (SimpleRNN)    (None, None, 800)         1280800

 dropout_5 (Dropout)         (None, None, 800)         0

 time_distributed_2 (TimeDis  (None, None, 57)         45657
 tributed)

 activation_2 (Activation)   (None, None, 57)          0

=================================================================
Total params: 2,012,857
Trainable params: 2,012,857
Non-trainable params: 0
```

ما بیش از ۲ میلیون پارامتر برای آموزش داریم. اما قبل از شروع روند آموزشِ طولانی، تمرین خوبی است که برخی از callbackها را برای پیگیری آمار و وضعیت‌های داخلی مدل در طول آموزش تنظیم کنیم. توابع callback شامل موارد زیر است:

- **نقطه وارسیِ (checkpoint) مدل**، که مدل را بعد از هر دوره ذخیره می‌کند تا بتوانیم آخرین مدل ذخیره شده را بارگیری کنیم و در صورت توقف غیرمنتظره، آموزش را از آنجا از سر بگیریم.
- **توقف زودهنگام**، که زمانی که تابع زیان، دیگر بهبود نمی‌یابد، آموزش را متوقف می‌کند.
- **بررسی نتایج تولیدِ متن به‌طور منظم**. ما می‌خواهیم ببینیم متن تولید شده چقدر معقول است چرا که زیان آموزش به اندازه کافی ملموس نیست.

این توابع به صورت زیر تعریف یا مقداردهی اولیه می‌شوند:

```
from keras.callbacks import Callback, ModelCheckpoint, EarlyStopping

filepath="weights/weights_layer_%d_hidden_%d_epoch_{epoch:03d}_loss_{loss
:.4f}.hdf5"

checkpoint = ModelCheckpoint(filepath, monitor='loss', verbose=1,
save_best_only=True, mode='min')
```

نقاط وارسی مدل، با شماره دوره و زیانِ آموزشی در نام فایل ذخیره می‌شود. ما همچنین زیان اعتبارسنجی را به‌طور همزمان نظارت می‌کنیم تا ببینیم که آیا کاهش آن برای ۵۰ دوره متوالی متوقف می‌شود یا خیر:

```
early_stop = EarlyStopping(monitor='loss', min_delta=0, patience=50,
verbose=1, mode='min')
```

در مرحله بعد، ما یک callback برای نظارت‌بر کیفیت لازم داریم. ابتدا یک تابع کمکی می‌نویسیم که متنی با هر طولی را با توجه به مدل RNN ما تولید می‌کند:

```
def generate_text(model, gen_length, n_vocab, index_to_char):
    """
    Generating text using the RNN model
    @param model: current RNN model
    @param gen_length: number of characters we want to generate
    @param n_vocab: number of unique characters
    @param index_to_char: index to character mapping
    @return:
    """
    # Start with a randomly picked character
    index = np.random.randint(n_vocab)
    y_char = [index_to_char[index]]
    X = np.zeros((1, gen_length, n_vocab))
    for i in range(gen_length):
        X[0, i, index] = 1.
        indices = np.argmax(model.predict(X[:, max(0, i - 99):i + 1, :])[0], 1)
        index = indices[-1]
        y_char.append(index_to_char[index])
    return ('').join(y_char)
```

این تابع، با کارکتری که به طور تصادفی انتخاب شده شروع می‌کند. سپس مدل ورودی هر یک از کارکترهای باقی‌مانده gen_length-1 را بر اساس کارکترهای تولید شده قبلی که طول آنها تا ۱۰۰ است (طول دنباله) پیش‌بینی می‌کند. اکنون می‌توانیم کلاس callback را تعریف کنیم که متن را برای هر N دوره تولید می‌کند:

```
class ResultChecker(Callback):
    def __init__(self, model, N, gen_length):
        self.model = model
        self.N = N
        self.gen_length = gen_length

    def on_epoch_end(self, epoch, logs={}):
        if epoch % self.N == 0:
            result = generate_text(self.model, self.gen_length, n_vocab,
index_to_char)
            print('\nMy War and Peace:\n' + result)
```

اکنون که همه اجزا آماده هستند، آموزش مدل را شروع می‌کنیم:

```
model.fit(X, Y, batch_size=batch_size, verbose=1, epochs=n_epoch,
                 callbacks=[ResultChecker(model, 10, 200), checkpoint, early_stop])
```

مولد برای هر ۱۰ دوره ۲۰۰ کاراکتر می‌نویسد. نتایج زیر برای دوره‌های ۱، ۱۱، ۵۱ و ۱۰۱ می‌باشد:

### Epoch 1:

```
Epoch 1/300
8000/31962 [======>.......................] - ETA: 51s - loss: 2.8891
31962/31962 [==============================] - 67s 2ms/step - loss: 2.1955
My War and Peace:
5 the count of the stord and the stord and the stord and the stord and the
stord and the stord and the stord and the stord and the stord and the stord
and the and the stord and the stord and the stord
```

```
Epoch 00001: loss improved from inf to 2.19552, saving model to
weights/weights_epoch_001_loss_2.19552.hdf5
```

Epoch 11:

```
Epoch 11/300
100/31962 [..............................] - ETA: 1:26 - loss: 1.2321
31962/31962 [==============================] - 66s 2ms/step - loss: 1.2493
My War and Peace:
?" said the countess was a strange the same time the countess was already
been and said that he was so strange to the countess was already been and
the same time the countess was already been and said
Epoch 00011: loss improved from 1.26144 to 1.24933, saving model to
weights/weights_epoch_011_loss_1.2493.hdf5
```

Epoch 51:

```
Epoch 51/300
31962/31962 [==============================] - 66s 2ms/step - loss: 1.1562
My War and Peace:
!!CDP!E.agrave!! to see him and the same thing is the same thing to him and
the same thing the same thing is the same thing to him and the sam thing
the same thing is the same thing to him and the same thing the sam
Epoch 00051: loss did not improve from 1.14279
```

Epoch 101:

```
Epoch 101/300
31962/31962 [==============================] - 67s 2ms/step - loss: 1.1736
My War and Peace:
= the same thing is to be a soldier in the same way to the soldiers and the
same thing is the same to me to see him and the same thing is the same to
me to see him and the same thing is the same to me
Epoch 00101: loss did not improve from 1.11891
```

آموزش در اوایل دوره ۲۰۳ متوقف می‌شود:

Epoch 00203: loss did not improve from 1.10864
Epoch 00203: early stopping

مدل، متن زیر را در دوره ۱۵۱ ایجاد می‌کند:

```
which was a strange and serious expression of his face and shouting and said
that the countess was standing beside him. "what a battle is a strange and
serious and strange and so that the countess was
```

جنگ و صلح ما تا حدودی خوب خوانده می‌شود. با این حال، شاید برای شما سوال پیش آید که آیا می‌توانیم با تغییر پارامترها در این مدل RNN بهتر عمل کنیم؟ پاسخ آری است، اما ارزشش را ندارد. چراکه آموزش یک مدل RNN برای حل مسائلی که نیازمند یادگیری وابستگی‌های بلندمدت هستند، کارایی خیلی عالی ندارد. معماری‌هایی مانند LSTM و GRU به‌طور خاص برای حل این مشکل طراحی شده‌اند.

## LSTM

در تئوری، RNNهای ساده قادر به یادگیریِ وابستگی‌هایِ بلند مدت هستند، اما در عمل، به دلیل مشکل محو گرادیان، آن‌ها خود را محدود به یادگیریِ وابستگیِ کوتاه مدت می‌کنند. در راستای مقابله با این محدودیت، شبکه **حافظه طولانی کوتاه مدت (Long short term memory)** یا به اختصار **LSTM** ارائه شد. LSTM می‌تواند وابستگی‌های بلندمدت را به دلیلِ وجودِ یک سلولِ حافظه‌یِ مخصوص در ساختارش، انجام دهد.

ایده کلیدی LSTM **سلول وضعیت (cell state)** است که اطلاعات می‌تواند به صراحت در آن نوشته یا حذف شود. این سلول وضعیت، برای زمان $t$ با عنوان $c_t$ در شکل **۵-۱** نشان داده شده است.

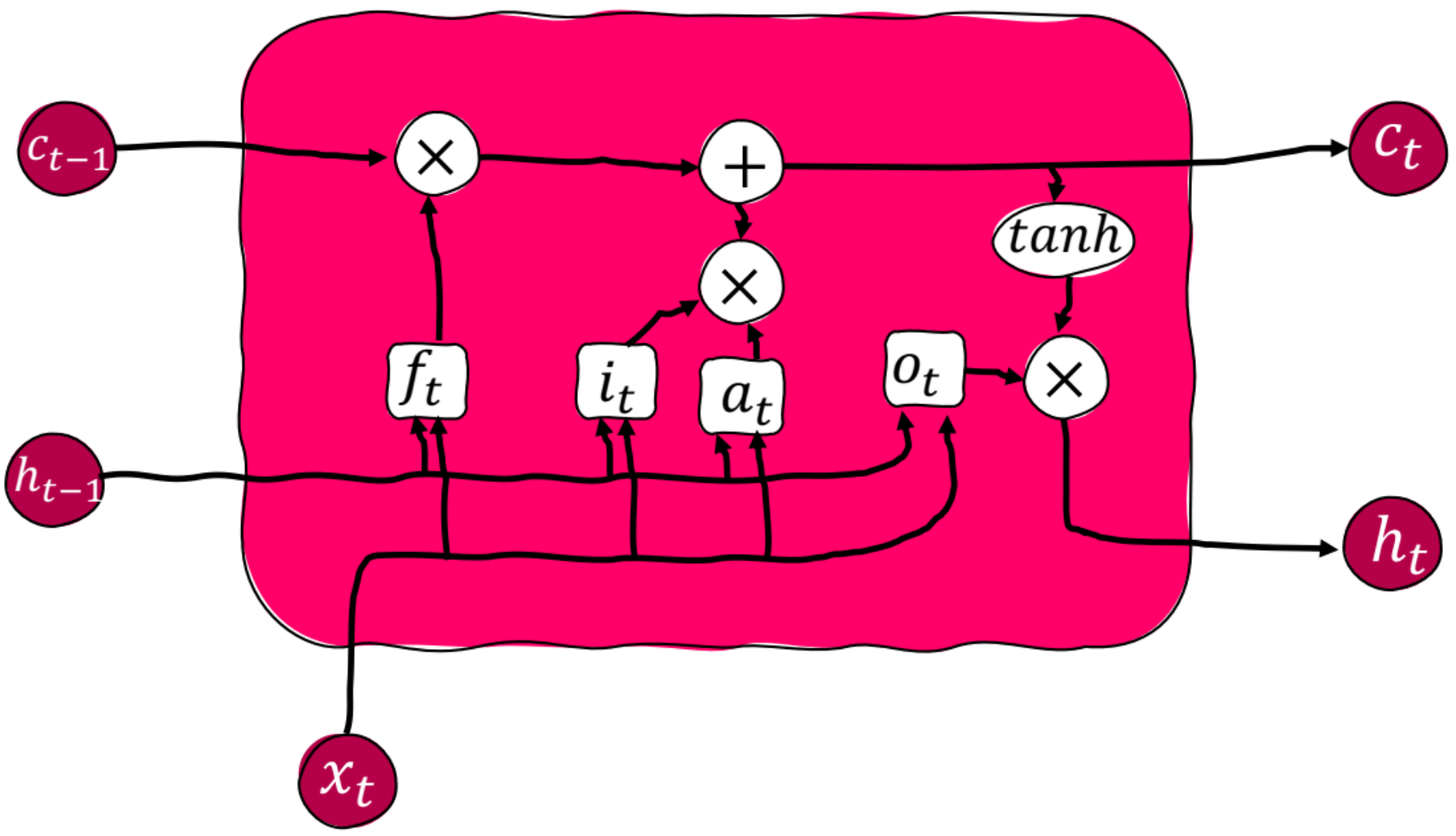

$$f_t = \sigma(W_f h_{t-1} + U_f x_t + b_f)$$
$$i_t = \sigma(W_i h_{t-1} + U_i x_t + b_i)$$
$$a_t = \tanh(W_c h_{t-1} + U_c x_t + b_f)$$
$$o_t = \sigma(W_o h_{t-1} + U_o x_t + b_o)$$
$$c_t = f_t * c_{t-1} + i_t * a_t$$
$$h_t = o_t * \tanh(c_t)$$

**شکل ۵-۱. ساختار یک LSTM**

سلول وضعیت LSTM تنها می‌تواند توسط دروازه‌های خاص تغییر کند که راهی برای انتقال اطلاعات از طریق آن است. یک LSTM معمولی از سه دروازه تشکیل شده است: **دروازه فراموش** ($f$)، **دروازه ورودی** ($i$) و **دروازه خروجی** ($o$).

دروازه اول در LSTM دروازه فراموشی است. این دروازه تصمیم می‌گیرد که آیا ما می‌خواهیم سلول وضعیت را پاک کنیم یا خیر. تصمیمِ دروازه فراموش براساس خروجی قبلی $h_{t-1}$ و ورودی فعلی $x_t$ است. از یک تابع sigmoid برای ایجاد خروجی با مقدار بین صفر و یک برای هر یک از عناصر در سلول وضعیت استفاده می‌شود. یک ضرب درایه‌ای بین خروجیِ دروازه‌یِ فراموشی و سلول وضعیت انجام می‌شود. مقدارِ یک در خروجی دروازه‌یِ فراموشی، به معنایِ حفظ کاملِ اطلاعات عنصر در سلول وضعیت است. در مقابل، صفر به معنای فراموش کردن کامل اطلاعات در عنصر سلول حالت است. این به این معنا است که LSTM می‌تواند اطلاعات بی‌ربط را از بردار سلول وضعیت خود دور بی‌اندازد. معادله‌ی دروازه‌یِ فراموشی به‌صورت زیر است:

$$f_t = \sigma(W_f h_{t-1} + U_f x_t + b_f)$$

دروازه بعدی تصمیم می‌گیرد که اطلاعات جدید به سلول حافظه اضافه شود. این کار در دو بخش انجام می‌شود: تصمیم بگیرد که کدام مقادیر را بروزرسانی کند و سپس ایجاد مقادیر برای بروزرسانی است. ابتدا از بردار $i_t$ برای انتخاب مقادیری از نامزدهای جدید بالقوه برای گنجاندن در سلول وضعیت استفاده می‌شود. بردار نامزد $a_t$ نیز ماتریس وزن مخصوص به خود را دارد و از حالت پنهان قبلی و ورودی‌ها برای تشکیل برداری با ابعاد مشابه سلول وضعیت استفاده می‌کند. برای ایجاد این بردار نامزد از یک تابع tanh به عنوان یک تابع غیرخطی استفاده می‌شود. این فرآیند در معادلات زیر نشان داده شده است:

$$i_t = \sigma(W_i h_{t-1} + U_i x_t + b_i)$$
$$a_t = \tanh(W_c h_{t-1} + U_c x_t + b_f)$$

دروازه‌ی فراموشی و ورودی نحوه‌یِ بروزرسانی سلول وضعیت را در هر مرحله زمانی مشخص می‌کنند. بروزرسانی سلول وضعیت در یک مرحله زمانی از طریق معادله زیر انجام می‌شود:

$$c_t = f_t * c_{t-1} + i_t * a_t$$

آخرین دروازه تصمیم می‌گیرد که خروجی چه باشد. خروجی نهایی یک سلول LSTM، حالت پنهان $h_t$ است. دروازه‌یِ خروجی $h_{t-1}$ و $x_t$ را به عنوان ورودی می‌گیرد. ابتدا، از یک تابع sigmoid برای محاسبه بردار با مقادیر بین صفر و یک استفاده می‌شود تا انتخابِ مقادیرِ سلولِ وضعیت در مرحله زمانی را انجام دهد. سپس مقدار سلول وضعیت را به یک لایه tanh می‌دهیم تا در نهایت مقدار آن را در خروجی لایه قبلی sigmoid ضرب کرده، تا قسمت‌های مورد نظر در خروجی به اشتراک گذاشته شوند. خروجی ۰ به این معنی است که بلوک سلولی هیچ اطلاعاتی را تولید نمی‌کند، در حالی که خروجی ۱ به این معنا است که حافظه کامل بلوک سلول به خروجی سلول منتقل می‌شود. معادلات زیر این روند را نشان می‌دهند:

$$o_t = \sigma(W_o h_{t-1} + U_o x_t + b_o)$$

$$h_t = o_t * \tanh(c_t)$$

حال چگونه LSTM از ما در برابر محو گرادیان محافظت می‌کند؟ توجه داشته باشید که اگر دروازه فراموش ۱ باشد و دروازه ورودی ۰ باشد، وضعیت سلول عینا مرحله به مرحله رونوشت می‌شود. تنها دروازه فراموشی می‌تواند به طور کامل حافظه سلول را پاک کند. در نتیجه، حافظه می‌تواند در مدت‌زمان طولانی بدون تغییر باقی بماند.

نحوه‌یِ بازشدن LSTM در طول زمان، در عمل در شکل زیر نشان‌داده شده‌است:

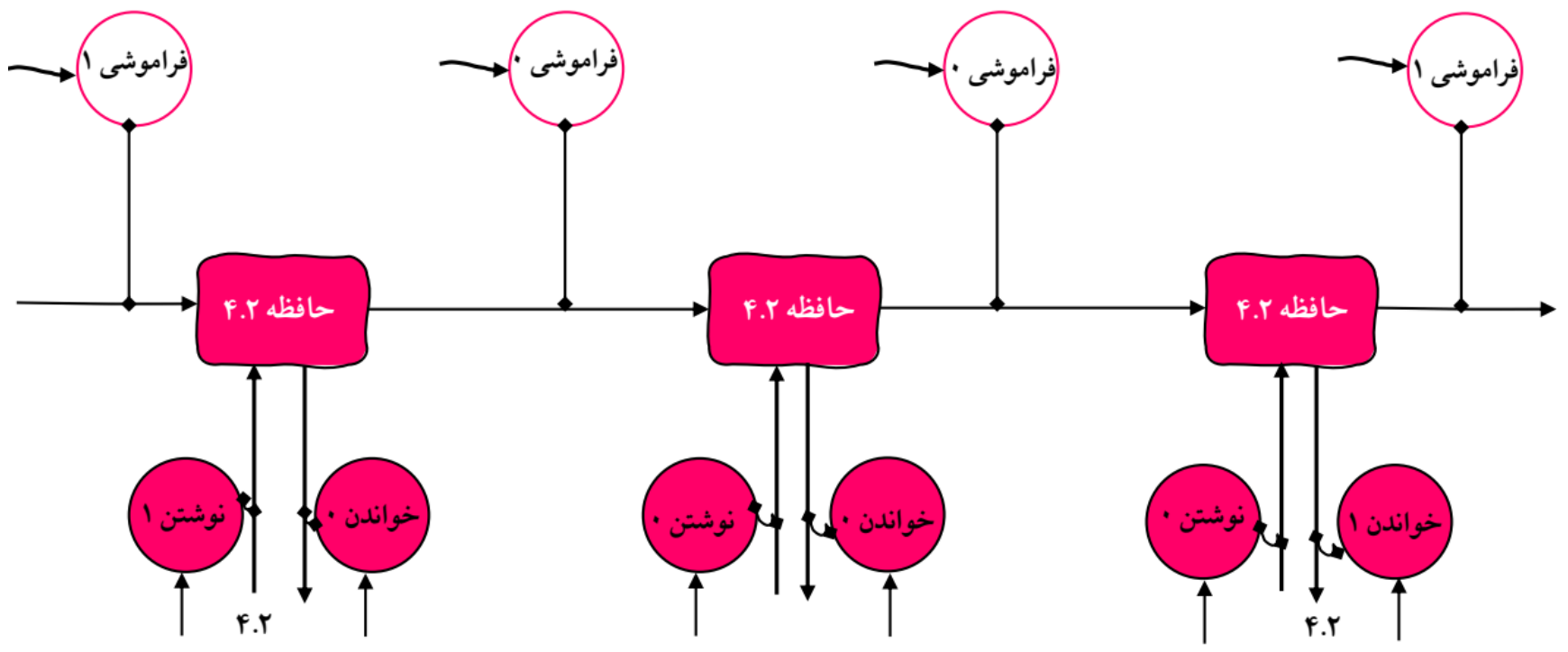

در ابتدا مقدار ۴.۲ به شبکه به عنوان ورودی داده می‌شود؛ دروازه ورودی به ۱ تنظیم شده است، بنابراین مقدار کامل ذخیره می‌شود. سپس برای دو مرحله بعد، دروازه فراموش شده به ۱ تنظیم شده است. بنابراین کل اطلاعات در طول این مراحل نگهداری می‌شود و هیچ اطلاعات جدیدی اضافه نمی‌شود، چراکه دروازه‌های ورودی به ۰ تنظیم شده‌اند. در نهایت، دروازه خروجی به ۱ تنظیم می‌شود و ۴.۲ تولید می‌شود و بدون تغییر باقی می‌ماند.

## تولید متن با LSTM

در مولدِ متنِ مبتنی‌بر LSTM، در مقایسه با مثال پیشین، طول دنباله را به ۱٦٠ کاراکتر افزایش می‌دهیم. از این‌رو مجموعه آموزشی X و Y را با مقدار جدید SEQ_LENGTH = 160 بازسازی می‌کنیم:

```
seq_length = 160
n_seq = int(n_chars / seq_length)

X = np.zeros((n_seq, seq_length, n_vocab))
Y = np.zeros((n_seq, seq_length, n_vocab))

for i in range(n_seq):
        x_sequence = raw_text[i * seq_length : (i + 1) * seq_length]
        x_sequence_ohe = np.zeros((seq_length, n_vocab))
        for j in range(seq_length):
                char = x_sequence[j]
                index = char_to_index[char]
                x_sequence_ohe[j][index] = 1.
        X[i] = x_sequence_ohe
        y_sequence = raw_text[i * seq_length + 1 : (i + 1) * seq_length + 1]
```

```
        y_sequence_ohe = np.zeros((seq_length, n_vocab))
        for j in range(seq_length):
                char = y_sequence[j]
                index = char_to_index[char]
                y_sequence_ohe[j][index] = 1.
        Y[i] = y_sequence_ohe
```

در مقایسه با مدل RNN قبلی، ما از مدلی با دو لایه بازگشتی حاوی ۸۰۰ نورون و حذف تصادفی با احتمال ۰/۴ استفاده می‌کنیم:

```
from keras.layers.recurrent import LSTM
batch_size = 100
n_layer = 2
hidden_units = 800
n_epoch= 300
dropout = 0.4
```

اکنون شبکه را ایجاد و کامپایل می‌کنیم:

```
model = Sequential()
model.add(LSTM(hidden_units, input_shape=(None, n_vocab),
return_sequences=True))
model.add(Dropout(dropout))
for i in range(n_layer - 1):
    model.add(LSTM(hidden_units, return_sequences=True))
    model.add(Dropout(dropout))
model.add(TimeDistributed(Dense(n_vocab)))
model.add(Activation('softmax'))
```

برای بهینه‌ساز، از RMSprop، با نرخ یادگیری ۰/۰۰۱ استفاده می‌کنیم:

```
optimizer = keras.optimizers.RMSprop(learning_rate=0.001, rho=0.9,
epsilon=1e-08, decay=0.0)

model.compile(loss= "categorical_crossentropy", optimizer=optimizer)
```

بیایید مدل LSTM را که به تازگی ساخته‌ایم، خلاصه کنیم:

```
model.summary()
```

```
_________________________________________________________________
Layer (type) Output Shape Param #
=================================================================
lstm_1 (LSTM) (None, None, 800) 2745600
_________________________________________________________________
dropout_1 (Dropout) (None, None, 800) 0
_________________________________________________________________
lstm_2 (LSTM) (None, None, 800) 5123200
_________________________________________________________________
dropout_2 (Dropout) (None, None, 800) 0
_________________________________________________________________
time_distributed_1 (TimeDist (None, None, 57) 45657
_________________________________________________________________
activation_1 (Activation) (None, None, 57) 0
=================================================================
Total params: 7,914,457
Trainable params: 7,914,457
Non-trainable params: 0
```

حدود ۸ میلیون پارامتر برای آموزش وجود دارد که تقریبا چهار برابر بیشتر از مدل RNN قبلی.

بیایید آموزش را شروع کنیم:

```
model.fit(X, Y, batch_size=batch_size, verbose=1, epochs=n_epoch,
                callbacks=[ResultChecker(model, 10, 200), checkpoint, early_stop])
```

مولد برای هر ۱۰ دوره متنی با ۵۰۰ کاراکتر می‌نویسد. نتایج زیر برای دوره‌های ۱۵۱، ۲۰۱ و ۲۵۱ است:

## Epoch 151:

```
Epoch 151/300
19976/19976 [==============================] - 250s 12ms/step - loss:
0.7300
My War and Peace:
ing to the countess. "i have nothing to do with him and i have nothing to
do with the general," said prince andrew.
"i am so sorry for the princess, i am so since he will not be able to say
anything. i saw him long ago. i am so sincerely that i am not to blame for
it. i am sure that something is so much talk about the emperor alexander's
personal attention."
"why do you say that?" and she recognized in his son's presence.
"well, and how is she?" asked pierre.
"the prince is very good to make

Epoch 00151: loss improved from 0.73175 to 0.73003, saving model to
weights/weights_epoch_151_loss_0.7300.hdf5
```

## Epoch 201:

```
Epoch 201/300
19976/19976 [==============================] - 248s 12ms/step - loss:
0.6794
My War and Peace:
was all the same to him. he received a story proved that the count had not
yet seen the countess and the other and asked to be able to start a tender
man than the world. she was not a family affair and was at the same time as
in the same way. a few minutes later the count had been at home with his
smile and said:
"i am so glad! well, what does that mean? you will see that you are always
the same."
"you know i have not come to the conclusion that i should like to
send my private result. the prin

Epoch 00201: loss improved from 0.68000 to 0.67937, saving model to
weights/weights_epoch_151_loss_0.6793.hdf5
```

## Epoch 251:

```
Epoch 251/300
19976/19976 [==============================] - 249s 12ms/step - loss:
0.6369
My War and Peace:
nd the countess was sitting in a single look on
her face.
"why should you be ashamed?"
"why do you say that?" said princess mary. "why didn't you say a word of
this?" said prince andrew with a smile.
"you would not like that for my sake, prince vasili's son, have you seen
the rest of the two?"
"well, i am suffering," replied the princess with a sigh. "well, what a
```

```
delightful norse?" he shouted.
the convoy and driving away the flames of the battalions of the first
day of the orthodox russian

Epoch 00251: loss improved from 0.63715 to 0.63689, saving model to
weights/weights_epoch_251_loss_0.6368.hdf5
```

در نهایت، در دوره ۳۰۰، آموزش با زیان ۰/۶۰۰۱ متوقف می‌شود.

مولدِ متن به لطف معماری LSTM قادر است داستانِ جنگ و صلح واقعی‌تر و جالب‌تر بنویسد. علاوه بر این، RNNهای LSTM برای تولید کاراکتر به متن محدود نمی‌شوند. آنها می‌توانند از هر داده کاراکتری، مانند HTML، LaTex و غیره یاد بگیرند.

## طبقه‌بندی چندبرچسبی متن با LSTM

در این بخش قصد داریم تا نحوه‌یِ ایجاد یک مدلِ طبقه‌بندِ متن، با خروجی‌های متعدد را شرح دهیم. ما به توسعه یک مدل طبقه‌بند متن خواهیم پرداخت که یک نظر متنی را تحلیل کرده و چندین برچسب مرتبط با نظر را پیش‌بینی می‌کند.

مجموعه داده مورد استفاده برای آموزش، شش برچسب خروجی برای هر نظر دارد: toxic، severe_toxic، obscene، threat، insult و identity_hate. یک نظر می‌تواند به همه این دسته‌ها یا زیرمجموعه‌ای از این دسته‌ها تعلق داشته باشد که آن را به یک مسئله طبقه‌بندی چندبرچسبی تبدیل می‌کند. این مجموعه داده را می‌توانید از این وب‌سایت[1] Kaggle دانلود کنید. در این مثال تنها از فایل "train.csv" استفاده خواهیم کرد.

ابتدا کتابخانه‌های مورد نیاز را وارد کرده و مجموعه داده را بارگذاری می‌کنیم. کد زیر کتابخانه‌های مورد نیاز را وارد می‌کند:

```
from numpy import array
from keras.preprocessing.text import one_hot
from keras.preprocessing.sequence import pad_sequences
from keras.models import Sequential
from keras.layers.core import Activation, Dropout, Dense
from keras.layers import Flatten, LSTM
from keras.layers import GlobalMaxPooling1D
from keras.models import Model
from keras.layers.embeddings import Embedding
from sklearn.model_selection import train_test_split
from keras.preprocessing.text import Tokenizer
from keras.layers import Input
from keras.layers.merge import Concatenate
import pandas as pd
import numpy as np
import re
import matplotlib.pyplot as plt
```

حال مجموعه داده را بارگذاری می‌کنیم:

```
toxic_comments = pd.read_csv("train.csv")
```

[1] https://www.kaggle.com/c/jigsaw-toxic-comment-classification-challenge/overview

کد زیر شکل مجموعه داده را نمایش می‌دهد:

```
print(toxic_comments.shape)
```

```
(159571, 8)
```

همان‌طور که مشاهده می‌شود، مجموعه داده شامل ۱۵۹۵۷۱ رکورد و ۸ ستون است.

با دستور زیر چند نمونه از داده‌ها را خروجی مشاهده می‌کنید:

```
toxic_comments.head()
```

| | id | comment_text | toxic | severe_toxic | obscene | threat | insult | identity_hate |
|---|---|---|---|---|---|---|---|---|
| **0** | 0000997932d777bf | Explanation\nWhy the edits made under my usern... | 0 | 0 | 0 | 0 | 0 | 0 |
| **1** | 000103f0d9cfb60f | D'aww! He matches this background colour I'm s... | 0 | 0 | 0 | 0 | 0 | 0 |
| **2** | 000113f07ec002fd | Hey man, I'm really not trying to edit war. It... | 0 | 0 | 0 | 0 | 0 | 0 |
| **3** | 0001b41b1c6bb37e | "\nMore\nI can't make any real suggestions on ... | 0 | 0 | 0 | 0 | 0 | 0 |
| **4** | 0001d958c54c6e35 | You, sir, are my hero. Any chance you remember... | 0 | 0 | 0 | 0 | 0 | 0 |

در مرحله بعد، تمام رکوردهایی که در آن هر ردیف حاوی مقدار تهی یا رشته خالی است را حذف می‌کنیم.

```
filter = toxic_comments["comment_text"] != ""
toxic_comments = toxic_comments[filter]
toxic_comments = toxic_comments.dropna()
```

ستون comment_text حاوی نظرات متنی است. بیایید یک نظر را چاپ کنیم و سپس برچسب‌های نظرات را ببینیم:

```
print(toxic_comments["comment_text"][168])
```

```
You should be fired, you're a moronic wimp who is too lazy to do research.
It makes me sick that people like you exist in this world.
```

حال با دستور زیر، نگاهی به برچسب‌های مرتبط با این نظر می‌اندازیم:

```
print("Toxic:" + str(toxic_comments["toxic"][168]))
print("Severe_toxic:" + str(toxic_comments["severe_toxic"][168]))
print("Obscene:" + str(toxic_comments["obscene"][168]))
print("Threat:" + str(toxic_comments["threat"][168]))
print("Insult:" + str(toxic_comments["insult"][168]))
print("Identity_hate:" + str(toxic_comments["identity_hate"][168]))
```

```
Toxic:1
Severe_toxic:0
Obscene:0
Threat:0
Insult:1
Identity_hate:0
```

حال بیایید تعداد نظرات را برای هر برچسب مصورسازی کنیم:

```
toxic_comments_labels = toxic_comments[["toxic", "severe_toxic", "obscene",
"threat", "insult", "identity_hate"]]
fig_size = plt.rcParams["figure.figsize"]
fig_size[0] = 10
```

```
fig_size[1] = 8
plt.rcParams["figure.figsize"] = fig_size
toxic_comments_labels.sum(axis=0).plot.bar()
```

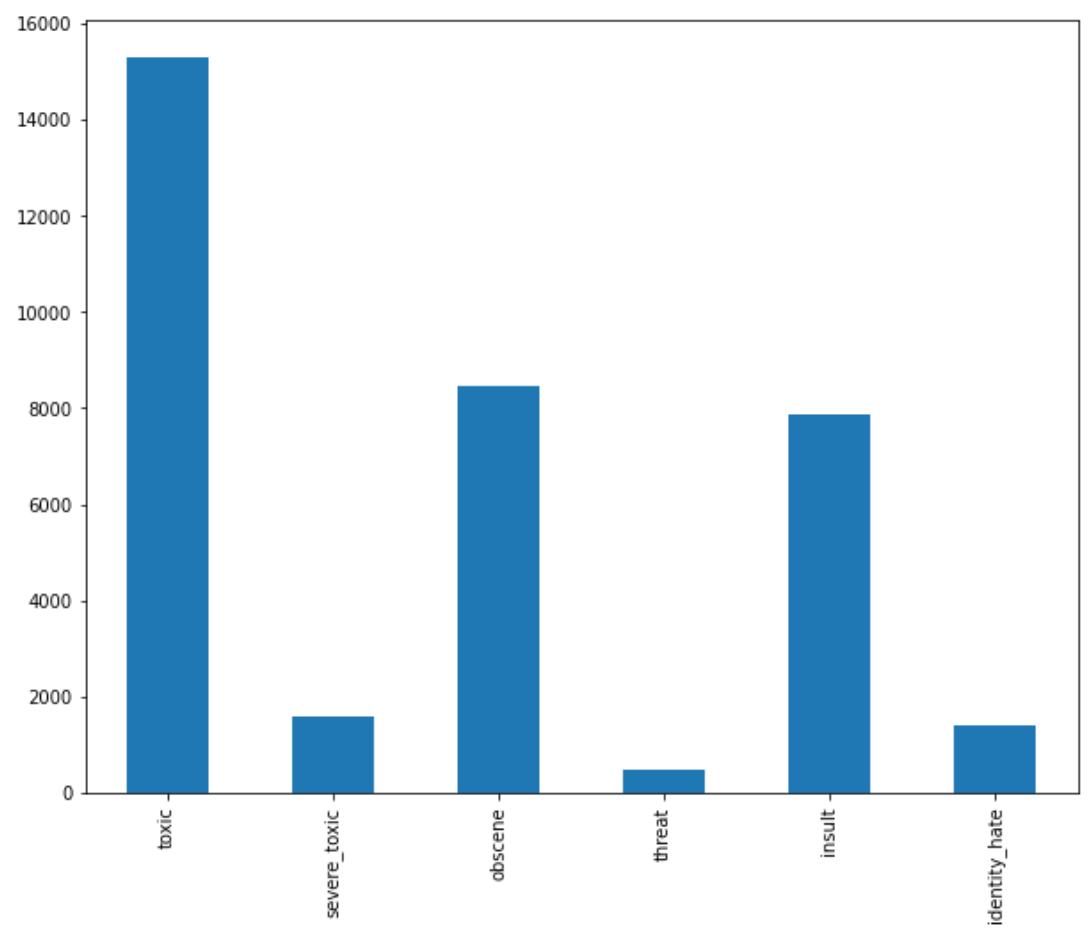

مشاهده می‌کنید که کلاس "toxic" بیشترین فراوانی را دارد. ما مجموعه داده‌یِ خود را با موفقیت تجزیه و تحلیل کردیم. در ادامه مدل طبقه‌بند چندبرچسبی را برای این مجموعه داده ایجاد خواهیم کرد.

به‌طور کلی، دو راه برای ایجاد مدل‌های طبقه‌بند چندبرچسبی وجود دارد: استفاده از یک لایه خروجی متصل کامل و استفاده از چندین لایه خروجی متصل کامل. در رویکرد اول، می‌توانیم از یک لایه متصل کامل با شش خروجی با تابع فعال‌ساز sigmoid و تابع زیان آنتروپی متقاطع دودویی استفاده کنیم. هر نورون در لایه متصل کامل، خروجیِ یکی از شش برچسب را نشان می‌دهد. همان‌طور که می‌دانیم، تابع فعال‌سازی sigmoid مقداری بین ۰ و ۱ برای هر نورون برمی‌گرداند. اگر مقدار خروجی هر نورون بزرگتر از ۰/۵ باشد، فرض می‌شود که نظر متعلق به کلاسی است که توسط آن نورون خاص نشان داده شده است.

در رویکرد دوم می‌توان یک لایه خروجی متصل کامل برای هربرچسب ایجاد کرد. برای این مثال، باید ٦ لایه متصل کامل در خروجی ایجاد کرد که هر لایه تابع sigmoid خود را خواهد داشت.

ما تنها از رویکرد اول برای این مجموعه داده استفاده می‌کنیم و یک مدل طبقه‌بند متن چند برچسبی را با یک لایه خروجی ایجاد خواهیم کرد. ابتدا، یک تابع ایجاد می‌کنیم تا به پاکسازی متن بپردازد:

```
def preprocess_text(sen):
    # Remove punctuations and numbers
    sentence = re.sub('[^a-zA-Z]', ' ', sen)
    # Single character removal
    sentence = re.sub(r"\s+[a-zA-Z]\s+", ' ', sentence)
    # Removing multiple spaces
    sentence = re.sub(r'\s+', ' ', sentence)
    return sentence
```

مرحله بعد مجموعه ورودی و خروجی خود را ایجاد می‌کنیم. ورودیِ نظر ستون comment_text است. ما تمام نظرات را در متغیر X ذخیره می‌کنیم. برچسب‌ها یا خروجی‌ها قبلا در toxic_comments_labels ذخیره شده‌اند. ما از آن مقادیر برای ذخیره‌یِ خروجی در متغیر y استفاده می‌کنیم. کد زیر این کار را انجام می‌دهد:

```
X = []
sentences = list(toxic_comments["comment_text"])
for sen in sentences:
    X.append(preprocess_text(sen))
y = toxic_comments_labels.values
```

در این مجموعه داده ما نیازی به انجام کدگذاری one-hot نداریم، چراکه برچسب‌هایِ خروجی ما قبلا تبدیل به بردارهای کدگذاری one-hot شده‌اند. در مرحله بعد، داده‌های خود را به مجموعه داده آموزشی و آزمون تقسیم می‌کنیم:

```
X_train, X_test, y_train, y_test = train_test_split(X, y, test_size=0.20,
random_state=42)
```

در ادامه نیاز به تبدیل ورودی خود به بردارهای عددی داریم. از این‌رو قبل از اینکه به ادامه این مثال بپردازیم بیاید تا در مورد جاساز کلمات (word embedding) بیشتر بدانیم. الگوریتم‌های یادگیری عمیق قادر به درک داده‌های متنی به‌صورت خام نیستند، از این رو باید متن را به‌گونه‌ای تبدیل کرد تا شبکه قادر به درک و پردازش آن‌ها باشد. جاساز کلمات روشی برای بازنمایی کلمات است که هدف آن نمایش معنایی کلمات در قالب بردارهایی حقیقی است، جایی که کلمات با معنی و زمینه مشابه با بردارهای مشابه نشان داده می‌شوند. این بردارهای عددی در مقایسه با رویکردهای آماری در پردازش زبان طبیعی برای تبدیل کلمات به اعداد، ابعاد کم‌تری دارند. هم‌چنین، این بردارهای عددی اگر به خوبی آموزش دیده شده باشند، توانایی این را دارند که ارتباطات معنایی و نحوی بین کلمات را نشان دهند. جاساز کلمات، سنگ بنای بسیاری از کارهای انجام گرفته در حوزه پردازش زبان طبیعی است که از یادگیری عمیق استفاده می‌کنند. جاساز کلمات را برای متون می‌توان از دو رویکرد متفاوت بدست آورد. در رویکرد اول، در حین آموزش شبکه هم‌زمان با کار اصلی یاد گرفته می‌شوند. در این روش، در ابتدا مقادیر عددی برای بردارها به‌صورت تصادفی تولید می‌شود و سپس در حین آموزش این مقادیر از طریق بهینه‌سازی مانند دیگر لایه‌های شبکه بروزرسانی می‌شوند. رویکرد دوم، از طریق آموزش دادن با الگوریتم‌های خاصی همانند fasttext و glove برروی مجموعه داده‌های بزرگ متنی و استفاده از وزن‌های بدست آمده از این الگوریتم‌ها است.

حال بیاید تا ورودی‌های متنی را به بردارهای جاساز تبدیل کنیم:

```
tokenizer = Tokenizer(num_words=5000)
tokenizer.fit_on_texts(X_train)
X_train = tokenizer.texts_to_sequences(X_train)
X_test = tokenizer.texts_to_sequences(X_test)
vocab_size = len(tokenizer.word_index) + 1
maxlen = 200
```

```
X_train = pad_sequences(X_train, padding='post', maxlen=maxlen)
X_test = pad_sequences(X_test, padding='post', maxlen=maxlen)
```

ما از جاسازی کلمه GloVe برای تبدیل ورودی‌های متن به همتایان عددی خود استفاده خواهیم کرد. برای بارگیری آن کد زیر را وارد کنید:

```
!wget http://nlp.stanford.edu/data/glove.6B.zip
!unzip glove*.zip
```

برای استفاده از آن به صورت زیر عمل می‌کنیم:

```
from numpy import array
from numpy import asarray
from numpy import zeros

embeddings_dictionary = dict()

glove_file = open('glove.6B.100d.txt', encoding="utf8")

for line in glove_file:
    records = line.split()
    word = records[0]
    vector_dimensions = asarray(records[1:], dtype='float32')
    embeddings_dictionary[word] = vector_dimensions
glove_file.close()

embedding_matrix = zeros((vocab_size, 100))
for word, index in tokenizer.word_index.items():
    embedding_vector = embeddings_dictionary.get(word)
    if embedding_vector is not None:
        embedding_matrix[index] = embedding_vector
```

سپس با کد زیر مدل خود را ایجاد می‌کنیم. مدل ما دارای یک لایه ورودی، یک لایه جاساز، یک لایه LSTM با ۱۲۸ نورون و یک لایه خروجی با ۶ نورون خواهد بود، چراکه ما ۶ برچسب در خروجی داریم.

```
deep_inputs = Input(shape=(maxlen,))
embedding_layer = Embedding(vocab_size, 100, weights=[embedding_matrix],
trainable=False)(deep_inputs)
LSTM_Layer_1 = LSTM(128)(embedding_layer)
dense_layer_1 = Dense(6, activation='sigmoid')(LSTM_Layer_1)
model = Model(inputs=deep_inputs, outputs=dense_layer_1)

model.compile(loss='binary_crossentropy', optimizer='adam', metrics=['acc'])
```

بیایید خلاصه مدل را چاپ کنیم:

```
print(model.summary())
```

```
Model: "model"
_________________________________________________________________
 Layer (type)                Output Shape              Param #
=================================================================
 input_1 (InputLayer)        [(None, 200)]             0

 embedding (Embedding)       (None, 200, 100)          14824300

 lstm (LSTM)                 (None, 128)               117248

 dense (Dense)               (None, 6)                 774

=================================================================
Total params: 14,942,322
Trainable params: 118,022
Non-trainable params: 14,824,300
```

می‌توان با استفاده از کد زیر معماری شبکه عصبی خود را به تصویر کشید:

```
from keras.utils.vis_utils import plot_model
plot_model(model, to_file='model_plot4a.png', show_shapes=True,
show_layer_names=True)
```

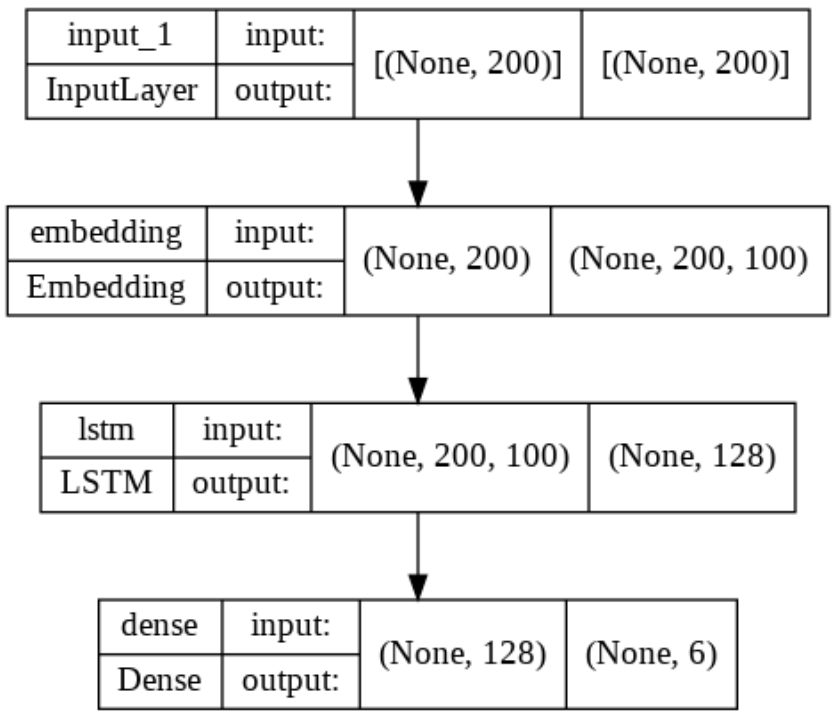

از شکل بالا می‌بینید که لایه خروجی فقط شامل ۱ لایه متصل کامل با ۶ نورون است. حال بیایید مدل خود را آموزش دهیم:

```
history = model.fit(X_train, y_train, batch_size=128, epochs=5, verbose=1,
validation_split=0.2)
```

```
Epoch 1/5
798/798 [==============================] - 20s 17ms/step - loss: 0.1193 - acc: 0.9684 - val_loss:
0.0739 - val_acc: 0.9941
Epoch 2/5
798/798 [==============================] - 13s 17ms/step - loss: 0.0643 - acc: 0.9927 - val_loss:
0.0599 - val_acc: 0.9943
Epoch 3/5
798/798 [==============================] - 13s 17ms/step - loss: 0.0572 - acc: 0.9938 - val_loss:
0.0573 - val_acc: 0.9935
Epoch 4/5
798/798 [==============================] - 13s 17ms/step - loss: 0.0548 - acc: 0.9939 - val_loss:
0.0566 - val_acc: 0.9943
Epoch 5/5
798/798 [==============================] - 14s 17ms/step - loss: 0.0523 - acc: 0.9940 - val_loss:
0.0542 - val_acc: 0.9942
```

حال بیایید مدل خود را در مجموعه آزمایشی ارزیابی کنیم:

```
score = model.evaluate(X_test, y_test, verbose=1)
print("Test Score:", score[0])
print("Test Accuracy:", score[1])
```

```
998/998 [==============================] - 6s 6ms/step - loss: 0.0529 - acc: 0.9938
Test Score: 0.05285229906439781
Test Accuracy: 0.9937959909439087
```

مدل ما به دقت ۹۹ درصد در مجموعه آزمون دست یافته است که بسیار عالی است. در نهایت، مقادیر زیان و دقت را برای مجموعه‌های آموزشی و آزمایشی ترسیم می‌کنیم تا ببینیم آیا مدل ما منجر به بیش‌برازش شده است یا خیر.

```
import matplotlib.pyplot as plt

plt.plot(history.history['acc'])
plt.plot(history.history['val_acc'])
```

```
plt.title('model accuracy')
plt.ylabel('accuracy')
plt.xlabel('epoch')
plt.legend(['train','test'], loc='upper left')
plt.show()
```

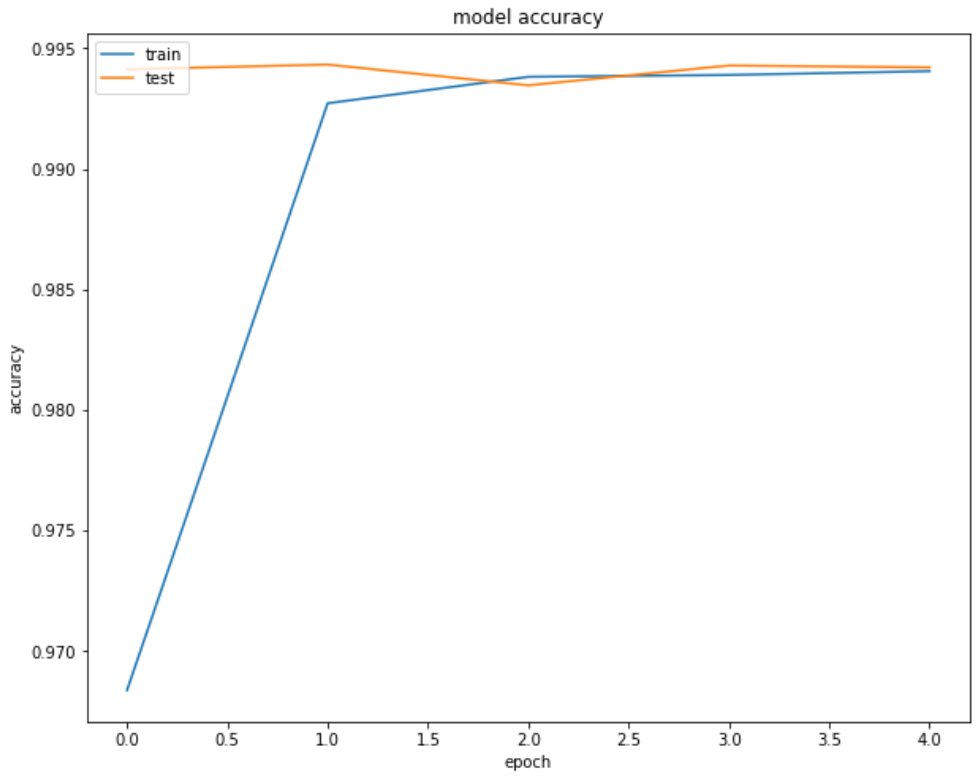

```
plt.plot(history.history['loss'])
plt.plot(history.history['val_loss'])

plt.title('model loss')
plt.ylabel('loss')
plt.xlabel('epoch')
plt.legend(['train','test'], loc='upper left')
plt.show()
```

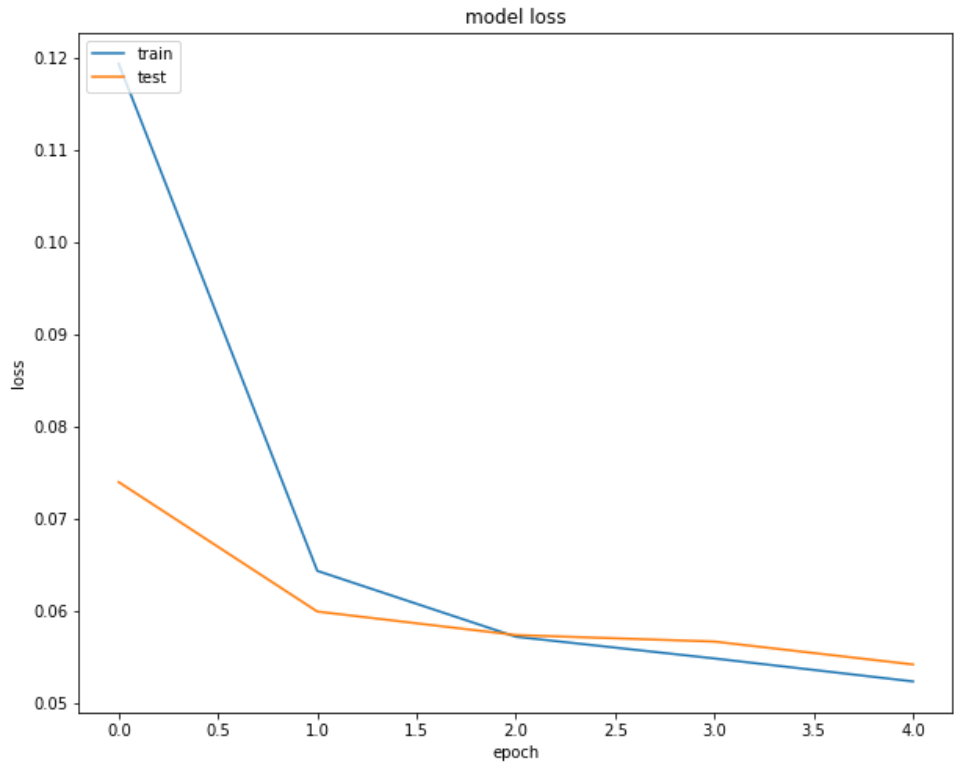

همچنان‌که در تصاویر بالا مشاهده می‌کنید، مدل منجر به بیش‌برازش نشده است.

## تحلیل احساسات با LSTM

با افزایش ابزارهای وب ۲ و ظهور رسانه‌های اجتماعی، زندگی امروز جوامع بشری با آن پیوند عمیقی خورده است. همین امر منجر به تولید حجم عظیمی از داده‌ها توسط کاربران این رسانه‌ها شده است. داده‌های متنی یکی از پرمصرف‌ترین‌ها است که می‌تواند برای بدست آوردن اطلاعات مهم در موضوعات مختلف مورد استفاده قرار گیرد. رسانه‌های اجتماعی در اشکال گوناگون

خود همانند انجمن‌ها، وبلاگ‌ها، میکروبلاگ‌ها، سایت‌های نظردهی و غیره روزانه منجر به تولید حجم وسیعی از داده‌ها می‌شوند. چنین داده‌هایی در قالب نظرات، نقدها، دیدگاه‌ها در مورد خدمات، شرکت‌ها، سازمان‌ها، رویدادها، افراد، مسائل و موضوعات می‌باشد. نظرات ارائه شده کاربران در شبکه‌های اجتماعی بسیار مهم و کاربردی هستند. در یک فروشگاه برخط، نظرات و دیدگاه‌های مختلف در مورد یک محصول می‌تواند سطح رضایت و کیفیت مشتری را منعکس سازد که این می‌تواند راهنمای بسیار خوبی برای سایر خریداران باشد. طبقه‌بندی و سازماندهی این حجم بسیار عظیم از نظرات در مورد یک موضوع خاص به‌صورت دستی کار آسانی نیست. از همین‌رو، نیاز به یک سیستم خودکار برای جمع‌آوری نظرات منجر به ظهور یک زمینه تحقیقاتی جدید به نام **تحلیل احساسات (sentiment analysis)** شد. تحلیل احساسات، زمینه مطالعاتی است که هدف اصلی آن شناسایی، استخراج و طبقه‌بندی احساسات، نظرات، نگرش‌ها، افکار، قضاوت‌ها، نقدها و دیدگاه‌ها نسبت به موجودیت‌ها، سازمان‌ها، رویدادها و غیره بدون تعامل انسانی در قالب دسته‌های مثبت، منفی و یا خنثی می‌باشد.

دو رویکرد متفاوتی که محققین برای طبقه‌بندی احساسات در یک متن استفاده می‌کنند، رویکردهای مبتنی‌بر واژگان و مبتنی‌بر یادگیری ماشین می‌باشد. رویکرد دیگری را هم می‌توان ترکیبی از این دو نظر گرفت. رویکرد مبتنی‌بر واژگان، متمرکز بر استخراج کلمات یا عباراتی است که می‌تواند فرآیند طبقه‌بندی را در جهت‌گیری معنایی خاصی هدایت کند. هر واژه دارای بار معنایی خاصی است که از طریق یک فرهنگ واژه از کلمات با بار احساسی مثبت و منفی که پیش‌تر امتیازبندی شده‌اند، استخراج می‌شود. با جمع امتیاز بار احساسی واژه‌ها یا شمارش تعداد واژه‌های با بار مثبت و منفی، قطبیت کلی جمله بدست می‌آید. رویکرد یادگیری ماشین را می‌توان در حالت‌های مختلفی برای مساله تحلیل احساسات آموزش داد و بکار برد. در حالت یادگیری بانظارت با یک مجموعه داده آموزشی که پیش‌تر برچسب خورده است مدل آموزش می‌بیند تا قادر به یادگیری شود و بتواند در مواجهه با داده‌های دیده نشده رفتاری مشابه با داده‌های آموزش دیده از خود نشان دهد.

طی سال‌های اخیر، به‌طور گسترده‌ای توسط محققین اثبات شده است که مدل‌های بازنمایی مبتنی‌بر یادگیری عمیق در مساله‌های مرتبط با طبقه‌بندی احساسات کارآیی بهتری دارند. اتخاذ رویکردهای یادگیری عمیق در تحلیل احساسات به دلیل توانایی بسیار بالای مدل‌های یادگیری عمیق در یادگیری ویژگی‌ها به‌صورت خودکار است که می‌تواند به دقت و عملکرد بهتری دست یابند. در بسیاری از زمینه‌های پردازش زبان طبیعی، استفاده از یادگیری عمیق سبب شده است نتایج از آنچه که در گذشته توسط روش‌های یادگیری ماشین و روش‌های آماری مورد استفاده قرار می‌گرفته است، فراتر رود.

حال که با تحلیل احساسات آشنایی پیدا کردید، بیاید تا به کمک شبکه LSTM یک مدل تحلیل احساسات در حوزه‌ی نقدهای فیلم‌های سینمایی بسازیم. برای این پیاده‌سازی، از

مجموعه داده IMDB استفاده می‌کنیم. مزیت این مجموعه داده این است که از قبل با کتابخانه مجموعه داده‌های Keras همراه است.

ابتدا مجموعه داده را از طریق کد زیر بارگیری می‌کنیم:

```
from keras.datasets import imdb
top_words = 5000
(X_train, y_train), (X_test, y_test) = imdb.load_data(num_words=top_words)
```

کد بالا همزمان با بارگیری ۵۰۰۰ کلمه برتر هر نقد، مجموعه داده را به دو مجموعه آموزشی و آزمایشی تقسیم می‌کند. حال بیایید تا نگاهی به مجموعه داده بیاندازیم:

```
X_train
array([list([1, 14, 22, 16, 43, 530, 973, 1622, 1385, 65, 458, 4468, 66, 3941, 4, 173, 36, 256, 5,
25, 100, 43, 838, 112, 50, 670, 22665, 9, 35, 480, 284, 5, 150, 4, 172, 112, 167, 21631, 336, 385,
39, 4, 172, 4536, 1111, 17, 546, 38, 13, 447, 4, 192, 50, 16, 6, 147, 2025, 19, 14, 22, 4, 1920,
4613, 469, 4, 22, 71, 87, 12, 16, 43, 530, 38, 76, 15, 13, 1247, 4, 22, 17, 515, 17, 12, 16, 626,
18, 19193, 5, 62, 386, 12, 8, 316, 8, 106, 5, 4, 2223, 5244, 16, 480, 66, 3785, 33, 4, 130, 12,
16, 38, 619, 5, 25, 124, 51, 36, 135, 48, 25, 1415, 33, 6, 22, 12, 215, 28, 77, 52, 5, 14, 407,
16, 82, 10311, 8, 4, 107, 117, 5952, 15, 256, 4, 31050, 7, 3766, 5, 723, 36, 71, 43, 530, 476, 26,
400, 317, 46, 7, 4, 12118, 1029, 13, 104, 88, 4, 381, 15, 297, 98, 32, 2071, 56, 26, 141, 6, 194,
7486, 18, 4, 226, 22, 21, 134, 476, 26, 480, 5, 144, 30, 5535, 18, 51, 36, 28, 224, 92, 25, 104,
4, 226, 65, 16, 38, 1334, 88, 12, 16, 283, 5, 16, 4472, 113, 103, 32, 15, 16, 5345, 19, 178, 32]),
       list([1, 194, 1153, 194, 8255, 78, 228, 5, 6, 1463, 4369, 5012, 134, 26, 4, 715, 8, 118,
       ...,
       list([1, 1446, 7079, 69, 72, 3305, 13, 610, 930, 8, 12, 582, 23, 5, 16, 484, 685, 54, 349,
11, 4120, 2959, 45, 58, 1466, 13, 197, 12, 16, 43, 23, 21469, 5, 62, 30, 145, 402, 11, 4131, 51,
575, 32, 61, 369, 71, 66, 770, 12, 1054, 75, 100, 2198, 8, 4, 105, 37, 69, 147, 712, 75, 3543, 44,
257, 390, 5, 69, 263, 514, 105, 50, 286, 1814, 23, 4, 123, 13, 161, 40, 5, 421, 4, 116, 16, 897,
13, 40691, 40, 319, 5872, 112, 6700, 11, 4803, 121, 25, 70, 3468, 4, 719, 3798, 13, 18, 31, 62,
40, 8, 7200, 4, 29455, 7, 14, 123, 5, 942, 25, 8, 721, 12, 145, 5, 202, 12, 160, 580, 202, 12, 6,
52, 58, 11418, 92, 401, 728, 12, 39, 14, 251, 8, 15, 251, 5, 21213, 12, 38, 84, 80, 124, 12, 9,
23]),
       list([1, 17, 6, 194, 337, 7, 4, 204, 22, 45, 254, 8, 106, 14, 123, 4, 12815, 270, 14437, 5,
16923, 12255, 732, 2098, 101, 405, 39, 14, 1034, 4, 1310, 9, 115, 50, 305, 12, 47, 4, 168, 5, 235,
7, 38, 111, 699, 102, 7, 4, 4039, 9245, 9, 24, 6, 78, 1099, 17, 2345, 16553, 21, 27, 9685, 6139,
5, 29043, 1603, 92, 1183, 4, 1310, 7, 4, 204, 42, 97, 90, 35, 221, 109, 29, 127, 27, 118, 8, 97,
12, 157, 21, 6789, 85010, 9, 6, 66, 78, 1099, 4, 631, 1191, 5, 2642, 272, 191, 1070, 6, 7585, 8,
2197, 70907, 10755, 544, 5, 383, 1271, 848, 1468, 12183, 497, 16876, 8, 1597, 8778, 19280, 21, 60,
27, 239, 9, 43, 8368, 209, 405, 10, 10, 12, 764, 40, 4, 248, 20, 12, 16, 5, 174, 1791, 72, 7, 51,
6, 1739, 22, 4, 204, 131, 9])],
      dtype=object)
```

اگر به داده‌های بالا نگاه کنید متوجه می‌شوید که داده‌ها قبلا پردازش شده‌اند. همه‌ی کلمات به اعداد صحیح نگاشت شده‌اند و اعداد صحیح نشان‌دهنده کلمات مرتب شده بر اساس فراوانی آن‌ها هستند. به عنوان مثال، ۴ نشان دهنده چهارمین کلمه پرکاربرد، ۵ پنجمین کلمه پرکاربرد و غیره است. عدد صحیح ۱ برای نشانگر شروع، عدد صحیح ۲ برای یک کلمه ناشناخته و ۰ برای padding رزرو شده است. اگر می‌خواهید خودتان به نقدها نگاهی بیندازید و ببینید مردم چه نوشته‌اند، می‌توانید این روند را نیز معکوس کنید:

```
word_index = imdb.get_word_index() # get {word : index}
index_word = {v : k for k,v in word_index.items()} # get {index : word}
index = 1
print(" ".join([index_word[idx] for idx in x_train[index]]))
print("positve" if y_train[index]==1 else "negetive")
```

```
the thought solid thought senator do making to is spot nomination assumed while he of
jack in where picked as getting on was did hands fact characters to always life
thrillers not as me can't in at are br of sure your way of little it strongly random
to view of love it so principles of guy it used producer of where it of here icon film
of outside to don't all unique some like of direction it if out her imagination below
keep of queen he diverse to makes this stretch stefan of solid it thought begins br
```

```
senator machinations budget worthwhile though ok brokedown awaiting for ever better were lugia diverse for budget look kicked any to of making it out bosworth's follows for effects show to show cast this family us scenes more it severe making senator to levant's finds tv tend to of emerged these thing wants but fuher an beckinsale cult as it is video do you david see scenery it in few those are of ship for with of wild to one is very work dark they don't do dvd with those them
negetive
```

از آنجایی که نقدها، از نظرِ طول بسیار متفاوت هستند، می‌خواهیم هر نقد را به ۵۰۰ کلمه اول برش دهیم. ما باید نمونه‌های متنی با طول یکسان داشته باشیم تا بتوانیم آن‌ها را به شبکه عصبی خود وارد کنیم. اگر نقدها کوتاه‌تر از ۵۰۰ کلمه باشند، آن‌ها را با صفر اضافه می‌کنیم. Keras برای این کار بسیار عالی است، چراکه مجموعه‌ای از روال‌های پیش‌پردازش را ارائه می‌دهد که می‌تواند این کار را براحتی برای ما انجام دهد:

```
word_index = imdb.get_word_index() # get {word : index}
index_word = {v : k for k,v in word_index.items()} # get {index : word}
index = 1
print(" ".join([index_word[idx] for idx in x_train[index]]))
print("positve" if y_train[index]==1 else "negetive")
```

برای درک بهتر بیاید یک نمونه از داده‌ها را به صورت تصادفی انتخاب کنیم و مشاهده کنیم که کد بالا چه کاری انجام داده است:

```
X_train[125]
```

```
array([    0,     0,     0,     0,     0,     0,     0,     0,     0,
           0,     0,     0,     0,     0,     0,     0,     0,     0,
           0,     0,     0,     0,     0,     0,     0,     0,     0,
           0,     0,     0,     0,     0,     0,     0,     0,     0,
           0,     0,     0,     0,     0,     0,     0,     0,     0,
           0,     0,     0,     0,     0,     0,     0,     0,     0,
           0,     0,     0,     0,     0,     0,     0,     1,    11,
           6,    58,    54,     4, 14537,     5,  6495,     4,  2351,
        1630,    71, 13202,    23,    26, 20094, 40865,    34,    35,
        9454,  1680,     8,  6681,   692,    39,    94,   205,  6177,
         712,   121,     4, 18147,  7037,   406,  2657,     5,  2189,
       61778,    26, 23906, 11420,     6,   708,    44,     4,  1110,
         656,  4667,   206,    15,   230, 13781,    15,     7,     4,
        4847,    36,    26, 54759,   238,   306,  2316,   190,    48,
          25,   181,     8,    67,     6,    22,    44,    15,   353,
        1659, 84675,  3048,     4,  9818,   305,    88, 11493,     9,
          31,     7,     4,    91, 12789, 53410,  3106,   126,    93,
          40,   670,  8759, 41931,     6,  6951,     4,   167,    47,
         623,    23,     6,   875,    29,   186,   340,  4447,     7,
           5, 44141,    27,  5485,    23,   220,   175,  2122,    10,
          10,    27,  4847,    26,     6,  5261,  2631,   604,     7,
        2118, 23310, 36011,  5350,    17,    48,    29,    71, 12129,
          18,  1168, 38886, 33829,  1918, 31235,  3255,  9977, 31537,
        9248,    40,    35,  1755,   362,    11,     4,  2370,  2222,
          56,     7,     4, 23052,  2489,    39,   609, 82401, 48583,
           6,  3440,   655,   707,  4198,  3801,    37,  4486,    33,
         175,  2631,   114,   471,    17,    48,    36,   181,     8,
          30,  1059,     4,  3408,  5963,  2396,     6,   117,   128,
          21,    26,   131,  4218,    11, 20663,  3826, 14524,    10,
          10,    12,     9,   614,     8,    97,     6,  1393,    22,
          44,   995,    84, 21800,  5801,    21,    14,     9,     6,
        4953,    22,    44,   995,    84,    93,    34,    84,    37,
         104,   507, 11076,    37,    26,   662,   180,     8,     4,
        5075,    11,   882,    71,    31,     8, 39022, 36011, 31537,
           5, 48583,    19,  1240, 31800,  1806, 11521,     5,  7863,
       28281,     4,   959,    62,   165,    30,     8,  2988,     4,
```

```
2772,  1500,     7,     4,    22,    24,  2358,    12,    10,
  10, 22993,   238,    43,    28,   188,   245,    19,    27,
 105,     5,   687,    48,    29,   562,    98,   615,    21,
  27,  8500,     9,    38,  2797,     4,   548,   139,    62,
9343,     6, 14053,   707,   137,     4,  1205,     7,     4,
6556,     9,    53,  2797,    74,     4,  6556,   410,     5,
  27,  5150,     8,    79,    27,   177,     8,  3126,    19,
  33,   222,    49, 22895,     7, 14090,   406,  5424,    38,
4144,    15,    12,     9,   165,  2268,     8,   106,   318,
 760,   215,    30,    93,   133,     7, 31537,    38,    55,
  52,    11, 26149, 17310,  1080, 24192, 68007,    29,     9,
6248,    78,   133,    11,     6,   239,    15,     9,    38,
 230,   120,     4,   350,    45,   145,   174,    10,    10,
22993,    47,    93,    49,   478,   108,    21,    25,    62,
 115,   482,    12,    39,    14,  2342,   947,     6,  6950,
   8,    27,   157,    62,   115,   181,     8,    67,   160,
   7,    27,   108,   103,    14,    63,    62,    30,     6,
  87,   902,  2152,  3572,     5,     6,   619,   437,     7,
   6,  4616,   221,   819,    31,   323,    46,     7,   747,
   5,   198,   112,    55,  3591], dtype=int32)
```

همانطور که در بالا می‌بینید، به دلیل اینکه این رکورد طولی کم‌تر از ۵۰۰ کلمه داشته است، تعدادی ۰ جلوی آن قرار گرفته است تا این رکورد طولی برابر با ۵۰۰ داشته باشد.

به طرز تعجب‌آوری کار پیش‌پردازش داده‌های ما به تمام رسید و اکنون می‌توانیم شروع به ساخت مدل خود کنیم:

```
from keras.models import Sequential
from keras.layers import Embedding
from keras.layers import LSTM, Dense
embedding_vector_length = 32
model = Sequential()
model.add(Embedding(top_words, embedding_vector_length,
input_length=max_review_length))
model.add(LSTM(100))
model.add(Dense(1, activation='sigmoid'))
model.compile(loss='binary_crossentropy',optimizer='adam',
metrics=['accuracy'])
print(model.summary())
```

```
Model: "sequential_2"
_________________________________________________________________
 Layer (type)                Output Shape              Param #
=================================================================
 embedding (Embedding)       (None, 500, 32)           160000

 lstm (LSTM)                 (None, 100)               53200

 dense (Dense)               (None, 1)                 101

=================================================================
Total params: 213,301
Trainable params: 213,301
Non-trainable params: 0
_________________________________________________________________
None
```

همان‌طور که پیش‌تر بیان شد، دو راه برای جاساز کلمات وجود دارد. در مثال قبلی ما از جاساز کلمات از پیش‌آموزش دیده استفاده کردیم. در این مثال از لایه Embedding استفاده می‌کنیم. لایه embedding یک جاساز کلمات را از مجموعه داده یاد می‌گیرد.

اکنون نوبت به آموزش مدل می‌رسد:

```
model.fit(X_train, y_train, validation_data=(X_test, y_test), epochs=5,
batch_size=32)
```

```
Epoch 1/5
782/782 [==============================] - 82s 103ms/step - loss: 0.4681 - accuracy:
0.7815 - val_loss: 0.3702 - val_accuracy: 0.8430
Epoch 2/5
782/782 [==============================] - 79s 101ms/step - loss: 0.3340 - accuracy:
0.8618 - val_loss: 0.3984 - val_accuracy: 0.8299
Epoch 3/5
782/782 [==============================] - 80s 102ms/step - loss: 0.2669 - accuracy:
0.8954 - val_loss: 0.3272 - val_accuracy: 0.8657
Epoch 4/5
782/782 [==============================] - 80s 102ms/step - loss: 0.2259 - accuracy:
0.9117 - val_loss: 0.3122 - val_accuracy: 0.8713
Epoch 5/5
782/782 [==============================] - 79s 101ms/step - loss: 0.1912 - accuracy:
0.9270 - val_loss: 0.3399 - val_accuracy: 0.8604
```

پس از اتمام آموزش مدل، نوبت ارزیابی کارایی مدل است:

```
scores = model.evaluate(X_test, y_test, verbose=0)
print("Accuracy: %.2f%%" % (scores[1]*100))
```

```
Accuracy: 86.04%
```

همان‌طور که مشاهده می‌شود مدل در عین سادگی توانسته است به دقتی حدود ۸۶ درصد دست یابد که با توجه به مسئله دشوار بسیار عالی است. با این حال، این مدل بهترین مدلِ ممکن نیست. به عنوان تمرین، می‌توانید با آزمایش برروی ابرپارمترهای مختلف نتایج را مشاهده کنید و یک مدل با کارایی بالا بسازید. همچنین، نمودار زیان و دقت را برای مدل‌ها رسم کنید و ببینید آیا مدل‌ها منجر به بیش‌برازش شده‌اند یا خیر. در مورد مدل بالا نظر شما در خصوص بیش‌برازش چیست؟

## خلاصه فصل

- **شبکه‌های عصبی بازگشتی، نواقصِ شبکه‌هایِ عصبیِ پیش‌خور را برطرف می‌کنند.**
- **RNNهای ساده قادر به یادگیریِ وابستگی‌هایِ بلند مدت نیستند.**
- **LSTM می‌تواند وابستگی‌های بلندمدت را به دلیلِ وجودِ یک سلولِ حافظه‌یِ مخصوص در ساختارش، انجام دهد.**

## آزمونک

**یک LSTM چه تعداد دروازه دارد و نقش هر یک از آن‌ها چیست؟**

# ۶

# شبکه متخاصم مولد

## اهداف یادگیری:

- تفاوت مدل مولد با مدل تفکیک‌گر
- آشنایی با شبکه متخاصم مولد
- آموزش به شیوه تخاصمی
- تولید ارقام دست‌نویس با شبکه متخاصم مولد

# مقدمه

در این فصل قصد داریم به معرفی خانواده‌ای از مدل‌های مولد بر اساس برخی مفاهیم تئوری بازی‌ها بپردازیم. ویژگیِ اصلی آن‌ها یک روش آموزشی متخاصم است که با هدفِ یادگیریِ تمایز بین نمونه‌های واقعی و جعلی توسط یک مولفه، در همان زمان، مولفه‌ی دیگری نمونه‌هایی را بیشتر و بیشتر شبیه به نمونه‌های آموزشی تولید می‌کند. به صورت خلاصه در این شبکه‌ها، یکی تولید می‌کند، دیگری نکته‌گیری می‌کند و در کنار هم و در یک همکاری کامل، نتایج بسیار خوبی بدست می‌آورند.

# مدل مولد چیست؟

به‌طور کلی دو نوع مدل اصلی در یادگیری ماشین وجود دارد: **مدل مولد (generative model)** و **مدل تفکیک‌گر (discriminative model)**. یک مدل تفکیک‌گر، همان‌طور که از نامش پیداست، سعی می‌کند داده‌ها را بین دو یا چند کلاس تفکیک کند. به‌طور کلی، مدل‌های تفکیک‌گر بر پیش‌بینی کلاس‌های داده با توجه به ویژگی‌های آن‌ها تمرکز می‌کنند. برعکس، مدل مولد سعی نمی‌کند ویژگی‌ها را به کلاس‌ها نگاشت کند، بلکه ویژگی‌هایی را تولید می‌کند که در یک کلاس خاص وجود دارد. یک راه آسان برای تشخیص یک مدل مولد از یک مدل تفکیک‌گر وجود دارد:

- مدل تفکیک‌گر به یافتن مرزها یا قوانینی برای تفکیک داده‌ها علاقه‌مند است.
- مدل مولد بر مدل‌سازیِ توزیع داده‌ها تمرکز دارد.

**تعریف ۱.۶ مدل مولد**

**یک مدل مولد نحوه تولید یک مجموعه داده را براساس یک مدل احتمالی توصیف می‌کند.**

مدل‌های مولد ابزار قدرتمندی برای بررسی توزیع داده‌ها یا تخمین چگالی مجموعه‌های داده هستند. مدل‌های مولد از یادگیری بدون‌نظارت پیروی می‌کنند که به‌طور خودکار الگوها یا بی‌نظمی‌های داده‌های مورد تجزیه و تحلیل را کشف می‌کند. این به تولید داده‌های جدیدی که عمدتا شبیه مجموعه داده اصلی است کمک می‌کند. به‌طور دقیق، هدف مدل‌های مولد یادگیریِ توزیعِ واقعیِ داده‌ها در مجموعه آموزشی، برایِ تولیدِ نقاطِ داده‌یِ جدید با برخی تغییرات است.

هدف اصلی انواع مدل‌های مولد یادگیری توزیعِ واقعیِ داده‌هایِ مجموعه آموزشی است تا نقاط داده جدید با تغییراتی تولید شوند. اما این امکان برای مدل وجود ندارد که توزیع دقیق داده‌های ما را بیاموزد و بنابراین توزیعی را مدل‌سازی می‌کنیم که مشابه توزیع داده‌های واقعی

است. برای این کار، ما از دانش شبکه‌های عصبی برای یادگیری تابعی استفاده می‌کنیم که می‌تواند توزیع مدل را به توزیع واقعی تقریب بزند.

مدل‌های مولد را می‌توان به عنوان یک کلاس از مدل‌ها تعریف کرد که هدف آن‌ها یادگیری نحوه‌ی تولید نمونه‌های جدیدی است که به نظر می‌رسد از همان مجموعه داده‌های آموزشی هستند. در طول مرحله آموزش، یک مدل مولد در تلاش است یک مساله تخمین چگالی را حل کند. در تخمین چگالی، مدل می‌آموزد تا یک تخمین تا حد امکان شبیه به تابع چگالی احتمال غیرقابل مشاهده بسازد. نکته مهم این است که، مدل مولد باید بتواند نمونه‌های جدیدی از توزیع را تشکیل دهد و نه فقط نمونه‌های موجود را رونوشت و ایجاد کند.

مدل‌های مولد طی دهه گذشته در خط مقدم یادگیری بدون‌نظارتِ عمیق قرار داشته‌اند. دلیل این امر این است که آن‌ها روشی بسیار کارآمد را برای تجزیه و تحلیل و درک داده‌های بدون برچسب ارائه می‌دهند.

## مدل‌های مولد و آینده هوش مصنوعی؟

سه دلیل کلی وجود دارد که چرا مدل‌های مولد را می‌توان کلیدِ بازگشاییِ شکلِ بسیار پیچیده‌تری از هوش مصنوعی در نظر گرفت که فراتر از آن چیزی است که مدل‌های تفکیک‌گر می‌توانند به آن دست یابند.

- اولا، صرفا از نقطه نظرِ تئوری ما نباید تنها به توانایی برتر در طبقه‌بندی داده‌ها بسنده کنیم، بلکه باید به دنبال درک کامل‌تری از نحوه تولید داده‌ها در وهله اول باشیم. بدون شک حل این مسئله بسیار دشوارتر در مقایسه روش‌های تفکیک‌گر است. با این حال، همان‌طور که خواهیم دید، بسیاری از تکنیک‌های مشابهی که باعث توسعه در مدل‌سازی تفکیک‌گر شده‌اند (همانند یادگیری عمیق)، می‌توانند توسط مدل‌های مولد نیز مورد استفاده قرار گیرند.
- دوم، این احتمال وجود دارد که مدل‌سازی مولد برایِ هدایتِ پیشرفت‌های آینده در زمینه‌های دیگر یادگیری ماشین، همانند یادگیری تقویتی، مهم‌تر و تاثیرگذارتر از هر چیز دیگری باشد. به عنوان مثال، می‌توانیم از یادگیری تقویتی برای آموزش یک ربات برای راه رفتن در یک زمین خاص استفاده کنیم. رویکرد کلی، ساخت یک شبیه‌سازی رایانه‌ای از زمین و سپس اجرای آزمایش‌های زیادی است که در آن عامل استراتژی‌های مختلف را امتحان می‌کند. با گذشت زمان، عامل می‌آموزد که کدام استراتژی‌ها موفق‌تر از سایرین هستند و بنابراین به تدریج بهبود می‌یابند. یک مشکل معمولی با این رویکرد این است که فیزیک محیط اغلب بسیار پیچیده است و باید در هر مرحله محاسبه شود تا اطلاعات

به عامل برای تصمیم‌گیری در مورد حرکتِ بعدیِ خود، بازگردانده شود. با این حال، اگر عامل بتواند محیط خود را از طریق یک مدل مولد شبیه‌سازی کند، نیازی به آزمایش استراتژی در شبیه‌سازی رایانه‌ای یا در دنیای واقعی نخواهد داشت، بلکه می‌تواند در محیط خیالی خود بیاموزد.

- در نهایت، اگر واقعا بخواهیم بگوییم که ماشینی ساخته‌ایم که شکلی از هوش را بدست آورده است که با هوش انسان قابل مقایسه است، مدل‌سازی مولد مطمئنا باید بخشی از راه‌حل باشد. یکی از بهترین نمونه‌های مدل مولد در دنیای واقعی، شخصی است که این کتاب را می‌خواند. لحظه‌ای به این فکر کنید که شما چه مدل مولد باورنکردنی هستید. می‌توانید چشمان خود را ببندید و تصور کنید که یک فیل از هر زاویه ممکن چه شکلی است. شما می‌توانید تعدادیِ پایان متفاوت و قابل قبول برای برنامه تلویزیونی مورد علاقه خود تصور کنید و همچنین شما می‌توانید هفته خود را با کار از طریق آینده در ذهن خود برنامه‌ریزی کرده و براساس آن اقدام کنید. نظریه عصب‌شناسی فعلی نشان می‌دهد که درک ما از واقعیت یک مدل تفکیک‌گر بسیار پیچیده نیست که بر روی ورودی حسی ما برای تولید پیش‌بینی‌هایی از آنچه تجربه می‌کنیم عمل می‌کند، بلکه در عوض یک مدل مولد است که از بدو تولد برای تولید شبیه‌سازی محیط اطرافمان که دقیقا با آینده تطبیق می‌کند، آموزش داده می‌شود.

## شبکه متخاصم مولد (Generative Adversarial Network)

به نحوه‌یِ یادگیری خود فکر کنید. شما چیزی را امتحان می‌کنید و بازخورد دریافت می‌کنید. شما استراتژی خود را تنظیم می‌کنید و دوباره تلاش می‌کنید. بازخورد ممکن است به شکل انتقاد، درد یا سود باشد. ممکن است از قضاوت خودتان در مورد اینکه چقدر خوب عمل کرده‌اید ناشی شود. با این حال، اغلب اوقات مفیدترین بازخورد، بازخوردی است که از طرف شخص دیگری می‌آید، چراکه فقط یک عدد یا احساس نیست، بلکه ارزیابی هوشمندانه‌ای است از اینکه چقدر کار را به خوبی انجام داده‌اید.

هنگامی که یک رایانه برای یک کار آموزش داده می‌شود، انسان معمولا بازخورد را در قالب تنظیم پارامترها یا الگوریتم‌ها ارائه می‌دهد. وقتی کار ساده‌ای مانند یادگیری ضرب دو عدد باشد، این بازخورد آسان‌تر است. شما می‌توانید براحتی و دقیقا به رایانه بگویید که چگونه اشتباه کرده است. حال آنکه با یک کار پیچیده‌تر، مانند ایجاد تصویری از گربه، ارائه‌یِ بازخورد دشوارتر می‌شود. آیا تصویر تار است، آیا بیشتر شبیه یک سگ است یا اصلا شبیه چیزی است؟ می‌توان آمارهای پیچیده‌ای را پیاده‌سازی کرد، اما ثبت تمام جزئیاتی که یک تصویر را واقعی جلوه می‌دهد، دشوار است. یک انسان می‌تواند تخمین بزند، چراکه ما تجربه زیادی در ارزیابی ورودی

بصری داریم، اما نسبتا کند هستیم و ارزیابی‌های ما می‌تواند بسیار ذهنی باشد. در عوض می‌توانیم یک شبکه عصبی را آموزش دهیم تا وظیفه تمایز بین تصاویر واقعی و تولید شده را بیاموزد. سپس، می‌توان به مولد تصویر (شبکه عصبی) و تمایزگر فرصت داد تا از یکدیگر یاد بگیرند و در طول زمان بهبود یابند. این دو شبکه که این بازی را انجام می‌دهند، یک شبکه متخاصم مولد هستند.

**شبکه‌های متخاصم مولد** یا به اختصار **GAN**، دسته‌ای از تکنیک‌های یادگیری ماشین هستند که از دو مدل آموزش داده شده به‌طور همزمان تشکیل شده‌اند: یکی **مولد** (**Generator**) که برای تولید داده‌های جعلی آموزش داده شده است و دیگری **تمایزگر** (**Discriminator**) که آموزش دیده است تا به تشخیص داده‌های جعلی از نمونه‌های واقعی بپردازد. کلمه مولد، هدف کلی مدل را نشان می‌دهد: *ایجاد داده‌های جدید*. داده‌هایی که یک GAN یاد می‌گیرد تولید کند به انتخاب مجموعه آموزشی بستگی دارد. برای مثال، اگر بخواهیم یک GAN تصاویری شبیه به داوینچی را ترکیب کند، از مجموعه داده آموزشی آثار هنری داوینچی استفاده می‌کنیم.

اصطلاح **تخاصم** (**adversarial**) به رقابتی پویا و بازی‌مانندِ بین دو مدلی که چارچوب GAN را تشکیل می‌دهند اشاره دارد: ***مولد*** و ***تمایزگر***. هدف مولد ایجاد نمونه‌هایی است که از داده‌های واقعی در مجموعه آموزشی قابل تشخیص نیستند. در مثال ما، این به معنای تولید نقاشی هایی است که دقیقا شبیه نقاشی‌های داوینچی هستند. هدف تمایزگر تشخیص نمونه‌های جعلی تولید شده توسط مولد از نمونه‌هایِ واقعیِ حاصل از مجموعه داده‌هایِ آموزشی است. در مثال ما، تمایزگر، نقش یک متخصص هنری را بازی می‌کند که اصالت نقاشی‌هایی را که تصور می‌شود متعلق به داوینچی است، ارزیابی می‌کند. این دو شبکه به طور مداوم در تلاش هستند تا یکدیگر را فریب دهند: هر چه مولد در ایجاد داده‌هایِ واقع‌بینانه بهتر باشد، تمایزگر باید در تشخیص نمونه‌های واقعی از نمونه‌های جعلی بهتر عمل کند.

در نهایت، کلمه شبکه‌ها کلاسِ مدل‌های یادگیریِ ماشینی را نشان می‌دهد که معمولا برای نشان دادن مولد و تمایزگر استفاده می‌شوند: **شبکه‌های عصبی**. بسته به پیچیدگی پیاده‌سازی GAN، می‌توانند از شبکه‌های عصبی پیش‌خور ساده تا شبکه‌های عصبی کانولوشنی یا حتی انواع پیچیده‌تری از آن‌ها باشند.

ریاضیات زیربنایی GANها پیچیده هستند. خوشبختانه، بسیاری از قیاس‌هایِ دنیایِ واقعی می‌توانند درک GANها را آسان‌تر کنند. در مثال قبلی در مورد یک جاعلِ هنری (مولد) صحبت کردیم که تلاش می‌کند یک متخصص هنری (تمایزگر) را فریب دهد. هر چه نقاشی‌های جعلی که جاعل می‌سازد متقاعدکننده‌تر باشد، متخصص هنری باید در تشخیص صحت آن‌ها بهتر عمل کند. این امر در وضعیت معکوس نیز صادق است: هر چه متخصص هنری در تشخیص واقعی بودن یک نقاشیِ خاص بهتر باشد، جاعل باید برای جلوگیری از گرفتار شدن، بهتر عمل کند.

به عبارت دقیق‌تر، هدفِ مولد، تولید نمونه‌هایی است که ویژگی‌هایِ مجموعه داده آموزشی را به تصویر می‌کشد، به‌طوری که نمونه‌هایی که تولید می‌کند از داده‌های آموزشی قابل تشخیص نیستند. مولد را می‌توان به عنوان یک مدل تشخیص شی در وضعیت معکوس در نظر گرفت. الگوریتم‌های تشخیص اشیا الگوهای موجود در تصاویر را یاد می‌گیرند تا محتوای یک تصویر را تشخیص دهند. در مقابل، به جای تشخیص الگوها، مولد یاد می‌گیرد که اساسا آن‌ها را از ابتدا ایجاد کند. در واقع، ورودی مولد اغلب بیش از یک بردار اعداد تصادفی نیست. مولد از طریق بازخوردهایی که از تمایزگر دریافت می‌کند، یاد می‌گیرد. هدف تمایزگر، تعیین این است که آیا یک نمونه‌یِ خاص واقعی (برگرفته از مجموعه داده آموزشی) یا جعلی (ایجاد شده توسط مولد) است. بر این اساس، هر بار که تمایزگر فریب خورده و یک تصویر جعلی را به عنوان واقعی طبقه‌بندی می‌کند، مولد، می‌داند که کاری را به خوبی انجام داده است. برعکس، هر بار که تمایزگر به درستی تصویر تولید شده توسط مولد را به عنوان جعلی رد می‌کند، مولد بازخوردی را دریافت می‌کند که باید بهبود یابد. تمایزگر نیز به بهبود خود ادامه می‌دهد. مانند هر طبقه‌بند دیگری، از اختلاف بین پیش‌بینی‌هایش با برچسب‌های واقعی (واقعی یا جعلی) یاد می‌گیرد. بنابراین، همانطور که مولد در تولید داده‌های واقعی بهتر می‌شود، تمایزگر در تشخیص داده‌های جعلی از واقعی بهتر می‌شود و هر دو شبکه به طور همزمان به پیشرفت خود ادامه می‌دهند.

در شبکه‌های متخاصم مولد، نویز به شبکه‌یِ عصبیِ مولد القا می‌شود که نمونه‌های جعلی از طریق آن ایجاد می‌شود. وظیفه شبکه‌ی تمایزگر شناسایی نمونه‌های جعلی تولید شده توسط شبکه مولد است. این با بررسی نمونه‌های آموزشی مشخص می‌شود تا ببینید که نمونه تولید شده چقدر با نمونه‌های واقعی متفاوت است. این شبکه‌ها مانند دو دشمن عمل می‌کنند که سعی در رقابت با یکدیگر دارند. در مراحل اولیه، شبکه تمایزگر براحتی می‌تواند نمونه‌های جعلی تولید شده توسط مولد را شناسایی کند. سپس شبکه مولدِ رقیب، سخت کار می‌کند تا تفاوت نمونه‌های جعلی تولید شده از داده‌های واقعی را کاهش دهد. آن‌ها سعی می‌کنند نمونه‌هایی را نزدیک به نمونه‌های آموزشی تولید کنند و این کار را برای شبکه تمایزگر کمی چالش‌برانگیز می‌کند. با این حال، هنوز شبکه تمایزگر تلاش می‌کند تا جعلی بودن داده‌های تولید شده را بیابد. هر دو شبکه با یکدیگر رقابت می‌کنند تا زمانی که شبکه تمایزگر تشخیص اینکه کدام یک از نمونه‌های تولید شده توسط شبکه مولد جعلی و کدام واقعی است، برایش دشوار شود.

## آموزش به شیوه تخاصمی

مولد و تمایزگر در یک GAN به روشی تخاصمی آموزش می‌بینند، یعنی در یک چارچوب **مجموع صفر** (**zero-sum**) با یکدیگر به رقابت می‌پردازند تا داده‌هایی شبیه توزیع داده‌های

واقعی تولید شود. هدف مولد در GAN تولید نمونه‌های است که به‌نظر می‌رسد از توزیع داده‌های واقعی می‌آیند، حتی اگر جعلی هستند و هدف تمایزگر، تشخیصِ جعلی یا واقعی بودن نمونه‌های تولیدی است.

از منظرِ بهینه‌سازی، هدفِ آموزشیِ مولد، افزایش خطاهای تمایزگر است. به عبارت دیگر، هر چه تعداد اشتباهات بیشتری توسط تمایزگر انجام شود، مولد عملکرد بهتری دارد. هدف تمایزگر کاهش خطای خود است. در هر تکرار، هر دو شبکه با استفاده از گرادیان کاهشی به سمت اهداف خود می‌روند. جالب اینجاست که هر شبکه در تلاش است تا شبکه دیگری را شکست دهد. مولد تلاش می‌کند تمایزگر را فریب دهد، در حالی که تمایزگر تلاش می‌کند فریب نخورد. در نهایت، داده‌های تولیدی (مانند تصاویر، صدا، ویدئو، سری‌های زمانی) از مولد می‌تواند پیچیده‌ترین تمایزگر را فریب دهد.

در عمل، مولد نمونه‌های تصادفی را از یک توزیع از پیش تعریف‌شده با توزیع به‌عنوان ورودی می‌گیرد و داده‌هایی را تولید می‌کند که به نظر می‌رسد از توزیع هدف می‌آیند.

به عنوان مثالی از تولید یک تصویر، یک مدل GAN را می‌توان در شکل زیر نشان داد:

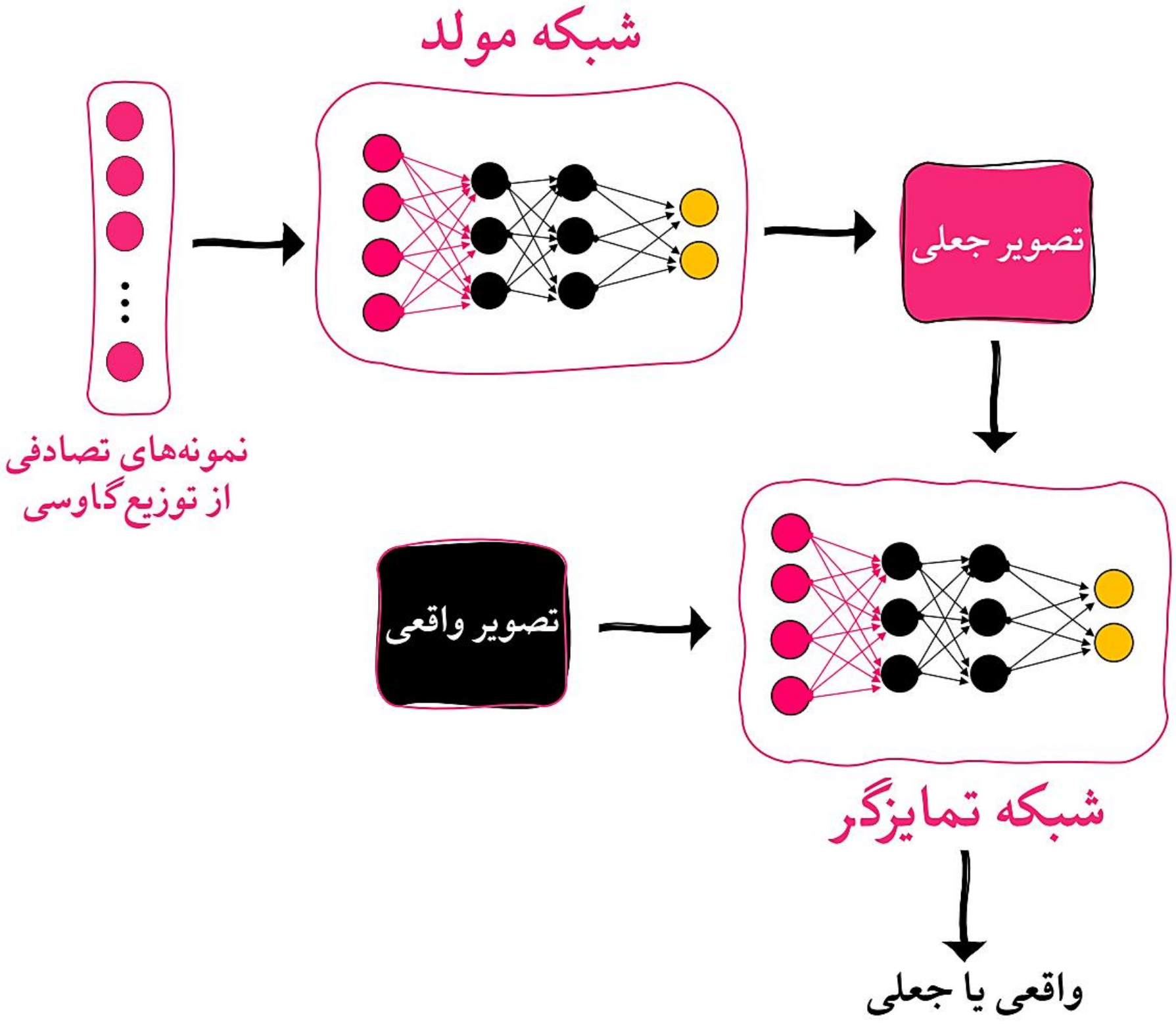

شکل ۶-۱. ساختار کلی یک شبکه متخاصم مولد.

مراحل زیر توسط GAN انجام می‌شود:

۱. شبکه مولد نمونه‌های تصادفی را از توزیع گاوسی می‌گیرد و تصاویر را ایجاد می‌کند.
۲. این تصاویر تولید شده سپس به شبکه تمایزگر داده می‌شوند.
۳. شبکه تمایزگر هم تصاویر تولید شده و هم تصاویری را که از مجموعه داده واقعی گرفته شده‌اند را می‌گیرد.
۴. تمایزگر احتمالات را در خروجی می‌دهد.
۵. زیانِ (تابع هزینه) مولد، بر اساسِ آنتروپیِ متقاطعِ تصاویرِ جعلی که توسط تمایزگر معتبر تلقی می‌شود، محاسبه می‌شود.
۶. زیانِ (تابع هزینه) تمایزگر بر اساسِ آنتروپیِ متقاطعِ تصاویرِ جعلی که جعلی تلقی می‌شوند، به اضافه آنتروپی متقاطع تصاویر واقعی که معتبر تلقی می‌شوند، محاسبه می‌شود.
۷. در هر دوره، هر دو شبکه به ترتیب برای به کمینه کردن زیان خود بهینه می‌شوند.
۸. در نهایت مولدِ به خوبی آموزش‌دیده، تصاویر را به عنوان خروجی نهایی تولید می‌کند که تصاویرِ ورودیِ واقعی را تقلید می‌کند.

## آموزش GAN

هدف اصلی ما این است که مولد تصاویر (داده‌های) واقعی تولید کند و چارچوب GAN وسیله‌ای برای این هدف است. قبل از پرداختن به جزئیات بیشتر، اجازه دهید به معرفی چند نماد بپردازیم:

- مولد را با $G(z,\theta_g)$ نشان می‌دهیم، جایی که $\theta_g$ وزن‌های شبکه هستند و $z$ **بردار نهفته (latent vector)** است که به عنوان ورودی مولد عمل می‌کند. آن را به عنوان یک مقدار بذر تصادفی (**random seed**) برای شروعِ فرآیندِ تولیدِ تصویر در نظر بگیرید. $z$ دارای توزیع احتمال $p_z(z)$ است که معمولاً **نرمال تصادفی (random normal)** یا **یکنواخت تصادفی (random uniform)** است. مولد نمونه‌های جعلی $x$ را با توزیع احتمال $p_g(x)$ خروجی می‌دهد. شما می‌توانید $p_g(x)$ را به عنوان توزیع احتمال داده‌های واقعی با توجه به مولد در نظر بگیرید.
- تمایزگر را با $D(x,\theta_d)$ نشان می‌دهیم که $\theta_d$ وزن شبکه است. داده‌های واقعی با توزیع $x\sim p_{data}(x)$ یا نمونه‌های تولید شده $x\sim p_g(x)$ را به عنوان ورودی می‌گیرد. تمایزگر یک طبقه‌بند دودویی است که براساس اینکه آیا تصویر ورودی واقعی (خروجی شبکه ۱) یا تولید شده (خروجی شبکه ۰) است، مقدار ۱ یا ۰ را خروجی می‌دهد.
- در طول آموزش، توابع زیان تمایزگر و مولد را به ترتیب با $J^{(D)}$ و $J^{(G)}$ نشان می‌دهیم.

آموزش GAN در مقایسه با آموزش یک شبکه‌یِ عصبیِ عمیقِ معمولی متفاوت است، چراکه ما دو شبکه داریم. می‌توانیم آن را به‌عنوان یک بازیِ مجموع-صفرِ کمین‌بیشِ متوالی دو بازیکن (مولد و تمایزگر) در نظر بگیریم:

۱. **متوالی (Sequential):** به این معنی که بازیکنان به نوبت پشت سر هم قرار می‌گیرند، شبیه به شطرنج (برخلاف هم‌زمان). ابتدا، تمایزگر سعی می‌کند $J^{(D)}$ را کمینه کند، اما تنها با تنظیم وزن‌های $\theta_d$ می‌تواند این کار را انجام دهد. سپس، مولد سعی می‌کند $J^{(G)}$ را به کمینه کند، اما فقط می‌تواند وزن‌های، $\theta_g$ را تنظیم کند. این فرآیند چندین بار تکرار می‌شود.

۲. **مجموع-صفر (Zero-sum):** به این معنی که سود یا زیان یک بازیکن دقیقا با سود یا زیان بازیکن مقابل متعادل می‌شود. یعنی مجموع ضرر مولد و تمایزگر همیشه ۰ است:

$$J^{(G)} = -J^{(D)}$$

۳. **کمین‌بیش (minimax):** به این معنی که استراتژی بازیکن اول (مولد) **کمینه کردنِ بیشینه‌یِ** امتیاز حریف (تمایزگر) است. وقتی تمایزگر را آموزش می‌دهیم، در تشخیص نمونه‌های واقعی و جعلی بهتر می‌شود (کمینه کردن $J^{(D)}$). در مرحله بعد، زمانی که مولد را آموزش می‌دهیم، سعی می‌کند تا سطحِ تمایزگر که به تازگی بهبود یافته است، بالا برود (ما $J^{(G)}$ را کمینه می‌کنیم، که معادل به بیشینه کردن $J^{(D)}$ است). این دو شبکه در حالِ رقابتِ دائمی هستند. ما بازی **minimax** را با موارد زیر نشان می‌دهیم که $V$ تابع هزینه است:

$$\boldsymbol{min_G max_D V(G,D)}$$

بیایید فرض کنیم که پس از چند مرحله آموزشی، هر دو $J^{(D)}$ و $J^{(G)}$ در کمینه محلی خواهند بود. سپس، راه‌حلِ بازیِ کمین‌بیش، **تعادل نش (Nash equilibrium)** نامیده می‌شود. تعادل نش زمانی اتفاق می‌افتد که یکی از بازیگران کنش خود را تغییر ندهد، صرف نظر از اینکه بازیگر دیگر چه کاری انجام دهد. تعادل نش در یک چارچوب GAN زمانی اتفاق می‌افتد که مولد آنقدر خوب شود که تمایزگر دیگر قادر به تشخیص نمونه‌های تولید شده و واقعی نباشد.

## آموزش تمایزگر

تمایزگر یک طبقه‌بند شبکه عصبی است و می‌توانیم آن را به روش معمول، با استفاده از گرادیان کاهشی و پس‌انتشار آموزش دهیم. با این حال، مجموعه آموزشی از بخش‌های مساوی نمونه‌های واقعی و تولید شده تشکیل شده است. بیایید ببینیم که چگونه می‌توان آن را در فرآیند آموزش گنجاند:

۱. بسته به نمونه ورودی (واقعی یا جعلی)، دو مسیر داریم:
- نمونه را از داده‌های واقعی $x \sim p_{data}$ انتخاب کرده و از آن برای تولید $D_x$ استفاده کنید.
- تولید نمونه جعلی $x \sim p_g$. در اینجا، مولد و تمایزگر به عنوان یک شبکه واحد کار می‌کنند. با یک بردار تصادفی $z$ شروع می‌کنیم که از آن برای تولید نمونه تولید شده $G_z$ استفاده می‌کنیم. سپس، از آن به عنوان ورودی تمایزگر برای تولید خروجی نهایی $D(G(z))$ استفاده می‌کنیم.

۲. محاسبه تابع زیان، که منعکس‌کننده‌ی دوگانگی داده‌های آموزشی است.
۳. گرادیان خطا را پس‌انتشار داده شده و وزن‌ها بروز می‌شوند. اگرچه این دو شبکه باهم کار می‌کنند، وزن‌های مولد $\theta_g$، قفل خواهند شد و ما فقط وزن‌های تمایزگر $\theta_d$ را بروز می‌کنیم. این تضمین می‌کند که ما عملکرد تمایزگر را بهبود می‌بخشیم.

برای درک از زیان تمایزگر، بیایید فرمول زیان آنتروپی متقاطع را یادآوری کنیم:

$$CE(p,q) = -\sum_{i=1}^{n} p_i(x)\log(q_i(x))$$

جایی که $q_i(x)$ احتمالِ تخمینِ خروجیِ متعلق به کلاس $i$ (از $n$ کلاس) و $p_i(x)$ احتمال واقعی است. برای سادگی، فرض می‌کنیم که فرمول را روی یک نمونه آموزشی اعمال می‌کنیم. در مورد طبقه‌بندی دودویی، این فرمول را می‌توان به صورت زیر ساده کرد:

$$CE(p,q) = -(p(x)\log q(x) + (1-p(x))\log(1-q(x))$$

می‌توانیم فرمول را برای یک ریزدسته از $m$ نمونه گسترش دهیم:

$$CE(p,q) = -\frac{1}{m}\sum_{j=i}^{m}\left(p(x_j)\log q(x_j) + (1-p(x_j))\log(1-q(x_j)\right)$$

با دانستن همه‌یِ این‌ها، حال بیایید زیان تمایزگر را تعریف کنیم:

$$J^{(D)} = -\frac{1}{2}\mathbb{E}_{x\sim p_{data}}\log(D(x)) - \frac{1}{2}\mathbb{E}_z\log\left(1-D(G(z))\right)$$

اگرچه پیچیده به نظر می‌رسد، با این حال فقط یک زیان آنتروپی متقاطع برای یک طبقه‌بند دودویی با برخی موارد خاص GAN است.

## آموزش مولد

ما مولد را با بهترکردنِ آن در فریب دادن متمایزگر آموزش خواهیم داد. برای انجام این کار، به هر دو شبکه نیاز داریم، مشابه روشی که ما تمایزگر را با نمونه‌های جعلی آموزش می‌دهیم:

۱. ما با یک بردار نهفته تصادفی $z$ شروع می‌کنیم و آن را از طریق مولد و تمایزگر تغذیه می‌کنیم تا خروجی $D(G(z))$ را تولید کنیم.
۲. تابع زیان، همان زیانِ تمایزگر است. با این حال، هدف ما در اینجا بیشینه کردن آن است، نه کمینه کردن. چرا که می‌خواهیم تمایزگر را فریب دهیم.
۳. در فاز عقب‌گرد، وزن‌های تمایزگر $\theta_d$ قفل هستند و فقط می‌توانیم $\theta_g$ را تنظیم کنیم. این به ما این امکان را می‌دهد به جای بدتر کردن تمایزگر، با بهتر کردن مولد زیان تمایزگر را بیشینه کنیم.

در این مرحله ما فقط از داده‌های تولید شده استفاده می‌کنیم. بخشی از تابع زیان که با داده‌های واقعی سروکار دارد همیشه ۰ خواهد بود. بنابراین، می‌توانیم آن را به صورت زیر ساده کنیم:

$$J^{(G)} = \mathbb{E}_z \log\left(1 - D\big(G(z)\big)\right)$$

در اوایل، زمانی که متمایزگر براحتی می‌تواند نمونه‌های واقعی و جعلی را تشخیص دهد، ($D\big(G(z)\big) \approx 0$)، گرادیان نزدیک به صفر خواهد بود. این امر منجر به یادگیری کمی از وزن‌ها می شود که این مشکل به عنوان **گرادیان کاهیده** (**diminished gradient**) شناخته می‌شود. ما می‌توانیم این مشکل را با استفاده از یک تابع زیان متفاوت حل کنیم:

$$J^{(G)} = -\mathbb{E}_z \log\left(D\big(G(z)\big)\right)$$

این زیان هنوز کمینه می‌شود، زمانی که $D\big(G(z)\big) \approx 1$ و در عین حال گرادیان بزرگ است (زمانی که مولد عملکرد ضعیفی دارد). با این زیان، بازی دیگر مجموع‌صفر نیست، اما تاثیر عملی بر چارچوب GAN نخواهد داشت.

## هر دو در کنار هم

با دانش جدیدمان، می‌توانیم هدف کمین‌بیشین را به طور کامل تعریف کنیم:

$$min_G max_D V(G,D) = \frac{1}{2}\mathbb{E}_{x\sim p_{data}} \log\big(D(x)\big) + \frac{1}{2}\mathbb{E}_z \log\left(1 - D\big(G(z)\big)\right)$$

به طور خلاصه، مولد سعی می کند هدف را کمینه کند، در حالی که تمایزگر سعی می‌کند آن را بیشینه کند. توجه داشته باشید، در حالی‌که تمایزگر باید زیان خود را کمینه کند، هدف کمین‌بیش منفی زیان تمایزگر است. بنابراین تمایزگر باید آن را بیشینه کند.

## تولید تصاویر جدید MNIST با GAN

در این بخش می‌خواهیم به پیاده‌سازی یک شبکه متخاصم مولد با استفاده از Keras و tensorflow بپردازیم. ابتدا کدهای مورد نیاز را وارد می‌کنیم:

```
from tensorflow.keras.datasets import mnist
from tensorflow.keras.layers import Input, Dense, Reshape, Flatten, Dropout
from tensorflow.keras.layers import BatchNormalization, Activation, ZeroPadding2D
from tensorflow.keras.layers import LeakyReLU
from tensorflow.keras.layers import UpSampling2D, Conv2D
from tensorflow.keras.models import Sequential, Model
from tensorflow.keras.optimizers import Adam
from tensorflow.keras import initializers
import matplotlib.pyplot as plt
import sys
import numpy as np
import tqdm
```

سپس، مجموعه داده MNIST را بارگذاری و نرمال می‌کنیم:

```
(X_train, _), (_, _) = mnist.load_data()
X_train = (X_train.astype(np.float32) - 127.5)/127.5
```

همانطور که احتمالا متوجه شدید، ما هیچ یک از برچسب‌ها یا مجموعه داده آزمایشی را بر نمی‌گردانیم. ما فقط از مجموعه داده آموزشی استفاده می‌کنیم. برچسب‌ها مورد نیاز نیستند، زیرا تنها برچسب‌هایی که استفاده خواهیم کرد ۰ برای جعلی و ۱ برای واقعی است. اینها تصاویر واقعی هستند، بنابراین به همه آن‌ها یک برچسب ۱ در تمایزگر اختصاص داده می‌شود.

ما از یک پرسپترون چند لایه استفاده خواهیم کرد و به آن تصویری به عنوان یک بردارِ مسطح با اندازه ۷۸۴ می‌دهیم، بنابراین داده‌های آموزشی را تغییر شکل می‌دهیم:

```
X_train = X_train.reshape(60000, 784)
```

اکنون باید یک مولد و تمایزگر بسازیم. هدف مولد دریافت یک ورودی نویزدار و تولید تصویری مشابه مجموعه داده آموزشی است. اندازه ورودی نویز توسط متغیر randomDim تعیین می‌شود. می توانید آن را به هر مقداری مقداردهی اولیه کنید. معمولا آن را روی ۱۰۰ تنظیم می‌کنند. برای پیاده‌سازی، ما مقدار ۱۰ را امتحان کردیم. این ورودی به یک لایه متراکم با ۲۵۶ نورون با فعال‌سازی LeakyReLU وارد می‌شود. در مرحله بعد یک لایه متصل کامل دیگر با ۵۱۲ نورون پنهان اضافه می‌کنیم، به دنبال آن لایه سوم پنهان با ۱۰۲۴ نورون و در نهایت لایه خروجی با ۷۸۴ نورون را اضافه می‌کنیم. می‌توانید تعداد نورون‌ها را در لایه‌های پنهان تغییر دهید و ببینید عملکرد چگونه تغییر می‌کند. **با این حال، تعداد نورون‌ها در واحد خروجی باید با تعداد پیکسل‌های موجود در تصاویر آموزشی مطابقت داشته باشد.** مولد مربوط به صورت زیر است:

```
randomDim = 10
generator = Sequential()
generator.add(Dense(256, input_dim=randomDim))
generator.add(LeakyReLU(0.2))
generator.add(Dense(512))
generator.add(LeakyReLU(0.2))
generator.add(Dense(1024))
generator.add(LeakyReLU(0.2))
generator.add(Dense(784, activation='tanh'))
```

به طور مشابه، ما یک تمایزگر ایجاد می‌کنیم. توجه داشته باشید که تمایزگر تصاویر را از مجموعه آموزشی یا تصاویر تولید شده توسط مولد می‌گیرد، بنابراین اندازه ورودی آن ۷۸۴ است. با این حال، خروجی تشخیص دهنده یک بیت است و ۰ نشان‌دهنده یک تصویر جعلی است (تولید شده توسط مولد) و ۱ نشان می‌دهد که تصویر از مجموعه داده آموزشی است:

```
discriminator = Sequential()
discriminator.add(Dense(1024, input_dim=784,
kernel_initializer=initializers.RandomNormal(stddev=0.02)))
discriminator.add(LeakyReLU(0.2))
discriminator.add(Dropout(0.3))
discriminator.add(Dense(512))
discriminator.add(LeakyReLU(0.2))
discriminator.add(Dropout(0.3))
discriminator.add(Dense(256))
discriminator.add(LeakyReLU(0.2))
discriminator.add(Dropout(0.3))
discriminator.add(Dense(1, activation='sigmoid'))
```

سپس، مولد و تمایزگر را با هم ترکیب می‌کنیم تا یک GAN تشکیل دهیم. در GAN، با تنظیم آرگومان trainable روی False، مطمئن می‌شویم که وزن‌های تمایزگر ثابت می‌شوند:

```
# Combined network
discriminator.trainable = False
ganInput = Input(shape=(randomDim,))
x = generator(ganInput)
ganOutput = discriminator(x)
gan = Model(inputs=ganInput, outputs=ganOutput)
```

تدبیر برای آموزش این دو این است که ابتدا تمایزگر را جداگانه آموزش می‌دهیم. از زیان آنتروپی متقاطع دودویی برای تمایزگر استفاده می‌کنیم. سپس، وزن‌های تمایزگر را freeze می‌کنیم و GAN ترکیبی را آموزش می‌دهیم. این منجر به آموزش مولد می‌شود:

```
discriminator.compile(loss='binary_crossentropy', optimizer=adam)
gan.compile(loss='binary_crossentropy', optimizer=adam)
```

حالا بیایید آموزش را اجرا شروع کنیم. برای هر دوره ابتدا نمونه‌ای از نویز تصادفی می‌گیریم، آن را به مولد می‌دهیم و مولد یک تصویر جعلی تولید می‌کند. ما تصاویر جعلی تولید شده و تصاویر آموزشی واقعی را در یک دسته با برچسب‌های خاص خود ترکیب می‌کنیم و از آن‌ها برای آموزش تمایزگر در دسته داده شده استفاده می‌کنیم:

```
dLosses = []
gLosses = []
def train(epochs=1, batchSize=128):
    batchCount = int(X_train.shape[0] / batchSize)
    print ('Epochs:', epochs)
    print ('Batch size:', batchSize)
    print ('Batches per epoch:', batchCount)

    for e in range(1, epochs+1):
        print ('-'*15, 'Epoch %d' % e, '-'*15)
        for _ in range(batchCount):
            # Get a random set of input noise and images
            noise = np.random.normal(0, 1, size=[batchSize, randomDim])
            imageBatch = X_train[np.random.randint(0, X_train.shape[0],
size=batchSize)]

            # Generate fake MNIST images
            generatedImages = generator.predict(noise)
            # print np.shape(imageBatch), np.shape(generatedImages)
            X = np.concatenate([imageBatch, generatedImages])

            # Labels for generated and real data
            yDis = np.zeros(2*batchSize)
            # One-sided label smoothing
            yDis[:batchSize] = 0.9

            # Train discriminator
            discriminator.trainable = True
            dloss = discriminator.train_on_batch(X, yDis)
```

حالا در همان حلقه for، ژنراتور را آموزش می‌دهیم. ما می خواهیم تصاویر تولید شده توسط مولد، توسط تمایزگر، واقعی تشخیص داده شوند، بنابراین از یک بردار تصادفی (نویز) به عنوان ورودی به مولد استفاده می‌کنیم. این یک تصویر جعلی تولید می‌کند و سپس GAN را طوری آموزش می‌دهد که متمایز کننده تصویر را واقعی درک کند (خروجی ۱):

```
            # Train generator
            noise = np.random.normal(0, 1, size=[batchSize, randomDim])
            yGen = np.ones(batchSize)
            discriminator.trainable = False
            gloss = gan.train_on_batch(noise, yGen)
```

در صورت تمایل، می‌توانید زیان مولد و تمایزگر و همچنین تصاویر تولید شده را ذخیره کنید. در مرحله بعد، ما زیان را برای هر دوره ذخیره می‌کنیم و پس از هر ۲۰ دوره تصاویر تولید می‌کنیم:

```
        # Store loss of most recent batch from this epoch
        dLosses.append(dloss)
        gLosses.append(gloss)

        if e == 1 or e % 20 == 0:
            saveGeneratedImages(e)
```

برای ترسیم زیان و تصاویر تولید شده ارقام دست‌نویس، دو تابع کمکی، plotLoss و saveGeneratedImages تعریف می‌کنیم. کد آن‌ها به شرح زیر است:

```
# Plot the loss from each batch
def plotLoss(epoch):
    plt.figure(figsize=(10, 8))
    plt.plot(dLosses, label='Discriminitive loss')
    plt.plot(gLosses, label='Generative loss')
    plt.xlabel('Epoch')
    plt.ylabel('Loss')
    plt.legend()
    plt.savefig('images/gan_loss_epoch_%d.png' % epoch)

# Create a wall of generated MNIST images
def saveGeneratedImages(epoch, examples=100, dim=(10, 10), figsize=(10,
10)):
    noise = np.random.normal(0, 1, size=[examples, randomDim])
    generatedImages = generator.predict(noise)
    generatedImages = generatedImages.reshape(examples, 28, 28)

    plt.figure(figsize=figsize)
    for i in range(generatedImages.shape[0]):
        plt.subplot(dim[0], dim[1], i+1)
        plt.imshow(generatedImages[i], interpolation='nearest',
cmap='gray_r')
        plt.axis('off')
    plt.tight_layout()
    plt.savefig('images/gan_generated_image_epoch_%d.png' % epoch)
```

حال بیاید تمام کدها را یکجا داشته باشیم:

```
from tensorflow.keras.datasets import mnist
from tensorflow.keras.layers import Input, Dense, Reshape, Flatten,
Dropout
from tensorflow.keras.layers import BatchNormalization, Activation,
ZeroPadding2D
from tensorflow.keras.layers import LeakyReLU
from tensorflow.keras.layers import UpSampling2D, Conv2D
from tensorflow.keras.models import Sequential, Model
from tensorflow.keras.optimizers import Adam
from tensorflow.keras import initializers
import matplotlib.pyplot as plt
import sys
import numpy as np
import tqdm

randomDim = 10

# Load MNIST data
(X_train, _), (_, _) = mnist.load_data()
X_train = (X_train.astype(np.float32) - 127.5)/127.5
X_train = X_train.reshape(60000, 784)

generator = Sequential()
generator.add(Dense(256, input_dim=randomDim)) #,
kernel_initializer=initializers.RandomNormal(stddev=0.02)))
generator.add(LeakyReLU(0.2))
generator.add(Dense(512))
generator.add(LeakyReLU(0.2))
generator.add(Dense(1024))
generator.add(LeakyReLU(0.2))
generator.add(Dense(784, activation='tanh'))
```

```
adam = Adam(learning_rate=0.0002, beta_1=0.5)

discriminator = Sequential()
discriminator.add(Dense(1024, input_dim=784,
kernel_initializer=initializers.RandomNormal(stddev=0.02)))
discriminator.add(LeakyReLU(0.2))
discriminator.add(Dropout(0.3))
discriminator.add(Dense(512))
discriminator.add(LeakyReLU(0.2))
discriminator.add(Dropout(0.3))
discriminator.add(Dense(256))
discriminator.add(LeakyReLU(0.2))
discriminator.add(Dropout(0.3))
discriminator.add(Dense(1, activation='sigmoid'))
discriminator.compile(loss='binary_crossentropy', optimizer=adam)

# Combined network
discriminator.trainable = False
ganInput = Input(shape=(randomDim,))
x = generator(ganInput)
ganOutput = discriminator(x)
gan = Model(inputs=ganInput, outputs=ganOutput)
gan.compile(loss='binary_crossentropy', optimizer=adam)

dLosses = []
gLosses = []

# Plot the loss from each batch
def plotLoss(epoch):
    plt.figure(figsize=(10, 8))
    plt.plot(dLosses, label='Discriminitive loss')
    plt.plot(gLosses, label='Generative loss')
    plt.xlabel('Epoch')
    plt.ylabel('Loss')
    plt.legend()
    plt.savefig('images/gan_loss_epoch_%d.png' % epoch)

# Create a wall of generated MNIST images
def saveGeneratedImages(epoch, examples=100, dim=(10, 10),
figsize=(10, 10)):
    noise = np.random.normal(0, 1, size=[examples, randomDim])
    generatedImages = generator.predict(noise)
    generatedImages = generatedImages.reshape(examples, 28, 28)

    plt.figure(figsize=figsize)
    for i in range(generatedImages.shape[0]):
        plt.subplot(dim[0], dim[1], i+1)
        plt.imshow(generatedImages[i], interpolation='nearest',
cmap='gray_r')
        plt.axis('off')
    plt.tight_layout()
    plt.savefig('images/gan_generated_image_epoch_%d.png' % epoch)

def train(epochs=1, batchSize=128):
    batchCount = int(X_train.shape[0] / batchSize)
    print ('Epochs:', epochs)
    print ('Batch size:', batchSize)
```

```
    print ('Batches per epoch:', batchCount)

    for e in range(1, epochs+1):
        print ('-'*15, 'Epoch %d' % e, '-'*15)
        for _ in range(batchCount):
            # Get a random set of input noise and images
            noise = np.random.normal(0, 1, size=[batchSize,
randomDim])
            imageBatch = X_train[np.random.randint(0,
X_train.shape[0], size=batchSize)]

            # Generate fake MNIST images
            generatedImages = generator.predict(noise)
            # print np.shape(imageBatch), np.shape(generatedImages)
            X = np.concatenate([imageBatch, generatedImages])

            # Labels for generated and real data
            yDis = np.zeros(2*batchSize)
            # One-sided label smoothing
            yDis[:batchSize] = 0.9

            # Train discriminator
            discriminator.trainable = True
            dloss = discriminator.train_on_batch(X, yDis)

            # Train generator
            noise = np.random.normal(0, 1, size=[batchSize,
randomDim])
            yGen = np.ones(batchSize)
            discriminator.trainable = False
            gloss = gan.train_on_batch(noise, yGen)

        # Store loss of most recent batch from this epoch
        dLosses.append(dloss)
        gLosses.append(gloss)

        if e == 1 or e % 20 == 0:
            saveGeneratedImages(e)


    # Plot losses from every epoch
    plotLoss(e)
```

اکنون می‌توانیم GAN را آموزش دهیم:

```
train(200, 128)
```

```
Epochs: 200
Batch size: 128
Batches per epoch: 468
--------------- Epoch 1 ---------------
--------------- Epoch 2 ---------------
--------------- Epoch 3 ---------------
--------------- Epoch 4 ---------------
--------------- Epoch 5 ---------------
...
--------------- Epoch 198 ---------------
--------------- Epoch 199 ---------------
--------------- Epoch 200 ---------------
```

در شکل زیر می‌توانید نمودار زیان مولد و تمایزگرِ GAN در حال یادگیری را مشاهده کنید:

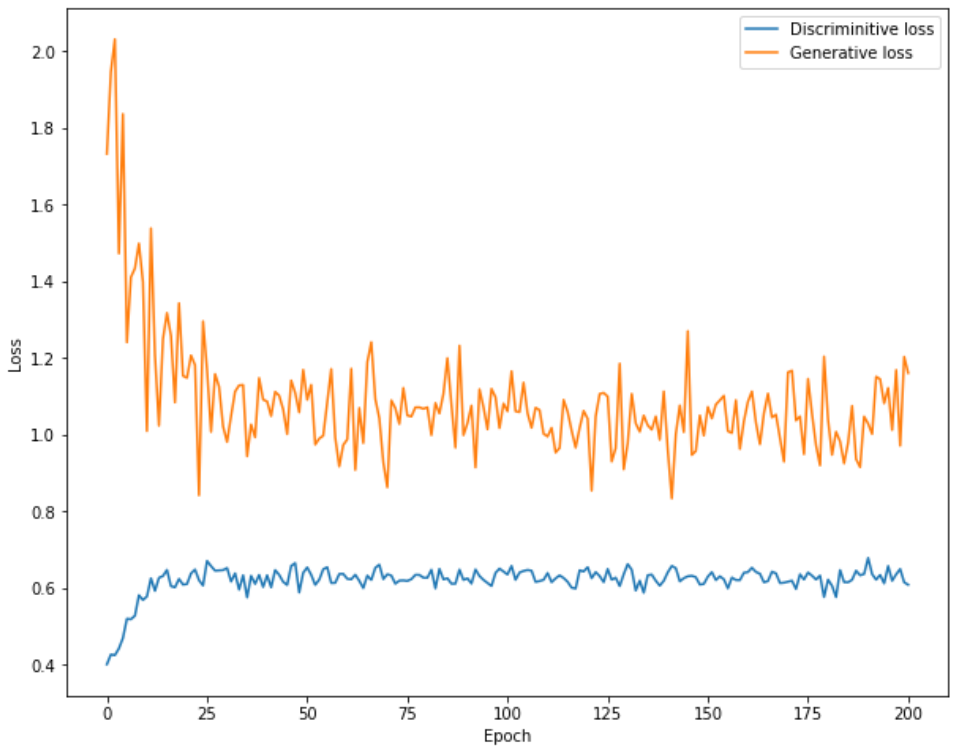

ارقام دست‌نویس تولید شده توسط GAN ما در دوره‌های مختلف به صورت زیر هستند:

Epoch 1:

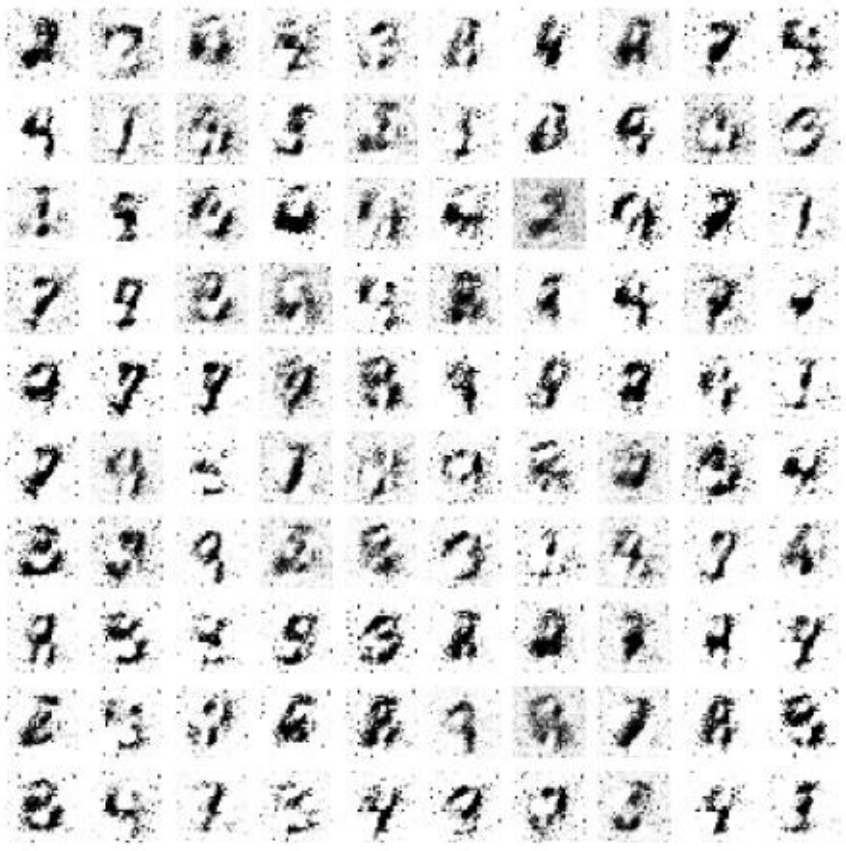

Epoch 20:

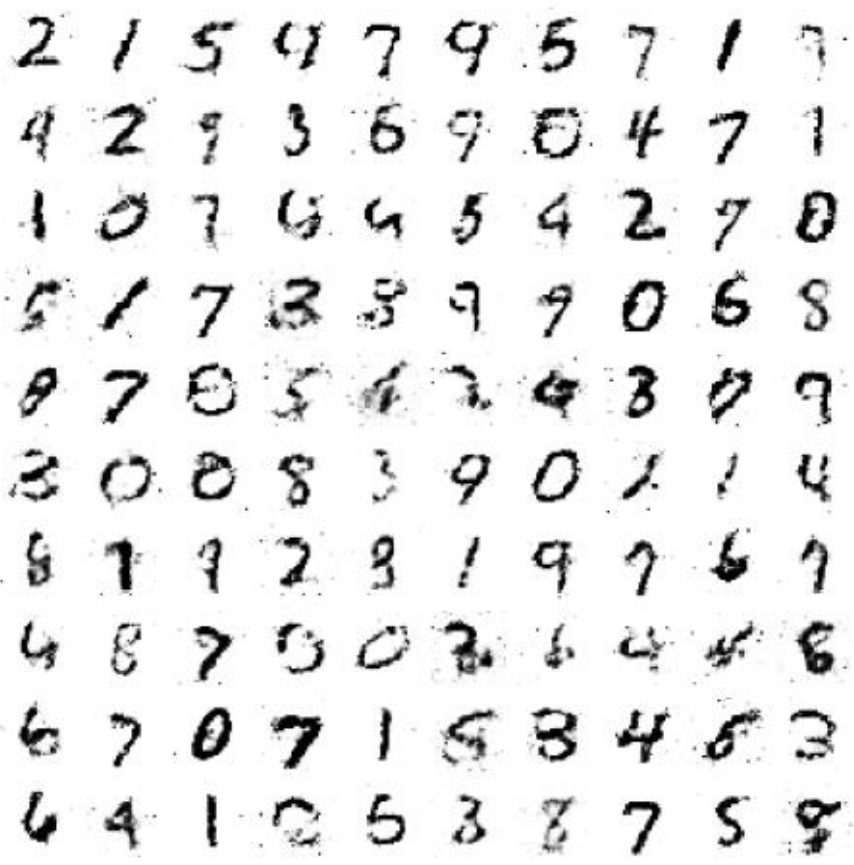

Epoch 100:

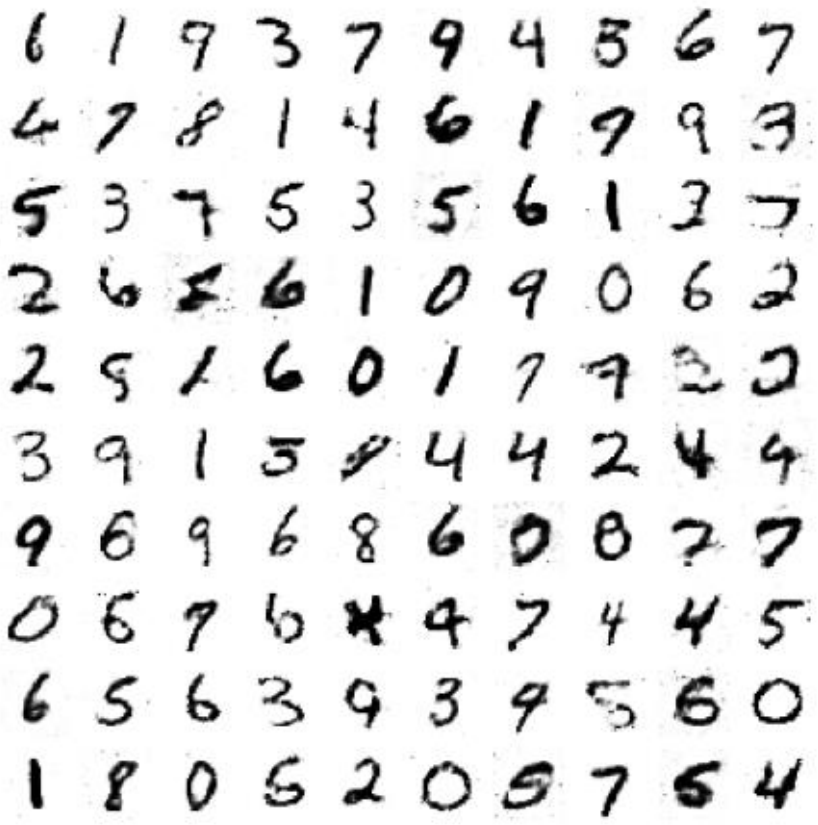

Epoch 200:

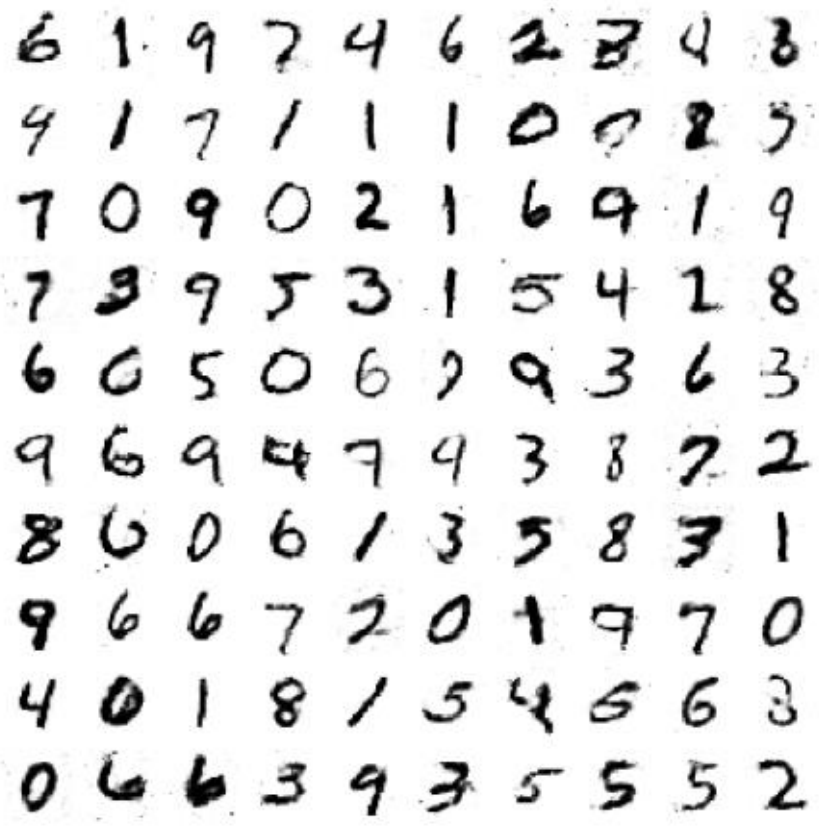

همان‌طور ک از شکل‌های قبلی می‌توانید ببینید، با افزایش دوره‌ها، ارقام دست‌نویس تولید شده توسط GAN بیشتر و بیشتر واقعی می‌شوند.

## چالش‌های مرتبط با آموزش شبکه‌های متخاصم مولد

هنگام کار با GAN چالش‌های زیادی وجود دارد. آموزش یک شبکه عصبی منفرد به دلیل تعدادِ زیادِ گزینه‌های درگیر، می‌تواند دشوار باشد: **معماری**، **توابع فعال‌سازی**، **الگوریتم بهینه‌سازی**، **نرخ یادگیری** و **نرخ حذف تصادفی**. شبکه‌های متخاصم مولد، همه این انتخاب‌ها را دو برابر می‌کنند و پیچیدگی‌های جدیدی را نیز اضافه می‌کنند. هم مولد و هم تمایزگر ممکن است تدبیرهایی را که قبلا در آموزش خود استفاده کرده بودند، فراموش کنند. این می‌تواند منجر به گرفتار شدن دو شبکه در چرخه‌یِ پایداریِ از راه‌حل‌ها شود که در طول زمان بهبود نمی‌یابد. یک شبکه ممکن است بر شبکه دیگر غلبه کند، به‌طوری که هیچ یک دیگر نتوانند یاد بگیرند. یا ممکن است مولد، بسیاری از فضایِ راه‌حلِ ممکن را کشف نکند و فقط از آن برای یافتن راه‌حل‌های واقع‌بینانه استفاده می‌کند. این وضعیت به عنوان **فروپاشی حالت (mode collapse)** شناخته می‌شود.

فروپاشی حالت زمانی رخ می‌دهد که مولد تنها زیرمجموعه کوچکی از حالت‌هایِ واقعی ممکن را یاد می‌گیرد. به عنوان مثال، اگر مسئله تولیدِ تصاویر گربه‌ها باشد، مولد می‌تواند یاد بگیرد که فقط تصاویری از گربه‌های مو کوتاه و مشکی را تولید کند. مولد تمام حالت‌های دیگر که شامل گربه‌هایی با رنگ‌های دیگر است را از دست می‌دهد.

راهبردهای بسیاری برای رسیدگی به این موضوع ارائه گردیده است، از جمله نرمال‌سازی دسته‌ای، افزودن برچسب‌ها به داده‌های آموزشی و غیره. افزودن برچسب به داده‌ها، یعنی تقسیم آن‌ها به دسته‌ها، تقریبا همیشه عملکرد GANها را بهبود می‌بخشد. به جای یادگیری ایجاد تصاویر از حیوانات خانگی به طور کلی، تولید تصاویری برای مثال از گربه‌ها، سگ‌ها، پرنده‌ها باید آسان‌تر باشد.

## خلاصه فصل

- **به‌طور کلی دو نوع مدل اصلی در یادگیری ماشین وجود دارد: مدل مولد و مدل تفکیک‌گر.**
- **مدل‌های تفکیک‌گر بر پیش‌بینی کلاس‌های داده با توجه به ویژگی‌های آن‌ها تمرکز می‌کنند.**
- **مدل مولد سعی نمی‌کند ویژگی‌ها را به کلاس‌ها نگاشت کند، بلکه ویژگی‌هایی را تولید می‌کند که در یک کلاس خاص وجود دارد.**
- **مدل‌های مولد از یادگیری بدون‌نظارت پیروی می‌کنند که به‌طور خودکار الگوها یا بی‌نظمی‌های داده‌های مورد تجزیه و تحلیل را کشف می‌کند.**

- هدف اصلی انواع مدل‌های مولد یادگیری توزیعِ واقعیِ داده‌هایِ مجموعه آموزشی است.

## آزمونک

۱. از منظر بهینه‌سازی، هدف آموزشی مولد و تمایزگر چیست؟

۲. تمایزگر با چه روشی آموزش می‌یابد؟

۳. ورودی مولد چه چیزی است؟

۴. فروپاشی حالت چیست؟

# Deep Learning

From Basics to Building Deep Neural Networks with Python

**Milad Vazan**